\RequirePackage{fix-cm}
\PassOptionsToPackage{table,dvipsnames}{xcolor}
\PassOptionsToPackage{noabbrev,nameinlink,capitalize}{cleveref}
\PassOptionsToPackage{section}{placeins}
\documentclass[]{selfevolagent}
\usepackage{microtype}
\usepackage{amsfonts}
\usepackage{xcolor}
\definecolor{lavender}{RGB}{230,230,250}
\definecolor{softlavender}{RGB}{238, 223, 255}
\definecolor{selfevolagent}{RGB}{242,240,250}
\usepackage{graphicx}
\usepackage{booktabs}
\usepackage{wrapfig}
\usepackage{float}
\usepackage{amsmath}
\usepackage{amsthm}
\usepackage{amssymb}

\usepackage{subcaption}
\usepackage{makecell}
\usepackage{algpseudocode}
\usepackage[linesnumbered,lined,boxed,commentsnumbered,ruled,longend]{algorithm2e}
\theoremstyle{plain}

\newcommand{\vx}{\mathbf{x}}
\newcommand{\vs}{\mathbf{s}}
\newcommand{\va}{\mathbf{a}}
\newcommand{\vc}{\mathbf{c}}
\newcommand{\vv}{\mathbf{v}}
\newcommand{\vw}{\mathbf{w}}

\theoremstyle{definition}

\theoremstyle{remark}

\usepackage{enumitem}
\usepackage{pifont}
\usepackage[edges]{forest}
\usetikzlibrary{arrows.meta}
\usepackage{tikz}
\usepackage[most]{tcolorbox}
\usepackage{fontawesome5}
\usepackage{url}
\usepackage{longtable}
\usepackage{tabularx}
\usetikzlibrary{
  positioning,
  calc,
  shapes.symbols,
  shapes.geometric,
  shapes.misc,
  backgrounds
}
\usepackage[T1]{fontenc}
\usepackage{lmodern}

\makeatletter
\@ifundefined{showhyphens}{}{%
  \expandafter\let\csname showhyphens \endcsname\showhyphens
}
\makeatother

\definecolor{selfevolagent_dark}{HTML}{37D2A6} 
\definecolor{selfevolagent_light}{HTML}{9BE9D3}
\definecolor{selfevolagent_lighter}{HTML}{CDF4E9}

\newcommand\blfootnote[1]{%
  \begingroup
  \renewcommand\thefootnote{}\footnote{#1}%
  \addtocounter{footnote}{-1}%
  \endgroup
}

\usepackage{svg}

\usepackage{xltabular}
\usepackage{caption}

\newcommand{\ghlink}[1]{\faIcon{github}\,\href{#1}{GitHub}}
\newcommand{\weblink}[1]{\faIcon{globe}\,\href{#1}{Website}}

\newcommand{\kemingwu}[1]{}
\newcommand{\kaichen}[1]{}
\newcommand{\zuhao}[1]{}
\newcommand{\xjqi}[1]{}
\newcommand{\shizun}[1]{}
\newcommand{\syx}[1]{}

\newcommand{\cmark}{\textcolor{green}{\ding{52}}}
\newcommand{\xmark}{\textcolor{red}{\ding{56}}}

\definecolor{rootcolor}{RGB}{101, 45, 144}
\definecolor{catcolor}{RGB}{255, 192, 0}
\definecolor{subcatcolor}{RGB}{237, 125, 49}
\definecolor{papercolor}{RGB}{68, 114, 196}

\newenvironment{highlightbox}[1]{%
  \begin{tcolorbox}[
    colback=selfevolagent_lighter!30,
    colframe=selfevolagent_light!90,
    colbacktitle=selfevolagent_light!90,
    coltitle=black,
    title={\bfseries\fontfamily{ppl}\selectfont #1},
    boxrule=1.5pt,
    arc=4pt,
    drop shadow,
    parbox=false,
    before skip=8pt,
    after skip=8pt,
    left=6pt,
    right=6pt,
  ]%
}{%
  \end{tcolorbox}%
}

\title{Visual Generation in the New Era: An Evolution from Atomic Mapping to Agentic World Modeling}

\author{Keming Wu$^{1,12,\dagger\text{\faCube}}$}
\author{Zuhao Yang$^{2,12,\dagger\text{\faCube}}$}
\author{Kaichen Zhang$^{3,12,\dagger}$}
\author{Shizun Wang$^{4,\dagger}$}
\author{Haowei Zhu$^{1,\dagger}$}
\author{Sicong Leng$^{2}$}
\author{Zhongyu Yang$^{2}$}
\author{Qijie Wang$^{1}$}
\author{Sudong Wang$^{11}$}
\author{Ziting Wang$^{2}$}
\author{Zili Wang$^{6}$}
\author{Hui Zhang$^{9}$}
\author{Haonan Wang$^{4}$}
\author{Hang Zhou$^{8}$}
\author{Yifan Pu$^{1}$}
\author{Xingxuan Li$^{7}$}
\author{Fangneng Zhan$^{10}$}
\author{Bo Li$^{2,12,}$}
\author{Lidong Bing$^{7}$}
\author{Yuxin Song$^{8,\ddagger}$}
\author{Ziwei Liu$^{2,12,\ddagger}$}
\author{Wenhu Chen$^{5,\ddagger}$}
\author{Jingdong Wang$^{8,\ddagger}$}
\author{Xinchao Wang$^{4,\ddagger}$}
\author{Xiaojuan Qi$^{3,\ddagger}$}
\author{Shijian Lu$^{2,\ddagger}$}
\author{Bin Wang$^{1,\ddagger}$}

\affiliation{
\vspace{0.5em}
\;\faCube \;Project Organizer.\;$^\dagger$Core Contributor.\; \;$^\ddagger$Senior Supervisor.

\vspace{0.5em}

\texttt{\textbf{Affiliations}: $^1$Tsinghua University, $^2$Nanyang Technological University, $^3$University of Hong Kong, $^4$National University of Singapore, $^5$University of Waterloo, $^6$StepFun, $^7$MiroMind, $^8$Baidu, $^9$Fudan University, $^{10}$Hong Kong University of Science and Technology, $^{11}$Hong Kong University of Science and Technology (Guangzhou), $^{12}$LMMs-Lab}}

\abstract{

Recent visual generation models have improved photorealism, typography, instruction following, and interactive editing, yet they raise a deeper question: what does it mean to become ``better''? Despite high fidelity and complex prompt following, systems still struggle with spatial reasoning, persistent state, long-horizon consistency, and causal understanding---excelling at appearance while falling short of structural, temporal, and causal coherence.
We argue the field must move beyond appearance synthesis to \emph{intelligent visual generation}: plausible visuals grounded in structure, dynamics, domain knowledge, and causal relations. We formalize this as a progression from single-pass rendering to controllable composition, persistent coherence, closed-loop interaction, and causal world modeling.
We introduce a five-level taxonomy---\emph{Atomic Generation}, \emph{Conditional Generation}, \emph{In-Context Generation}, \emph{Agentic Generation}, and \emph{World-Modeling Generation}---from passive renderers to interactive, agentic, world-aware generators, where each level nests prior capabilities and adds a qualitatively new competence. We analyze drivers of this shift---diffusion-to-flow matching, unified understanding-and-generation models, improved visual representations, supervised fine-tuning and preference-based post-training, reward modeling, large-scale data curation and synthetic data distillation, and sampling acceleration and distillation for real-time deployment---as collectively advancing a shared trajectory toward intelligent visual generation rather than isolated improvements.
Evaluations often overestimate progress by privileging perceptual quality over structural, temporal, and causal weaknesses. We therefore pair benchmark review with in-the-wild stress tests and expert-constrained visual case studies that map failure modes to taxonomy levels to locate the frontier. Reframing generation as a path toward broader visual intelligence, this roadmap outlines a research agenda for controllable, interactive, physically grounded systems. We hope this roadmap provides the community with a capability-centered lens and practical insights for understanding, evaluating, and advancing the next generation of intelligent visual generation systems.

}

\date{\today}
\metadata[\faEnvelope\ Main Author Contact ]{wukeming0608@gmail.com, yang0756@e.ntu.edu.sg, zhan0564@e.ntu.edu.sg, shizun.wang@u.nus.edu}
\metadata[{\raisebox{-0.2ex}{\includegraphics[height=1em]{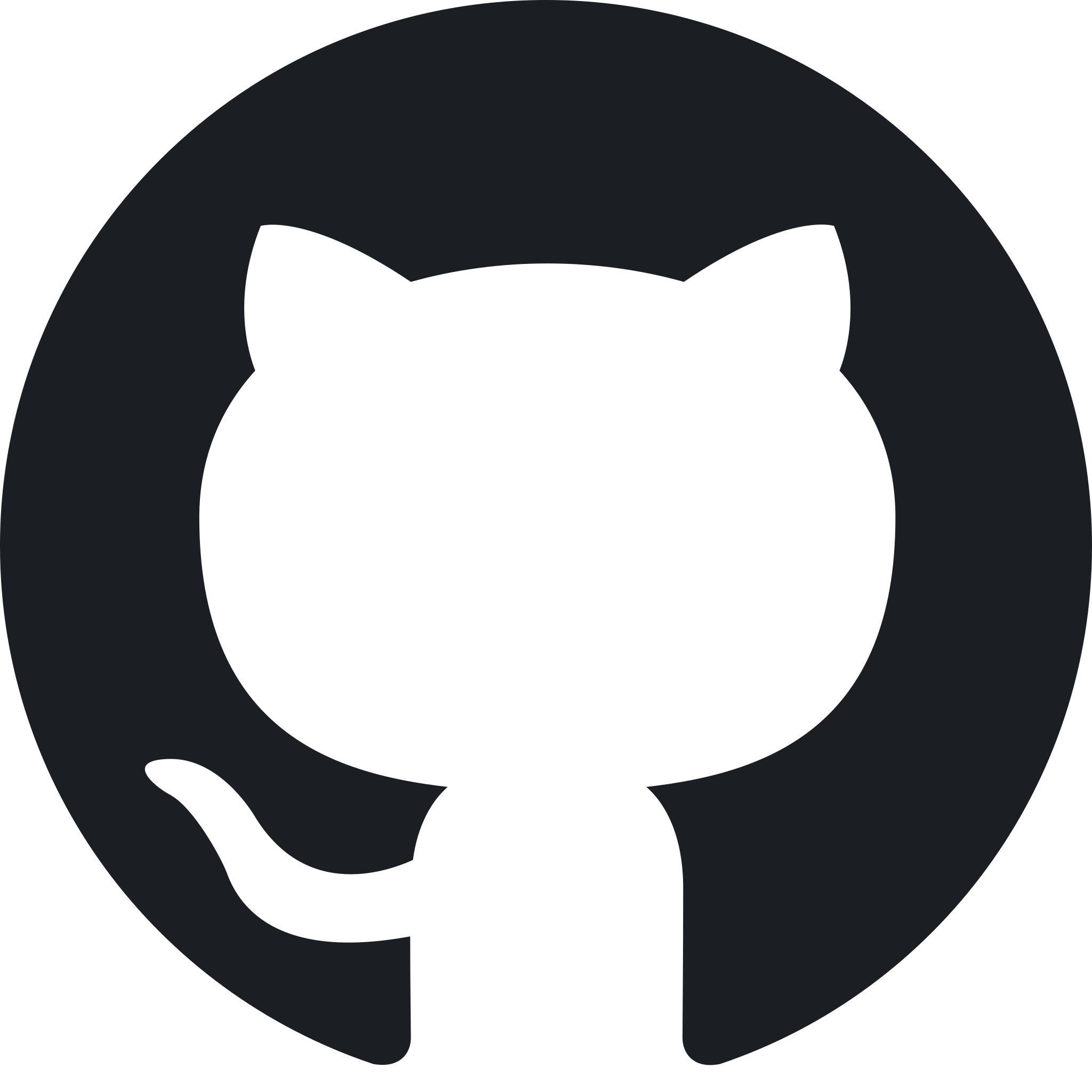}}\ Project Page}]{\url{https://github.com/EvolvingLMMs-Lab/Evolving-Visual-Generation}} 

\metadata[\faBullhorn\ Declaration]{This manuscript is a preprint and an evolving roadmap. Its analysis, taxonomy, references, and viewpoints may be corrected, updated, or revised in future versions as the field develops and new evidence or community feedback emerges.}

\begin{document}

\maketitle

\clearpage
\tableofcontents

\clearpage
\section{Introduction}
\label{sec:introduction}

\blfootnote{Note: If you identify your own or other papers relevant to this roadmap that have not been discussed (we apologize for any such omissions due to the rapidly expanding literature), please feel free to contact us via email or raise an issue on \href{https://github.com/EvolvingLMMs-Lab/Awesome-New-Era-Visual-Gen}{GitHub}.}

\vspace{0.6em}
\begin{tcolorbox}[
  colback=selfevolagent_light!20,
  colframe=selfevolagent_light!80,
  colbacktitle=selfevolagent_light!80,
  coltitle=black,
  title={\bfseries\fontfamily{ppl}\selectfont{Key Questions}},
  boxrule=2pt,
  arc=5pt,
  drop shadow,
  parbox=false,
  before skip=5pt,
  after skip=5pt,
  left=5pt,
  right=5pt,
]
\begin{itemize}[leftmargin=*]

\item[\ding{182}] \textbf{Capability Taxonomy:}
What does it mean for a visual generation model to become ``more intelligent,'' and how can we organize progress from atomic rendering to world-modeling generation?

\item[\ding{183}] \textbf{Modeling Mechanisms:}
How do diffusion, flow matching, autoregressive modeling, hybrid AR--diffusion systems, and unified multimodal architectures each change the trade-off between fidelity, controllability, reasoning, and efficiency?

\item[\ding{184}] \textbf{Training and Data Engines:}
Why are modern gains increasingly driven by data density, VLM-based relabeling, continued training, SFT, preference alignment, reward models, and acceleration rather than by parameter scaling alone?

\item[\ding{185}] \textbf{Applications as Constraints:}
How do applications such as personalization, layout control, typography, editing, domain adaptation, and embodied prediction reveal increasingly explicit requirements for structure, memory, and state consistency?

\item[\ding{186}] \textbf{Evaluation and Stress Testing:}
Why do current metrics overestimate progress, and how can in-the-wild stress tests expose failures in spatial logic, physical reasoning, identity preservation, text fidelity, and causal grounding?

\item[\ding{187}] \textbf{Agentic and World-Modeling Frontiers:}
What separates today’s strong renderers from closed-loop visual agents and playable world models, and how might visual chain-of-thought, tool use, verification, and world simulation define the next stage?

\end{itemize}
\end{tcolorbox}

\vspace{1em}
\vspace{-0.4em}

\paragraph{The Paradigm Shift: From Static Generation to World Simulation.}
Visual generation has evolved from a narrowly defined text-to-image problem into a much broader interface for composing, editing, and increasingly \emph{simulating} visual worlds. Early milestones in generative modeling established the feasibility of mapping language or latent codes to photorealistic pixels, and latent diffusion systems such as Stable Diffusion~\citep{rombach2022stablediffusion} made high-quality synthesis scalable and widely accessible. Yet this classical framing largely treated image generation as a one-shot distribution-matching problem: given a prompt, sample an image that looks plausible. In the latest generation of foundation models, this perspective is no longer sufficient. Rectified-flow and flow-matching objectives~\citep{liu2022flow,lipman2023flowmatchinggenerativemodeling}, transformer-native generators such as MM-DiT~\citep{esser2024stablediffusion3}, and increasingly unified multimodal systems have transformed generation from passive rendering into a controllable process that can absorb reference images, preserve identity, reason over structure, and participate in downstream decision loops. The central question, therefore, is no longer only how to produce sharper pixels, but how to measure and improve the \emph{intelligence} of visual models: their ability to obey constraints, maintain long-range consistency, close the loop with feedback, and ultimately approximate causal dynamics. This shift motivates a new lens for the field, one that treats visual generation not merely as aesthetic synthesis, but as a stepping stone toward agentic visual intelligence and world simulation.

\paragraph{The ``Nano Banana \& GPT-Image'' Era: New Capabilities and New Challenges.}
Frontier systems exemplified by Nano Banana, GPT-Image, and strong open counterparts such as Qwen-Image and Z-Image~\citep{wu2025qwen,cai2025z} illustrate why this new lens is necessary. On the capability side, modern models exhibit far stronger long-prompt comprehension, typography, instruction following, high-resolution fidelity, reference-based editing, and fluid transitions between generation and manipulation. At the same time, unified understanding-and-generation frameworks such as X-Omni, BLIP3o-NEXT, and UAE indicate that visual models are beginning to operate over a shared multimodal space in which perception, reasoning, and rendering increasingly inform one another~\citep{geng2025xomni,chen2025blip3onext,yan2025uae}. Their training pipelines have also changed qualitatively: success now depends not only on scale, but on active data curation, VLM-driven relabeling, synthetic data distillation, and post-training alignment via DPO, GRPO, and learned reward models~\citep{liu2025flow}. However, these gains coexist with persistent blind spots. Models that excel at photorealism still fail on puzzle-like spatial reconstruction, multi-step state-transition editing, persistent character identity, physical reasoning, and embodied predictive tasks. In other words, current systems often achieve strong \emph{semantic plausibility} without robust \emph{causal competence}. This mismatch explains why conventional metrics such as FID, CLIP-based alignment, or isolated benchmark averages are increasingly insufficient: the field now needs richer task-specific evaluation and in-the-wild stress testing to reveal where present-day models remain confined to statistical correlation rather than genuine visual reasoning.

\begin{figure}[!htbp]
\centering
\includegraphics[width=0.9\textwidth]{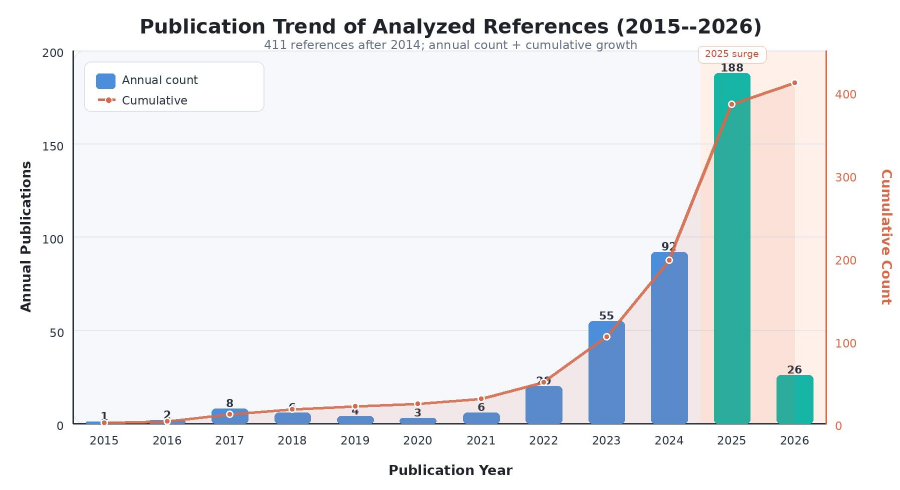}
\caption{\textbf{Publication trend of the 411 post-2014 references analyzed.} Annual publication count (bars) and cumulative total (line) are computed from the updated bibliography after truncating entries before 2015. The curve reveals an exponential acceleration since 2022, with 2025 alone contributing 188 papers (45.7\%). This surge reflects the rapid maturation of visual generation research driven by diffusion transformers, unified multimodal models, and post-training alignment techniques.}
\label{fig:pub_trend}
\end{figure}

\paragraph{Scope and Contributions.}
This work focuses on the rapidly evolving landscape of modern visual generation and editing in the foundation-model era. Rather than cataloging works only by architecture family or application label, we organize the literature around the broader transition from \emph{atomic mapping} to \emph{agentic intelligence}. Our goal is to explain not only \emph{what} has improved, but also \emph{what kind of capability} is being improved as visual systems move from one-shot rendering toward controllable, persistent, interactive, and physically grounded generation. The main contributions of this roadmap are as follows:
\begin{itemize}[leftmargin=*]
\item We propose a capability-oriented five-level taxonomy of visual intelligence, spanning Atomic Generation, Conditional Generation, In-Context Generation, Agentic Generation, and World-Modeling Generation, to clarify what it means for a visual model to become ``more intelligent.''
\item We synthesize the major technical drivers behind this evolution, including the transition from diffusion to flow matching, the rise of unified understanding-and-generation systems, the redesign of visual representations, and the importance of pre-training, post-training alignment, reward modeling, and efficiency engineering.
\item We organize the expanding application space into a coherent progression from conditional generation and domain-specific adaptation to reasoning-driven editing and embodied interaction, highlighting how practical tasks increasingly demand structure, memory, state tracking, and closed-loop control.
\item We complement benchmark-based review with a stress-testing perspective, using real-world case studies to show why high aesthetic quality does not imply mastery of spatial logic, physical consistency, or long-horizon visual reasoning.
\end{itemize}
Together, these perspectives aim to answer the motivating question of this roadmap: what design principles underlie the surprising capabilities of modern ``Nano Banana''- and ``GPT-Image''-style systems, and what still separates them from true playable world models?

\begin{figure}[!htbp]
\centering
\includegraphics[width=0.9\textwidth]{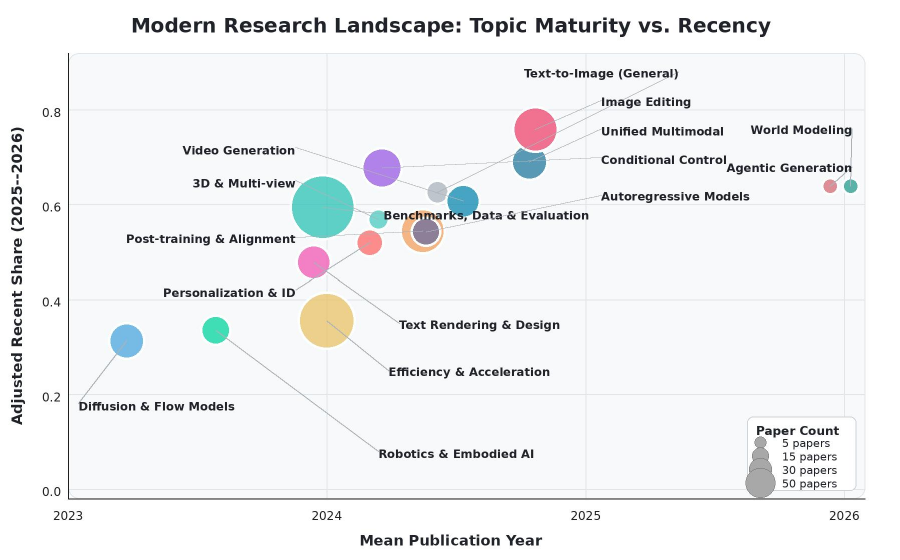}
\caption{\textbf{Modern research landscape: topic maturity vs.\ recency.} Each bubble represents a research category whose mean publication year falls in the modern visual-generation period, positioned by its mean publication year ($x$-axis) and an adjusted 2025--2026 recency share ($y$-axis), with bubble size proportional to paper count. The recency score is smoothed toward the corpus-level average so that very small emerging topics are not artificially pinned to 1.0. Topics in the upper-right quadrant (e.g., image editing, unified multimodal models, world modeling, and agentic generation) are both recent and rapidly growing, while comparatively mature modern foundations such as diffusion \& flow models anchor the lower-left.}
\label{fig:research_landscape}
\end{figure}

To ground the scope and timeliness of this work, we present two complementary analyses of the analyzed references. \Cref{fig:pub_trend} summarizes the post-2014 publication timeline, revealing an exponential acceleration since 2022 with 2025 alone contributing nearly half of the modern cited works. \Cref{fig:research_landscape} maps research categories by maturity and recency, highlighting the field's ongoing pivot from foundational modeling toward application-driven and alignment-oriented research.

\paragraph{Roadmap of the Paper.}
The remainder of this work is organized as follows. \Cref{sec:evolution} introduces our five-level taxonomy of visual intelligence. \Cref{sec:method} reviews model foundations, architectural components, and cross-cutting methodological trends, with particular emphasis on the diffusion-to-flow-matching transition and unified understanding-generation systems. \Cref{sec:train} covers pre-training, post-training, and inference acceleration. \Cref{sec:resource_and_infra} summarizes datasets, data-construction methodologies, benchmarks, and supporting infrastructure. \Cref{sec:applications} covers major application regimes and evolving frontiers, including controllable generation, domain adaptation, editing, and embodied settings. We then move beyond standard benchmarks to analyze real-world stress tests that expose failure modes in spatial structuring, physical reasoning, and identity consistency, before discussing future directions in \Cref{sec:frontier} and concluding in \Cref{sec:conclusion}.
\section{The Evolution of Visual Intelligence}
\label{sec:evolution}

\paragraph{Overview: The Hierarchy of Generative Capabilities.}

As foundation models move visual generation beyond isolated image synthesis, it becomes increasingly important to articulate \emph{what capability is being improved} when we claim a model is ``more intelligent.'' Inspired by OpenAI's staged roadmap for general AI, we propose a domain-specific \textbf{5-Level Taxonomy for Visual Intelligence} that characterizes models by the \emph{type of competence} they exhibit in generation, rather than by architecture or task label.

\textbf{Why a leveled taxonomy?} The five levels reflect a \emph{nested expansion}: each subsumes its predecessors and adds one qualitatively new competence. Two operational axes separate them. \emph{How much a single forward pass can absorb}: L1 takes only a prompt; L2 adds one explicit condition; L3 absorbs rich context within one pass. \emph{Whether the system orchestrates multiple calls}: L4 introduces an external controller that orchestrates multiple passes; L5 further requires that generation be anchored in physical and causal world knowledge. Each level is made precise below and summarized in \Cref{tab:5levels}.

\begin{figure}[!htbp]
    \centering
    \includegraphics[width=\textwidth]{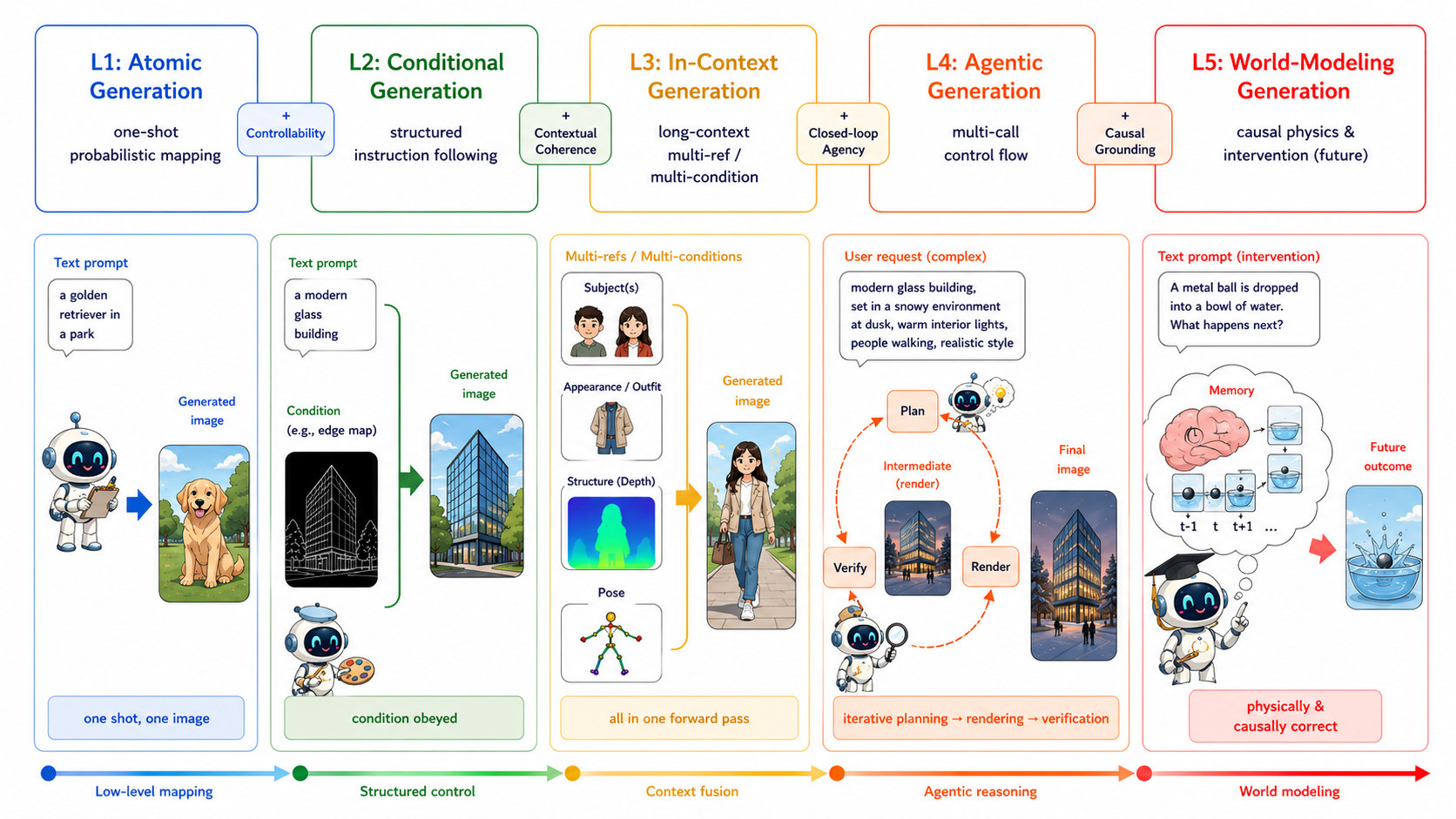}
    \caption{\textbf{Overview of the five-level taxonomy of visual intelligence: progression and concrete examples.} The figure unifies an abstract progression strip (top, with capability axes \textit{Controllability}, \textit{Contextual Coherence}, \textit{Closed-loop Agency}, and \textit{Causal Grounding}) with one concrete input/output example per level (bottom). \textbf{L1}: atomic text-to-image generation. \textbf{L2}: single-condition controlled generation. \textbf{L3}: multi-reference and multi-condition composition in one forward pass. \textbf{L4}: agentic generation with a planner--render--verify loop. \textbf{L5}: world-modeling generation respecting physical and domain rules. The five-level structure is a thread that runs through this survey---see \Cref{tab:5levels} for one-line definitions, the operational distinction between L3 and L4, representative methods, and key challenges.}
    \label{fig:visual_intelligence_overview}
\end{figure}

Progress along this taxonomy is not one-dimensional: it is a transition from \emph{passive statistical rendering} to \emph{goal-directed, physically informed visual intelligence}. This clarifies why many photorealistic systems still fail at persistent identity, long-horizon reasoning, or real-world interaction---they remain at lower capability levels. The taxonomy threads through this work: \Cref{sec:method,sec:train,sec:resource_and_infra,sec:applications} can each be read as mechanisms pushing models along this trajectory; \Cref{sec:stress_test} stress-tests each level with real-world cases; \Cref{sec:frontier} discusses what separates the current frontier from L5 simulation. See \Cref{fig:visual_intelligence_overview,tab:5levels}.

\subsection{Level 1: Atomic Generation}
\label{subsec:level1}

\textbf{Definition.} Models at this level perform a direct, one-shot mapping from an input condition---typically a text prompt---to a visual output. Their role is to approximate the training distribution and produce visually convincing appearances, rather than to maintain precise structure, memory, or causal consistency. Each generation is independent.

\textbf{Representative Methods.} On the diffusion side, DDPM~\citep{ho2020denoising} introduced iterative denoising rivaling GAN samples without adversarial-training instability; Latent Diffusion Models (LDM / Stable Diffusion)~\citep{rombach2022stablediffusion} scaled it via a compressed latent space; DiT~\citep{peebles2023scalable} replaced the U-Net with a transformer that scales reliably with compute. On the autoregressive side, DALL-E~\citep{ramesh2021zero}, LlamaGen~\citep{sun2024autoregressive}, and VAR~\citep{tian2024visual} show that next-token / next-scale prediction also yields strong one-shot images. Earlier GAN systems (e.g., StackGAN~\citep{zhang2017stackgan}) established prompt-to-pixel feasibility but at limited resolution.

\textbf{Key Challenge.} The central limitation of L1 is \emph{uncontrolled variation}: the model generates what ``looks right'' according to the learned distribution, but the user has no mechanism to specify precise spatial layout, object count, identity, or structural constraints. A prompt like ``a red cube to the left of a blue sphere'' may produce a visually pleasing image where the spatial relation is violated, because the model optimizes for distributional plausibility rather than compositional correctness. We stress-test this failure mode in \Cref{sec:stress_test}, Dimension~I, where jigsaw reconstruction tasks reveal that current models default to semantic hallucination rather than geometric reasoning (\Cref{subsubsec:jigsaw_case}).

\subsection{Level 2: Conditional Generation}
\label{subsec:level2}

\textbf{Definition.} This level introduces explicit structural or multimodal constraints---such as depth maps, edge sketches, segmentation layouts, reference images, or identity embeddings---into the generative process. The model moves beyond unconstrained sampling and becomes a \emph{controllable visual constructor} that can faithfully translate prompts, controls, and references into coherent visual results. While the output now respects user-specified conditions, each generation remains a single isolated transaction: the model has no mechanism to carry forward state or memory from one generation to the next.

\textbf{Representative Methods.} ControlNet~\citep{zhang2023controlnet} injects spatial conditions (edges, depth, pose) into a frozen diffusion backbone via a trainable copy. IP-Adapter~\citep{ye2023ip} and InstantID~\citep{wang2024instantid} inject reference-image features through cross-attention for identity control; DreamBooth~\citep{ruiz2023dreambooth} and Textual Inversion~\citep{gal2022image} extend the model's concept vocabulary from a few reference images. Layout-to-image methods like GLIGEN~\citep{li2023gligen} and CreatiLayout~\citep{CreatiLayout} take bounding boxes or region semantics as input. On the architecture side, SD3~\citep{esser2024stablediffusion3} builds L2 conditioning into the backbone via MM-DiT. The shared property: the output is a \emph{constrained} sample satisfying user-specified conditions, not a free sample.

\textbf{Key Challenge.} The central challenge at L2 is \emph{spatial precision and attribute binding}: ensuring that the model places the right content in the right region with the right attributes, rather than merely producing an image globally consistent with the prompt. Even with explicit layout constraints (e.g., ``left box: red hat; right box: blue scarf''), current models frequently exhibit attribute leakage (the scarf inherits the hat's color) or spatial confusion (left-right reversal). We examine these failure modes in detail in \Cref{sec:stress_test}, Dimension~I.

\subsection{Level 3: In-Context Generation}
\label{subsec:level3}

\textbf{Definition.} L3 is still a \emph{single forward pass}, but one that absorbs \emph{rich context}: multiple references, multiple conditions, or an accumulated history of prior generations and edits. The model has no external controller---memory is implicit in the input. This framing reveals why multi-turn editing belongs at L3, not at L4: each round is a single forward pass over a growing input sequence (image$_0$, edit$_1$, image$_1$, edit$_2$, $\dots$), with the model producing image$_k$ in one shot. The challenge is analogous to the difference between answering a single question (L2) and sustaining a coherent multi-turn conversation where earlier statements constrain later ones---still one model, one forward pass, just with more context.

\textbf{Representative Methods.} L3 capability appears in three forms, all sharing the operational signature: one forward pass over rich input. \textit{(i)~Multi-reference / multi-condition composition.} The model fuses several references (subject, outfit, scene) and structural conditions (depth, pose) into one coherent output---the regime that recent unified backbones target most directly. \textit{(ii)~Multi-turn editing with cumulative context.} SEED-Data-Edit~\citep{ge2024seed} and ImgEdit~\citep{ye2025imgedit} let users chain successive edits (``change background,'' then ``add sunglasses,'' then ``shorten hair''); each turn is one forward pass over the accumulated history, requiring pixel-level fidelity in unedited regions---a property favoring VAE-based representations over purely semantic encoders. Cross-panel storytelling extends this: StoryMaker~\citep{zhou2024storymaker} and Visual Persona~\citep{nam2025visual} keep character identity stable across panels via multi-pose supervision and full-body identity representations. \textit{(iii)~Reasoning trace inside one pass.} ReasonGen-R1~\citep{zhang2025reasongen} and T2I-R1~\citep{jiang2025t2i} interleave a textual reasoning trace with image generation inside a single AR sequence---still one forward pass, with no external controller deciding when to stop, placing them at L3.

\textbf{Key Challenge.} The defining difficulty at L3 is \emph{coherence preservation under cumulative context}: the model must update what the user asks (expression, pose, background) while keeping everything else pixel-faithful as the input context grows. Errors accumulate---a small identity drift in one editing step compounds across subsequent steps, and even VAE reconstruction error becomes visible after several round-trips. We examine these failure modes in \Cref{sec:stress_test}, Dimension~IV, and the specific case of cross-frame identity preservation in \Cref{subsubsec:video_re-rendering_case}.

\textbf{Method vs.\ probe.} The L3 classification above is about \emph{method}: a naive multi-turn system stays at L3 because each round is one forward pass with no controller deciding what to verify or roll back. The same task surface, however, is the cleanest \emph{probe} for L4 \emph{capability}: a system that detects mid-sequence identity drift and silently restores, or triggers re-rendering on context conflict, exhibits L4 even though mechanically each round is still one forward pass. \Cref{sec:stress_test}, Dimension~IV uses this dual nature as its probe design.

\subsection{Level 4: Agentic Generation}
\label{subsec:level4}

\textbf{Definition.} At L4, generation becomes one action inside a control loop rather than the final output. Following the LLM literature \citep{anthropic2024agents}, we use \emph{agentic} in the strong sense: a system qualifies as L4 when it (i)~\textbf{dynamically decides} the next action based on observations rather than following a pre-scheduled pipeline, (ii)~operates inside an \textbf{environment feedback loop} where outputs change state and the new state shapes subsequent inputs, (iii)~\textbf{persists toward a long-horizon goal} across multiple forward passes, and (iv)~\textbf{decides when to terminate}. The L3-vs-L4 boundary is therefore about \emph{autonomy over the trajectory}, not the number of user turns.

\textbf{Representative Methods.} Following Anthropic's distinction between \emph{workflows} (LLMs and tools orchestrated through pre-defined code paths) and \emph{agents} (LLMs that dynamically direct their own processes and tool use) \citep{anthropic2024agents}, current L4-related systems span a spectrum.

\textit{Workflow side.} GEMS~\citep{he2026gems} composes a fixed planner$\to$decomposer$\to$verifier$\to$refiner loop with trajectory-level memory; Gen-Searcher~\citep{feng2026gen} extends it with web search and evidence collection. Verification and retrieval become first-class building blocks, but the control flow is engineered, not chosen by the model.

\textit{Agent side.} JarvisArt~\citep{lin2025jarvisart} dynamically selects from 200+ Lightroom tools based on user intent and intermediate state. CoT-VLA~\citep{zhao2025cotvla} and UniPi~\citep{du2023unipi} are the embodied analog: the model predicts visual states conditioned on actions and uses these predictions to choose subsequent actions in a robotic policy loop.

Most current visual L4 systems sit on the workflow end; full agentic systems---where the model chooses its own actions toward a long-horizon goal in a single closed loop---remain rare. Two boundary clarifications: (a)~unified multimodal backbones (Transfusion~\citep{zhou2024transfusion}, BAGEL~\citep{deng2025bagelemerging}, OmniMamba~\citep{zou2025omnimamba}) are \emph{not} L4 agents themselves but provide the shared state space on top of which planning and tool use become easier; (b)~boundary with L5: at L4, generation is a \emph{means} in a policy loop (CoT-VLA predicts visual states for downstream control); at L5, generation itself must encode causal world dynamics.

\textbf{Key Challenge.} The defining challenge at L4 is \emph{grounded verification under intervention}: the system must verify not only the final image but each intermediate \emph{decision}---what to search, what tool to invoke, when to stop---against whether it is the right next action for the current state. The harder a system pushes from workflow- toward agent-side, the more its trajectory depends on its own decision quality. Memory and retrieval (GEMS, Gen-Searcher) extend competence yet expose new failure modes: noisy evidence, brittle verifier signals, trajectories optimizing proxies rather than the goal. Tool-grounded settings (JarvisArt) demand correct executable edits with content preservation and local precision, not just plausible pixels. L4 systems fail not because they cannot render, but because they cannot yet \emph{reliably decide, verify, and correct} inside an open-ended loop. \Cref{sec:stress_test} Dimensions~III and IV surface this through fragile reasoning traces and silent drift in multi-turn editing.

\begin{table}[!htp]
\centering
\caption{\textbf{Comparative taxonomy of visual intelligence.} Our five-level framework organizes progress in visual generation as a nested expansion of capability, drawing parallels to OpenAI's general AI roadmap. Each level subsumes its predecessors while introducing a qualitatively new competence, a distinct set of representative methods, and a defining challenge that separates statistical plausibility from genuine visual intelligence. The \textit{One-Line Definition} column captures the operational signature of each level---most importantly, that L3 remains a single forward pass over rich context, while L4 is the regime where multiple forward passes are stitched by an external control flow.}
\label{tab:5levels}
\resizebox{\textwidth}{!}{%
    \begin{tabular}{@{}c l l l l l l@{}}
        \toprule[1pt]
        \textbf{Level} & \textbf{Visual Paradigm} & \textbf{One-Line Definition} & \textbf{Core Characteristic} & \textbf{Representative Methods} & \textbf{Key Challenge} & \textbf{AI Counterpart} \\ \midrule
        \textbf{L1} & \textbf{Atomic Generation} & \makecell[l]{One forward pass without \\ explicit constraints; sample a \\ plausible image from text alone} & \makecell[l]{Stochastic Plausibility \\ (Distribution Matching)} & \makecell[l]{DDPM, DiT, LDM/SD, \\ LlamaGen, VAR} & \makecell[l]{Uncontrolled variation; \\ no spatial precision} & \makecell[l]{OpenAI L1 \\ (Chatbots)} \\ \midrule
        \textbf{L2} & \textbf{Conditional Generation} & \makecell[l]{One forward pass with a \\ single explicit condition \\ (structure / reference / \\ source-image $+$ instruction)} & \makecell[l]{Compositional \\ Controllability} & \makecell[l]{ControlNet, IP-Adapter, \\ GLIGEN, SD3} & \makecell[l]{Attribute binding \& \\ spatial precision} & \makecell[l]{Transition toward \\ Reasoning} \\ \midrule
        \textbf{L3} & \textbf{In-Context Generation} & \makecell[l]{One forward pass over rich \\ context: multi-reference, \\ multi-condition, accumulated \\ history, or multi-output} & \makecell[l]{Contextual Coherence \\ (Visual Logic \& Memory)} & \makecell[l]{SEED-Data-Edit, ImgEdit, \\ StoryMaker, Visual Persona} & \makecell[l]{Identity preservation \\ under state change} & \makecell[l]{OpenAI L2 \\ (Reasoners)} \\ \midrule
        \textbf{L4} & \textbf{Agentic Generation} & \makecell[l]{Multiple forward passes \\ with control flow: plan, \\ verify, invoke tools, refine} & \makecell[l]{Closed-loop Agency \\ (Perception-Action-Feedback)} & \makecell[l]{GEMS, Gen-Searcher, \\ JarvisArt, CoT-VLA} & \makecell[l]{Grounded verification \\ \& self-correction} & \makecell[l]{OpenAI L3 \\ (Agents)} \\ \midrule
        \textbf{L5} & \makecell[l]{\textbf{World-Modeling} \\ \textbf{Generation} \textit{(future)}} & \makecell[l]{Generation anchored by an \\ internalized world model; \\ discriminative tasks and \\ domain-structured generation \\ emerge as consequences} & \makecell[l]{Causal Simulation \\ (Physics \& Intervention)} & \makecell[l]{Genie 2, GameNGen, \\ Oasis, UniSim, GAIA-1} & \makecell[l]{Causal faithfulness; \\ physical plausibility} & \makecell[l]{OpenAI L4 \\ (Innovators)} \\
        \bottomrule[1pt]
    \end{tabular}%
    }
\end{table}

\subsection{Level 5: World-Modeling Generation \textit{(future)}}
\label{subsec:level5}

\textbf{Definition.} L5 is reached when a model internalizes stable representations of dynamics, physics, and intervention effects---a \emph{world simulator}, not just an appearance generator. Where L4 agents reason from learned correlations, L5 demands prediction of what \emph{would actually happen} under a specific action in a specific physical context. The most fully realized L5 demonstrations today appear in interactive video and embodied settings; we treat them as case studies of L5 \emph{capability}, not as methodological focus.

\textbf{Representative Methods.} \emph{Neural game engines} provide the most vivid demonstrations: Genie~2~\citep{parker2024genie2} generates 3D environments learned from passive video, with persistent object behavior; GameNGen~\citep{valevski2024gamenngen} replaces DOOM's hand-coded engine with a diffusion model rendering frames conditioned on player actions at interactive frame rates; Oasis~\citep{oasis2024} scales this to Minecraft. In embodied settings, UniSim~\citep{yang2024unisim} is a universal visual simulator for manipulation and navigation; GAIA-1~\citep{hu2024gaia1} generates controllable driving scenarios conditioned on vehicle actions. Image-side, L5 also encompasses \emph{world-knowledge grounding}: rendering scenarios that respect gravity, fluid dynamics, material properties, or domain rules---probed by benchmarks like PhyBench~\citep{meng2024phybench} and WISE~\citep{niu2025wise}.

\textbf{Key Challenge.} The decisive challenge is \emph{causal faithfulness}: whether the generated future state reflects the causal consequences of an intervention, not merely its statistical correlates. A photorealistic rollout that ignores the commanded steering or misrepresents contact dynamics is worse than useless for downstream policy learning---it corrupts the training signal. Current models learn correlations without internalizing physical laws and fail under unusual mechanisms or extreme maneuvers. Evaluation itself is a fundamental barrier at L5: judging whether a scene is \emph{physically correct} and \emph{causally consistent} requires benchmarks beyond perceptual quality---verifying how a scene \emph{works}, not what it \emph{looks like}. We examine this gap through fluid-dynamics counterfactuals (\Cref{subsubsec:fluid_dynamics_case}), action-conditioned driving (\Cref{subsubsec:driving_collision_case}), robotic manipulation (\Cref{subsubsec:robot_grasp_case}), and trajectory synthesis (\Cref{subsubsec:trajectory_synthesis}).

\section{Model, Architecture, and Method}\label{sec:method}

This section answers three questions about how a frontier visual generation system is built. \Cref{sec:model_arch} asks how generation is mathematically modeled, tracing the historical progression from GANs through diffusion, flow matching, autoregressive, and hybrid AR$+$diffusion paradigms as a sequence of moves where each generation removes a key bottleneck of the previous one. \Cref{sec:arch_components} asks what components make up a system and how they cooperate, decomposing every modern model into an encoder/tokenizer, a backbone, a condition module, and a multimodal fusion module---a decomposition that increasingly turns generation and editing into two data flows through the same architecture. \Cref{sec:closed_source_frontier} asks what can be inferred when the strongest systems disclose only behavior rather than architecture: it turns the convergence and bifurcation observed in open reports into a future-facing discussion of closed-source agent loops, stronger upstream VLMs, and the system-level gaps that open models still need to close. The reader can treat this section as an entry point into the internals of today's frontier open- and closed-source systems, anchored to the L1--L5 capability levels of \Cref{sec:evolution}.

\subsection{Foundational Generative Paradigms}
\label{sec:model_arch}

\begin{figure*}[!htbp]
    \centering
    \includegraphics[width=\textwidth]{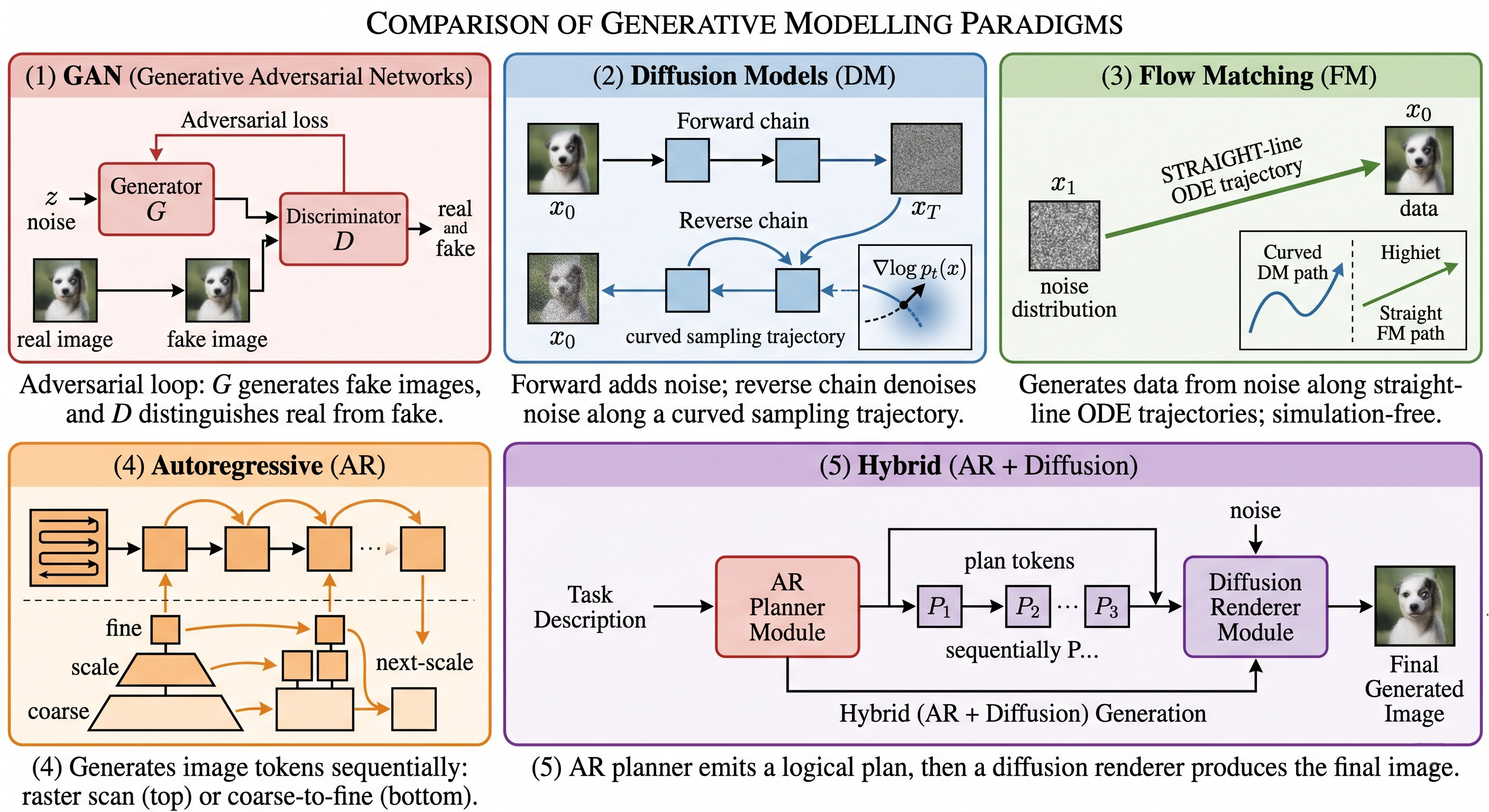}
    \caption{\textbf{Five generative-modeling paradigms at a glance.} \textbf{(1) GAN}: a generator $G$ maps noise $z$ to a fake image while a discriminator $D$ separates real from fake under an adversarial loss. \textbf{(2) Diffusion Model (DM)}: a forward chain $x_0 \to x_T$ adds noise; a learned reverse chain denoises along a curved sampling trajectory. \textbf{(3) Flow Matching (FM)}: noise is transported to data along a straight-line ODE trajectory, simulation-free to train and integrable in few steps. \textbf{(4) Autoregressive (AR)}: an image is discretized into tokens predicted next-token (raster) or next-scale (coarse-to-fine pyramid). \textbf{(5) Hybrid (AR + Diffusion)}: an AR planner emits plan tokens that condition a diffusion renderer producing the final image.}
    \label{fig:paradigm_overview}
\end{figure*}

The history of generative modeling for images can be read as a sequence of paradigms, each one defined by the limitation of the previous generation it sets out to remove. \emph{Generative adversarial networks} (GANs)~\citep{goodfellow2014generativeadversarialnetworks} first demonstrated that a neural network could turn random noise directly into perceptually convincing images, but their adversarial training was notoriously unstable, prone to mode collapse, and difficult to scale beyond curated single-domain datasets. \emph{Diffusion models}~\citep{ho2020denoising,song2021scorebasedgenerativemodelingstochastic} answered exactly that limitation: by replacing the adversarial game with iterative denoising under a tractable likelihood objective, they made training stable and easy to scale to web-scale data; once paired with cross-attention conditioning as in LDM~\citep{rombach2022stablediffusion}, the same recipe also became amenable to text conditioning---at the cost of hundreds of sequential network evaluations per sample. \emph{Flow matching} and rectified flow~\citep{liu2022flow,lipman2023flowmatchinggenerativemodeling} then attacked precisely that sampling bottleneck by learning a straight-line transport between noise and data, which is simulation-free to train and integrable in only a handful of steps. In parallel, the rise of large language models motivated an \emph{autoregressive} reading of vision: LlamaGen~\citep{sun2024autoregressive} and VAR~\citep{tian2024visual} discretize images into tokens and predict them next-token or next-scale, while Chameleon~\citep{chameleon2024} and Emu3~\citep{wang2024emu3} push the same protocol to a fully unified multimodal token stream---trading some per-pixel fidelity for a single architecture that natively reasons across text and image. A prominent frontier direction is now \emph{hybridization}~\citep{chen2025blip3onext,cao2025hunyuanimage}: an autoregressive model decides \emph{what} to generate at the level of plan tokens, while a diffusion or flow model decides \emph{how} to render the corresponding pixels, combining AR's instruction-following with diffusion's per-pixel fidelity. \Cref{fig:paradigm_overview} summarizes the five paradigms side by side, allowing the reader to compare their core mechanisms before we examine each in detail.

\begin{figure*}[!htbp]
    \centering
    \includegraphics[width=\textwidth]{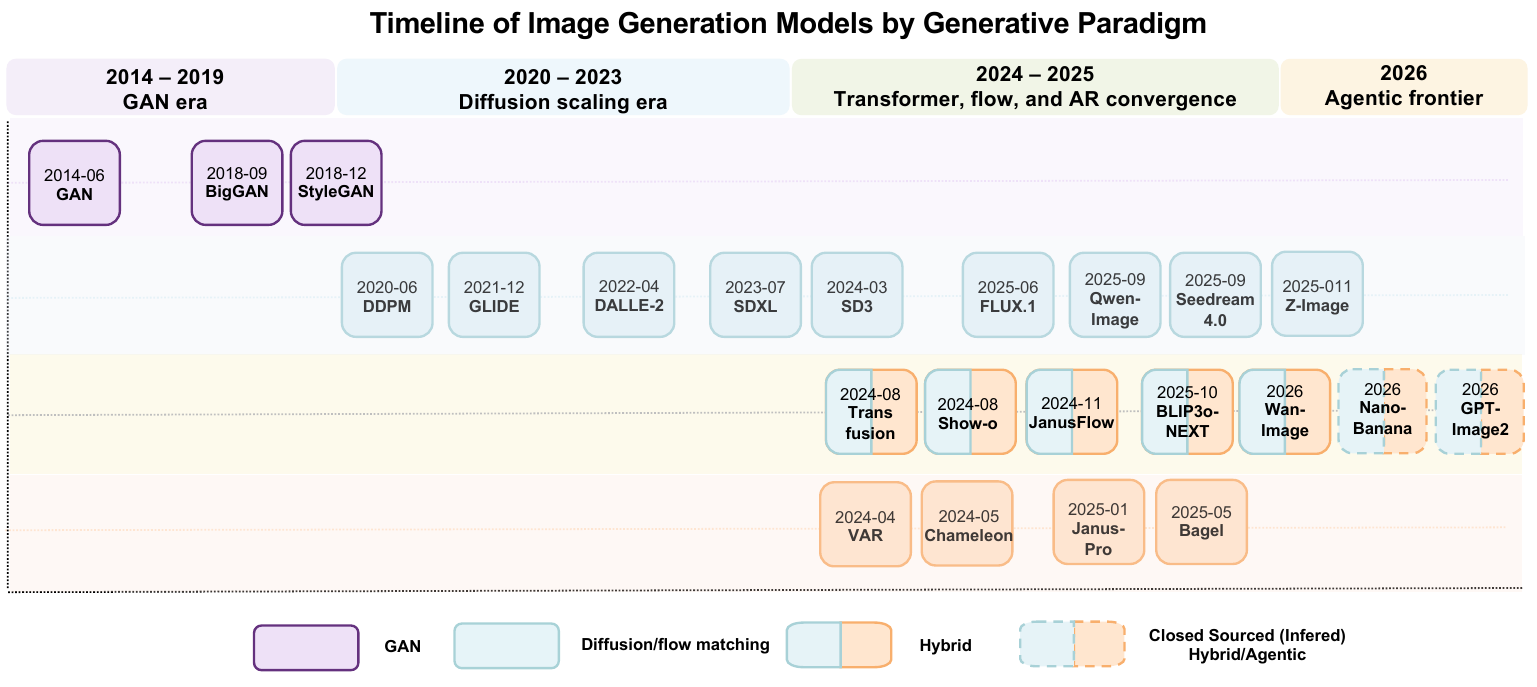}
    \caption{\textbf{Timeline of representative image generation models by generative paradigm.} Nodes are positioned by first public release month when available and grouped into GAN, diffusion / flow-matching, autoregressive, and hybrid categories. Hybrid nodes use both autoregressive and diffusion / flow colours to emphasize that recent frontier systems increasingly combine semantic planning with high-fidelity rendering.}
    \label{fig:model_timeline}
\end{figure*}

\subsubsection{GAN}

\paragraph{Overview.}
Generative Adversarial Networks (GANs)~\citep{goodfellow2014generativeadversarialnetworks} synthesize images by training a generator $G$ and a discriminator $D$ in a minimax game; while no longer the frontier paradigm, they remain industrially relevant for one-step inference and fine-grained latent control.

\paragraph{Mathematical Formulation.}
The original GAN objective is the minimax problem
\begin{align}
\min_G \max_D \; V(D,G)
&= \mathbb{E}_{x \sim p_{\text{data}}}[\log D(x)]
   + \mathbb{E}_{z \sim p_z}[\log(1 - D(G(z)))] ,
\label{eq:gan}
\end{align}
which reaches equilibrium when $p_G = p_{\text{data}}$. In practice, the generator update is replaced by the non-saturating loss $\min_G \mathbb{E}_{z}[-\log D(G(z))]$ to avoid vanishing gradients early in training, and the WGAN family~\citep{arjovsky2017wassersteingan} swaps the Jensen--Shannon divergence for the Earth-Mover distance under a Lipschitz constraint to further stabilize the dynamics.

\paragraph{Representative Work.}
Subsequent research advanced GANs along four directions, each addressing a distinct limitation of the original minimax recipe. \emph{Training stability} is improved by Wasserstein and Lipschitz-controlled objectives (WGAN-GP~\citep{gulrajani2017improvedtrainingwassersteingans}, LSGAN~\citep{mao2017squaresgenerativeadversarialnetworks}, SN-GAN~\citep{miyato2018spectralnormalizationgenerativeadversarial}). \emph{Architectural scaling} progresses from convolutional DCGAN~\citep{radford2016unsupervisedrepresentationlearningdeep} through ProGAN~\citep{salimans2022progressive} and BigGAN~\citep{brock2019largescalegantraining} to the style-based StyleGAN family~\citep{karras2019stylebasedgeneratorarchitecturegenerative}, which introduced disentangled fine-grained control. \emph{Conditional control} extends GANs to label-, text-, or image-driven generation (Conditional/InfoGAN~\citep{chen2016infoganinterpretablerepresentationlearning}, CycleGAN~\citep{zhu2020unpairedimagetoimagetranslationusing}, Pix2Pix~\citep{isola2017image}). \emph{Fast inference} keeps GANs industrially relevant in the diffusion era through transformer-scale designs (GigaGAN~\citep{kang2023gigagan}), modernized adversarial objectives such as R3GAN~\citep{huang2025r3gan}, or by fusing adversarial losses with diffusion training to produce one-step samplers.

\paragraph{Summary.}
GANs have receded from the prompt-following frontier but remain the recipe of choice when one-shot synthesis speed and fine-grained latent control matter more than the broader instruction fidelity that diffusion-based systems now lead.

\subsubsection{Diffusion and Flow Matching}

\paragraph{Mathematical Formulation.}
Both diffusion models and flow matching belong to the broader family of continuous-time generative models. They share a unifying principle: learning a deterministic or stochastic transformation between a simple tractable distribution $p_0$ (e.g., Gaussian noise) and the complex data distribution $p_1$.

Diffusion models originally framed this as reversing a gradual noising process. A forward process corrupts data $x_1$ into noise $x_0$ over time, and the model learns a denoising network $\epsilon_\theta(x_t, t)$ to reverse it:
\begin{equation}
\min_\theta \mathbb{E}_{x_1, \epsilon, t} \left[ \| \epsilon - \epsilon_\theta(x_t, t) \|^2 \right].
\end{equation}
Crucially, as established by score-based SDEs~\citep{song2021scorebasedgenerativemodelingstochastic}, this reverse process can be described by a \emph{Probability Flow ODE}. This revealed that diffusion models are implicitly learning a vector field that defines the trajectory from noise to data.

Flow Matching (FM) and Rectified Flow directly generalize this ODE perspective. Instead of relying on a predefined forward noising process, FM constructs a vector field that transports samples along explicit, often straighter, probability paths. Given $x_0 \sim p_0$ and $x_1 \sim p_1$, a simple linear interpolation is defined as:
\begin{equation}
x_t = (1 - t)\, x_0 + t\, x_1, \qquad t \sim \mathcal{U}[0,1].
\end{equation}
The target velocity along this path is straightforwardly $v_{\text{target}}(x_t, t) = x_1 - x_0$. The model directly learns this velocity field $v_\theta(x_t,t)$ via a simulation-free regression objective:
\begin{equation}
\mathcal{L}_{\text{FM}}(\theta) = \mathbb{E}_{x_0, x_1, t} \left[ \bigl\| v_\theta(x_t, t) - (x_1 - x_0) \bigr\|^2 \right].
\end{equation}
During inference, both paradigms generate data by solving their respective ordinary differential equations starting from $x_0 \sim p_0$.

\paragraph{Representative Work.}
DDPM~\citep{ho2020denoising} re-established likelihood-based image generation by learning a denoiser that inverts a fixed forward noising process, but its strong sample quality came with a major practical drawback: high-fidelity synthesis required hundreds to thousands of reverse steps. DDIM~\citep{songDDIM} showed that this sampling cost is separable from the training objective. By constructing a non-Markovian deterministic reverse trajectory, it retained DDPM training while reducing inference cost by an order of magnitude, making explicit that sampler design is an independent axis of innovation.

Score-based SDE models~\citep{song2021scorebasedgenerativemodelingstochastic} pushed this insight further by recasting diffusion as a continuous-time stochastic process with an equivalent probability-flow ODE. This continuous-time view unified discrete diffusion, ODE solvers, and likelihood computation under a single framework, turning sampling into a numerical integration problem with an explicit accuracy--compute trade-off. Rectified Flow~\citep{liu2022flow} and Flow Matching~\citep{lipman2023flowmatchinggenerativemodeling} then shifted the focus from solving increasingly curved reverse trajectories to learning straighter transport paths directly. In particular, Flow Matching removed the need for expensive ODE simulation during training and made ODE-based generators practical at modern scale.

Modern systems increasingly pair these objectives with transformer backbones. SD3~\citep{esser2024stablediffusion3} shows that rectified-flow objectives compose naturally with the transformer scaling recipe, yielding strong prompt fidelity and typography when combined with MM-DiT (the modality-aware DiT variant; we discuss its architecture in \Cref{sec:arch_backbone}). Diff2Flow~\citep{schusterbauer2025diff2flowtrainingflowmatching} further narrows the gap between diffusion and flow by converting pretrained diffusion checkpoints into flow-matching generators through schedule alignment. Taken together, the field has evolved from discrete denoising chains to a broader family of continuous transport models whose main practical question is no longer whether diffusion or flow is better in principle, but how to co-design objective, sampler, and backbone for low-NFE high-fidelity generation.

\subsubsection{Autoregressive Models}

\paragraph{Mathematical Formulation.}
Autoregressive (AR) generative models produce sequences by predicting each token conditioned on all previously generated tokens. Given a sequence $x_{1:T} = (x_1, \ldots, x_T)$, an AR model factorizes the joint distribution as
\begin{equation}
p_\theta(x_{1:T}) = \prod_{t=1}^T p_\theta(x_t \mid x_{<t}),
\end{equation}
where $x_{<t} = (x_1, \ldots, x_{t-1})$ and $\theta$ denotes model parameters. In practice, this factorization is commonly implemented using a decoder-only Transformer trained to maximize the log-likelihood of observed sequences. During inference, generation proceeds sequentially by sampling from $p_\theta(x_t \mid x_{<t})$ until a termination condition, such as an end-of-sequence token, is reached.

\paragraph{Representative Work.}
Recent progress in autoregressive visual generation is better illustrated by image and multimodal systems than by text-only language models. LlamaGen~\citep{sun2024autoregressive} shows that a vanilla decoder-only transformer trained with next-token prediction on discrete image tokens can match diffusion baselines when paired with an effective tokenizer, sufficient scale, and strong data. VAR~\citep{tian2024visual} makes the AR recipe more efficient by replacing raster-scan prediction with next-scale prediction in a coarse-to-fine hierarchy, reducing the effective sequence length and becoming the first AR model to surpass DiT in both FID and generation speed.

A second wave extends the same next-token objective from pure image synthesis to unified multimodal modeling. Chameleon~\citep{chameleon2024} and Emu3~\citep{wang2024emu3} treat images and text as a single mixed-modal token stream, showing that early-fusion autoregression can support both understanding and generation without a dedicated diffusion decoder. Janus-Pro~\citep{chen2025janusprounifiedmultimodalunderstanding}, BAGEL~\citep{deng2025bagelemerging}, and OmniMamba~\citep{zou2025omnimamba} scale this idea further through better training data, larger unified token streams, and more efficient sequence backbones. The resulting picture is that AR models are most compelling when the goal is not only image synthesis, but also sharing a common token protocol with language and multimodal reasoning. We discuss the architectural mechanisms that enable this multimodal unification---token-stream design, encoder decoupling, and fusion strategies---in \Cref{sec:arch_fusion}.

\subsubsection{Hybrid AR + Diffusion/Flow Models}
\label{sec:hybrid_ar_diff}

\paragraph{Formulation.}
Hybrid systems explicitly separate semantic planning from visual rendering. A common factorization introduces an intermediate plan $z$:
\begin{equation}
p_{\theta}(x \mid c) = \int p_{\phi}(z \mid c)\, p_{\psi}(x \mid z, c)\, dz,
\end{equation}
where an autoregressive model predicts the plan $z$---for example, a mixed-modal token sequence, semantic layout, or continuous latent prefix---and a diffusion or flow model renders high-fidelity visual details conditioned on that plan. The motivation is pragmatic: AR models naturally support causal reasoning, long-context instruction following, and tight integration with language, while diffusion/flow models remain stronger at spatially coherent, high-fidelity synthesis.

\begin{figure*}[!htbp]
    \centering
    \includegraphics[width=0.9\textwidth]{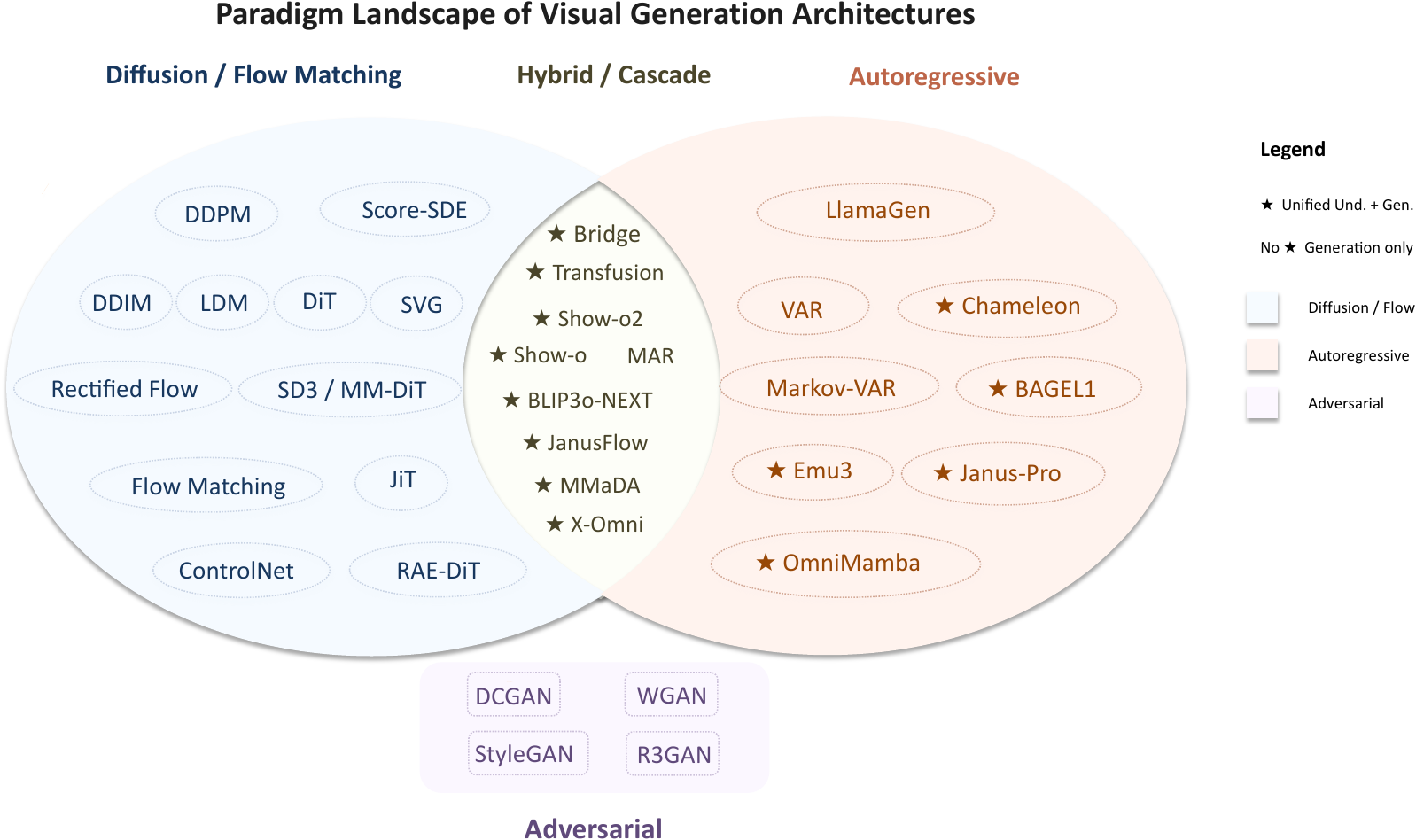}
    \caption{Paradigm landscape of representative visual generation architectures. Methods are positioned by their generative mechanism: diffusion/flow matching (\textit{left}), autoregressive (\textit{right}), and adversarial (\textit{bottom}). The central overlap hosts hybrid architectures that combine diffusion and autoregressive components. $\bigstar$ denotes unified models that jointly support visual understanding and generation.}
    \label{fig:paradigm_landscape}
\end{figure*}

\paragraph{Representative Work.}
Transfusion~\citep{zhou2024transfusion} is a canonical early-fusion design: it interleaves discrete text tokens and continuous image latents within one transformer, using cross-entropy for text and diffusion loss for image patches. MonoFormer~\citep{zhao2024monoformer} pushes the shared-backbone idea further by using a pretrained LLM-style transformer for both AR text generation and latent diffusion image generation, switching only the attention mask and prediction head: causal attention with an AR head for text, and bidirectional attention with a diffusion head for noised VAE latents.
Show-o~\citep{xie2024showo} reaches a similar goal through attention-mask switching, unifying AR text generation and discrete diffusion image generation in one backbone; Show-o2~\citep{xie2025showo2} updates the image branch to flow matching in a 3D causal VAE space. JanusFlow~\citep{ma2024janusflow} combines an AR LLM with rectified flow through decoupled visual encoders and representation alignment, while BLIP3o-NEXT~\citep{chen2025blip3onext} adopts an explicit cascade in which an AR stage plans semantics and a diffusion stage renders pixels from AR hidden states. MAR~\citep{li2024mar} and NextStep-1~\citep{team2025nextstep1} make the coupling even tighter by retaining an AR backbone but replacing the categorical token head with token-wise diffusion over continuous latents.

Across these methods, the shared conclusion is not that AR should replace diffusion, or vice versa, but that the two families solve different subproblems. Hybridization is therefore a central design pattern, not a niche exception---and the same ``AR plans, diffusion renders'' division reappears at the system level, prefiguring the L4 agentic-generation regime of \Cref{sec:evolution} and the closed-source agent loop discussed in \Cref{sec:closed_source_frontier}.

\begin{table*}[!htp]
\centering
\caption{\textbf{Component-level morphology of representative visual generation architectures.} A \cmark{} marks a defining design choice; \xmark{} indicates the component is not used. The table is intentionally schematic: a checkmark denotes the \emph{primary} component of each method, not the only possible configuration. AR = Autoregressive; Rect.\ Flow = Rectified Flow. This table presents methods through a \emph{modeling-paradigm} lens (objective $+$ backbone $+$ representation $+$ conditioning); the complementary \Cref{tab:unified_methods} re-examines a partly overlapping set of methods through a \emph{multimodal-fusion} lens, so paper overlap reflects two views of the same systems rather than redundancy.}
\label{tab:arch_morphology}
\scriptsize
\setlength{\tabcolsep}{3pt}
\renewcommand{\arraystretch}{1.20}
\resizebox{\textwidth}{!}{%
    \begin{tabular}{@{}l l ccc cccc ccccc@{}}
        \toprule[1pt]
        \multirow{2}{*}[-8ex]{\textbf{Method}} & \multirow{2}{*}[-8ex]{\textbf{Objective}} &
        \multicolumn{3}{c}{\textbf{Backbone}} &
        \multicolumn{4}{c}{\textbf{Representation}} &
        \multicolumn{5}{c}{\textbf{Conditioning}} \\
        \cmidrule(lr){3-5} \cmidrule(lr){6-9} \cmidrule(lr){10-14}
        & &
        \rotatebox{70}{Conv/U-Net} & \rotatebox{70}{Transformer} & \rotatebox{70}{SSM} &
        \rotatebox{70}{Pixel} & \rotatebox{70}{VAE latent} & \rotatebox{70}{SSL/RAE latent} & \rotatebox{70}{Discrete VQ} &
        \rotatebox{70}{Cross-attn} & \rotatebox{70}{AdaLN / Embed.} & \rotatebox{70}{Dual-stream} & \rotatebox{70}{Unified token} & \rotatebox{70}{AR$\rightarrow$Diff} \\
        \midrule
        LDM~\citep{rombach2022stablediffusion}           & Diffusion    & \cmark & \xmark & \xmark & \xmark & \cmark & \xmark & \xmark & \cmark & \xmark & \xmark & \xmark & \xmark \\
        DiT~\citep{peebles2023scalable}                   & Diffusion    & \xmark & \cmark & \xmark & \xmark & \cmark & \xmark & \xmark & \xmark & \cmark & \xmark & \xmark & \xmark \\
        SD3 / MM-DiT~\citep{esser2024stablediffusion3}             & Rect.\ Flow  & \xmark & \cmark & \xmark & \xmark & \cmark & \xmark & \xmark & \xmark & \cmark & \cmark & \xmark & \xmark \\
        RAE-DiT~\citep{zheng2025rae}                      & Diffusion    & \xmark & \cmark & \xmark & \xmark & \xmark & \cmark & \xmark & \xmark & \cmark & \xmark & \xmark & \xmark \\
        SVG~\citep{shi2025svg}                            & Diffusion    & \xmark & \cmark & \xmark & \xmark & \xmark & \cmark & \xmark & \xmark & \cmark & \xmark & \xmark & \xmark \\
        JiT~\citep{li2025back}                            & Diffusion    & \xmark & \cmark & \xmark & \cmark & \xmark & \xmark & \xmark & \xmark & \cmark & \xmark & \xmark & \xmark \\
        R3GAN~\citep{huang2025r3gan}                      & Adversarial  & \xmark & \cmark & \xmark & \xmark & \cmark & \xmark & \xmark & \xmark & \xmark & \xmark & \xmark & \xmark \\
        LlamaGen~\citep{sun2024autoregressive}            & AR           & \xmark & \cmark & \xmark & \xmark & \xmark & \xmark & \cmark & \xmark & \xmark & \xmark & \xmark & \xmark \\
        VAR~\citep{tian2024visual}                        & AR           & \xmark & \cmark & \xmark & \xmark & \xmark & \xmark & \cmark & \xmark & \xmark & \xmark & \xmark & \xmark \\
        Janus-Pro~\citep{chen2025janusprounifiedmultimodalunderstanding} & AR & \xmark & \cmark & \xmark & \xmark & \xmark & \xmark & \cmark & \xmark & \xmark & \xmark & \cmark & \xmark \\
        Bridge~\citep{wang2025bridge}                     & AR           & \xmark & \cmark & \xmark & \xmark & \xmark & \xmark & \cmark & \xmark & \xmark & \xmark & \cmark & \xmark \\
        OmniMamba~\citep{zou2025omnimamba}                & AR           & \xmark & \xmark & \cmark & \xmark & \xmark & \xmark & \cmark & \xmark & \xmark & \xmark & \cmark & \xmark \\
        Transfusion~\citep{zhou2024transfusion}           & Hybrid       & \xmark & \cmark & \xmark & \xmark & \cmark & \xmark & \xmark & \xmark & \xmark & \xmark & \cmark & \xmark \\
        Monoformer~\citep{zhao2024monoformer} & Hybrid & \xmark & \cmark & \xmark & \xmark & \cmark & \xmark & \xmark & \xmark & \xmark & \cmark & \cmark & \xmark \\
        
        BLIP3o-NEXT~\citep{chen2025blip3onext}            & Hybrid       & \xmark & \cmark & \xmark & \xmark & \xmark & \xmark & \cmark & \xmark & \cmark & \xmark & \cmark & \cmark \\
        \bottomrule[1pt]
    \end{tabular}%
}
\end{table*}

\begin{figure*}[!htbp]
    \centering
    \includegraphics[width=0.9\textwidth]{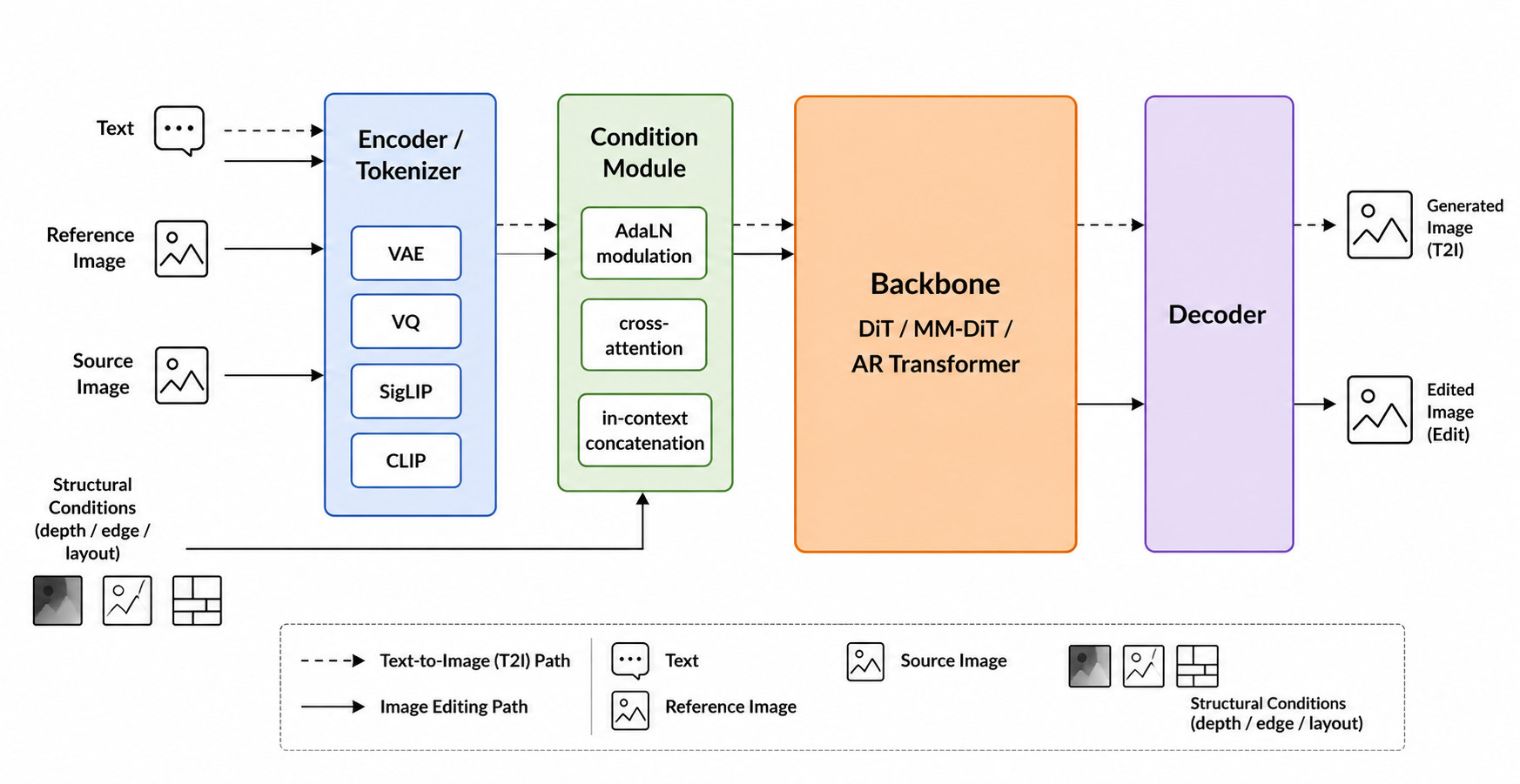}
    \caption{\textbf{The unified architecture for visual generation and editing.} A modern visual generation / editing pipeline contains an encoder / tokenizer, condition routing, a generative backbone, and an output decoder arranged left-to-right. \textbf{(1) Encoder / Tokenizer} accepts heterogeneous inputs (text prompt, optional reference image, optional source image, optional structural conditions such as depth / edge / layout) via VAE, VQ, SigLIP, or CLIP. \textbf{(2) Condition Module} routes the encoded condition into the backbone through AdaLN modulation, cross-attention, or in-context concatenation. \textbf{(3) Backbone} (DiT, MM-DiT, or AR-transformer) does the heavy compute. \textbf{(4) Decoder} (VAE / VQ / pixel) produces the final image. The decoder is shown as the output endpoint, but its design is tied to the representation choice; the four axes analyzed below are representation, backbone, condition routing, and multimodal fusion. The same pipeline supports both \emph{T2I} (text $\to$ generated image, blue path) and \emph{editing} (text $+$ source image $\to$ edited image, red path)---only the inputs entering the encoder and condition module differ.}
    \label{fig:unified_arch}
\end{figure*}
\subsection{Architecture Components}
\label{sec:arch_components}

Beyond cataloging components, this subsection delivers a single message: \emph{generation and editing have collapsed into the same architecture}. As recently as 2024, image editing was almost always a separate paper with its own backbone fork; within roughly a year, the clear majority of frontier open- and closed-source systems present T2I and editing as two data flows through one model, where the task is determined entirely by what enters the encoder and condition module rather than by a different network (\Cref{fig:unified_arch}). Three camps achieve this unification differently. The \emph{DiT camp}~\citep{seedream2025seedream,wu2025qwen,cai2025z,jdjoyaiimage,team2025longcat,mao2026wan} injects reference images as VAE-latent tokens disambiguated by 3D-RoPE and a different timestep, then trusts MM-DiT to absorb both flows. The \emph{AR camp}~\citep{cao2025hunyuanimage,team2026longcat} treats every image as a token sequence so that T2I and editing become different contexts to the same next-token predictor. The \emph{hybrid camp}~\citep{chen2025blip3onext} factorizes into an MLLM planner and a DiT visualizer that share an attention backbone. Each camp resolves a different open trade-off: DiT systems inherit a mature distillation toolchain that dominates the few-step inference frontier, while discrete-AR systems argue that their finite-MDP structure makes reinforcement learning more natural. The active question in this corner of the field is no longer \emph{whether} to unify, but \emph{which mechanism} delivers the better Pareto frontier.

\begin{highlightbox}{Community Message: Edit Unification in One Year}

A year ago Seedream~3.0~\citep{gao2025seedream} (early 2025) treated image editing as a separate paper from text-to-image; today, by Wan-Image~\citep{mao2026wan} and LongCat-Next~\citep{team2026longcat} (spring 2026), \emph{60\% of recent frontier reports ship a fully unified architecture}\footnote{The cohort consists of ten public 2025--2026 image-generation tech reports we read side by side: Seedream~3.0~\citep{gao2025seedream}, Seedream~4.0~\citep{seedream2025seedream}, HunyuanImage~3.0~\citep{cao2025hunyuanimage}, Qwen-Image~\citep{wu2025qwen}, Z-Image~\citep{cai2025z}, LongCat-Image~\citep{team2025longcat}, FireRed-Image-Edit-1.0~\citep{team2026firered}, JoyAI-Image~\citep{jdjoyaiimage}, LongCat-Next~\citep{team2026longcat}, and Wan-Image~\citep{mao2026wan}.} where T2I, I2I, and instruction-based editing share the same backbone. The debate has shifted from \emph{whether} to unify to \emph{which mechanism} unifies best: a DiT-style triplet (VAE + RoPE + timestep injection of the reference) versus an AR-style ``image-as-token, task decided by context'' framing, with hybrid bets like Wan-Image splitting the difference. The largest open trade-off is \textbf{distillation-friendliness vs.\ RL-friendliness}: AR-style unification is easier to RL but harder to compress to few steps; DiT-style unification is the reverse. This single trade-off shapes which design will dominate the 2026 cohort, and is currently the most consequential unsolved question in image-generation architecture.
\end{highlightbox}

The four axes below are largely orthogonal (\Cref{tab:arch_morphology}): the same backbone can sit on top of VAE latents or discrete tokens, and the same condition module serves both diffusion and autoregressive backbones, which is what most new methods exploit by innovating on one or two axes while inheriting established designs on the others. The output decoder is still part of the pipeline, but because its VAE / VQ / pixel choice is tied to representation, we discuss it under encoder/tokenizer rather than as a standalone axis.

\begin{enumerate}[leftmargin=*,itemsep=2pt]
    \item \textbf{Encoder / Tokenizer} (\S\ref{sec:arch_encoder}): the entry point of the pipeline. It maps raw pixels into a compact representation space---continuous latents or discrete tokens---where generation becomes computationally tractable.
    \item \textbf{Backbone network} (\S\ref{sec:arch_backbone}): the central compute unit that operates in the encoded space, parameterizing the denoising step (diffusion), velocity prediction (flow matching), next-token generation (autoregressive), or a staged combination of these primitives (hybrid models).
    \item \textbf{Condition module} (\S\ref{sec:arch_condition}): translates external control signals (text prompts, spatial maps, reference images) into features that steer the backbone's output toward user intent.
    \item \textbf{Multimodal fusion module} (\S\ref{sec:arch_fusion}): determines \emph{where} and \emph{how} information from different modalities interacts within the backbone, from simple cross-attention injection to fully unified token streams.
\end{enumerate}

\subsubsection{Encoder, Tokenizer, and Representation Space}
\label{sec:arch_encoder}

Generating images directly in pixel space is computationally prohibitive: a $512 \times 512$ RGB image has $786{,}432$ dimensions. The encoder (or tokenizer) is the \emph{entry point} of the generation pipeline---it maps raw pixels into a compact representation space where the backbone can operate efficiently. The encoder thus controls a fundamental trade-off: aggressive compression speeds up generation but may lose fine-grained visual details; conservative compression preserves fidelity but limits scalability. The choice of representation is therefore not merely a preprocessing detail: it determines which objectives are natural (reconstruction, next-token prediction, or noise/velocity prediction), how efficiently the backbone scales, and whether understanding and generation can share a common token protocol.

\paragraph{Continuous Latents: VAE and Beyond.}
The dominant approach for diffusion-based models is the \emph{variational autoencoder} (VAE). LDM~\citep{rombach2022stablediffusion} trains a KL-regularized autoencoder that compresses images by $8\times$ spatially, achieving a near-optimal balance between complexity reduction and perceptual fidelity. Generation then occurs entirely in this latent space, reducing the backbone's computational burden by $64\times$.

A recent trend moves beyond trainable encoders toward \emph{representation-first} designs that leverage frozen pretrained models. RAE~\citep{zheng2025rae} replaces the VAE encoder with frozen discriminative features (e.g., DINOv2, SigLIP) and trains only a lightweight decoder, finding that DiT models converge substantially faster on these semantically rich latents. SVG~\citep{shi2025svg} pushes this further by removing the VAE entirely, performing diffusion directly on frozen DINOv2 features with a residual branch for pixel recovery. REPA~\citep{yu2024representation} takes a complementary approach: rather than replacing the encoder, it adds an auxiliary alignment loss during training that pulls noisy intermediate denoiser states toward clean-image representations from a pretrained encoder, improving training efficiency and generation quality without any architectural change.

\paragraph{Discrete Tokens: VQ for Autoregressive and Unified Models.}
Autoregressive generation requires a discrete vocabulary. Vector-quantized (VQ) tokenizers map image patches to codebook entries, enabling standard next-token prediction. The codebook quality directly limits generation fidelity: VAR~\citep{tian2024visual} requires a multi-scale VQ-VAE producing hierarchical token maps at multiple resolutions so the backbone can perform its next-scale prediction.

An active frontier is \emph{multimodal tokenization}---designing token spaces that serve both vision and language. TokenFlow~\citep{qu2024tokenflow} uses a dual-codebook design to decouple semantic and pixel-level information while preserving alignment between them. ``Vision as a Dialect''~\citep{han2025vision} introduces TA-Tok, whose codebook is projected from an LLM vocabulary so that image and text tokens share a discrete semantic space. Bridge~\citep{wang2025bridge} proposes a semantic-to-pixel discrete representation: a short semantic-token prefix followed by a longer pixel-token suffix, guiding the model to plan semantics before filling details. Janus-Pro~\citep{chen2025janusprounifiedmultimodalunderstanding} resolves the same invariance-versus-variance tension through separate visual encoders---SigLIP for understanding and VQ tokens for generation---behind a shared AR backbone. UAE~\citep{yan2025uae} pushes the bottleneck further by treating text as the shared interface between image understanding and image generation. Industrial-scale autoregressive systems push tokenization further still with \emph{residual VQ} (RVQ), which interleaves multiple codebook lookups per patch to recover fidelity that a single codebook would lose: LongCat-Next's dNaViT tokenizer~\citep{team2026longcat} stacks eight residual codebooks of size $16{,}384$ at $28\times$ spatial compression and dispenses with a VAE entirely, supplying both image and (separately quantized) audio tokens directly to the AR backbone.

\paragraph{Can We Skip the Tokenizer Entirely?}
JiT~\citep{li2025back} challenges the encoder paradigm by feeding raw pixel patches directly to a transformer. While this sacrifices latent-space efficiency, it eliminates reconstruction artifacts entirely---an appealing property when pixel-level fidelity matters (e.g., medical imaging or document generation).

\paragraph{Representation Trend.}
Viewed together, recent representation designs trace a clear progression: VAE latents minimize compute for diffusion, discrete codebooks enable AR scaling and multimodal token sharing, SSL-aligned latents inject stronger semantics into diffusion backbones, and tokenizer-free pixel representations eliminate reconstruction artifacts altogether. Representation choice has therefore become a co-evolving part of the model design rather than a fixed front-end. In practice, the VAE itself remains unsettled at the frontier: FLUX-VAE~\citep{labs2025flux1kontextflowmatching}, the Wan-2.1 VAE adopted by Qwen-Image~\citep{wu2025qwen} (single-encoder dual-decoder, decoder fine-tuned for text-rich domains), self-trained 4-channel RGBA designs (Wan-Image~\citep{mao2026wan}, with $16\times$ compression and 48 latent channels), and VAE-free RVQ tokenizers (LongCat-Next) all coexist across 2026 reports, indicating that compression ratio, latent dimension, and reconstruction-fidelity trade-offs still carry unresolved choices.

\subsubsection{Backbone Networks}
\label{sec:arch_backbone}

Given the encoded representation---whether continuous latents from a VAE or discrete tokens from a VQ codebook---the backbone is the central compute unit that performs the actual generation: denoising (diffusion), velocity prediction (flow matching), or next-token generation (autoregressive). The choice of backbone determines the model's \emph{inductive bias} (local vs.\ global reasoning), \emph{scaling behavior} (how quality improves with compute), and \emph{inference speed}. The field has evolved through three generations, each unlocking new capabilities.

\paragraph{From U-Net to Transformer: The Scaling Transition.}
Early diffusion models adopted the convolutional U-Net architecture, whose hierarchical feature pyramids naturally match multi-resolution image statistics. LDM (Stable Diffusion)~\citep{rombach2022stablediffusion} made this practical at scale by performing diffusion in a compressed latent space rather than pixel space, with text conditioning injected via cross-attention layers. However, U-Nets lack the predictable compute--quality scaling later observed in transformer backbones.

DiT~\citep{peebles2023scalable} addressed this by replacing the U-Net entirely with a plain vision transformer operating on patchified latents. The key finding was \emph{monotonic FID improvement with increasing FLOPs}, establishing that transformer-based diffusion follows predictable scaling laws analogous to those in language modeling. Conditioning is provided through \emph{adaptive layer normalization} (AdaLN), which regresses scale and shift parameters from a timestep-class embedding---lighter than cross-attention and sufficient for class-conditional generation.

SD3~\citep{esser2024stablediffusion3} extended this with the \emph{MM-DiT} (Modality-aware DiT) architecture, which maintains separate weight streams for text and image tokens while enabling bidirectional information exchange via joint attention. This dual-stream design substantially improved text comprehension and typography fidelity, establishing the architectural template adopted by subsequent industrial systems such as the FLUX family~\citep{labs2025flux1kontextflowmatching}. Two recent variants probe the limits of the dual-stream choice. Z-Image~\citep{cai2025z} is a notable counter-move: its S3-DiT (Single-Stream DiT) collapses MM-DiT's dual streams back into a single attention path, arguing that a stronger upstream VLM and tighter data curation make modality-specific weights redundant---an explicit test of whether the MM-DiT gain is structural or compensatory. A complementary backbone-internal hybrid keeps MM-DiT layers in the lower stack and switches to single-stream DiT in the upper stack once cross-modal binding has been established (FLUX.1-dev style, adopted by LongCat-Image~\citep{team2025longcat} with a 10:20 ratio of MM-DiT to single-stream blocks).

\paragraph{Autoregressive Backbones: Unifying Language and Vision.}
A parallel line of work reframes image generation as a sequence prediction problem, enabling direct reuse of language model architectures. LlamaGen~\citep{sun2024autoregressive} demonstrated that a standard decoder-only transformer with next-token prediction on discrete image tokens can match diffusion baselines given sufficient scale, validating visual scaling laws for the autoregressive paradigm. VAR~\citep{tian2024visual} introduced a more efficient generation order: instead of raster-scan token prediction, it performs \emph{next-scale prediction} in a coarse-to-fine hierarchy, reducing effective sequence length from $\mathcal{O}(n^2)$ to $\mathcal{O}(n\log n)$---the canonical AR backbone alternative to raster-scan, with the FID and speed comparison against DiT discussed earlier in \Cref{sec:model_arch}.

\paragraph{Emerging Architectures.}
Beyond transformers, several alternative backbones have emerged. OmniMamba~\citep{zou2025omnimamba} builds on the Mamba-2 state-space model, reporting up to $119.2\times$ speedup and $63\%$ GPU-memory reduction relative to Transformer-based counterparts on long-sequence generation---particularly relevant for high-resolution generation where sequence lengths explode, a regime that becomes the binding constraint at L3 in-context generation in \Cref{sec:evolution}. At the other extreme, JiT~\citep{li2025back} demonstrates that pixel-space transformers operating \emph{without any tokenizer} can achieve competitive quality when switching from noise prediction to direct $x$-prediction, trading latent-space efficiency for elimination of reconstruction artifacts. Hybrid model families are also gaining traction, but their main novelty lies in how they factorize planning and rendering rather than in a wholly new backbone primitive; we therefore discuss them at the paradigm level in \Cref{sec:hybrid_ar_diff}.

\paragraph{Sparse-Expert Backbones.}
A second axis along which 2026 frontier systems split is parameter sparsity. HunyuanImage~3.0~\citep{cao2025hunyuanimage} routes through a 64-expert, top-8 mixture-of-experts with 80B total / 13B active parameters, and LongCat-Next~\citep{team2026longcat} applies the same sparse-expert recipe to a discrete-AR backbone with 68.5B total / A3B active parameters. The bet is capacity without proportional inference cost; the cost is router stability and a heavier training pipeline. Sparse-expert vision backbones remain rare relative to dense designs---Qwen-Image~\citep{wu2025qwen}, Z-Image~\citep{cai2025z}, Seedream~\citep{seedream2025seedream}, and LongCat-Image~\citep{team2025longcat} all stay dense---indicating that the MoE-versus-dense trade-off is not yet settled at the frontier, and that the MoE bet is currently concentrated in systems that already commit to AR-style backbones where sparse routing composes naturally with token-level prediction.

\subsubsection{Condition Module}
\label{sec:arch_condition}

A generative model without conditioning produces random samples from the learned distribution. The condition module is what makes generation \emph{controllable}: it translates external signals---text prompts, spatial or visual conditions, reference images, or class labels---into features that steer the backbone's output toward user intent. To achieve this systematically without disrupting the underlying modeling objective, modern architectures decouple this process into two fundamental stages: \emph{Condition Feature Extraction} (how to encode the external signal) and \emph{Feature Injection} (how to fuse it into the backbone).

\begin{figure}[!htbp]
    \centering
    \includegraphics[width=\textwidth]{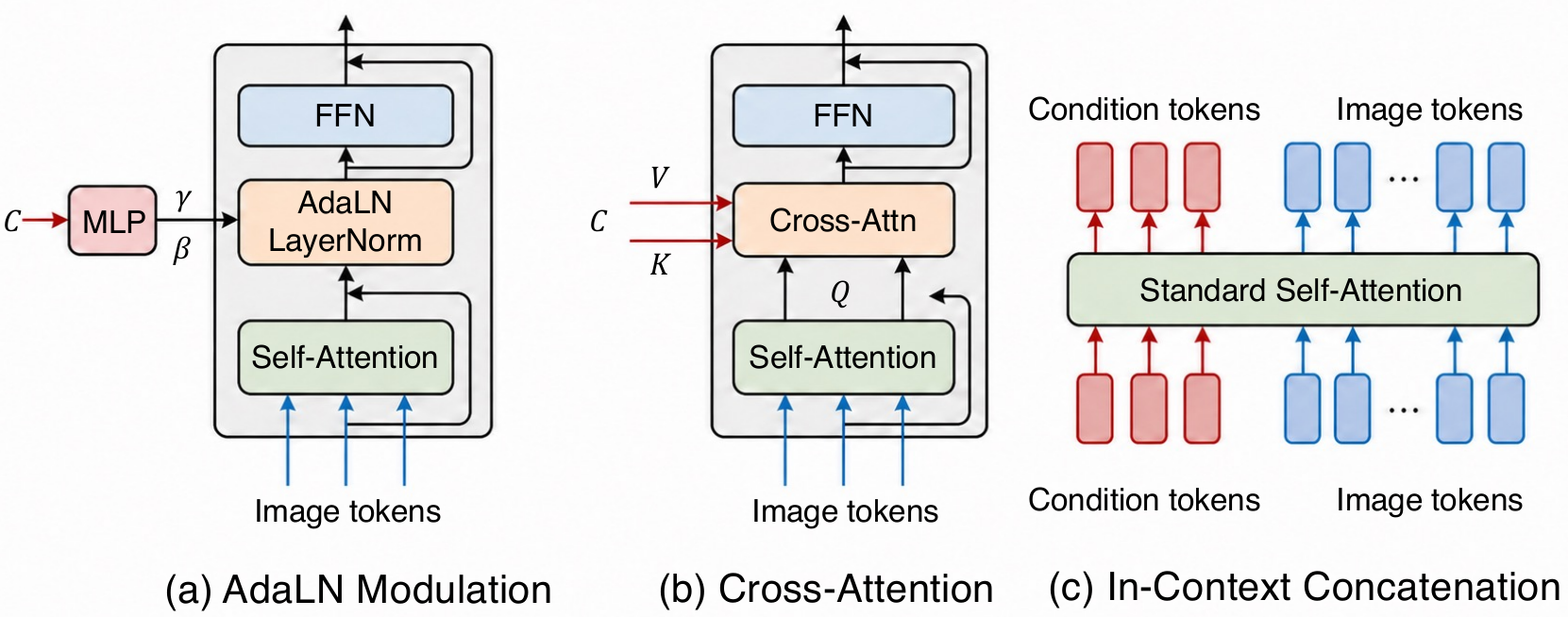}
    \caption{\textbf{Three condition injection routes in modern transformer-based generators.} \textbf{(a) AdaLN modulation}: condition $c$ predicts scale and shift parameters of LayerNorm via a small MLP (used in DiT). \textbf{(b) Cross-attention}: condition $c$ flows into the K/V of an additional cross-attention layer (used in Stable Diffusion's U-Net, IP-Adapter). \textbf{(c) In-context concatenation}: condition tokens are prepended to image tokens in the same sequence and processed by self-attention (used in MM-DiT, OmniGen, Bridge). }
    \label{fig:condition_routes}
\end{figure}

\paragraph{Condition Feature Extraction.}
The first step is transforming raw conditional inputs into a latent representation compatible with the generative backbone. Depending on the complexity and modality of the condition, three dominant extraction paradigms have emerged:
\begin{itemize}
    \item \textbf{Adapter on Copied Backbone:} For spatially dense conditions (e.g., depth maps, edge maps, or poses), preserving exact spatial alignment is critical. Methods like ControlNet~\citep{zhang2023controlnet} and SCEdit~\citep{jiang2024scedit} freeze the pretrained backbone and attach a trainable, initialized-to-zero copy to extract spatial features. ControlNet++~\citep{controlnet++} further refines this by incorporating a discriminative reward model to enforce cycle consistency between input conditions and generated outputs, ensuring higher fidelity in spatial extraction.
    CoLoGen~\citep{song2026cologen} frames unified conditional generation as a representation-conflict problem, separating concept-level semantics from localization-level spatial control instead of forcing all conditions through a single shared adapter.
    \item \textbf{Frozen Vision Foundation Models (VFMs):} When the condition is a reference image requiring semantic or identity preservation, models typically leverage frozen pretrained VFMs (e.g., CLIP, SigLIP, or VAEs). 
    Lumina-mGPT~\citep{liu2024luminamgpt} relies on a VQ-VAE-style image tokenizer to convert images into discrete tokens, enabling a decoder-only mGPT to autoregressively model interleaved text--image sequences.
    Explanatory Instructions~\citep{shen2024explanatory} builds on a token-based AR VLM with an image tokenizer/de-tokenizer, using linguistic transformations between tokenized input-output images to unify diverse vision tasks.
    IP-Adapter~\citep{ye2023ip} extracts global CLIP image embeddings, while recent unified frameworks (OmniGen~\citep{wu2025omnigen2}, Qwen-Image~\citep{wu2025qwen}, Z-Image~\citep{cai2025z}) extract VAE latents and SigLIP patch features in parallel to represent visual prompts. The two streams are complementary: VLM-derived semantic features carry instruction intent (\emph{what} to draw), while VAE-derived pixel features preserve fine-grained appearance (\emph{how} it should look)---a division of labor that the Qwen-Image report identifies as the source of its TI2I gains.
    \item \textbf{Adapter on Vision-Language Models (VLMs):} As editing tasks become more complex, simple visual or textual extraction is insufficient. Recent methods employ powerful VLMs to perform reasoning before conditioning. Systems like Step-1X-Edit~\citep{liu2025step1xedit}, Query-Kontext~\citep{song2025query} and Qwen-Image-Edit~\citep{wu2025qwen} use VLMs to parse complex instructions and extract reasoning-aware conditional features. Similarly, Bridge~\citep{wang2025bridge} introduces semantic planning tokens generated prior to rendering, guiding the backbone with high-level semantic intent.
\end{itemize}

\paragraph{Feature Injection Mechanisms.}
Once the conditional features are extracted, they must be routed into the backbone. The choice of injection mechanism dictates the trade-off between expressiveness, computational cost, and modularity:
\begin{itemize}
    \item \textbf{Feature-wise Addition:} The most direct approach for spatially aligned features is to add the extracted condition directly to the intermediate feature maps. ControlNet and SCEdit utilize this mechanism, bypassing complex attention calculations while ensuring strict spatial constraint propagation.
    \item \textbf{Cross-Attention:} Popularized by LDM~\citep{rombach2022stablediffusion}, this projects conditioning embeddings into key-value pairs that attend to the backbone's queries. While IP-Adapter and InstantID~\citep{wang2024instantid} use separate cross-attention layers, SD3~\citep{esser2024stablediffusion3} employs dual-stream joint attention, allowing the image and text branches to attend to each other bidirectionally, which substantially improves typography and prompt adherence.
    \item \textbf{In-Context Self-Attention:} In unified autoregressive and DiT-based models, conditioning is treated as a token-level protocol. Extracted condition tokens are prepended or interleaved with generation tokens and processed jointly via shared self-attention. This is the core mechanism in OmniGen, Qwen-Image, and Z-Image. OmniMamba~\citep{zou2025omnimamba} enhances this with decoupled modality vocabularies and task-specific LoRA adapters. MMaDA~\citep{yang2025mmada} shows that the same token-level protocol can also be realized in a unified diffusion backbone, where modality-agnostic tokens are denoised jointly rather than decoded causally.
\end{itemize}

\paragraph{Reference-Image Injection in Unified Models.}
When the same model serves both T2I and editing, the source image must be routed into the backbone in a way that does not collide with the noised target. Five mechanisms have emerged across recent frontier reports, and they are stackable rather than mutually exclusive. \emph{(i)} \textbf{VAE-latent token concat}: source latents are concatenated with noised target latents along the sequence dimension (Seedream~4.0~\citep{seedream2025seedream}, Qwen-Image, JoyAI-Image~\citep{jdjoyaiimage}, FireRed~\citep{team2026firered}). \emph{(ii)} \textbf{VLM/MLLM dual-stream}: the source additionally passes through a VLM (Qwen2.5-VL, SigLIP~2) whose semantic features supplement the text condition (Qwen-Image, Z-Image-Edit~\citep{cai2025z}). \emph{(iii)} \textbf{3D-RoPE temporal-axis separation}: reference and target tokens share spatial $(x,y)$ RoPE coordinates but differ on a temporal axis, letting attention disambiguate them (Z-Image-Edit, FireRed, Wan-Image~\citep{mao2026wan}). \emph{(iv)} \textbf{Different timestep conditioning}: the clean reference uses $t{=}0$ while the noisy target uses a normal timestep, exposing the cleanliness gap to the denoiser. \emph{(v)} \textbf{Context-as-prefix}: in pure-AR systems (HunyuanImage~3.0~\citep{cao2025hunyuanimage}, LongCat-Next~\citep{team2026longcat}), the source-image tokens simply precede target tokens as context, with no special routing. The mechanisms are routinely combined: FireRed pairs (i) with (iii), and Wan-Image stacks (i), (iii), and (iv).

\paragraph{Global Modulation and Inference-Time Mechanisms.}
Beyond spatial and token-level injection, models rely on specialized mechanisms to handle global context and inference-time signal amplification:
\begin{itemize}
    \item \textbf{Adaptive Layer Normalization (AdaLN):} Introduced in DiT~\citep{peebles2023scalable}, AdaLN handles global conditions such as timesteps, class labels, and pooled text embeddings. It regresses scale and shift parameters from the condition to modulate feature statistics, providing a lightweight alternative to attention-based injection. SD3~\citep{esser2024stablediffusion3} combines AdaLN for timestep and class conditioning with attention-based injection for text, achieving both efficiency and expressiveness.
    \item \textbf{Classifier-Free Guidance (CFG):} CFG~\citep{ho2022classifier} is an \emph{inference-time} strategy rather than an architectural module. By jointly training conditional and unconditional branches, the model linearly extrapolates score estimates during sampling to amplify the conditioning signal. Increasing the guidance scale improves prompt adherence at the cost of reduced diversity. CFG remains the primary tool for controlling the fidelity-diversity trade-off across virtually all modern generation pipelines.
\end{itemize}

\paragraph{Inference-Time Prompt Rewriting.} Beyond architectural injection mechanisms, frontier systems increasingly augment the text condition itself before it ever reaches the encoder. The Prompt Engineering (PE) module---typically a small fine-tuned VLM such as Qwen3-VL-2B in Wan-Image, Seed1.5-VL in Seedream~3.0/4.0~\citep{gao2025seedream,seedream2025seedream}, or a Gemini-class router---rewrites a user's short query into a long detailed prompt aligned with the training distribution, performs task routing (T2I vs.\ editing vs.\ multi-reference), and optionally injects a chain-of-thought reasoning trace; Wan-Image's dual-mode design, for instance, uses a non-CoT 2B model for short prompts and a CoT 30B-A3B model for complex requests. PE is therefore not an architectural component of the generation backbone; it is an \emph{inference-time conditioning preprocessor} that ensures the gen model receives prompts of the form it was trained on. This explains why high-quality short-prompt performance often correlates more with PE module quality than with backbone capacity---a non-trivial industrial observation.

\subsubsection{Multimodal Fusion Module}
\label{sec:arch_fusion}

As visual generation systems increasingly integrate language understanding and image synthesis within a single model, a critical architectural question emerges: \emph{at what stage should information from different modalities begin to interact?} The fusion module answers this question. Following the taxonomy introduced by Transfusion~\citep{zhou2024transfusion}, we distinguish between \emph{late fusion} (modalities processed independently until a final combination stage) and \emph{early fusion} (modalities interacting within the backbone's internal layers). The trend is decisively toward early fusion, but the degree of parameter sharing varies widely---from fully shared transformers to architectures that maintain separated processing streams while still enabling early-stage cross-modal interaction.

\Cref{tab:unified_methods} summarizes representative unified architectures along this fusion axis. We intentionally keep alignment-only methods out of this summary: once the question shifts from \emph{where modalities meet} to \emph{how a unified model is post-trained to balance understanding and generation}, the emphasis moves from architecture to alignment, which is discussed in \Cref{subsec:post_training}.

\begin{table*}[t]
\centering
\caption{\textbf{Representative unified understanding-and-generation architectures.} \emph{Fusion strategy} indicates where modalities first interact; \emph{visual repr.} denotes the generation-side representation; \emph{gen.\ mechanism} specifies how images are synthesized. Alignment-only methods are discussed separately in \Cref{subsec:post_training}. This table is the multimodal-fusion-lens companion to the modeling-paradigm-lens \Cref{tab:arch_morphology}; the two views overlap on shared papers by design.}
\label{tab:unified_methods}
\scriptsize
\setlength{\tabcolsep}{3.5pt}
\renewcommand{\arraystretch}{1.15}
\resizebox{\textwidth}{!}{%
\begin{tabular}{@{}l c l l l l@{}}
\toprule
\textbf{Method} & \textbf{Year} & \textbf{Fusion Strategy} & \textbf{Visual Repr.} & \textbf{Gen.\ Mechanism} & \textbf{Key Innovation} \\
\midrule
Chameleon~\citep{chameleon2024}            & 2024 & Unified token stream     & Discrete VQ              & AR next-token         & Early-fusion mixed-modal training recipe \\
Transfusion~\citep{zhou2024transfusion}    & 2024 & Unified token stream     & Continuous latent        & AR + Diffusion        & Hybrid loss: CE (text) + diffusion (image) \\
Show-o~\citep{xie2024showo}                & 2024 & Unified token stream     & Discrete VQ              & AR + Discrete diff.   & Attention-mask switching per modality \\
Emu3~\citep{wang2024emu3}                  & 2024 & Unified token stream     & Discrete VQ              & AR next-token         & Pure next-token prediction for mixed-modal data \\
JanusFlow~\citep{ma2024janusflow}          & 2024 & Decoupled encoder        & Continuous latent        & AR + Rect.\ Flow      & LLM backbone harmonized with rectified flow \\
Bridge~\citep{wang2025bridge}              & 2025 & Separated-stream early fusion & Discrete semantic+pixel & AR next-token   & Mixture-of-Transformers with semantic-to-pixel planning \\
Janus-Pro~\citep{chen2025janusprounifiedmultimodalunderstanding} & 2025 & Decoupled encoder & VQ (gen) + SigLIP (und)  & AR next-token  & Separate visual pathways with a shared AR backbone \\
BAGEL~\citep{deng2025bagelemerging}        & 2025 & Unified token stream     & Interleaved multimodal   & AR decoder-only       & Trillion-token scale for unified reasoning and generation \\
Show-o2~\citep{xie2025showo2}              & 2025 & Hybrid heads             & 3D causal VAE            & AR + Flow matching    & Flow-matching image head in shared causal backbone \\
BLIP3o-NEXT~\citep{chen2025blip3onext}     & 2025 & AR$\rightarrow$Diff cascade & Discrete$\rightarrow$cont.  & AR$\rightarrow$Diffusion & AR semantic planning followed by diffusion rendering \\
OmniMamba~\citep{zou2025omnimamba}         & 2025 & Unified token stream     & Discrete VQ              & AR (SSM backbone)     & Linear-time state-space backbone for long contexts \\
MMaDA~\citep{yang2025mmada}                & 2025 & Unified diffusion        & Modality-agnostic        & Diffusion (all mod.)  & Diffusion-native multimodal unification \\
UAE~\citep{yan2025uae}                     & 2025 & Encoder-decoder          & Text as bottleneck       & AR enc.$\rightarrow$dec. & Text bottleneck linking understanding and generation \\
HunyuanImage~3.0~\citep{cao2025hunyuanimage} & 2026 & Unified token stream     & Continuous latent (32-ch) & AR + Flow head        & Industrial-scale Transfusion: 80B/13B-active MoE \\
Wan-Image~\citep{mao2026wan}               & 2026 & MLLM planner + DiT visualizer & VAE (4-ch RGBA)     & AR$\rightarrow$Flow      & Bidirectional attention extends across an image series \\
JoyAI-Image~\citep{jdjoyaiimage}           & 2026 & Decoupled MLLM brain + MMDiT & VLM hidden + VAE     & AR-conditioned diffusion & 8B MLLM conditions a 16B MMDiT via hidden states \\
\bottomrule
\end{tabular}%
}
\end{table*}

\paragraph{Late Fusion: Modular but Limited.}
The simplest approach treats text as an \emph{external condition} injected into a vision-only backbone. LDM~\citep{rombach2022stablediffusion} exemplifies this: frozen CLIP text embeddings are projected into cross-attention layers, creating a one-directional flow from text to image. This design is modular---the text encoder and image generator can be developed independently---but the lack of bidirectional interaction limits the model's ability to handle complex prompts that require joint text-image reasoning. MetaQueries~\citep{pan2025metaqueries} is a more modern bridge of the same flavor: it preserves a strong multimodal understanding model and transfers its latent state to a diffusion decoder through learnable queries, improving generation without yet making the backbone truly unified.

\paragraph{Early Fusion with Separated Streams.}
A growing family of models achieves early fusion while maintaining modality-specific processing pathways. SD3~\citep{esser2024stablediffusion3} exemplifies this via its MM-DiT architecture: text and image tokens pass through modality-specific projection weights but share self-attention within each transformer block, enabling bidirectional cross-modal interaction at every layer. This dual-stream design substantially improves text comprehension and typography fidelity.

Recent unified models extend this principle through specialized mechanisms. \emph{Mixture-of-Transformers} (MoT), as used in Bridge~\citep{wang2025bridge}, assigns modality-specific expert parameters within a shared backbone, enabling early interaction through shared attention while preserving per-modality specialization in the feed-forward layers. Janus-Pro~\citep{chen2025janusprounifiedmultimodalunderstanding} takes a different approach to the same principle: it employs separate visual encoders---SigLIP for understanding, VQ for generation---while routing both through a shared AR backbone, resolving the \emph{invariance-variance tension} that arises because understanding benefits from view-invariant embeddings while generation requires view-variant spatial detail. UAE~\citep{yan2025uae} reaches a similar goal through an encoder-decoder view, using text as a shared bottleneck that lets the model translate between understanding and generation rather than forcing both into a single visual representation.

\paragraph{Early Fusion with Fully Shared Parameters.}
The most integrated approach routes all modality tokens through a single shared backbone with minimal modality-specific branching. Chameleon~\citep{chameleon2024}, Emu3~\citep{wang2024emu3}, and BAGEL~\citep{deng2025bagelemerging} all follow this recipe with decoder-only transformers, differing mainly in scale, tokenizer design, and data recipe. OmniMamba~\citep{zou2025omnimamba} pursues the same level of sharing with a linear-time SSM backbone, trading the transformer's quadratic attention cost for linear scaling on long multimodal contexts. MMaDA~\citep{yang2025mmada} shows that this principle is not exclusive to AR models: a unified diffusion backbone can also denoise modality-agnostic token streams in a fully shared manner.

\paragraph{Hybrid Fusion: Combining Paradigms.}
Many recent systems combine autoregressive and diffusion components in complementary roles, using the AR stage for semantic planning and the diffusion/flow stage for high-fidelity rendering. Transfusion~\citep{zhou2024transfusion} mixes discrete text tokens and continuous image latents in a single sequence, applying causal attention to text and bidirectional attention to image patches with a hybrid cross-entropy / diffusion loss---a defining example of multi-modality early fusion. Show-o and Show-o2 keep a single shared backbone but swap the image-generation head from discrete diffusion to flow matching as the latent representation evolves. JanusFlow and BLIP3o-NEXT instead make the division of labor explicit: the AR component reasons over multimodal context, while the flow or diffusion component renders the final image. HunyuanImage~3.0~\citep{cao2025hunyuanimage} pushes the Transfusion recipe to industrial scale, pairing an 80B/13B-active MoE AR backbone with a flow-matching image head, and is the largest open-weight instantiation of the AR-plus-flow design point to date.

\begin{highlightbox}{Convergence Snapshot: Architecture Across Frontier Tech Reports}

Reading ten frontier tech reports released between April 2025 and April 2026 side-by-side reveals that the architectural search space has \emph{narrowed sharply}. Four near-consensus choices stand out: \emph{(i)} Qwen-VL family is the de-facto bilingual text encoder (used in 6 of 10 reports), displacing T5 / CLIP for non-trivial Chinese rendering and long-prompt comprehension; \emph{(ii)} MM-DiT or its hybrid variants is the dominant backbone, with U-Net effectively retired at the frontier; \emph{(iii)} flow matching with rectified-flow targets has displaced the $\epsilon$- and $v$-prediction objectives common in earlier diffusion training, for both training stability and few-step distillability; and \emph{(iv)} jointly training T2I + I2I + edit on a single architecture is now default---pure T2I-only systems are rare, as detailed in the Edit Unification snapshot earlier in \Cref{sec:method}. The takeaway: the next year of progress is unlikely to come from yet another backbone redesign, but from data, captions, and post-training, as discussed in \Cref{sec:train,sec:resource_and_infra}.
\end{highlightbox}

A complementary frontier extends bidirectional attention not across modalities but across multiple generated images. Wan-Image~\citep{mao2026wan} treats an entire image series (T2S, TI2S) as a single attention context, so each output image attends to the others and inherits their identity, lighting, and style; this produces cross-image consistency that no per-image single-pass system can match, and reframes ``edit'' as one element of a broader \emph{compositional visual creation} task closer to L3 narrative consistency in \Cref{sec:evolution}.

\paragraph{From Architectural Unification to Alignment.}
Fusion design determines where modalities meet, but it does not guarantee that understanding and generation improve each other symmetrically. That problem is increasingly handled in post-training rather than in the architecture itself. We therefore discuss alignment-focused methods such as HermesFlow, UAE's Unified-GRPO stage, MMaDA's UniGRPO, and X-Omni's RL refinement in \Cref{subsec:post_training} rather than treating them as separate generative paradigms here.

\begin{highlightbox}{Where the Field Is Splitting: Architecture Trade-offs}

Convergence is not uniform---several axes are still actively contested across frontier reports. \emph{(i)} \textbf{Diffusion vs.\ Autoregressive vs.\ Hybrid:} Wan-Image~\citep{mao2026wan} is an explicit hybrid bet (AR planner + flow renderer); HunyuanImage~3.0~\citep{cao2025hunyuanimage} is AR-native with a flow head, and LongCat-Next~\citep{team2026longcat} is pure discrete AR; Z-Image~\citep{cai2025z}, Qwen-Image~\citep{wu2025qwen}, Seedream~\citep{seedream2025seedream}, and LongCat-Image~\citep{team2025longcat} stay diffusion-native; and Lumina-mGPT and Janus-Pro~\citep{chen2025janusprounifiedmultimodalunderstanding} occupy a separate AR-only branch. \emph{(ii)} \textbf{MoE vs.\ Dense:} HunyuanImage~3.0 (80B / 13B active) and LongCat-Next (68.5B / A3B) wager on sparse experts, whereas Qwen-Image, Z-Image, and Seedream stay dense. A complementary convergence is parameter-count moderation: Z-Image (6.15B) and LongCat-Image (6B) both argue---and benchmark---that careful data curation and stronger upstream encoders close the quality gap to 20B+ dense systems, shifting the scaling bet from parameters to data and curriculum. \emph{(iii)} \textbf{VAE choice has not converged at all}: FLUX-VAE~\citep{labs2025flux1kontextflowmatching}, Wan-VAE, self-trained VAEs, and even VAE-free designs all coexist. These splits matter because they expose the real \emph{trade-off frontier}---distillation-friendliness, RL-friendliness, throughput, and edit fidelity each pull in different directions, and no single combination dominates yet.
\end{highlightbox}

\subsection{The Closed-Source Frontier: A Speculative Reading}
\label{sec:closed_source_frontier}

The frontier of visual generation is no longer set entirely by the open-source models discussed above. Closed-source systems---most visibly Google's Nano Banana\footnote{\url{https://gemini.google/overview/image-generation/}} and OpenAI's GPT-Image-2\footnote{\url{https://openai.com/index/introducing-chatgpt-images-2-0/}}---deliver a noticeably tighter user experience on multi-turn editing, long-form prompts, and structured-content rendering, yet publish almost nothing about how they are built.

This subsection makes four falsifiable conjectures about how the closed-source stack departs from the open-source recipe, then grounds them in user-visible failure modes that the open community currently does not match. Mapped onto the L1--L5 taxonomy of \Cref{sec:evolution}, these conjectures imply that today's open-vs-closed gap is concentrated in the system architecture around the renderer rather than in the renderer itself: the closed systems most plausibly realize \emph{L4 agentic generation} (multi-call control flow), whereas the strongest open-source systems discussed here expose a single forward pass and are therefore L3-bounded \emph{by construction}.

\paragraph{(i) Text-encoding path: a frontier-grade VLM upstream of the renderer.}
Closed systems handle long compositional instructions, multi-turn references to previously generated images, and casual chatty rephrasing in a way that is hard to explain with a frozen T5\footnote{T5: a transformer text-encoder family widely used as the upstream text encoder by Stable Diffusion 1--3 and many subsequent open-source T2I systems.} or even a small VLM. We conjecture the text condition is produced by---or routed through---a frontier-grade VLM (a Gemini-class or GPT-4o-class model) that can plan and rephrase before emitting the prompt actually consumed by the renderer. This is the same Prompt-Engineering-module idea documented in open systems such as Wan-Image~\citep{mao2026wan} (Qwen3-VL-2B) and Seedream~\citep{seedream2025seedream} (Seed1.5-VL), pushed to a much stronger upstream model.

\paragraph{(ii) Visual-encoding path: a dual-path encoder for fidelity and semantics.}
Multi-turn and multi-image edits in closed systems preserve fine pixel-level detail---texture, hair, micro-typography---across many rounds while still executing context-aware modifications. A purely semantic encoder (CLIP / SigLIP-style) would lossily summarize the source and force the renderer to re-hallucinate fine detail on every round, whereas a purely VAE-style encoder would lack the semantic priors needed to localize edit intent. The behavior is most consistent with a dual-path encoder pairing a VAE-style branch for pixel-addressable fidelity with a semantic branch (SigLIP-style or representation-aligned) for edit-intent localization---the \emph{DiT-camp} unification recipe of \Cref{sec:arch_components}.

\paragraph{(iii) Generation-time understanding: from preprocessing to control signal.}
Beyond input encoders, closed systems suggest a qualitative shift in how understanding interacts with generation at runtime. GPT-Image-2's behavior on code-centric rendering/editing and structured mathematical figures (\Cref{fig:code_generation_case,fig:math_structure_case}) implies the renderer must preserve symbolic dependencies, respect layout-level constraints, and revise local visual elements according to their semantic and functional roles---understanding is no longer an upstream preprocessing step but a generation-time control signal that participates as a persistent constraint during multi-step synthesis and editing. This contrasts with the open-source paradigm, where semantic understanding is typically supplied by frozen encoders, prompt-engineering modules, or representation alignment, rather than as a runtime constraint inside the rendering loop.

\begin{figure*}[!htbp]
    \centering
    \includegraphics[width=0.9\textwidth]{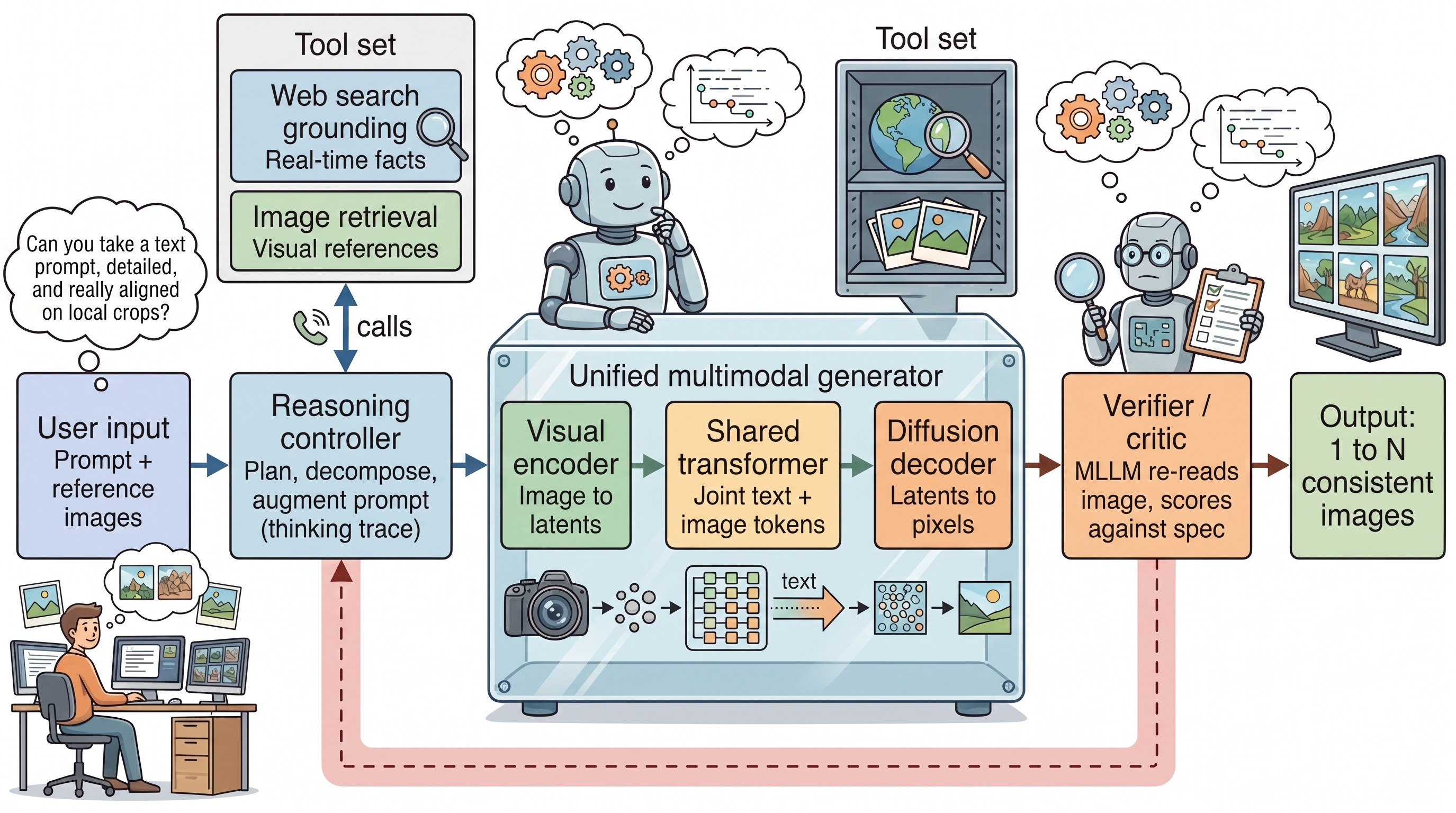}
    \caption{\textbf{Speculative agent-loop architecture for closed-source frontier image generation.} The diagram visualizes conjecture (iv): a frontier VLM ingests the user instruction (and any reference image) and emits a planned prompt; a renderer (likely diffusion) generates a candidate image; a verifier checks the candidate against the plan and either accepts it or routes it back to a refiner that can call a small toolbox (e.g., text-rendering, font alignment, search grounding) before re-rendering. \emph{This figure is tentative and will be revised} as more evidence becomes available; treat it as a hypothesis-shaped sketch rather than a confirmed pipeline.}
    \label{fig:closed_source_agent}
\end{figure*}

\paragraph{(iv) System level: an external agent loop, not a single forward pass.}
Above the renderer itself, behaviors such as silent self-correction of broken outputs, partial regeneration when only a region needs updating, and tool-like calls (e.g.,~aligning text rendering to a known font, or grounding a chemical structure in a search result) are most easily explained if the public API hides an agent loop with a planner, a verifier, a refiner, and a small toolbox, rather than a single forward pass through a monolithic model (\Cref{fig:closed_source_agent}).

\paragraph{Open vs.\ closed gap, by failure mode.}
Four failure modes recur when stress-testing open-source models against Nano Banana and GPT-Image-2; each maps onto one or more of (i)--(iv).

\paragraph{Multi-turn edit drift.}
Open systems show steeper identity, lighting, and composition drift after three or more rounds, whereas the closed systems' degradation curve is markedly flatter (see \Cref{sec:stress_test}, Dimension IV, for multi-turn editing case studies). The gap is consistent with both \emph{(ii)}---a high-fidelity VAE keeps the source pixel-addressable across turns---and \emph{(iv)}---a verifier in the agent loop catches and re-renders failed turns silently before the user ever sees them.

\paragraph{Joint long-form, multi-condition adherence.}
Requests such as ``keep the dress, change the background to a snowy mountain at sunset, add three skiers in the lower-left, render the sign in Chinese'' are a second axis where open models more often drop one constraint silently. The gap is consistent with \emph{(i)}---an upstream VLM resolves and re-emits the constraint set in a renderer-friendly form---and \emph{(iv)}---a verifier checks each constraint in the loop and triggers re-rendering on miss.

\paragraph{Domain-knowledge generation.}
Chemical structures, electrical schematics, multi-step diagrams, and dense bilingual long text are a third axis where closed systems pull ahead. The gap is consistent with \emph{(i)}---world-knowledge retrieval and planning before rendering---\emph{(iii)}---understanding holding symbolic constraints during the renderer step---and \emph{(iv)}---tool-like grounding calls (font alignment, search) inside the agent loop.

\paragraph{Reasoning-augmented requests.}
Tasks that require the model to think before drawing (e.g., ``solve this geometry problem on the image'', see \Cref{subsubsec:physics_solver_case}) are the fourth axis. The gap is essentially \emph{(iii)} + \emph{(iv)}: understanding actively rewriting pixels inside an L4 closed loop, conceptually impossible to match with a single L3 forward pass.

Crucially, none of (i)--(iv) requires architectural innovation that the open-source community lacks; each is a system-engineering bet rather than a fundamentally new model. The open question for the next year is therefore less about narrowing a per-pixel quality gap, and more about reproducing the \emph{system architecture} around the renderer.

\begin{highlightbox}{Community Message: Closed-Source Frontier $\approx$ Production L4 Agentic Generation}

Mapped onto the L1--L5 taxonomy of \Cref{sec:evolution}, the inferred Nano-Banana / GPT-Image-2 stack is precisely \emph{L4 agentic generation realized in production}: a planner / verifier / refiner controller decides multiple forward passes around the renderer (\Cref{tab:5levels}). Open-source systems that currently expose a single forward pass to the user are L3-bounded \emph{by construction}---no amount of renderer improvement, data scaling, or RL on the renderer alone can promote them across the L3$\to$L4 boundary. The community's most consequential next move is therefore not another half-point of FID, but the open-source reproduction of the agent loop \emph{around} the renderer.
\end{highlightbox}

\section{Training and Inference}\label{sec:train}

\begin{figure*}[!htbp]
    \centering
    \includegraphics[width=\textwidth]{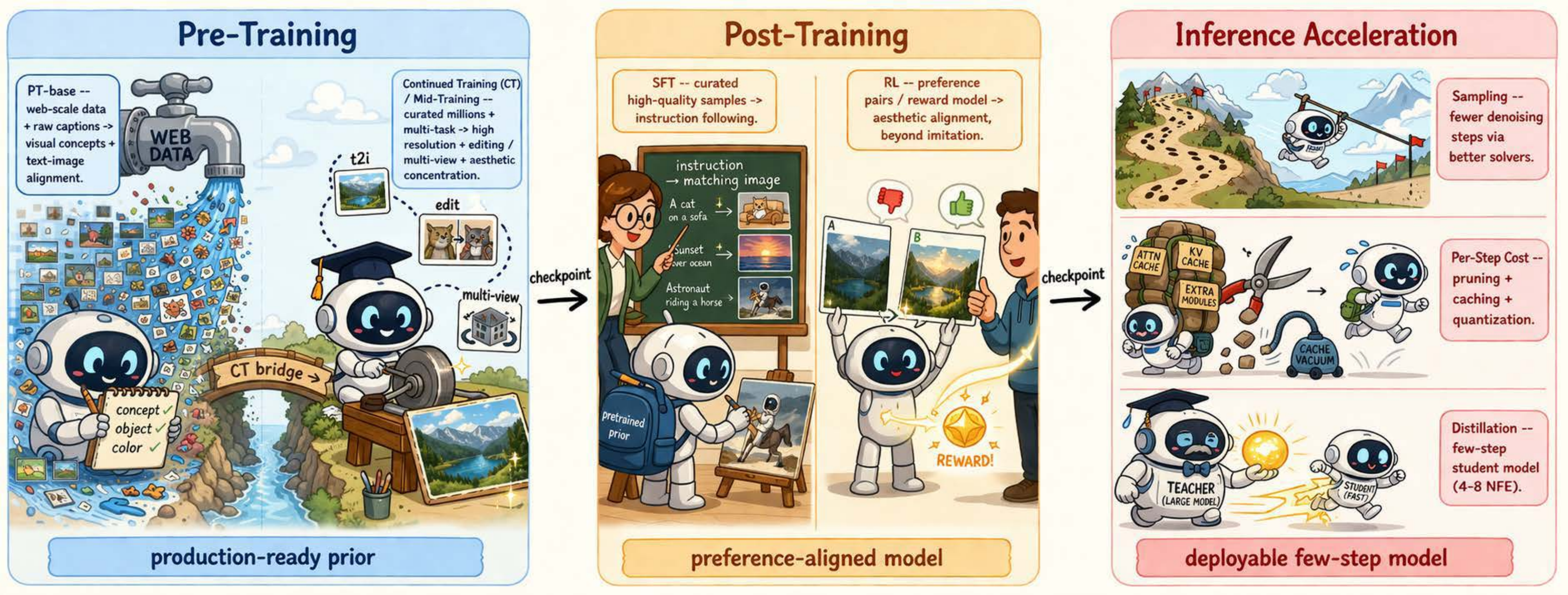}
    \caption{\textbf{Overview of the three-stage training and inference pipeline for visual generation models.} \textbf{Pre-training} learns a broad visual prior from web-scale data, then a continued-training (CT) sub-stage graduates the model to production resolutions and downstream tasks (text-to-image, editing, multi-view). \textbf{Post-training} aligns the prior with human preference through supervised fine-tuning followed by reinforcement learning. \textbf{Inference acceleration} reduces deployment cost through faster ODE/SDE solvers, per-step compute reduction (pruning, caching, quantization), and few-step distillation.}
    \label{fig:training_pipeline}
\end{figure*}

{Training pipelines determine how the model architectures discussed in \Cref{sec:method} are translated into scalable visual capabilities. Along the capability trajectory defined in \Cref{sec:evolution}, this process can be organized into three stages. \textbf{Pre-training} (\S\ref{subsec:pre-train}) establishes the base capability through large-scale data curation, VLM-driven annotation, curriculum learning, and a continued-training (CT) bridge that graduates the model toward higher resolutions and downstream tasks. \textbf{Post-training} (\S\ref{subsec:post_training}) transforms the pretrained generator into a user-aligned model through supervised fine-tuning (SFT) and reinforcement learning. \textbf{Inference acceleration} (\S\ref{subsec:acceleration}) addresses the deployment bottleneck of inference speed through faster samplers, per-step efficiency improvements, and distillation for few-step generation. \Cref{fig:training_pipeline} provides an overview of this pipeline.}

\subsection{Pre-training Methodologies}
\label{subsec:pre-train}

Pre-training serves as the bedrock of visual intelligence, where the model learns to map semantic concepts to visual manifestations. Recent industrial-grade models, exemplified by Qwen-Image~\citep{wu2025qwen} and Z-Image~\citep{cai2025z}, reveal a consensus: the bottleneck has shifted from parameter scaling to \textbf{Data Density} and \textbf{Training Efficiency}. We organize this phase around four recurring axes: data construction and curation, synthetic annotation and captioning, systems--architecture co-design, and a staged curriculum spanning pre-training and continued training, as summarized in \Cref{tab:pretrain_summary}.

This industrial landscape is now broader than any single pair of case studies. Seedream~3.0/4.0~\citep{gao2025seedream,seedream2025seedream} emphasize bilingual typography, defect-aware data construction, mixed-resolution training, and deployment-aware acceleration. HunyuanImage~3.0~\citep{cao2025hunyuanimage} pushes in a different direction: a native multimodal MoE system that couples aggressive filtration of a $>$10B-image pool with Chain-of-Thought-style multimodal supervision and heavy post-training. LongCat-Image~\citep{team2025longcat} represents a compact 6B-class path focused on Chinese--English text rendering, photorealism, and an end-to-end pre-training/SFT/DPO recipe, while LongCat-Next~\citep{team2026longcat} explores a divergent paradigm in the same family: a 68.5B sparse-MoE model that lexicalizes vision and audio as discrete tokens for a fully autoregressive multimodal backbone. Wan-Image~\citep{mao2026wan} pushes in yet another direction, coupling an MLLM Planner and a DiT Visualizer in a shared-attention rectified-flow stack that natively supports T2I, I2I, and image-series generation. FireRed-Image-Edit-1.0~\citep{team2026firered}, by contrast, reflects an editing-centric branch in which the training pipeline is designed from the outset around instruction-following edits, identity preservation, and later preference optimization. JoyAI-Image~\citep{jdjoyaiimage} adds a spatial-intelligence-first route: instead of loosely coupling perception and generation, it builds a unified MLLM--MMDiT stack around the explicit goal of improving geometry-aware understanding, controllable editing, and view-consistent generation through shared data engines and multi-stage training. The overall trend is clear: modern visual foundation models are differentiating not only by scale, but by how they allocate data quality, supervision, and system budget across the full lifecycle.

\begin{table*}[t]
\centering
\caption{\textbf{Representative industrial training recipes in modern visual generation.} Rather than being exhaustive, this table highlights how recent systems differentiate themselves by choosing different primary bottlenecks to optimize: typography, efficiency, unification, editing, or spatial intelligence.}
\label{tab:industrial_training_recipes}
\footnotesize
\setlength{\tabcolsep}{3.5pt}
\renewcommand{\arraystretch}{1.12}
\begin{tabularx}{\textwidth}{@{}>{\raggedright\arraybackslash}p{2.9cm}>{\raggedright\arraybackslash}p{2.4cm}>{\raggedright\arraybackslash}p{4.0cm}>{\raggedright\arraybackslash}X@{}}
\toprule
\textbf{Model family} & \textbf{Primary emphasis} & \textbf{Representative design choice} & \textbf{Pipeline lesson} \\
\midrule
Seedream 3.0/4.0 \citep{gao2025seedream,seedream2025seedream} & Typography + deployment & Defect-aware data construction, mixed-resolution training, acceleration-aware design & High-fidelity bilingual rendering depends as much on data engines and curricula as on backbone scale. \\
Qwen-Image \citep{wu2025qwen} & General-purpose industrial scaling & Waterfall filtering, VLM relabeling, multi-stage SFT/post-training & Large dense models benefit from aggressive curation and staged alignment rather than parameter count alone. \\
Z-Image \citep{cai2025z} & Efficiency-first diffusion & Single-stream DiT, OCR-first captioning, compact 6B deployment recipe & Strong quality can be preserved in compact models when the training recipe is co-designed for efficiency. \\
HunyuanImage 3.0 \citep{cao2025hunyuanimage} & Native multimodal unification & Sparse MoE backbone, reasoning-style supervision, aggressive post-training & Unification shifts the bottleneck from aesthetics alone to balancing reasoning and generation under a shared budget. \\
LongCat-Image \citep{team2025longcat} & Compact bilingual generation & 6B-class full-stack pre-training/SFT/DPO recipe & Strong staged curation and later alignment can keep smaller models competitive in rendering and photorealism. \\
LongCat-Next \citep{team2026longcat} & Discrete-token unification & 68.5B sparse-MoE autoregressive backbone with shared vision/audio/text tokenization & A discrete-AR multimodal stack offers natural finite-MDP RL semantics at the cost of harder distillation. \\
Wan-Image \citep{mao2026wan} & Unified planning + visualization & MLLM Planner + DiT Visualizer in shared attention; Generation-CT with multi-task mix & Coupling understanding and generation under one attention stack expands the task surface to T2I, I2I, and image series. \\
FireRed-Image-Edit-1.0 \citep{team2026firered} & Editing-first optimization & Balanced generation/editing corpus, editing-oriented post-training & If editing is the product target, the pipeline should be built around editing from the outset. \\
JoyAI-Image \citep{jdjoyaiimage} & Spatial intelligence in unified models & Qwen3-VL-based MLLM + Wan-2.1-VAE + 16B MMDiT + OpenSpatial engine & Spatial reasoning can be treated as a pipeline-wide objective rather than a downstream benchmark add-on. \\
\bottomrule
\end{tabularx}
\end{table*}

\begin{table*}[t]
\centering
\caption{\textbf{Key axes of modern visual foundation model pre-training across recent tech reports.} The table abstracts recurring strategies from frontier industrial systems rather than treating Qwen-Image and Z-Image as the only representative cases.}
\label{tab:pretrain_summary}
\scriptsize
\setlength{\tabcolsep}{3.5pt}
\renewcommand{\arraystretch}{1.16}
\begin{tabularx}{\textwidth}{@{}>{\raggedright\arraybackslash}p{2.6cm}>{\raggedright\arraybackslash}p{3.0cm}>{\raggedright\arraybackslash}p{4.6cm}>{\raggedright\arraybackslash}X@{}}
\toprule
\textbf{Dimension} & \textbf{Key strategy} & \textbf{Representative reports} & \textbf{Pipeline implication} \\
\midrule
Data construction & Waterfall / defect-aware filtering; long-tail curation & Qwen-Image~\citep{wu2025qwen}; Z-Image~\citep{cai2025z}; Seedream~3.0/4.0~\citep{gao2025seedream,seedream2025seedream}; HunyuanImage~3.0~\citep{cao2025hunyuanimage} & Web-scale data is useful only after aggressive quality, semantic, diversity, and failure-aware filtering; long-tail concepts increasingly require active retrieval and re-sampling rather than passive scraping. \\
\midrule
Synthetic annotation & Structured VLM relabeling; OCR-first / multi-level captions & Qwen-Image~\citep{wu2025qwen}; Z-Image~\citep{cai2025z}; LongCat-Image~\citep{team2025longcat}; JoyAI-Image~\citep{jdjoyaiimage} & Captions are becoming an engineered supervision interface: they separate attributes, preserve readable text, encode bilingual typography, and expose spatial relations that raw web metadata misses. \\
\midrule
Systems and architecture & Producer--consumer input pipelines; hybrid parallelism; dense/MoE co-design & Qwen-Image~\citep{wu2025qwen}; Z-Image~\citep{cai2025z}; HunyuanImage~3.0~\citep{cao2025hunyuanimage}; JoyAI-Image~\citep{jdjoyaiimage} & Training efficiency is no longer separable from architecture: token layout, activation sparsity, parallelism, and data loading jointly determine feasible model scale and utilization. \\
\midrule
Curriculum learning & Resolution graduation; typography/task staging & Z-Image~\citep{cai2025z}; Qwen-Image~\citep{wu2025qwen}; Seedream~3.0/4.0~\citep{gao2025seedream,seedream2025seedream}; LongCat-Image~\citep{team2025longcat} & Models commonly learn coarse composition at lower resolution before high-resolution refinement, while text-heavy, bilingual, or task-specific samples are staged to avoid early overfitting. \\
\midrule
Continued training / CT & High-quality million-scale CT; editing and multi-view expansion & Seedream~4.0~\citep{seedream2025seedream}; FireRed-Image-Edit-1.0~\citep{team2026firered}; LongCat-Image~\citep{team2025longcat}; JoyAI-Image~\citep{jdjoyaiimage}; Wan-Image~\citep{mao2026wan}; LongCat-Next~\citep{team2026longcat}; Z-Image~\citep{cai2025z} & Modern tech reports increasingly insert a CT / mid-training bridge between generic pre-training and narrow SFT, using curated data to graduate models toward production resolution, editing, multi-view, and spatially grounded capabilities. \\
\bottomrule
\end{tabularx}
\end{table*}

\subsubsection{Data Construction and Curation}
Modern data pipelines have evolved from simple heuristic filters to dynamic, model-in-the-loop engines.

\paragraph{Waterfall Filtering.}
Qwen-Image establishes a rigorous multi-stage filtering protocol to progressively refine data quality. The pipeline begins with basic physical filters (Resolution, Aspect Ratio, Broken Files) and progresses to semantic gates (Image-Text Alignment via CLIP/SigLIP, Aesthetic Scoring). Specific filters target statistical anomalies, such as extreme Entropy (removing empty images), Luma/Saturation (removing artificial artifacts), and Texture complexity. Finally, deduplication via graph-based community detection (as seen in Z-Image) or perceptual hashing (Qwen-Image) is employed to remove redundancy, ensuring the model sees diverse unique samples rather than memorizing duplicates.

This filtration philosophy now appears to be broadly shared across industrial systems. HunyuanImage~3.0~\citep{cao2025hunyuanimage} likewise reports a multi-stage filtering pipeline over a raw pool of more than 10 billion images, retaining less than half after semantic, quality, and diversity constraints are enforced. Seedream~3.0~\citep{gao2025seedream} adopts a defect-aware data engine and dual-axis sampling scheme to reduce caption noise, visual corruption, and typography-specific failure cases. JoyAI-Image~\citep{jdjoyaiimage} adds a complementary perspective: beyond generic quality filtering, it constructs an OpenSpatial engine that turns 3D scans and web videos into box-centric spatial supervision, suggesting that for unified models, ``data quality'' increasingly means not only cleaner images but also stronger geometric grounding. The common message is that web scale alone is no longer a differentiator; the differentiator is how aggressively the pipeline can turn noisy web corpora into a semantically balanced and failure-aware training set.

\paragraph{Active Curation and Knowledge Topology.}
Z-Image introduces a more dynamic "Active Curation" paradigm. Instead of passive filtering, it employs a World Knowledge Topological Graph constructed from Wikipedia and internal captions. By mapping training data to this topological graph, the system achieves semantic balancing by identifying under-represented concepts (the "long-tail") and dynamically up-sampling them. Furthermore, an active learning loop creates a self-improving data engine by using the model's own failure cases (e.g., inability to render specific cultural dishes) to trigger targeted retrieval and annotation.

\subsubsection{Synthetic Annotation and Captioning}
Raw web captions are dominated by alt-text, SEO metadata, and incidental descriptions that tend to be short, noisy, and focused on context rather than visual content. They rarely enumerate object attributes, spatial arrangements, or stylistic details at the granularity a generator needs for complex compositions. A critical insight is therefore that web-scale quantity cannot substitute for descriptive density, and the industry has standardized on using large Vision-Language Models (VLMs) to re-annotate data with dense, structured captions.

\paragraph{Structured Annotation and Visual CoT.}
Qwen-Image utilizes VLMs (e.g., Qwen2.5-VL) to output structured JSON metadata, explicitly separating the visual caption from attributes like "Image Style," "Watermark Presence," and "OCR Text." This prevents the model from baking in unwanted artifacts as semantic features. Z-Image's captioner employs a Visual Chain-of-Thought (CoT) strategy: it first explicitly performs OCR to recognize all text in the image, and \textit{then} generates the caption. This causal sequencing significantly improves the model's text rendering capability by grounding the caption in recognized glyphs.

Other recent systems suggest that this ``caption bottleneck'' is now a first-order design target. Seedream~3.0~\citep{gao2025seedream} attributes a substantial portion of its gains in bilingual text rendering and prompt fidelity to improved data construction, including defect-aware filtering and stronger alignment between textual supervision and visual layout. LongCat-Image~\citep{team2025longcat} similarly treats Chinese--English rendering as a first-class objective throughout pre-training, mid-training, and SFT rather than as a downstream afterthought. JoyAI-Image~\citep{jdjoyaiimage} extends this recipe with multi-level captioning and OCR-aware bilingual descriptions, using Qwen3-VL-8B-Instruct to generate short, long, extended-long, and structured captions so that the same pipeline can absorb both general semantics and spatially grounded textual supervision. Across these systems, the role of captions has shifted: they are no longer passive metadata, but an engineered interface that determines how much structure the model can absorb from the world.

\paragraph{Targeted Synthesis.}
To solve the "long-tail" distribution of text rendering (especially for non-Latin scripts), Qwen-Image synthesizes data via three strategies: pure rendering (clean background), compositional rendering (text on objects), and complex rendering (UI/Slide layouts).

\subsubsection{Systems Engineering}
Publicly documented visual generation systems already span dense models in the 6B--20B range, while the scale of leading closed-source systems is often undisclosed and may include much larger sparse MoE architectures with substantially more total parameters than active parameters per token. Across these regimes, training strong foundation models requires sophisticated distributed engineering to maximize model flops utilization (MFU) under memory, communication, and data-loading constraints.

\paragraph{Decoupled Pipelines and Hybrid Parallelism.}
To eliminate data loading bottlenecks, Qwen-Image adopts a Ray-inspired architecture where "Producer" nodes (CPU-heavy) perform asynchronous preprocessing and caching, while "Consumer" nodes (GPU-heavy) focus purely on gradient updates. This decoupling ensures 0\% GPU idle time due to I/O. Both models employ advanced parallelism. Qwen-Image combines Data Parallelism (DP) with Tensor Parallelism (TP) and Head-wise parallelism to fit large MMDiT models. Z-Image utilizes FSDP2 for optimizer sharding and sequence-length aware dynamic batching to minimize padding waste during multi-resolution training.

Recent open systems span a much wider systems spectrum than this 6B--20B range might suggest. LongCat-Image~\citep{team2025longcat} pursues an efficiency-first design in the 6B class, avoiding the memory overhead of very large sparse models while still targeting competitive typography and photorealism. HunyuanImage~3.0~\citep{cao2025hunyuanimage}, in contrast, adopts an MoE architecture with more than 80B total parameters but only 13B active parameters per token, showing that sparse activation and systems co-design are becoming a viable alternative to dense scaling. JoyAI-Image~\citep{jdjoyaiimage} highlights another systems lesson at the post-understanding stage: dynamic sequence packing and decoupled learning rates are used during spatial-specialized SFT of the MLLM to preserve pretrained vision representations while allowing the language backbone to adapt more aggressively. The broader lesson is that ``training systems'' can no longer be separated cleanly from ``model design'': the architecture dictates the parallelization strategy, and the systems budget feeds back into what architectural choices are practical.

\paragraph{Architecture-Efficient Designs.}
System efficiency is increasingly co-designed with model architecture rather than handled as a separate engineering layer. Compact dense models such as Z-Image reduce training and serving cost through a Scalable Single-Stream DiT (S3-DiT), which concatenates text and image tokens into a unified sequence with 3D Unified Rotary Positional Embeddings (RoPE). In contrast, MoE-based systems such as HunyuanImage~3.0 trade a larger total parameter count for sparse activation, reducing active computation per token while preserving capacity. Unified multimodal stacks such as JoyAI-Image further show that architecture choices on the understanding side---for example, MLLM conditioning, dynamic sequence packing, and decoupled learning rates---directly shape the feasible training system. The common lesson is that architectural efficiency and systems efficiency must be optimized jointly: token layout, activation sparsity, modality fusion, and parallelization strategy together determine the practical training budget.

\subsubsection{Staged Curriculum: From Pre-training to Continued Training}
\label{subsec:curriculum}

{Modern pipelines reject static training in favor of a progressive curriculum that spans pre-training and a subsequent \emph{continued-training} (CT) stage, also called \emph{mid-training} in recent literature. The two stages form a single spectrum rather than separate phases. Pre-training alone is too coarse for production use---it operates on noisy web-scale data at modest resolution to learn broad concepts---while supervised fine-tuning is too narrow to exercise the full generative prior. CT bridges the gap by inheriting the flow-matching objective and image-only data regime from pre-training, but operating on a tighter, higher-quality corpus (typically millions rather than billions of samples), graduating the model to production-grade resolutions, and introducing additional task heads such as image editing and multi-view consistency. Across ten frontier tech reports analyzed (April 2025--April 2026), eight explicitly insert this stage---Seedream~3.0/4.0~\citep{gao2025seedream,seedream2025seedream}, FireRed-Image-Edit-1.0~\citep{team2026firered}, LongCat-Image~\citep{team2025longcat}, JoyAI-Image~\citep{jdjoyaiimage}, Wan-Image's Generation-CT~\citep{mao2026wan}, LongCat-Next's mid-training~\citep{team2026longcat}, and Z-Image's Omni-pre-training~\citep{cai2025z}---while Qwen-Image~\citep{wu2025qwen} and HunyuanImage~3.0~\citep{cao2025hunyuanimage} fold equivalent behavior into late pre-training sub-stages rather than naming a separate phase. We discuss the resulting curriculum along three axes: resolution graduation, task expansion and edit specialization timing, and data tightening.}

\paragraph{Resolution Graduation.}
{Training typically starts at low resolutions (e.g., $256^2$ px) to learn semantic composition and color quickly---consuming more than half of the total compute in Z-Image~\citep{cai2025z}---before progressively scaling to higher resolutions ($512^2 \rightarrow 1024^2$) for detail refinement. CT is where models cross the production-resolution threshold, often through several intermediate sub-stages with mixed aspect ratios. Seedream~4.0~\citep{seedream2025seedream} continues training in the $1024^2$--$4096^2$ range, JoyAI-Image~\citep{jdjoyaiimage} steps from 208P through 512P to 1024P, and Wan-Image's Generation-CT~\citep{mao2026wan} reaches up to 2K natively and 4K through a downstream Refiner. Z-Image-Edit~\citep{cai2025z} runs a similar $512^2 \rightarrow 1024^2$ schedule on a dedicated edit checkpoint. The shared engineering rationale is that low-resolution training is the cheapest place to learn composition, while a comparatively small high-resolution tail is where detail and aesthetic-grade rendering are acquired. Skipping the early low-resolution stage forces detail-level optimization on top of incomplete composition priors and tends to inflate compute without proportional quality gains.}

\paragraph{Task Expansion and Edit Specialization Timing.}
{A central design choice is when to introduce image editing and multi-view tasks. Three timings are visible across the analyzed reports, each with its own trade-off. \textit{PT-mixed} schedules co-train T2I and editing from early pre-training: Z-Image's Omni-pre-training~\citep{cai2025z} packs T2I, I2I, and weakly-aligned image pairs into a single curriculum, while Qwen-Image's I2I reconstruction~\citep{wu2025qwen} is added not for reconstruction itself but to align the MMDiT image stream with the Qwen2.5-VL semantic space, so that editing capability emerges as a byproduct of representation alignment. The benefit is that editing arises with no extra stage; the risk is interference with the dominant T2I objective. \textit{CT-introduced} schedules keep pre-training pure and activate multi-task learning at the CT stage: Seedream~4.0's CT~\citep{seedream2025seedream} primarily improves instruction following for editing, while Wan-Image's Generation-CT~\citep{mao2026wan} mixes T2I, I2I, T2S, and TI2S tasks at a roughly $7{:}2{:}0.5{:}0.5$ ratio. This timing is structurally cleaner and easier to schedule, at the cost of an explicit transition stage and the bookkeeping of stage-wise task ratios. \textit{Fork-after-main} schedules keep T2I as the main run and train an independent edit checkpoint from a CT or SFT branch: LongCat-Image-Edit~\citep{team2025longcat}, FireRed-Image-Edit-1.0~\citep{team2026firered}, JoyAI-Image-Edit~\citep{jdjoyaiimage}, and Z-Image-Edit~\citep{cai2025z} all follow this pattern, paying double-checkpoint storage cost in exchange for full isolation between the two task distributions. The fork-after-main pattern is becoming the default whenever a team treats editing as a distinct product rather than a side capability of the base model.}

\paragraph{Data Tightening for the PT--SFT Bridge.}
{CT corpora are built by raising the filters used at pre-training rather than by replacing the data engine. Aesthetic, image-quality-assessment (IQA), watermark, and AIGC-detection thresholds are tightened so that the training pool drops from web-scale billions to curated millions. LongCat-Image~\citep{team2025longcat} runs a dedicated IQA pipeline together with an AIGC detector that is also reused adversarially during reinforcement learning, while Wan-Image's stage-by-stage threshold tightening~\citep{mao2026wan} progressively raises the bar across CT sub-stages. Seedream~3.0/4.0~\citep{gao2025seedream,seedream2025seedream} extend their defect-aware data engine---originally built for pre-training---to also remove typography and edit-specific failure cases at the CT stage. The role of filtering therefore shifts as the curriculum advances: pre-training filters preserve diversity in a noisy web-scale pool, whereas CT filters trade volume for fidelity, retaining only samples that match the production resolution and aesthetic targets that downstream SFT will further specialize.}

\begin{highlightbox}{Convergence Snapshot: Training Recipe Has Standardized}

Reading the same ten frontier tech reports released between April 2025 and April 2026 along the training-pipeline axis reveals that the macro recipe has \emph{narrowed just as sharply as the architecture}. Four near-consensus choices stand out: \emph{(i)} the pipeline has converged on a four-stage skeleton---\textbf{PT $\to$ CT $\to$ SFT $\to$ RL}---and every analyzed system instantiates the full four stages; \emph{(ii)} the PT--SFT bridge is structural rather than optional, with eight reports naming an explicit continued-training (CT) / mid-training stage and the remaining two~\citep{wu2025qwen,cao2025hunyuanimage} folding equivalent behavior into late pre-training sub-stages; \emph{(iii)} SFT typically closes with merging across task-specialist checkpoints (text rendering, photorealism, edit) rather than shipping a single fine-tuned model; and \emph{(iv)} RL has settled on a DPO + GRPO combination as its outer wrapper, even as the specific algorithmic variant within RL continues to fragment (discussed in the RL snapshot below). The takeaway: a research lab joining the frontier in 2026 no longer needs to discover a training recipe---it needs to execute the converged one well, with differentiation now confined to caption quality, CT data filters, RL variants, and merging recipes rather than to the macro skeleton itself.
\end{highlightbox}

\subsection{Post-training}
\label{subsec:post_training}

Post-training transforms a broad pretrained generator into a system that better reflects human intent, preference, and deployment requirements. Within the four-stage training pipeline summarized in \Cref{fig:training_pipeline}, post-training comprises the final two stages. \textbf{Stage I} uses supervised fine-tuning to bootstrap instruction following from curated high-quality data. \textbf{Stage II} uses preference-based optimization and reward modeling to move beyond the ceiling of supervised imitation. The remainder of this subsection follows that two-stage organization.

\subsubsection{Supervised Fine-Tuning}
\label{subsec:sft}

Stage~I uses \textbf{supervised fine-tuning (SFT)} to convert a broad pretrained generator into a functional, user-aligned system capable of interpreting nuanced human intent. While pre-training provides the visual vocabulary and semantic associations learned from noisy web-scale data, SFT establishes the baseline for instruction following, complex text rendering, and high-fidelity image editing. We organize the discussion around four recurring issues: the fidelity--diversity trade-off, data quality and curation, architecture-specific recipes, and the limits that motivate preference-based alignment.

\paragraph{Mode-Seeking Trade-Off.}
{Unlike NLP SFT, which targets a single ``correct'' response format, image generation faces an inherently multimodal target: thousands of valid images can satisfy the same prompt. The standard SFT objective minimizes a reverse KL divergence $D_{\mathrm{KL}}(q\|p)$ that is intrinsically \emph{mode-seeking}---it heavily penalizes mass placed where the target has none, but tolerates ignoring existing modes. As SFT deepens, output diversity therefore collapses toward the most frequent training patterns, producing sharper but more aesthetically conventional images. Industrial pipelines mitigate this primarily through model merging across specialized SFT runs~\citep{cai2025z,team2025longcat}---linear weight interpolation that recovers Pareto-optimal capability balance---rather than through algorithmic regularization at the loss level.}

\paragraph{Data Curation.}

The prevailing consensus is that the quality ceiling of SFT is set by the training data rather than the model's parameter count. This has led to the development of sophisticated data profiling and refinement engines.

\subparagraph{VLM Relabeling.}
Web-crawled datasets (e.g., LAION, DataComp) often contain captions that are uninformative, mismatched, or grammatically poor, limiting the model's ability to learn complex semantic bindings. To address this, industrial systems such as DALL-E~3 and Recap-DataComp-1B employ a ``recaptioning'' pipeline: a powerful VLM---often GPT-4V or a high-capacity open model like Qwen2.5-VL---generates dense, descriptive captions for every image. These enhanced captions provide the diffusion model with a much richer signal regarding object properties, spatial arrangements, and stylistic nuances, enabling the model to follow complex multi-attribute queries that would baffle a model trained on raw web data.

\subparagraph{Multi-Stage Filtering and Curation.}
Modern SFT pipelines, exemplified by Qwen-Image~\citep{wu2025qwen} and Z-Image~\citep{cai2025z}, typically adopt a three-phase data refinement strategy:
\begin{itemize}[leftmargin=*,itemsep=2pt]
    \item \textbf{Phase 1 --- Alignment} ($>$100M samples): Large-scale filtered web data for general semantic mapping.
    \item \textbf{Phase 2 --- Refinement} (10M--50M): VLM-relabeled data to improve instruction adherence and attribute binding.
    \item \textbf{Phase 3 --- Specialization} (100K--1M): Expert-curated data targeting high-value capabilities, including professional typography (bilingual text rendering), identity preservation (same person across poses/outfits), and geometric reasoning (perspective, vanishing points, structural coherence).
\end{itemize}
The drastic reduction in volume from Phase~1 to Phase~3 reflects a key insight: \emph{a small amount of expert-quality data in the final stage has disproportionate impact} on the capabilities that users notice most. For instruction-based editing tasks, a similar data-centric principle applies: InstructPix2Pix~\citep{brooks2023instructpix2pix} bootstrapped editing capability from 450K synthetically generated triplets, while SEED-Data-Edit~\citep{ge2024seed} demonstrated that incorporating 52K expert-edited real photographs and 21K multi-turn sequences substantially improves identity preservation and sequential editing consistency (see \Cref{sec:applications} for a detailed discussion of editing methods).

\paragraph{Architecture-Specific Recipes.}

The choice of underlying architecture profoundly influences SFT methodology. While diffusion-based models remain the dominant paradigm, alternatives are emerging as strong contenders for specific tasks:
\begin{itemize}[leftmargin=*,itemsep=2pt]
    \item \textbf{MMDiT (Qwen-Image):} A 20B-parameter Multi-Modal Diffusion Transformer with dual encoding---separate semantic (VLM) and reconstructive (VAE) paths---enabling a balance between visual fidelity preservation and semantic edit execution during SFT.
    \item \textbf{S3-DiT (Z-Image):} A 6B-parameter Scalable Single-Stream DiT that concatenates text, semantic, and image VAE tokens into a unified sequence. Its broader pipeline pairs SFT with a separate Decoupled-DMD distillation stage that produces the 8-step Z-Image-Turbo variant (see \Cref{subsec:distillation}).
    \item \textbf{Native Multimodal MoE (HunyuanImage~3.0~\citep{cao2025hunyuanimage}):} A scale-first alternative that unifies language modeling, image understanding, and image generation within a sparse MoE backbone. Its training pipeline combines progressive pre-training with multimodal reasoning traces, illustrating that SFT is no longer only about prompt following but also about preserving cross-modal competence under a shared parameter budget.
    \item \textbf{Compact Bilingual DiT (LongCat-Image~\citep{team2025longcat}):} A 6B-class design that prioritizes Chinese--English rendering, photorealism, and full-stack practicality. The key lesson here is that a compact model paired with strong staged curation and later DPO can remain competitive without resorting to extreme parameter counts.
    \item \textbf{Editing-First Diffusion Transformer (FireRed-Image-Edit-1.0~\citep{team2026firered}):} Instead of adapting a generic text-to-image backbone after the fact, FireRed treats editing as the primary objective. Its training corpus is aggressively filtered from a much larger pool into balanced generation/editing data, and later post-training is tailored to instruction following, identity preservation, and multi-condition editing fidelity.
    \item \textbf{Spatial-Intelligence-First Unified Stack (JoyAI-Image~\citep{jdjoyaiimage}):} A unified framework that combines a spatially enhanced Qwen3-VL-8B-Instruct interface, Wan-2.1-VAE, and a 16B MMDiT. Its progressive recipe---first strengthening multimodal spatial understanding, then training high-fidelity generation with MLLM-derived priors, and finally optimizing precise instruction-based editing---shows how SFT can be used to explicitly propagate understanding-side gains into generation and editing.
    \item \textbf{Masked Generative Transformers:} Models like EditMGT shift toward localized decoding---predicting masked tokens rather than performing full diffusion denoising---which inherently preserves non-relevant regions during editing, enabling faster SFT convergence on localized edit datasets.
    \item \textbf{Reasoning- and Agent-Trajectory Models:} The most recent innovation supervises not only prompt-to-image pairs but also intermediate reasoning and action traces. ReasonGen-R1~\citep{zhang2025reasongen} uses rationale-annotated SFT data for deliberate image generation. GEMS~\citep{he2026gems} adds planner/decomposer/verifier/refiner trajectories together with memory and skill selection, while Gen-Searcher~\citep{feng2026gen} trains on search-intensive traces that teach the model when and how to retrieve external knowledge before rendering. In editing, JarvisArt~\citep{lin2025jarvisart} pairs CoT annotations with executable Lightroom operation records, showing that SFT can teach an agent not just what image to output, but how to plan and act inside a closed loop. Collectively, these recipes push training toward L4 (Agentic Generation) by supervising planning, retrieval, verification, and tool use as first-class targets.
\end{itemize}

Taken together, these examples show that SFT is increasingly architecture-dependent. The question is no longer simply how to fine-tune a generic generator, but what kind of generator one is fine-tuning in the first place: a dense DiT, a single-stream compact model, a sparse MoE, an editing-first system, or a reasoning-augmented autoregressive hybrid. Different backbones expose different bottlenecks, and consequently require different SFT recipes.

\paragraph{Limits of SFT.}

Despite the significant advancements described above, SFT alone is insufficient to reach the peak of human preference. As models undergo intensive SFT, they eventually reach a performance plateau where adding more labeled data fails to meaningfully improve visual quality or instruction adherence. Furthermore, SFT acts primarily as an ``imitator'' of the training distribution: it lacks the capacity to optimize for global, subjective objectives such as aesthetic appeal, emotional resonance, or the subtle quality trade-offs that distinguish ``good'' from ``excellent'' images. Two images may both follow the instruction correctly but differ in spatial coherence, lighting plausibility, or stylistic preference---distinctions that supervised labels cannot capture. This limitation motivates the transition from supervised learning to the preference-based optimization techniques discussed in the next subsection.

\subsubsection{Reinforcement Learning}
\label{subsubsec:pref_alignment}

Beyond supervised imitation, the central question in diffusion alignment is not simply whether one should add an external reward, but how supervision should enter a generation process unfolded over many denoising steps. Should alignment optimize a terminal image-level score, backpropagate reward gradients through the denoising chain, or learn directly from pairwise preferences? Over the last two years, the field has progressively sharpened this question from generic reward maximization to process-aware credit assignment. \Cref{tab:posttrain_summary} summarizes the main paradigms discussed below.

This issue is especially pronounced for \emph{unified understanding-and-generation models}. Once a model shares parameters across multimodal understanding and image synthesis, post-training must do more than improve aesthetics: it must rebalance instruction following, reasoning, and rendering quality under a shared parameter budget. HermesFlow~\citep{yang2025hermesflow}, UAE~\citep{yan2025uae}, MMaDA~\citep{yang2025mmada}, X-Omni~\citep{geng2025xomni}, and the instruction-following stage of BLIP3o-NEXT~\citep{chen2025blip3onext} all exemplify this shift from \emph{generative quality alignment} to \emph{capability balancing} in multimodal generators.

\begin{table*}[t]
\centering
\caption{\textbf{Post-training alignment paradigms for visual generation.} The field has evolved from terminal-reward RL to direct preference optimization and finer-grained credit assignment, while reward models provide the supervisory interface consumed by these optimization schemes.}
\label{tab:posttrain_summary}
\footnotesize
\setlength{\tabcolsep}{4pt}
\renewcommand{\arraystretch}{1.16}
\begin{tabularx}{\textwidth}{@{}>{\raggedright\arraybackslash}p{3.2cm}>{\raggedright\arraybackslash}p{3.4cm}>{\raggedright\arraybackslash}X@{}}
\toprule
\textbf{Alignment paradigm} & \textbf{Representative works} & \textbf{Core mechanism} \\
\midrule
Policy-gradient RL & DDPO, DPOK & Treats reverse denoising as an MDP and optimizes terminal reward with policy updates under distribution constraints \\
Reward Backpropagation & AlignProp & Backpropagates reward gradients through the denoising trajectory, reducing policy-gradient variance but increasing memory cost \\
Direct Preference Optimization & Diffusion-DPO, VideoDPO & Learns directly from preferred-versus-rejected outputs instead of first fitting an explicit reward model \\
Group-Relative Preference Optimization & Flow-GRPO, UniGRPO, Gen-Searcher & Uses online group sampling and relative advantages to optimize preference-aligned outputs through exploration \\
Dense Reward / Credit Assignment & Dense Reward View & Distributes supervisory signal across denoising steps instead of treating the full trajectory as a black-box terminal outcome \\
Reward Modeling & HPSv3, MPS, VisionReward, VIEScore, EditReward & Supplies discriminative, generative, or task-specific supervisory signals for both generation and editing \\
\bottomrule
\end{tabularx}
\end{table*}

\paragraph{Common Formulation.}
We refer readers to \Cref{sec:model_arch} for the mathematical formulations of diffusion and flow matching objectives. For post-training, the key abstraction is to view the reverse denoising process as a Markov Decision Process (MDP) $(\mathcal{S}, \mathcal{A}, P, R)$: the state $\vs_t$ includes the current noisy image $\vx_t$ and the conditioning prompt $\vc$, the action $\va_t$ is one denoising step, and the supervisory signal is often observed only after the final image $\vx_0$ is produced. A generic alignment objective can therefore be written as
\begin{equation}
    \max_{\pi_\theta} \mathbb{E}_{\vx \sim \pi_\theta} [r(\vc, \vx)] - \beta \mathbb{D}_{\text{KL}}(\pi_\theta || \pi_{\text{ref}}).
\end{equation}
This formulation makes the central bottleneck explicit: the supervision is usually terminal, but responsibility for image quality is distributed across a long denoising chain. The remaining question is therefore not only how to optimize the objective, but how to assign credit along the trajectory.

\paragraph{RL-Based Alignment: DDPO, DPOK, and AlignProp.}
DDPO~\citep{black2023training} was the first method to systematically cast reverse diffusion as a sequential decision problem and optimize the final reward directly, opening a post-training route for diffusion models analogous to RLHF in language models. Its main contribution is conceptual: once denoising is treated as a policy, any computable target---aesthetics, prompt alignment, compressibility, or task-specific utility---can in principle become a reward. However, DDPO also inherits the classical weaknesses of policy-gradient RL: expensive online sampling, high-variance updates, and coarse terminal supervision over a long chain of denoising steps.

DPOK~\citep{fan2023dpok} retains this RL framing but focuses on making text-to-image fine-tuning more stable. Its key lesson is that alignment should be viewed as a constrained distribution shift rather than brute-force reward maximization: KL regularization and careful online updates are needed to raise the target reward without destroying base-model quality or diversity. AlignProp~\citep{prabhudesai2023alignprop} explores the opposite repair. Instead of tolerating noisy policy gradients, it backpropagates reward gradients directly through the denoising trajectory, making use of diffusion's differentiable structure. This substantially reduces variance but increases memory and implementation cost; the original arXiv manuscript was later withdrawn after the authors noted that its content had been absorbed by subsequent work. Together, these methods define the first phase of diffusion alignment: once reverse diffusion is recognized as a long-horizon process, the main challenge becomes how to inject external supervision into that process stably.

\paragraph{Preference Optimization: Diffusion-DPO and GRPO.}
A second phase shifts the problem from \emph{how to maximize a scalar reward} to \emph{how to learn from relative preferences}. This transition is especially natural in image generation, where human judgments are often more reliable as pairwise comparisons than as absolute scores. Diffusion-DPO~\citep{wallace2024diffusion} adapts direct preference optimization to diffusion models by replacing explicit reward maximization with a preference loss over winner-loser pairs $\mathcal{D} = \{(\vc, \vx_w, \vx_l)\}$:
\begin{equation}
    \mathcal{L}_{\text{DPO}} = - \mathbb{E}_{(\vx_w, \vx_l) \sim \mathcal{D}} \left[ \log \sigma \left( \beta \log \frac{\pi_\theta(\vx_w)}{\pi_{\text{ref}}(\vx_w)} - \beta \log \frac{\pi_\theta(\vx_l)}{\pi_{\text{ref}}(\vx_l)} \right) \right].
\end{equation}
The conceptual shift is important: alignment no longer depends on first fitting a separate reward model, but instead updates the generator directly to increase the relative likelihood of preferred outputs. Recent variants such as \textit{VideoDPO}~\citep{liu2025videodpo} extend this logic from single images to trajectory-level preferences. In unified settings, DPO-style post-training is increasingly used to rebalance understanding and generation rather than merely boost aesthetics. HermesFlow~\citep{yang2025hermesflow}, for instance, introduces Pair-DPO with homologous preference pairs plus self-play iterative optimization so that a shared model can improve both tasks together.

GRPO should be understood as the online counterpart of this preference-based view rather than as a disconnected topic. Instead of optimizing fixed offline pairs, it samples a group of $G$ outputs $\{\vx_0^i\}_{i=1}^G$ for the same prompt $\vc$ and computes a relative advantage within the group:
\begin{equation}
\hat{A}^i = \frac{R(\vx^i_0, \vc) - \mathbb{E}[R]_{\text{group}}}{\text{std}(R)_{\text{group}}}.
\end{equation}
The policy is then updated via a clipped objective similar to PPO but without a value function:
\begin{equation}
\mathcal{J}_{\text{GRPO}}(\theta) = \mathbb{E}_{\{\vx^i\}} \left[ \frac{1}{G}\sum_{i=1}^{G} \left( \min \left( r^i(\theta) \hat{A}^i, \text{clip}(r^i(\theta), 1-\epsilon, 1+\epsilon)\hat{A}^i \right) - \beta \mathbb{D}_{\text{KL}} \right) \right].
\end{equation}
In this sense, DPO and GRPO are best regarded as two regimes within the same preference-alignment family: DPO is \emph{offline pairwise preference optimization}, whereas GRPO is \emph{online group-relative preference optimization}. For flow-based generators, however, GRPO requires stochastic exploration even though standard Flow Matching relies on deterministic ODE sampling. Methods such as Flow-GRPO~\citep{liu2025flow} and DanceGRPO~\citep{xue2025dancegrpo} therefore convert the ODE into a stochastic SDE that preserves the marginal distribution:
\begin{equation}
    \mathrm{d} \vx_t = \left[\vv_\theta(\vx_t) + \frac{\sigma_t^2}{2t}\left(\vx_t + (1-t)\vv_\theta(\vx_t)\right)\right] \mathrm{d} t + \sigma_t \mathrm{d} \vw.
\end{equation}
This enables online exploration while retaining computational efficiency through strategies such as denoising reduction (e.g., using $T \approx 10$ steps). The same group-relative logic is now being adapted to unified multimodal generators. UAE~\citep{yan2025uae} introduces Unified-GRPO for staged encoder-decoder co-optimization; MMaDA~\citep{yang2025mmada} applies UniGRPO to a unified diffusion backbone; X-Omni~\citep{geng2025xomni} uses reinforcement learning to reduce discrete-token artifacts in autoregressive image generation; and BLIP3o-NEXT~\citep{chen2025blip3onext} refines its AR-to-diffusion cascade with GRPO-based alignment. Agentic systems make this shift even more explicit: Gen-Searcher~\citep{feng2026gen} combines SFT with agentic RL and dual text/image rewards to train a search policy, while JarvisArt~\citep{lin2025jarvisart} introduces GRPO-R for multimodal photo retouching with reasoning traces, tool selection, and executable editing actions. Collectively, these works show that GRPO is evolving from a sampler-level optimization tool into a general recipe for online preference alignment in unified generators.

\paragraph{GRPO Variants in Frontier Tech Reports.}
{The same online preference-alignment framing has produced a rich zoo of GRPO variants in frontier industrial systems within a single year, each addressing a different bottleneck of running RL on flow-matching or autoregressive image generators. \textit{MixGRPO} and \textit{SRPO}~\citep{cao2025hunyuanimage}, used in HunyuanImage~3.0, trade per-step granularity for stability when applied to MoE backbones with sparse activation. \textit{MPO}~\citep{team2025longcat}, introduced in the LongCat family, eliminates the group-synchronization bottleneck that limits training throughput when GRPO is applied at large scale. \textit{DenseGRPO}~\citep{mao2026wan} densifies credit assignment along the denoising trajectory in Wan-Image, addressing the same terminal-reward limitation discussed below. \textit{DiffusionNFT}~\citep{team2026firered,jdjoyaiimage}, used in FireRed-Image-Edit and JoyAI-Image, drops the requirement for paired preferences and allows alignment from non-paired reward signals such as scalar quality scores. \textit{Flow-GRPO}~\citep{wu2025qwen,jdjoyaiimage} adapts the GRPO update rule to flow-matching's deterministic ODE structure---used by Qwen-Image and JoyAI-Image---and \textit{ReFMA}~\citep{mao2026wan} pushes this adaptation further by integrating reward modeling into the flow-matching loss itself. \textit{LongCat-Next}~\citep{team2026longcat} argues a more philosophical point: discrete autoregressive generation is a natural finite Markov decision process, so GRPO applied to discrete tokens is structurally cleaner than GRPO applied to continuous-time flows. The proliferation of variants reflects three live research axes---process-aware credit assignment, group synchronization under sparse activation, and supervision without paired preferences---rather than a settled algorithm.}

\paragraph{Credit Assignment Beyond Terminal Rewards.}
Even after moving from scalar rewards to relative preferences, a deeper issue remains: if the entire reverse diffusion chain is treated as a black box, the supervisory signal is still attached only to the final image. \textit{A Dense Reward View on Aligning Text-to-Image Diffusion with Preference}~\citep{yang2024dense} argues that this is too coarse. Early denoising steps largely determine global semantics and layout, whereas later steps refine local details; preference supervision should therefore be distributed along the trajectory rather than applied as a uniform whole-trajectory label. This perspective reframes diffusion alignment as a process-level credit-assignment problem rather than a simple choice between RL and DPO. In that sense, the main line from DDPO to recent work is not merely ``from RL to DPO,'' but a steady move toward supervision that is more aware of the time structure of denoising.

\paragraph{Reward Models as Supervisory Interfaces.}
The optimization schemes above differ in how they use supervision, but they all depend on a supervisory interface. Discriminative reward models such as HPSv3~\citep{ma2025hpsv3}, MPS~\citep{zhang2024learning}, and VisionReward~\citep{xu2024visionreward} predict holistic or decomposed scores for aesthetics, semantic fidelity, and safety. Generative or VLM-as-a-judge approaches such as VIEScore~\citep{ku2024viescore}, UniPic 2.0~\citep{wei2025skywork}, RewardDance~\citep{wu2025rewarddance}, and OneReward~\citep{gong2025onereward} instead derive supervision from reasoning traces or token probabilities of large multimodal models. Editing introduces an additional difficulty because the scorer must verify \emph{cross-image consistency}---whether the requested region changed correctly while the rest of the image was preserved. Specialized models such as EditReward~\citep{wu2025editreward}, EditScore~\citep{luo2025editscore}, and the reward decomposition used in JarvisArt~\citep{lin2025jarvisart} therefore evaluate instruction following, preservation, and executable tool correctness jointly.

Despite this progress, current reward interfaces still struggle with explicit spatial grounding. Many rely on global feature matching or final-answer judgments, which can obscure which regions and which denoising steps are responsible for success or failure. This is precisely why denser credit assignment and process-aware supervision are becoming central themes in recent diffusion alignment research.

\begin{highlightbox}{Where the Field Is Splitting: RL Algorithms Are Fragmenting Fast}

RL has become a non-negotiable stage of frontier post-training, but the \emph{specific} algorithm has \emph{not} converged: in a single year (2025--2026), at least seven distinct GRPO variants have appeared, each solving a different facet of the ``how to do online RL stably on a flow-matching or AR image generator'' problem. \textbf{MixGRPO} and \textbf{SRPO} (HunyuanImage~3.0) trade granularity for stability; \textbf{MPO} (LongCat) tackles the group-synchronization bottleneck that hurts MoE training; \textbf{DenseGRPO} (Wan-Image) densifies credit assignment along the denoising trajectory; \textbf{DiffusionNFT} (FireRed / JoyAI) drops the requirement for paired preferences entirely; \textbf{Flow-GRPO} (Qwen / JoyAI) and \textbf{ReFMA} (Wan) adapt the GRPO update rule to flow-matching's ODE structure. The pattern is consistent: process-aware credit assignment, group synchronization, and pairing-free supervision are the three live axes, and the lab that nails their combination first will set the next-year baseline.
\end{highlightbox}

\subsection{Inference Acceleration}
\label{subsec:acceleration}

Once model quality has been established through pre-training and post-training, inference latency becomes the main deployment bottleneck. Real-time applications such as interactive editing, agentic visual generation (\Cref{subsec:agentic_visual_generation}), and embodied simulation (\Cref{subsec:level5}) all require generation speeds that standard multi-step diffusion cannot provide. We therefore organize acceleration into three complementary directions. First, \textbf{sampling acceleration} reduces the number of denoising steps through better ODE/SDE solvers. Second, \textbf{per-step efficiency} reduces the computational cost of each denoising evaluation through pruning and feature caching. Third, \textbf{distillation} learns new student generators that collapse many steps into a few learned shortcuts. \Cref{tab:acceleration_summary} provides an overview.

This deployment-first focus is visible across recent industrial systems. Seedream~3.0~\citep{gao2025seedream} explicitly co-designs training and acceleration, coupling a stronger base model with faster inference schemes. LongCat-Image-Edit-Turbo~\citep{team2025longcat} pushes the open-source editing stack toward 8-step generation with roughly one order of magnitude speedup over its base editor. FireRed-Image-Edit~\citep{team2026firered} similarly emphasizes optimized inference and editing efficiency as first-class product constraints rather than downstream engineering details. JoyAI-Image~\citep{jdjoyaiimage}, while less centered on a standalone turbo variant, highlights a different deployment constraint: if spatial intelligence is a core product target, acceleration cannot be optimized independently from the understanding backbone that generates the conditioning signal. The broader message is that acceleration is no longer a secondary optimization layer; it is part of the model design target itself.

\begin{table*}[t]
\centering
\caption{\textbf{Overview of sampling acceleration and efficiency techniques.} The first four rows address the NFE reduction problem (how many steps are needed); the bottom two rows address per-step cost reduction (how expensive each step is). Both are critical for deployment.}
\label{tab:acceleration_summary}
\scriptsize
\setlength{\tabcolsep}{4pt}
\renewcommand{\arraystretch}{1.2}
\begin{tabular}{@{}l l >{\raggedright\arraybackslash}p{7.3cm}@{}}
\toprule
\textbf{Strategy} & \textbf{Representative Approaches} & \textbf{Key Insight} \\
\midrule
ODE/SDE Solvers & DDIM, DPM-Solver, DPM-Solver++, Restart & Reduces NFEs via specialized numerical integrators that exploit trajectory structure \\
Trajectory Distillation & Progressive Distill., Shortcut Models & Learns to collapse many steps into few by matching the teacher's ODE trajectory \\
Consistency Models & CM, sCM, rCM & Maps any point on the trajectory directly to clean data via self-consistency constraint \\
Distribution Matching & DMD/DMD2, ADD, SiD & Aligns student and teacher distributions via adversarial or variational objectives; on-policy exploration \\
\midrule
\multicolumn{3}{@{}l}{\textit{Orthogonal efficiency techniques (see \S\ref{subsec:efficiency}):}} \\
Pruning \& Compression & Diff-Pruning, LD-Pruner, LayerMerge & Removes structural redundancy using diffusion-aware importance metrics \\
Inference Caching & DeepCache, ToCa, TaylorSeer, TAP & Bypasses redundant compute by reusing or predicting temporal features \\
\bottomrule
\end{tabular}
\end{table*}

\subsubsection{Sampling Acceleration}
\label{subsec:sampling}

\paragraph{Diffusion Dynamics.}
Diffusion generative models operate by inverting a progressive noising process. In the discrete setting this forward chain is typically written as:
\begin{equation}
\label{eq:forward_process}
q(\mathbf{x}_{1:T}\mid\mathbf{x}_0)=\prod_{t=1}^T q(\mathbf{x}_t\mid\mathbf{x}_{t-1}),\qquad
q(\mathbf{x}_t\mid\mathbf{x}_{t-1})=\mathcal{N}\!\big(\mathbf{x}_t;\sqrt{\alpha_t}\mathbf{x}_{t-1},(1-\alpha_t)\mathbf{I}\big),
\end{equation}
and generation requires applying a learned denoiser $\epsilon_\theta(\mathbf{x},t)$ repeatedly. Viewing diffusion in continuous time leads to two related formulations that underpin most fast-sampling work: the score-driven stochastic differential equation (SDE):
\begin{equation}
\mathrm{d}\mathbf{x}=f(\mathbf{x},t)\,\mathrm{d}t + g(t)\,\mathrm{d}\mathbf{w}_t,
\end{equation}
and its deterministic probability-flow ODE \citep{songDDIM}:
\begin{equation}
\frac{\mathrm{d}\mathbf{x}}{\mathrm{d}t}=f(\mathbf{x},t)-\tfrac{1}{2}g(t)^2\,\nabla_{\mathbf{x}}\log p_t(\mathbf{x}).
\label{eq:probflow}
\end{equation}

\paragraph{Numerical Solvers and Distillation.}
Fast-sampling research reduces the number of function evaluations (NFEs) required to traverse these dynamics while preserving sample fidelity. One family of approaches designs specialized numerical integrators that exploit structure in (\ref{eq:probflow}): DDIM recasts discrete sampling as an ODE-like deterministic procedure \citep{songDDIM}, and later integrators (DPM-Solver and follow-ups) analytically integrate linear parts while approximating nonlinear components to achieve high-order convergence with very few steps \citep{lu2022dpm,lu2022dpm++,zheng2023dpmsolvervF,zhang2022fast}. Complementary work improves stochastic integrators (e.g., multistep Adams-style or variance-controlled SDE solvers) to reuse past evaluations and better manage injected noise \citep{xue2024sa,guo2024gaussian,xu2023restart}. Distillation-based methods (progressive distillation and successors) take a different route: they learn to ``collapse'' many denoising steps into a small number of learned super-steps, trading extra training for drastically reduced inference NFEs \citep{salimans2022progressive,meng2022on,yin2024improved}. In practice, solver improvements are attractive for being training-free and applicable to off-the-shelf models, while distillation yields very compact samplers when additional training is acceptable. Hybrid strategies (adaptive stepping, solver-aware distillation, and stochastic–deterministic schedules) seek to combine the robustness of stochastic samplers with the efficiency of tailored ODE integrators \citep{jolicoeur2021gotta,liu2022pseudo,karras2022elucidating}.

\subsubsection{Per-Step Cost Reduction}
\label{subsec:efficiency}

{While \Cref{subsec:sampling} addresses \emph{how many steps} are needed, this subsection tackles the complementary question of \emph{how expensive each step is}. The two axes are orthogonal and multiplicatively compose: halving NFEs via distillation and halving per-step cost via pruning yields a $4\times$ speedup. We organize per-step acceleration into two pillars discussed below: structural pruning and compression, and inference-time feature caching.}

\paragraph{Structural Pruning.}
\label{subsubsec:pruning}

\subparagraph{Diffusion-Aware Pruning.}
Pruning and related compression methods target the per-step cost of the denoiser $\epsilon_\theta$, and hence multiply the wall-clock gains achieved by any timestep reduction. Unlike typical classification or generation models, diffusion denoisers must operate across many noise levels, so pruning criteria that are \emph{diffusion-aware} perform better: for example, Diff-Pruning estimates parameter importance by Taylor expansions aggregated over timesteps to account for temporal contributions to the diffusion loss \citep{fang2023structural}. For latent diffusion architectures, evaluating the effect of removing operators in the model's latent manifold (LD-Pruner) yields pruning decisions that better preserve perceptual quality \citep{castells2024ld}.

\subparagraph{Structured Compression.}
Practical structured strategies remove channels, blocks, or entire layers to produce hardware-efficient kernels, while algorithmic formulations such as TinyFusion, LayerMerge and LAPTOPDiff cast layer selection as a constrained optimization or one-shot additive pruning problem to meet latency, FLOP budgets with predictable impact \citep{fang2025tinyfusion, kim2024layermerge,zhang2024laptop}. Pruning is frequently combined with orthogonal techniques, e.g., post-training quantization, low-rank factorization, token and attention reduction, to further reduce memory and compute~\citep{10377259,shang2023post,kim2025ditto,bolya2023tomesd,kim2024tofu,yuan2024ditfastattn}. Important open challenges include timestep-conditional sparsity (parameters that are active only at certain noise levels), understanding how pruning perturbations affect solver stability at low NFEs, and jointly optimizing pruning and sampler schedules so that per-step compression and step reduction are co-designed rather than applied independently.

\paragraph{Feature Caching.}
\label{subsubsec:caching}

\subparagraph{Spatial-Temporal Redundancy.}
Feature caching at inference leverages temporal redundancy in the denoising sequence to avoid re-computing stable intermediate activations. Let \(\mathbf{x}_t\) denote the latent at diffusion timestep \(t\) and write a typical denoiser as a composition
\begin{equation}
\epsilon_{\theta}(\mathbf{x}_t,t) \;=\; \Psi_\theta\big(\Phi(\mathbf{x}_t,t)\big),
\label{eq:denoiser-decomp}
\end{equation}
where \(\Phi(\cdot,t)\) extracts expensive intermediate features (e.g. transformer token embeddings or convolutional feature maps) and \(\Psi_\theta(\cdot)\) denotes the remaining, relatively cheaper head that maps features to the noise prediction. Caching-based acceleration targets \(\Phi\): if \(\Phi(\mathbf{x}_t,t)\) varies slowly across adjacent timesteps, one can reuse previously computed features \(\Phi(\mathbf{x}_{t-\Delta},t-\Delta)\) to form a surrogate \(\tilde{\Phi}(\mathbf{x}_t,t)\) and compute
\[
\tilde{\epsilon}_\theta(\mathbf{x}_t,t) \;=\; \Psi_\theta\big(\tilde{\Phi}(\mathbf{x}_t,t)\big),
\]
thereby saving the cost of re-evaluating \(\Phi\). Early reuse-based systems detect and exploit spatial–temporal redundancy: DeepCache and related designs cache convolutional feature maps and reuse them when visual content changes little between timesteps \citep{ma2024deepcache}. 
However, DeepCache uses a hand-crafted cache scheduling strategy to control the intervals between consecutive full computation steps, which is often suboptimal. 
DiP-GO and L2C learn layer-wise caching patterns in a differentiable manner, using lightweight predictors to determine which layers can be skipped \citep{zhu2024dip, ma2024learning}. Token-level caching methods (e.g., ToCa and DiTFastAttn) refine caching methods by operating at finer granularity, caching per-token activations and selectively recomputing only those tokens whose features exceed a change threshold \citep{chen2024delta-dit,liu2024timestep,zou2024accelerating, yuan2024ditfastattn}. 
Formally, with a binary reuse mask \(m_{i,t}\in\{0,1\}\) for token (or spatial location) \(i\),
\begin{equation}
\tilde{\Phi}_i(\mathbf{x}_t,t) \;=\; m_{i,t}\,\Phi_i(\mathbf{x}_{t-\Delta},t-\Delta)\;+\;(1-m_{i,t})\,\Phi_i(\mathbf{x}_t,t),
\end{equation}
and only indices with \(m_{i,t}=0\) incur the full \(\Phi\)-compute.
However, they use a fixed sparsity rate per layer. DiffSparse further improves on this by learning per-layer per-step sparsity patterns differentiably \citep{zhu2026diffsparse}.

\subparagraph{Feature Forecasting.}
Forecasting-based methods aim to go beyond retrieval: instead of directly reusing cached features, they predict the feature at a future timestep. TaylorSeer proposes a Taylor-like expansion to predict features in (continuous) time:
\begin{equation}
\Phi(\mathbf{x}_{t+\Delta},t+\Delta) \approx \Phi(\mathbf{x}_t,t) + \Delta\,\partial_t\Phi(\mathbf{x}_t,t) + \tfrac{1}{2}\Delta^2\partial_t^2\Phi(\mathbf{x}_t,t) + \mathcal{O}(\Delta^3).
\label{eq:taylor}
\end{equation}
TaylorSeer estimates time-derivatives of features and uses low-order expansions for multi-step prediction.
FreqCa decomposes features into low- and high-frequency components and applies frequency-aware interpolation (e.g. Hermite-style interpolation) for the high-frequency part. SpeCa adopts a draft-then-verify workflow where a fast predictor proposes \(\hat{\Phi}\) and a lightweight verifier decides whether to accept or recompute \citep{liuReusingForecastingAccelerating2025,liu2025freqca,liu2025speca}. TAP employs a probe-then-predict strategy to assign the optimal predictor to each token during the sampling process \citep{zhu2026tap}.
In general a designed predictor \(P\) maps historical features to a future estimate
\begin{equation}
\hat{\Phi}(\mathbf{x}_{t+\Delta},t+\Delta) \;=\; P\big(\Phi(\mathbf{x}_{t},t), \Phi(\mathbf{x}_{t-\delta},t-\delta),\dots\big),
\label{eq:predictor}
\end{equation}
and the expected error is controlled by a tolerance \(\varepsilon\):
\[
\big\|\Phi(\mathbf{x}_{t+\Delta},t+\Delta) - \hat{\Phi}(\mathbf{x}_{t+\Delta},t+\Delta)\big\| \le \varepsilon.
\]

Reusing and forecasting methods trade computation for approximation error. Pure reuse deteriorates when the timestep gaps widen or when \(\mathbf{x}_t\) moves through regions with rapid feature dynamics. Forecasting mitigates this degradation but relies on the assumption that changes between consecutive timesteps are relatively smooth; otherwise, it can produce substantial prediction errors.
Draft-then-verify schemes mitigate risk by only accepting predictions under a confidence threshold, but the verifier itself adds overhead and may trigger full recomputation in edge cases \citep{liu2024timestep,chen2024delta-dit}. 
Practical systems therefore combine token-level adaptivity (to limit wasted recomputation), multi-step predictors (to bridge larger gaps), and verification with conservative acceptance criteria to balance speed and fidelity.

A remaining challenge is integrating caching with aggressive timestep reduction and pruning: predictors trained on dense schedules may fail under distilled samplers or pruned denoisers, and cache-induced approximation errors can interact with solver discretization errors in nontrivial ways. 

\subsubsection{Distillation for Few-Step Generation}
\label{subsec:distillation}

While the ODE/SDE solvers discussed in \Cref{subsec:sampling} reduce NFEs without additional training, distillation takes a fundamentally different approach: it \emph{trains} a student model to reproduce the teacher's output quality in dramatically fewer steps. This makes distillation the dominant strategy for achieving single-digit or even single-step generation---a capability of intense industrial interest. We elevate distillation to a standalone subsection to reflect its growing importance and the diversity of recent approaches.

\begin{figure}[!htbp]
    \centering
    \includegraphics[width=\textwidth]{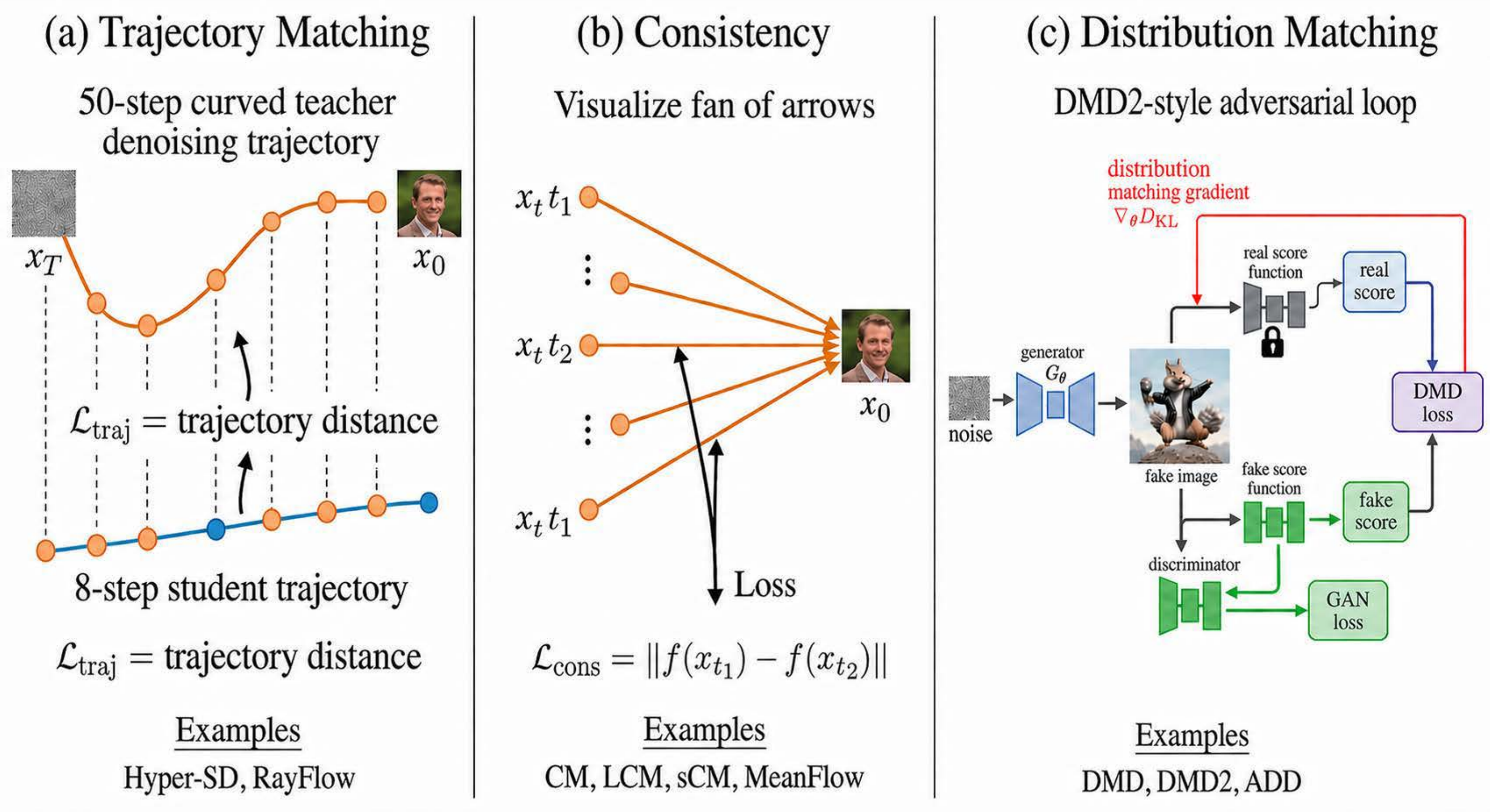}
    \caption{\textbf{Three families of distillation for diffusion / flow-matching models.} \textbf{(a) Trajectory Matching}: student learns to follow the teacher's per-step denoising trajectory at coarser granularity (e.g., 50-step teacher $\to$ 8-step student). Used in Hyper-SD, RayFlow, Wan-Image. \textbf{(b) Consistency}: any-step prediction maps directly to the same clean output, enforced by a self-consistency loss. Used in CM, sCM, rCM, LCM. \textbf{(c) Distribution Matching}: student's output distribution matches the teacher's via a fake/real score adversarial loop. Used in DMD, DMD2, ADD, MeanFlow.}
    \label{fig:distillation_paradigms}
\end{figure}

\paragraph{Objective Formulation.}
Diffusion models generate samples through iterative denoising, which poses computational challenges for real-time applications.
Distillation accelerates diffusion or flow image generation by \emph{learning a student sampling pipeline} whose behavior matches that of a teacher under substantially lower inference cost.
Let $f_{\theta}$ denote the teacher pipeline parameterized by pretrained weights $\theta$ and typically frozen during distillation, which maps a randomized noise latent $\mathbf{z}$ and condition $c$ to an image sample, and let $f_{\phi}$ be the student counterpart.
A generic distillation objective minimizes a discrepancy between teacher and student behaviors,
\begin{equation}
\min_{\phi}\;\mathbb{E}_{\mathbf{z},c}\Big[d\big(f_{\phi}(\mathbf{z},c),\,f_{\theta}(\mathbf{z},c)\big)\Big],
\label{eq:distill_generic}
\end{equation}
while enforcing a clear cost gap $\mathcal{C}(f_{\phi}) \ll \mathcal{C}(f_{\theta})$ at inference time.
Here $d(\cdot,\cdot)$ may compare final samples, intermediate states, or distributions depending on the method, and $\mathcal{C}(\cdot)$ abstracts the end-to-end inference cost, including the number of neural function evaluations (NFEs), per-step compute and memory footprint, and overhead from auxiliary modules such as guidance and decoding.

\paragraph{Matching Strategies.}
Existing methods fall into three categories. \textit{Trajectory matching} approaches learn mappings along the probability flow ODE, including Progressive Distillation~\citep{salimans2022progressive}, flow maps parameterized by average velocity~\citep{geng2025mean}, and Shortcut Models~\citep{frans2024one}. \textit{Consistency models}~\citep{song2023consistency} map any point on the trajectory directly to clean data through the self-consistency constraint $f_\theta(x_t, t) = f_\theta(x_{t'}, t')$. The continuous-time extension sCM~\citep{lu2024simplifying} uses a tractable objective with Jacobian-vector products, while rCM~\citep{zheng2025rcm} incorporates score distillation as regularization to improve quality at scale. \textit{Distribution matching} methods align student and teacher distributions through adversarial training~\citep{sauer2024adversarial} or variational score distillation~\citep{yin2024one,yin2024improved}. For video generation, TMD~\citep{nie2026tmd} introduces a decoupled architecture with a main backbone and flow head to enable efficient few-step sampling. Key challenges include scaling JVP computation to 10B+ parameter models~\citep{zheng2025rcm}, error accumulation in consistency-based methods, and balancing quality with diversity in distribution matching approaches.

\paragraph{Frontier System Recipes.}
{Industrial systems realize the three matching strategies above through specific recipes that reflect their broader architectural choices. Seedream~3.0~\citep{gao2025seedream} adopts a trajectory-distillation route based on Hyper-SD and RayFlow, achieving a $4$--$8\times$ speedup over the multi-step teacher; Seedream~4.0~\citep{seedream2025seedream} compounds this with adversarial distillation post-training (ADP and ADM) plus 4/8-bit hybrid quantization and speculative decoding for a $>10\times$ FLOP reduction at 2K output. Z-Image's Decoupled DMD~\citep{cai2025z} sits in the distribution-matching family and produces the 8-step Z-Image-Turbo variant; the same paper's DMDR extension folds reinforcement learning directly into the distillation objective rather than treating alignment and acceleration as separate stages. HunyuanImage~3.0~\citep{cao2025hunyuanimage} uses MeanFlow distillation to bring its 80B MoE backbone down to 4--8 NFE, illustrating that distribution-style distillation extends to sparse architectures. Wan-Image~\citep{mao2026wan} combines trajectory-style and DMD-style objectives in a hybrid that achieves more than $8\times$ acceleration. The common thread is that distillation has become a co-trained primary objective rather than a post-hoc compression step: flow-matching is partly preferred over legacy diffusion because its rectified-flow ODE distills more cleanly into few-step students.}

\begin{highlightbox}{Convergence Snapshot: Distillation Has Become Mandatory, Not Optional}

For any production-grade visual generation system, distillation is no longer an optional acceleration afterthought---it is a \emph{required} stage of the training pipeline. Sub-second 2K generation has become the 2025--2026 deployment target reported by leading commercial systems, and \emph{six of the ten frontier tech reports surveyed} document an explicit distillation phase with detailed recipes: trajectory-style \textbf{Hyper-SD / RayFlow} (Seedream~3.0~\citep{gao2025seedream}, $4$--$8\times$); compound \textbf{ADP + ADM + quantization} (Seedream~4.0~\citep{seedream2025seedream}, $>10\times$); decoupled-DMD (Z-Image's~\citep{cai2025z} 8-step Turbo); MeanFlow (HunyuanImage~3.0~\citep{cao2025hunyuanimage}, NFE 4--8); trajectory + DMD hybrid (Wan-Image~\citep{mao2026wan}, $>8\times$); and the LongCat-Image~\citep{team2025longcat} pipeline. The remaining four reports either ship distilled checkpoints without disclosing the recipe or report acceleration in less detail, but none avoids it. The conceptual shift is just as important as the engineering one: distillation is no longer a post-hoc compression of a frozen teacher, it is a \emph{co-trained primary objective} that shapes architecture choices upstream (e.g., flow-matching is preferred partly because rectified-flow ODEs distill more cleanly). Researchers who treat distillation as ``something the deployment team handles later'' are now training models that the deployment team cannot ship.
\end{highlightbox}

\section{Resources and Infrastructure}
\label{sec:resource_and_infra}

\subsection{Data Construction Methodologies}

The rapid advancement of visual generation and editing models has been fundamentally driven by the availability and quality of training data. From early text-to-image generation systems to sophisticated instruction-based image editors, the evolution of these models reflects a parallel evolution in dataset construction methodologies. As diffusion models, autoregressive transformers, and multimodal large language models (MLLMs) continue to push the boundaries of visual synthesis, the role of training data has become increasingly critical and complex.

\begin{figure}
    \centering
    \includegraphics[width=1\linewidth]{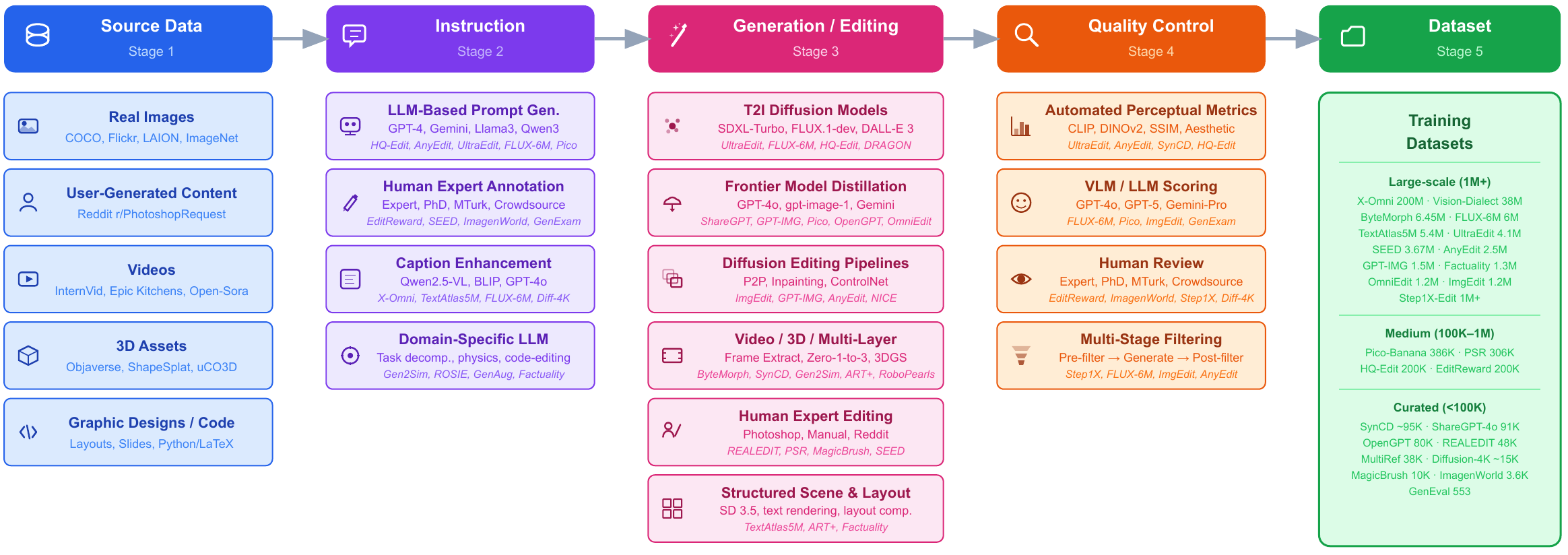}
    \caption{A unified five-stage pipeline summarizing dataset construction methodologies across analyzed works. The pipeline progresses from source data acquisition (Stage 1), through instruction generation (Stage 2), image generation or editing (Stage 3), and quality control (Stage 4), to the final training datasets (Stage 5) ranging from curated collections of hundreds of samples to large-scale datasets exceeding one million pairs.}
    \label{fig:data}
\end{figure}

Constructing high-quality training data for visual generation and editing can be decomposed into a five-stage pipeline: (1) source data acquisition, (2) instruction and caption construction, (3) image generation or editing, (4) quality control and filtering, and (5) final dataset assembly with splits and packaging. In practice, most datasets span two to four of these stages, for example, datasets grounded in real photographs skip Stage~3, while fully synthetic pipelines may collapse Stages 1 and 3 into a single model call. The following subsections discuss each stage independently, highlighting recurring design patterns and cross-cutting trade-offs. \Cref{fig:data} provides a schematic overview of the pipeline and its variants. Before detailing each stage, we first contextualize the paradigm shift from passive web-scale collection to actively engineered synthetic data that gave rise to this pipeline architecture.

\subsubsection{Paradigm Shift: From Web Scraping to Synthetic Engines}
\label{sec:paradigm}

The training data landscape for visual generation has undergone a fundamental shift---from \emph{passive collection}, where practitioners scraped and filtered what already existed on the web, to \emph{active engineering}, where purpose-built pipelines produce exactly the training signal that models require.

\textbf{Passive Collection.}
The first wave of large-scale generative models was trained on web-crawled image--text corpora---LAION-5B~\citep{schuhmann2022laion}, COYO-700M~\citep{kakaobrain2022coyo-700m}, Conceptual Captions~\citep{sharma2018conceptual}, DataComp-1B~\citep{gadre2023datacomp}---providing billions of image--alt-text pairs that powered systems such as Stable Diffusion and DALL-E~2. The appeal was scale, but the paradigm hit three walls: alt-text captions were noisy and shallow, capping text--image alignment quality; filtering was limited to coarse heuristics (resolution, aesthetic scores, deduplication) that missed semantic defects; and most critically, the web contains no natural (source, instruction, target) triplets---there is no ``editing web'' to scrape.

\textbf{Active Engineering.}
The current paradigm inverts this philosophy: rather than finding data that happens to exist, practitioners \emph{engineer} the training signal using foundation models as construction tools. Frontier models are distilled into reusable training sets~\citep{chen2025sharegpt, qian2025pico}; open-source models are composed with segmentation, inpainting, and control modules into editing pipelines~\citep{zhao2024ultraedit, ye2025imgedit, liu2025step1x}; videos are mined for temporal before/after pairs~\citep{chang2025bytemorph, zeng2025editworld}; and code or 3D rendering provides verifiable ground truth for specialized domains~\citep{zhuo2025factuality, kumari2025generating}. Three forces drove this transition: foundation models mature enough to produce training-quality outputs, editing tasks that demand paired data absent from natural corpora, and VLMs/LLMs that serve as scalable annotation and quality-control tools. The result is \emph{multi-mechanism composition}---state-of-the-art datasets combine multiple generation, annotation, and filtering strategies into integrated pipelines (Sections~\ref{sec:source}--\ref{sec:quality}).

This shift has created a \emph{data flywheel}: stronger models yield higher-quality synthetic data that trains still stronger models---91K frontier-quality samples can now match millions of web-scraped ones~\citep{chen2025sharegpt}. Yet the flywheel depends on distilling proprietary frontier models, tying the ecosystem's quality ceiling to their access terms. As open-source alternatives close the gap, the tension between closed-API distillation and reproducible open pipelines remains an active force shaping the field's data infrastructure. A minority position is worth flagging: Z-Image~\citep{cai2025z} declines to distill from any proprietary model on the principle that doing so risks a closed feedback loop in which error accumulation and homogenization cap the student at the teacher's distribution---an argument that gains weight as the open-source ceiling rises.

\subsubsection{Source Data Acquisition}\label{sec:source}

The first stage determines the raw material from which all subsequent processing derives. The choice of source data is consequential: it shapes the distributional characteristics, domain coverage, and ultimately the capability boundaries of the trained model. Practitioners navigate a fundamental trade-off among three axes---\emph{realism} (how faithfully the data reflects natural image statistics), \emph{controllability} (how precisely individual factors such as viewpoint or lighting can be varied), and \emph{scale} (how cheaply millions of diverse samples can be obtained). We identify five principal data modalities below, each occupying a different region of this trade-off space.

\textbf{Real Images.}
Large-scale image--caption corpora remain the most common starting point, offering high realism at the cost of limited control over scene factors. Key repositories include MS~COCO~\citep{lin2014microsoft}, Flickr~\citep{young2014image}, NoCaps~\citep{agrawal2019nocaps}, VizWiz~\citep{gurari2018vizwiz}, TextCaps~\citep{sidorov2020textcaps}, LAION-Aesthetics~\citep{schuhmann2022laion}---a curated subset of the LAION corpus filtered by predicted aesthetic scores---, and ImageNet~\citep{deng2009imagenet}. UltraEdit~\citep{zhao2024ultraedit} anchors its editing pipeline to roughly 1.6M real image--caption pairs drawn from eight such sources to mitigate biases inherent in text-to-image models. ImgEdit~\citep{ye2025imgedit} pre-filters LAION-Aesthetics to retain only images with shorter-side resolution $\geq$1280\,px and aesthetic score $>$4.75, yielding a high-fidelity seed pool. X-Omni~\citep{geng2025x} applies design-specific filtering to COYO-700M~\citep{kakaobrain2022coyo-700m}, DataComp-1B~\citep{gadre2023datacomp}, and LAION-2B~\citep{schuhmann2022laion}, distilling 200M images suitable for unified generation and understanding. AnyEdit~\citep{yu2025anyedit} draws from MS~COCO (123K), MVImgNet~\citep{yu2023mvimgnet} (31K), and CC3M~\citep{sharma2018conceptual} (519K) to ensure category breadth. Across these works, real-image anchoring is a recurring strategy that injects distributional realism absent from purely synthetic pipelines. Recent industrial pre-training corpora operate at substantially larger scale but with aggressive selection: HunyuanImage~3.0~\citep{cao2025hunyuanimage} retains under 45\% of a 10B+ raw pool to obtain roughly 5B images; LongCat-Image~\citep{team2025longcat} curates 1.2B samples through MD5/SigLIP-based deduplication and a six-method aesthetic ensemble; FireRed-Image-Edit~\citep{team2026firered} compresses 1.6B raw samples (900M T2I + 700M edit) down to roughly 100M after multi-level deduplication and post-filtering---a $\sim$6\% retention rate that the authors frame as a deliberate quality-over-scale choice; and Wan-Image~\citep{mao2026wan} discloses 13.27T tokens for generation pre-training alone. These numbers reframe ``billions of pairs'' from a vague descriptor into a comparable axis along which industrial pipelines now diverge by an order of magnitude.

\textbf{User-Generated Content.}
Community platforms provide ecologically valid editing requests that reflect genuine user intentions. Reddit's r/PhotoshopRequest is the dominant source: REALEDIT~\citep{sushko2025realedit} collects 48K training pairs spanning 2012-2021, while PSR~\citep{taesiri2025understanding} harvests 82,976 unique requests and 305,806 human-edited images from 2013-2025. SEED-Data-Edit~\citep{ge2024seed} supplements its automated pipeline with 52K real-world photographs posted by amateur photographers seeking professional edits, and RefBlend~\citep{chen2025multiref} curates 1,000 real-world multi-reference samples from Reddit. These sources offer natural-language instructions and diverse editing distributions that synthetic prompt generators struggle to replicate.

\textbf{Videos.}
Temporal sequences supply naturally paired frames that depict scene changes, object motion, or viewpoint shifts---providing before/after pairs without the manual construction cost that image-only pipelines incur. InternVid~\citep{wang2023internvid}, a large-scale video--text corpus, has become a popular source: ByteMorph-6M~\citep{chang2025bytemorph} extracts sequential frames from it to construct 6.45M non-rigid motion editing pairs, while EditWorld~\citep{zeng2025editworld} mines it with a complementary objective, selecting frame pairs that maximize spatial or visual variance under strong identity consistency. Beyond web video, domain-specific footage also proves valuable: Masquerade~\citep{lepert2025masquerade} leverages Epic Kitchens~\citep{damen2018scaling}, extracting 675K frames depicting human manipulation activities for robot-learning augmentation; ImgEdit~\citep{ye2025imgedit} harvests 160K motion-change pairs from Open-Sora Plan~\citep{zheng2024open} in-house videos; and DreamGen~\citep{jang2025dreamgen} aggregates 438.1M frames from ten robot, simulation, and human-activity datasets for world-model pre-training. While video-derived pairs are abundant and temporally coherent, their semantic scope is constrained by whatever changes naturally occur in the source footage, limiting fine-grained control over edit type and magnitude.

\textbf{3D Assets.}
Rendered 3D objects enable precise, disentangled control over viewpoint, lighting, and geometry---factors that are entangled and difficult to isolate in 2D photographs. SynCD~\citep{kumari2025generating} renders 75,000 rigid assets from Objaverse~\citep{deitke2023objaverse} with depth-conditioned generation to produce multi-view consistent customization data. Gen2Sim~\citep{katara2024gen2sim} lifts 2D diffusion outputs into textured 3D meshes via Zero-1-to-3~\citep{liu2023zero} and Score Distillation Sampling for sim-to-real robotics transfer. RoboPearls~\citep{tao2025robopearls} retrieves assets from ShapeSplat~\citep{ma2024shapesplat} and uCO3D~\citep{liu24uco3d} to build editable 3D Gaussian Splatting simulations. The chief limitation of 3D-rendered data is asset diversity: current repositories such as Objaverse, while large, skew toward isolated objects and lack the compositional complexity of real-world scenes.

\textbf{Graphic Designs and Code.}
Structured visual artifacts---layouts, slides, rendered programs---serve as source data for domains where photographic imagery is insufficient. ART+~\citep{pu2025art} extracts semantic layouts from 800K multi-layer graphic designs, providing compositional scaffolds for transparent-image generation. Factuality Matters~\citep{zhuo2025factuality} executes approximately 2M Python and \LaTeX{} drawing programs to produce code-aligned structured visuals (charts, diagrams, mathematical figures). TextAtlas5M~\citep{wang2025textatlas5m} combines SlideShare presentations, ArXiv papers, Amazon book covers, and Common Crawl pages to cover dense text-rendering scenarios. Vision as a Dialect~\citep{han2025vision} sources user prompts from JourneyDB~\citep{sun2023journeydb} and Midjourney-Prompts to guide generation-centric pretraining. Industrial pipelines extend the same logic to plug specific knowledge gaps: Seedream~4.0~\citep{seedream2025seedream} extracts figures from in-house textbook, research-article, and novel PDFs and additionally synthesizes formula images from OCR output and \LaTeX{} source---a redesign motivated by the empirical finding that purely top-down resampling underrepresents fine-grained, knowledge-centric concepts such as instructional diagrams and mathematical expressions. These structured sources fill critical gaps---typography, diagrammatic reasoning, compositional layout---that natural-image corpora leave largely unaddressed, though models trained primarily on such data may require adaptation to generalize to photographic content.

\paragraph{Cross-source complementarity.}
No single data modality satisfies all requirements simultaneously. Real images provide distributional realism but offer no control over scene factors; user-generated content captures genuine editing intent but is scarce and noisy; videos supply temporally coherent pairs at scale but constrain edit semantics to whatever changes naturally occur; 3D assets afford disentangled manipulation at the cost of limited scene diversity; and structured designs cover specialized domains that photographs leave unaddressed. Recognizing these complementarities, most recent large-scale efforts adopt \emph{multi-source fusion}: UltraEdit~\citep{zhao2024ultraedit} draws from eight real-image corpora, ImgEdit~\citep{ye2025imgedit} combines photographs with video-derived motion pairs, and DreamGen~\citep{jang2025dreamgen} aggregates ten heterogeneous datasets spanning real video, simulation, and human-activity footage. This strategy mitigates the biases inherent in any single source and broadens the resulting model's generalization capacity. The raw assets assembled in this stage, however, are rarely usable directly---they must be paired with textual conditioning signals whose construction we discuss next.

\subsubsection{Instruction Construction}\label{sec:instruction}

Once source data has been acquired, the next stage produces the textual signals---prompts, editing instructions, captions, and annotations---that condition downstream generation and evaluation. Among all pipeline stages, this one arguably carries the highest leverage: caption quality directly bounds how well a model can learn text--image alignment, and instruction specificity determines the granularity of controllable editing. Effective annotations share three desirable properties: \emph{specificity} (precise enough to constrain generation), \emph{diversity} (covering the breadth of visual concepts and editing operations), and \emph{faithfulness} (accurately reflecting the visual content they describe). Three complementary construction strategies have emerged---VLM-based annotation, text-only LLM generation, and human labeling---each trading off differently among cost, scale, and quality. A fourth paradigm---constructing instructions as part of frontier-model distillation pipelines---employs the same LLM-based strategies described below for its instruction stage, while its generation and refinement stages are discussed in \Cref{sec:generation}.

\begin{highlightbox}{Community Message: Caption Progress Has Outpaced Architecture Progress}

The dominant training lever of the past two years has \emph{not} been a new backbone or a new objective---it has been the steady upgrade of caption / VLM-driven relabeling pipelines. With architecture and compute roughly fixed, caption quality now decides the capability ceiling of the resulting model: Z-Image's~\citep{cai2025z} ``6B competitive with 20B+'' result and LongCat-Image's~\citep{team2025longcat} edge over larger contemporaries both rest on aggressive caption pipelines plus tight data curation, not on parameter count. The community-level reading is unambiguous: \textbf{architecture has converged; the next phase of progress lives in data and caption.} The corollary for new entrants is practical---investing the next FTE-quarter in a stronger captioning / relabeling pipeline almost always beats investing it in an architecture variant.
\end{highlightbox}

\textbf{VLM-Based Annotation.}
Vision--language models (VLMs) have become the primary tool for producing textual descriptions of visual content, as they can directly observe images and generate grounded descriptions---an operation that text-only models fundamentally cannot perform. Two scenarios drive their use. \emph{From-scratch annotation} is required when source data carries no textual description at all. Video-derived editing pairs are a prominent case: ByteMorph-6M~\citep{chang2025bytemorph} and ImgEdit~\citep{ye2025imgedit} both employ GPT-4o-class models to generate motion descriptions for frame pairs extracted from video corpora, while EditWorld~\citep{zeng2025editworld} adopts a two-stage approach where Video-LLaVA first produces video-level descriptions that GPT-3.5 then transforms into structured editing instructions. For real-world editing pairs that lack text, REALEDIT~\citep{sushko2025realedit} uses LLaVA-Next to caption input images and GPT-4o to describe outputs, and AnyEdit~\citep{yu2025anyedit} employs VILA to caption source and target images across 25 editing types. FLUX-Reason-6M~\citep{fang2025flux} introduces Generation Chain-of-Thought (GCoT), a structured annotation format that decomposes the intended generation into explicit reasoning steps---planning composition, specifying spatial layout, and describing visual attributes---synthesized by combining category-specific captions from Qwen-VL with multi-stream reasoning, representing a move toward richer, process-aware textual conditioning. \emph{Re-captioning} addresses a complementary problem: many large-scale corpora ship with noisy alt-text that is inadequate for conditioning modern generative models. X-Omni~\citep{geng2025x} undertakes the largest such effort discussed here, replacing captions for 200M images from COYO, DataComp, and LAION with dense descriptions from Qwen2.5-VL-72B. TextAtlas5M~\citep{wang2025textatlas5m} deploys an ensemble of Qwen2-VL, BLIP, and Intern-VL2 with cross-validation via LLM semantic-similarity checks and perplexity filtering for text-heavy images. Diffusion-4K~\citep{zhang2025diffusion} and Vision as a Dialect~\citep{han2025vision} similarly recaption specialized corpora using GPT-4o and Qwen2.5-VL, respectively. While VLM-based annotation is now indispensable for grounding textual signals in visual content, it carries inherent trade-offs: VLMs can hallucinate details absent from the image, and processing hundreds of millions of images with large models incurs substantial computational cost.

\emph{Multi-granularity captioning} has emerged as the dominant industrial pattern for resolving the tension between detailed-prompt training signal and short-prompt user behavior at inference. Rather than producing a single caption per image, recent frontier reports generate a small set of descriptions at different lengths and structures and sample among them during training. HunyuanImage~3.0~\citep{cao2025hunyuanimage} defines four bilingual descriptive tiers ranging from roughly 30-word summaries to roughly 1{,}000-word exhaustive depictions, augmented by stylistic and named-entity fields and grounded by an OCR agent and a named-entity agent through a bidirectional verification loop. LongCat-Image's Multi-Granularity Captioning framework~\citep{team2025longcat} produces four levels (entity, phrase, composition, photographic) with the most detailed photographic-level description sampled at probability 0.65 to keep world-knowledge density as the dominant training signal, while the entity, phrase, and composition tiers receive probabilities 0.05, 0.10, and 0.20 respectively. Z-Image's Z-Captioner~\citep{cai2025z} generates five caption types per image---long, medium, short, tags, and a simulated-user-prompt that mimics incomplete real-world inputs---with OCR results kept in their original language to prevent translated text leaking into rendered outputs. Qwen-Image~\citep{wu2025qwen} adopts a dual-track design that combines raw web alt-text, recaptioned descriptions, and a fused split, with structured JSON metadata fields (image type, image style, watermark list, abnormal-element flags) extracted in a single Qwen2.5-VL-72B pass. JoyAI-Image~\citep{jdjoyaiimage} produces four caption types (short, long, extended-long, structured JSON) under a two-stage OCR-conditioned pipeline gated by text-coverage, language-integrity, and anti-hallucination filters. FireRed-Image-Edit~\citep{team2026firered} additionally schedules a stage-specific caption mix---approximately 55/40/5 for original/structured/instructive captions in pre-training shifting to 10/45/45 by SFT---so that instructive supervision rises in lockstep with the model's growing instruction-following demand. Wan-Image~\citep{mao2026wan} routes each image into one of five primary categories and draws from a global pool of 25 attributes, assigning category-specific subsets (14 dimensions for photorealistic, 11 for non-photorealistic, 11 for text-centric, 10 for charts, 7 for multi-image compositions). The shared lesson is that caption quality and granularity, rather than raw caption count, now bound how much an architecture can learn from a fixed image corpus---a finding that aligns with the empirical claims in the highlight box above.

\textbf{Text-only LLM Generation.}
When visual grounding is unnecessary---for instance, when crafting diverse prompts for text-to-image generation or expanding a small seed set of editing instructions---text-only LLMs offer a faster and cheaper alternative. The dominant pattern is \emph{seed-and-expand}: a small set of human-authored examples serves as in-context demonstrations, and the LLM generates diverse variations at scale. UltraEdit~\citep{zhao2024ultraedit} has human raters compose hundreds of seed instructions on COCO images, which LLMs expand to roughly 10K examples; HQ-Edit~\citep{hui2024hq} applies GPT-4 in a Self-Instruct loop, expanding 293 seed triplets to roughly 100K diptych prompts; and FLUX-Reason-6M~\citep{fang2025flux} uses Gemini-2.5-Pro to craft 200 seed prompts that Qwen3-32B then creatively diversifies (complementing its VLM-based captioning described above). An alternative approach is \emph{constraint-based generation}, where the LLM produces instructions within a predefined taxonomy or rule set rather than from seed examples. ShareGPT-4o-Image~\citep{chen2025sharegpt} defines a six-dimensional attribute space (objects, background, style, lighting, camera viewpoints, composition) over 1,000 ImageNet categories and uses Gemini-Pro-2.5 to compose attributes into natural-language prompts. OpenGPT-4o-Image~\citep{chen2025opengpt} builds a hierarchical taxonomy with structured resource pools (Object, Relation, Qualifier) and template-based generation spanning 51 fine-grained sub-capabilities. Pico-Banana-400K~\citep{qian2025pico} introduces a dual-model reformulation strategy, using Gemini-2.5-Flash for detailed instructions and Qwen2.5-7B for concise human-style rewrites across a 35-type editing taxonomy. At a smaller scale, AnyEdit~\citep{yu2025anyedit} uses Llama3-8B with task-specific constraints and iterative self-enhancement to produce instructions spanning 25 editing types, and TIIF-Bench~\citep{wei2025tiif} leverages GPT-4o to systematically decompose prompts into objects and attributes, defining 36 compositional combinations. Note that several of these works---HQ-Edit, ShareGPT-4o-Image, OpenGPT-4o-Image, and Pico-Banana-400K---embed their instruction generation within larger frontier-model distillation pipelines; their generation and refinement stages are discussed in \Cref{sec:generation}. Specialized domains impose further constraints beyond generic prompt engineering: Gen2Sim~\citep{katara2024gen2sim} uses GPT-4 to generate temporal task decompositions, physics parameters, and Python reward functions for robotic environments; ROSIE~\citep{yu2023scaling} employs GPT-3 with few-shot prompting for semantically coherent robot-scene augmentation; and Factuality Matters~\citep{zhuo2025factuality} leverages GPT-5 to produce paired code-editing instructions with chain-of-thought reasoning for structured visual programs. Text-only generation excels at scaling instruction diversity cheaply, but its outputs are ungrounded in visual content---a limitation that typically requires downstream VLM verification or human review to catch misalignments.

\textbf{Human Annotation.}
Where precision, nuance, or subjective judgment are paramount, human annotators remain indispensable despite their limited scalability. Annotation efforts span three broad purposes. For \emph{preference labeling}, EditReward~\citep{wu2025editreward} engages expert annotators following standardized protocols---including pilot studies, calibration sessions, and continuous cross-checking---to label over 200K preference pairs on a 4-point Likert scale. For \emph{professional editing}, SEED-Data-Edit~\citep{ge2024seed} employs Photoshop professionals to produce 52K real-world edits and 21K multi-turn sequences of up to five rounds, while Step1X-Edit~\citep{liu2025step1x} implements Redundancy-Enhanced Annotation with multi-round review and Stylized Annotation via Contextual Examples to reduce hallucination in annotator outputs. For \emph{evaluation benchmarks}, ImagenWorld~\citep{sani2026imagenworld} recruits 22 expert graduate students providing 20,000 fine-grained annotations (three per image, capped at 200 per annotator per week) with Krippendorff's $\alpha$ of 0.51--0.78, GenExam~\citep{wang2025genexam} relies on three PhD annotators for cross-validated scoring after initial GPT-5 drafting, and GenEval~\citep{ghosh2023geneval} crowdsources 6,000 annotations on 1,200 images through Amazon Mechanical Turk. Human annotation sets the quality ceiling for each of these tasks but cannot scale to millions of examples, and is therefore typically reserved for high-stakes subsets or evaluation data.

\paragraph{Cross-strategy complementarity.}
In practice, most dataset construction efforts combine multiple annotation strategies. UltraEdit~\citep{zhao2024ultraedit} exemplifies this hybrid approach: human annotators author seed instructions, LLMs expand them to scale, and the resulting text conditions image generation. Similarly, FLUX-Reason-6M~\citep{fang2025flux} pairs text-only seed expansion with VLM-based captioning, and AnyEdit~\citep{yu2025anyedit} combines LLM-generated instructions with VLM-produced image captions. The emerging division of labor is clear: text-only LLMs handle breadth (generating diverse instructions cheaply), VLMs ensure faithfulness (grounding descriptions in actual visual content), and humans provide depth (quality-critical labeling that neither model type can reliably perform). This complementarity extends to frontier-model distillation pipelines (\Cref{sec:generation}), which integrate instruction generation, image synthesis, and post-hoc VLM refinement into a unified workflow.

\subsubsection{Generation and Editing}
\label{sec:generation}

Given source assets and textual conditioning from the preceding stages, the generation stage produces the actual images or image pairs that constitute the training data. This is where dataset quality is most visibly determined: the choice of generation mechanism shapes output fidelity, edit precision, and the diversity of the resulting training signal. Practitioners navigate a fundamental trade-off among \emph{output quality}, \emph{scale}, \emph{reproducibility} (whether the pipeline can be replicated without proprietary access), and \emph{cost}. We organize the analyzed approaches into five mechanisms based on how the target image is produced, noting that many large-scale efforts combine multiple mechanisms within a single dataset.

\textbf{Frontier Model Distillation.}
A rapidly dominant paradigm treats proprietary frontier models---GPT-4o-Image, gpt-image-1, Gemini-2.5-Flash-Image, DALL-E 3---as black-box teachers whose outputs are harvested into reusable training data. Because these models currently define the quality ceiling for both generation and editing, distilling their outputs provides a shortcut to high-quality data without replicating their internal capabilities.

In practice, frontier distillation follows a three-stage pipeline. The first stage, \emph{instruction preparation}, crafts the prompts or editing commands sent to the API: ShareGPT-4o-Image~\citep{chen2025sharegpt} samples from a six-dimensional attribute space, OpenGPT-4o-Image~\citep{chen2025opengpt} builds a hierarchical taxonomy spanning 51 fine-grained sub-capabilities with difficulty calibration, Pico-Banana-400K~\citep{qian2025pico} uses dual-model reformulation across a 35-type editing taxonomy, and HQ-Edit~\citep{hui2024hq} expands 293 human-authored seed triplets to roughly 100K diptych prompts via GPT-4 Self-Instruct (instruction construction details are discussed in \Cref{sec:instruction}). The second stage, \emph{model invocation}, executes the generation or edit: ShareGPT-4o-Image captures outputs directly from GPT-4o-Image; GPT-IMAGE-EDIT-1.5M~\citep{wang2025gpt} harvests edits from gpt-image-1 with automated regeneration on failure; Pico-Banana-400K calls Gemini-2.5-Flash-Image with retry logic; and HQ-Edit~\citep{hui2024hq} pioneers \emph{diptych generation} via DALL-E 3, producing input and output images side-by-side within a single canvas to ensure visual consistency between the pair. The third stage, \emph{post-generation refinement}, uses VLMs to realign instructions with the actual visual outputs---GPT-IMAGE-EDIT-1.5M employs GPT-4o to rewrite editing instructions based on the observed source--target difference, and HQ-Edit applies GPT-4V to describe the actual visual changes and revise instructions accordingly. OmniEdit~\citep{wei2024omniedit} takes a complementary approach, distilling outputs from multiple specialist editing models and using GPT-4o scoring across seven dimensions to select the best results.

This paradigm offers the highest output quality among automated methods but carries notable constraints: generation costs can be substantial (Pico-Banana-400K reports approximately \$100K in API expenditure), output diversity depends critically on prompt design, and terms-of-service restrictions may limit redistribution of harvested data.

\textbf{Open-Source Model Pipelines.}
A broad family of approaches relies on open-weight diffusion and flow-matching models---FLUX, SDXL, Stable Diffusion---composed with auxiliary tools such as segmentation models, ControlNet, and inpainting modules. While typically yielding lower per-sample quality than frontier distillation, these pipelines are freely reproducible, support fine-grained engineering control, and incur no API costs.

\emph{Text-to-image generation.}
Several datasets are built by generating standalone images from text rather than editing existing ones. FLUX-Reason-6M~\citep{fang2025flux} generates 8M candidates using FLUX.1-dev from creatively diversified prompts, retaining 6M after multi-stage VLM filtering---an effort requiring 15,000 A100 GPU-days. Vision as a Dialect~\citep{han2025vision} uses FLUX.1-schnell in two modes: Gen23M generates images from VLM-recaptioned prompts for text-to-image pretraining, while Gen15M produces \emph{paired} images from the same prompt with different random seeds to construct image-to-image training data without explicit editing---a simple but effective strategy for bridging understanding and generation. DRAGON~\citep{bertazzini2025dragon} pursues a distinct objective---forensic detection training---by synthesizing 2.6M images across 25 different diffusion architectures, where cross-model diversity rather than individual image quality drives the design.

\emph{Image Editing pipelines.}
When the goal is to produce (source, instruction, target) editing tuples, practitioners compose multiple open-source tools into task-specific workflows. The defining characteristic is \emph{pipeline specialization}: different edit types require different tool combinations, and most systems maintain separate sub-pipelines per category. A common pattern is \emph{attention-based editing}: UltraEdit~\citep{zhao2024ultraedit} employs SDXL-Turbo with Prompt-to-Prompt attention control, using 2--4 diffusion steps for free-form edits and 3--7 for region-based edits with soft masks, achieving roughly 100$\times$ faster throughput than InstructPix2Pix; EditWorld~\citep{zeng2025editworld} similarly leverages cross-attention map extraction for keyword localization, combined with inpainting, IP-Adapter for identity preservation, and ControlNet for structural guidance. At larger scale, \emph{multi-pipeline systems} maintain distinct workflows per edit type: ImgEdit~\citep{ye2025imgedit} implements ten sub-pipelines combining FLUX, SDXL, IP-Adapters, ControlNet, and depth/canny LoRA; AnyEdit~\citep{yu2025anyedit} spans 25 editing types with adaptive tool selection; and Step1X-Edit~\citep{liu2025step1x} deploys 11 category-specific pipelines using Florence-2, SAM-2, FLUX-Fill, and ControlNet, generating 20M candidates from which only 1M (5\%) survive filtering. Other works target specific editing subtasks: SEED-Data-Edit~\citep{ge2024seed} maintains two branches---mask-based object manipulation via GroundingDINO, SAM, and LaMa, and appearance changes via Plug-and-Play diffusion; NICE~\citep{pakdamansavoji2025improving} applies scene surgery for robot-learning augmentation; and ART+~\citep{pu2025art} targets multi-layer transparent images using MultiLayerFLUX with RMBG-2.0 alpha extraction. Recent industrial pipelines push the multi-pipeline pattern further along orthogonal axes. LongCat-Image-Edit~\citep{team2025longcat} aggregates four sources---open-source datasets (OmniEdit, OmniGen2, NHREdit), expert-model synthesis, video-frame mining, and interleaved web corpora---and runs each editing pair through GPT-4o instruction-rewriting that generates 3--5 bilingual paraphrase variants per sample. FireRed-Image-Edit~\citep{team2026firered} structures its data-production engine around three explicit strategies---instructional control via task-aligned expert libraries, structured control driven by SAM-3 masks and DWPose keypoints, and model-free template-based synthesis using graphics engines and deterministic algorithmic filters---paired with task-inversion and task-splitting augmentations that double effective data volume without new generation. Z-Image-Edit~\citep{cai2025z} introduces an ``efficient graphical representation'' trick: from one source image plus $N$ edited versions, $2\cdot\binom{N}{2}+1$ paired tuples are derived by exhaustive pairing, multiplying training pairs at zero generation cost while inverse pairs (edited-to-original) anchor the model to clean targets.

\textbf{Temporal Pair Extraction.}
Video sequences provide a natural source of before/after pairs: consecutive frames capture real-world changes---object motion, viewpoint shifts, lighting transitions---without any generative model producing the edit itself. This mechanism yields temporally coherent pairs with high realism at low cost, but the range of editable semantics is limited to whatever changes naturally occur in the footage.

Two approaches have emerged. In \emph{generative frame extraction}, an image-to-video model synthesizes a video from a static image, and frame pairs are extracted from the result. ByteMorph-6M~\citep{chang2025bytemorph} follows this approach with layered compositing, producing 6.45M non-rigid motion editing pairs. DreamGen~\citep{jang2025dreamgen} fine-tunes a video world model on a small number of real robot demonstrations and generates novel trajectory videos as augmented training data. In \emph{real-video mining}, pairs are extracted directly from existing corpora. EditWorld~\citep{zeng2025editworld} selects frame pairs from InternVid that maximize spatial variance while maintaining identity consistency, complementing its diffusion-based editing branch discussed above. ImgEdit~\citep{ye2025imgedit} harvests 160K motion-change pairs from in-house videos, and Step1X-Edit~\citep{liu2025step1x} extracts motion pairs from the Koala-36M corpus using RAFT optical flow estimation. HunyuanImage~3.0~\citep{cao2025hunyuanimage} adds a more elaborate video-mining filter cascade---shot-boundary detection to isolate single-scene segments, camera-motion classification to drop heavy viewpoint shifts, object detection and semantic segmentation to retain canonical transformation keyframes, and motion-blur rejection at the end---and feeds the resulting pairs into a dedicated image-difference captioner that consumes both frames plus the bracketing two-frame video clip to ground its description in temporal context.

\textbf{Human Curation and Editing.}
Where authenticity and ecological validity are paramount, human-produced data remains irreplaceable. This mechanism yields training pairs reflecting genuine user intentions and aesthetic judgments---qualities that automated pipelines struggle to replicate---but cannot scale beyond tens of thousands of samples.

Community-sourced data offers the most naturalistic editing distributions. REALEDIT~\citep{sushko2025realedit} collects 48K training pairs from Reddit's r/PhotoshopRequest spanning 2012--2021, preserving authentic human-made edits with no AI involvement in image production. PSR~\citep{taesiri2025understanding} harvests the same community at larger scale, yielding 83K requests and 306K human-edited images over 2013--2025 with fine-grained annotations across 15 action types and 3 creativity levels. Professional editing offers higher per-sample quality: SEED-Data-Edit~\citep{ge2024seed} employs Photoshop experts for 52K edits and 21K multi-turn sequences of up to five rounds---the only large-scale dataset capturing iterative human editing trajectories. Diffusion-4K~\citep{zhang2025diffusion} takes a curation-only approach, manually filtering real 4K photographs for artifacts and pairing them with GPT-4o captions to address the scarcity of high-resolution training images.

\textbf{Programmatic and 3D Rendering.}
For domains requiring pixel-level precision or disentangled factor control, deterministic rendering provides guarantees that stochastic generative models cannot match. Code-executed and 3D-rendered images are exactly reproducible, support manipulation of individual scene variables, and produce verifiable ground truth---properties critical for text rendering, diagrammatic reasoning, multi-view consistency, and robotic simulation.

On the code-execution side, Factuality Matters~\citep{zhuo2025factuality} executes approximately 2M Python and \LaTeX{} drawing programs to render source images, then uses GPT-5 to generate code-level edits that produce target images with precise, verifiable state transitions. TextAtlas5M~\citep{wang2025textatlas5m} programmatically renders text into designated image regions using 8,700 fonts atop Stable Diffusion 3.5-generated backgrounds, ensuring exact typographic ground truth. On the 3D side, SynCD~\citep{kumari2025generating} renders depth maps from 75K Objaverse assets and applies Masked Shared Attention for depth-conditioned generation, producing multi-view consistent customization data. Gen2Sim~\citep{katara2024gen2sim} lifts 2D diffusion outputs into textured 3D meshes via Zero-1-to-3 and Score Distillation Sampling for robotic simulation. RoboPearls~\citep{tao2025robopearls} reconstructs real videos into editable 3D Gaussian Splatting scenes, enabling object manipulation within physically grounded simulations. Masquerade~\citep{lepert2025masquerade} bridges real and simulated embodiment by compositing rendered robot models onto real video frames via HaMeR tracking, SAM2 segmentation, and E2FGVI inpainting. JoyAI-Image's SpatialEdit data engine~\citep{jdjoyaiimage} applies the same logic to camera- and object-level editing supervision: Branch~A renders Blender scenes under a static camera while applying controlled translations, rotations, and scalings to a foreground asset (compositing it back via inpainting), and Branch~B parameterizes a 3-DOF camera motion (yaw, pitch, distance) around a focus object to produce source-target pairs with globally consistent scene content but different viewing conditions---providing geometrically unambiguous supervision that natural video pairs cannot supply.

\paragraph{Cross-mechanism complementarity.}
In practice, most large-scale data construction efforts combine multiple mechanisms. EditWorld~\citep{zeng2025editworld} maintains parallel diffusion-editing and video-extraction branches; ImgEdit~\citep{ye2025imgedit} and Step1X-Edit~\citep{liu2025step1x} supplement their editing pipelines with video-derived motion pairs; and SEED-Data-Edit~\citep{ge2024seed} spans automated pipelines, professional human editing, and community-sourced pairs---the broadest methodological coverage among analyzed works. The emerging division of labor mirrors that observed in instruction construction (\Cref{sec:instruction}): frontier distillation provides the quality ceiling, open-source pipelines deliver reproducible breadth, temporal extraction supplies naturalistic dynamics, human editing anchors ecological validity, and programmatic rendering guarantees domain-specific precision. The relative emphasis among these mechanisms is ultimately determined by the target application, available budget, and the desired balance among quality, diversity, and cost.
\subsubsection{Quality Control}
\label{sec:quality}

Once data is generated and annotated, quality control determines what enters the final training set. This stage is critical because generative pipelines---whether diffusion-based editing or frontier-model distillation---inevitably produce outputs that are misaligned with instructions, contain visual artifacts, or fail to preserve source identity. Without rigorous filtering, these failure modes propagate directly into model training. The central challenge is balancing \emph{scalability} against \emph{judgment reliability}: lightweight perceptual metrics can process millions of samples cheaply but miss semantic errors, while human review catches subtle failures but cannot scale beyond thousands of samples per annotator. We organize the filtering strategies observed across the analyzed works into four categories that form a natural progression from low-level perception to high-level semantic understanding to human judgment, with increasing cost and reliability at each level.

\textbf{Automated Perceptual and Semantic Metrics.}
The most scalable filtering layer operates on perceptual similarity and aesthetic quality, capturing low-level fidelity without requiring semantic understanding. Reference-based metrics---such as SSIM for structural similarity and DINOv2 cosine similarity for deep feature-level correspondence---measure how closely a generated image matches its source or reference. These can be applied as hard thresholds (e.g., SynCD~\citep{kumari2025generating} requires aesthetic score ${>}6$ and DINOv2 similarity ${>}0.7$) or as ranking criteria within a best-of-$N$ selection framework (e.g., UltraEdit~\citep{zhao2024ultraedit} generates 100 candidates per sample and selects the top-ranked output across SSIM, DINOv2, and CLIP similarities). For editing tasks specifically, CLIP Directional Similarity---which measures whether the change between source and target in CLIP embedding space aligns with the textual instruction---provides a task-aware signal beyond generic perceptual fidelity~\citep{zhao2024ultraedit}. Complementary heuristics target specific failure modes: resolution and aesthetic score pre-filtering removes low-quality seeds~\citep{ye2025imgedit}, deformation-based rejection discards spatially distorted outputs~\citep{hui2024hq}, and composite preference scores such as Multi-dimensional Preference Scoring (MPS) assess overall realism of synthetic images~\citep{bertazzini2025dragon}. These automated metrics are cheap and parallelizable, making them the natural first-pass filter; however, they are fundamentally limited to low-level properties and cannot assess whether an edit correctly follows a complex natural-language instruction.

A specialized but increasingly central instance of this layer is \emph{AIGC contamination filtering}. As web crawls accumulate AI-generated images, leaving them inside a real-image pre-training pool has been reported to cause concrete pathologies. LongCat-Image~\citep{team2025longcat} attributes a ``plastic'' or ``greasy'' texture in generated outputs to even a small AIGC fraction, and reports that such contamination drives premature convergence to a narrow local optimum that ``severely limits the model's potential to achieve higher levels of realism during subsequent fine-tuning''---purging AIGC entirely from pre-training and mid-training, and then repurposing the same detector adversarially as one of four reward models during RL. HunyuanImage~3.0~\citep{cao2025hunyuanimage} compounds per-image AIGC detection with \emph{source-level removal}: any data source whose AIGC ratio crosses an internal threshold is dropped wholesale, on the rationale that per-image classifiers miss residual contamination from heavily polluted feeds. Z-Image~\citep{cai2025z}, FireRed-Image-Edit~\citep{team2026firered}, and Wan-Image~\citep{mao2026wan} each ship a dedicated AIGC classifier as a first-pass operator alongside watermark, greasy-texture, and compression-artifact detectors. The cross-paper convergence is informative: AIGC filtering has moved from a sanitation afterthought to a structural quality control with mechanistic justifications grounded in optimization dynamics rather than aesthetic preference.

\textbf{VLM and LLM Scoring.}
Perceptual metrics cannot judge whether an edit faithfully executes its instruction at a semantic level---for instance, whether ``replace the dog with a cat'' actually produces a cat rather than a visually plausible but semantically incorrect animal. Vision-language models (VLMs) and large language models (LLMs) fill this gap by serving as automated judges capable of evaluating instruction compliance, content preservation, and overall coherence. A representative design is Pico-Banana-400K~\citep{qian2025pico}, which uses Gemini-2.5-Pro to score each sample on four weighted criteria---Instruction Compliance (40\%), Seamlessness (25\%), Preservation (20\%), and Technical Quality (15\%)---with a composite acceptance threshold. Similarly, FLUX-Reason-6M~\citep{fang2025flux} employs multi-stage VLM filtering that scores relevance across six characteristics including typographic quality. For text-centric domains, LLMs assess prompt quality and content richness: GenExam~\citep{wang2025genexam} uses GPT-5 to rate text richness, domain relevance, and complexity, while TextAtlas5M~\citep{wang2025textatlas5m} validates textual accuracy through perplexity analysis. Several other works incorporate VLM/LLM scoring as one component within broader pipelines~\citep{liu2025step1x, wei2024omniedit, ye2025imgedit, taesiri2025understanding}. The key advantage is the ability to evaluate high-level semantic properties that perceptual metrics miss entirely; the key limitation is that VLM judges themselves can hallucinate or exhibit systematic biases, and their per-sample inference cost is orders of magnitude higher than perceptual metrics.

\textbf{Human Review.}
When automated methods---both perceptual and model-based---reach their reliability ceiling, human review provides the final quality gate. This is particularly necessary for subjective quality dimensions, nuanced instruction compliance, and domain-specific correctness that current VLMs cannot reliably assess. Structured annotation protocols ensure consistency: EditReward~\citep{wu2025editreward} employs expert annotators with 4-point Likert scales across Instruction Following and Visual Quality dimensions, while ImagenWorld~\citep{sani2026imagenworld} uses 22 expert annotators (3 independent ratings per image) and measures inter-annotator agreement via Krippendorff's $\alpha$ (a statistical measure of rater consistency), achieving values of 0.51--0.78 across tasks. Domain-specific review addresses failure modes that generic metrics cannot capture: Diffusion-4K~\citep{zhang2025diffusion} manually filters for motion blur and focus artifacts in ultra-high-resolution photographs, and RoboEngine~\citep{yuan2025roboengine} applies multi-round verification for its robotics segmentation masks. The fundamental limitation is throughput---ImagenWorld caps annotators at 200 samples per week to prevent fatigue, making human review practical only as a final-stage filter on pre-screened candidates rather than a first-pass mechanism.

\textbf{Multi-Stage Filtering Pipelines.}
In practice, the most effective quality control cascades the preceding mechanisms into a unified pipeline, using cheap automated metrics for coarse filtering, VLM scoring for semantic verification, and optional human review for final validation. This design lets each level operate at its natural cost-quality trade-off point. Step1X-Edit~\citep{liu2025step1x} exemplifies the aggressive end: filtering 20M candidates down to 1M (5\% retention) through a cascade of specialized models (Florence-2, SAM-2, Flux-Fill, ControlNet), MLLM scoring, and human validation. BLIP3o-NEXT~\citep{chen2025blip3o} similarly chains automated stages---low-resolution removal, watermark exclusion, CLIP alignment filtering, and caption quality thresholds---to progressively refine its training pool. AnyEdit~\citep{yu2025anyedit} combines instruction validation pre-filters with image quality post-filters spanning heuristic rules, perceptual metrics, and VLM assessment. At the other end, FLUX-Reason-6M~\citep{fang2025flux} retains 75\% of its 8M initial images through progressive VLM filtering, reflecting a fundamentally different philosophy. ImgEdit~\citep{ye2025imgedit} chains six stages from LAION pre-filtering through YOLO-World grounding, SAM2 segmentation, and GPT-4o evaluation. The spread in retention ratios---from 5\% to 75\%---reveals two distinct generation-filtering strategies: \emph{aggressive overgeneration} with strict multi-stage selection, which maximizes output quality but demands substantially more compute for initial generation; versus \emph{targeted generation} with lighter curation, which is more efficient but requires higher-quality generation upfront. The choice between these strategies depends on the relative cost of generation versus filtering and the quality floor required by the downstream task. Industrial pipelines add a second axis on top of retention ratio---\emph{stage-wise threshold tightening}: Wan-Image~\citep{mao2026wan} applies the same set of more than 20 scoring operators across pre-training, continual-training, and SFT stages but raises the aesthetic and clarity thresholds at each successive stage, and JoyAI-Image~\citep{jdjoyaiimage} reports IQA retention falling from roughly 34\% at 512p training to roughly 20\% at 1024p training. The temporal axis complements the per-dataset retention spread: data purity does not have a single target but a schedule that tightens as the model transitions from broad coverage to fine-grained specialization.

\subsubsection{Discussion: Scale, Quality, and Efficiency Trade-offs}
\label{sec:tradeoffs}

\textbf{The Scale--Quality Spectrum.}
The analyzed datasets span a wide operational range, from scale-first approaches (Step1X-Edit~\citep{liu2025step1x} generating 20M candidates to retain 1M, FLUX-Reason-6M~\citep{fang2025flux} producing 8M images filtered to 6M, DreamGen~\citep{jang2025dreamgen} aggregating 438M frames) to quality-first strategies (Diffusion-4K~\citep{zhang2025diffusion} curating 12K ultra-high-resolution images, GenExam~\citep{wang2025genexam} distilling 1K expert-level prompts). Empirical evidence consistently favors quality-aware scale: massive generation paired with rigorous multi-stage filtering produces superior results compared to either extreme alone. ImgEdit~\citep{ye2025imgedit} achieves its best-in-class GPT-4o score of 4.71/5 with 1.2M samples precisely because each sample survives a six-stage automated pipeline, while Step1X-Edit's~\citep{liu2025step1x} aggressive 20:1 retention ratio confirms that brute-force generation with stringent selection outperforms moderate generation with lenient filtering.

\textbf{Hybrid Approaches Dominate State of the Art.}
The most successful datasets are never purely synthetic or purely curated; they integrate multiple source types and construction methods. SEED-Data-Edit~\citep{ge2024seed} combines 3.5M automated samples with 52K real-world expert edits and 95K multi-turn sequences. UltraEdit~\citep{zhao2024ultraedit} anchors 4.1M synthetic edits to real images from COCO and Flickr to mitigate text-to-image model biases. TextAtlas5M~\citep{wang2025textatlas5m} validates mixed synthetic-real training with dramatic CER reduction (0.98$\rightarrow$0.33--0.35), demonstrating that complementary data sources yield performance gains unattainable by either alone. This convergence suggests that hybrid construction is not merely convenient but structurally necessary for robust generalization.

\textbf{Frontier Distillation as Dominant Paradigm.}
A striking pattern across the analyzed works is the reliance on frontier models as implicit teachers. ShareGPT-4o-Image~\citep{chen2025sharegpt} (91K samples) and Pico-Banana-400K~\citep{qian2025pico} (386K samples, $\sim$\$100K USD) demonstrate that distilling frontier model capabilities achieves competitive performance at modest dataset scale, effectively trading compute-time inference cost for reusable training signal. However, this paradigm implies a quality ceiling set by the teacher model and raises questions about long-term sustainability as open-source models narrow the capability gap, potentially reducing the marginal value of proprietary distillation.

\textbf{Industrial Frontier Convergence.}
Independent reports from frontier industry labs over the past year converge on three operational answers to the scale--quality--cost trilemma. First, \emph{aggressive overgeneration with strict selection} dominates over moderate generation with lenient filtering: HunyuanImage~3.0~\citep{cao2025hunyuanimage} retains under 45\% of a 10B+ raw pool, FireRed-Image-Edit~\citep{team2026firered} reduces 1.6B raw samples to roughly 100M ($\sim$6\% retention), and Step1X-Edit~\citep{liu2025step1x} sits at 5\%---collectively confirming that frontier teams treat raw-pool size as a knob to fund the curation pipeline rather than a target to maximize. Second, \emph{6B with stronger data can match 20B+ with more parameters}: Z-Image (6.15B) and LongCat-Image (6B) report competitive performance against contemporaries at 20B (Qwen-Image) and 80B-MoE (HunyuanImage), and both attribute the result to multi-granularity captioning, AIGC purging, and stage-stratified subsets rather than parameter scaling~\citep{cai2025z,team2025longcat}. Third, the field is split on \emph{whether to distill from proprietary frontier models}: most analyzed reports use such distillation as a quality ceiling, while Z-Image~\citep{cai2025z} declines on the principle that closed feedback loops induce error accumulation and homogenization---leaving the long-run quality ceiling of fully open pipelines an open question that the next generation of releases will be best positioned to answer.

\textbf{Efficiency and Diminishing Returns.}
Computational costs vary dramatically: UltraEdit~\citep{zhao2024ultraedit} achieves $\sim$100$\times$ speedup via SDXL-Turbo, while FLUX-Reason-6M~\citep{fang2025flux} requires 15,000 A100 GPU days. ShareGPT-4o-Image~\citep{chen2025sharegpt} shows that 91K frontier-quality samples can substitute for millions of lower-quality ones, suggesting diminishing returns beyond a quality-dependent threshold. Domain-specific datasets reinforce this finding from a different angle: RoboEngine~\citep{yuan2025roboengine} (3.8K images), GenExam~\citep{wang2025genexam} (1K prompts), and Diffusion-4K~\citep{zhang2025diffusion} (12K images) all achieve strong results at modest scale, confirming that specialized tasks reward depth over breadth. More broadly, AnyEdit's~\citep{yu2025anyedit} counterfactual scene generation for tail concepts demonstrates that methodological sophistication---targeted augmentation of underrepresented regions of the data distribution---amplifies scale benefits beyond what simple filtering alone can achieve.

\subsection{Evaluation and Human Preference}
\label{sec:benchmarks}

Evaluating visual generation has evolved well beyond single-number metrics like FID and IS, which capture distributional quality but reveal nothing about \emph{where} a model fails. Modern benchmarks instead decompose generation quality into distinct \textbf{cognitive ability dimensions}---compositional reasoning, world knowledge, text rendering, identity preservation, and others---each probing a specific facet of visual intelligence. This shift mirrors the move in NLP from perplexity to task-specific leaderboards, and is driven by a practical need: a model with strong overall FID may still fail catastrophically at counting objects or rendering legible text.

In parallel, \emph{how} we evaluate has also transformed: the methodology has shifted from hand-crafted heuristic metrics to VLM-based automated judges and large-scale human preference arenas. Arena-based platforms such as LM Arena and GenAI Arena collect anonymous human preferences to produce Elo-based global rankings (\Cref{subsubsec:arena})---the gold standard for overall model comparison, though unable to diagnose \emph{which} cognitive abilities drive the ranking. The dimension-specific benchmarks discussed below fill exactly this gap: they tell practitioners not just which model wins, but what cognitive abilities separate the winners from the rest. We organize the following discussion by these ability dimensions, covering both generation and editing benchmarks within each (see \Cref{sec:stress_test} for complementary stress-testing experiments along similar dimensions).

\subsubsection{Paradigm Shift: From Heuristics to VLM-as-a-Judge}
\label{subsubsec:eval_paradigm}

The evaluation methodology for visual generation has undergone a fundamental transition---from \emph{fixed heuristic metrics} that measure low-level statistical properties to \emph{VLM-based judges} that assess high-level semantic correctness.

\textbf{The Heuristic Era.}
Early evaluation relied on distribution-level metrics---FID for distributional distance, IS for quality and diversity, CLIP score for text-image alignment---complemented by task-specific measures such as DINO similarity for subject fidelity, SSIM and LPIPS for perceptual similarity in editing, and CER for text rendering accuracy. These metrics remain valuable as cheap, reproducible first-pass filters, but they share a fundamental limitation: they capture only \emph{what they were designed to measure}. A CLIP score cannot tell whether a generated molecule obeys valence rules, and SSIM cannot judge whether an edit correctly followed a complex instruction.

\textbf{The VLM-as-a-Judge Era.}
Capable vision-language models (GPT-4o, Gemini, Qwen2.5-VL) have enabled a qualitatively different paradigm: rather than computing a fixed function of pixel values, a VLM judge reads the prompt, examines the generated image, and answers specific evaluation questions---providing fine-grained, dimension-aware semantic assessment. This approach now powers most recent benchmarks, from QA-style scoring~\citep{chen2025r2ibench} to conditional question dependency graphs~\citep{OneIG-Bench} to multi-dimensional rating~\citep{ImagenWorld}. A more recent refinement is \emph{MLLM-as-judge with Chain-of-Thought reasoning}, in which the judge produces an explicit reasoning trace over a fixed set of pre-extracted key points before assigning a score---HunyuanImage~3.0's SSAE benchmark~\citep{cao2025hunyuanimage} uses this protocol over 3{,}500 manually verified key points to stabilize cross-model comparison. However, VLM judges introduce new failure modes: they can hallucinate, show systematic bias toward AI-generated content~\citep{taesiri2025understanding}, and achieve only moderate agreement with human judges (typically 0.57--0.75 Spearman). The limitation is sharpest on long-tail content---LongCat-Image~\citep{team2025longcat} explicitly falls back from MLLM scoring to PPOCRv5 for evaluating rare Chinese characters because ``MLLMs often fail on rare characters,'' a concrete instance of the broader pattern that VLM judges inherit the same blind spots as the generative models they evaluate. The current best practice is therefore a \emph{cascaded} approach---heuristic metrics for coarse filtering, VLM judges for semantic assessment, and human evaluation for calibration---mirroring the multi-stage quality control pipelines discussed in \Cref{sec:quality}.

\subsubsection{Instruction Following and Editing Fidelity}

The ability to accurately understand and execute complex instructions---whether a detailed generation prompt or a multi-step editing request---is a cross-cutting cognitive ability that underpins practical usability. This dimension evaluates not a specific visual skill, but the model's capacity for precise language grounding and faithful execution.

For generation, TIIF-Bench~\citep{TIIF-Bench} provides hierarchical evaluation spanning Basic Following (attributes, relations, reasoning), Advanced Following (combined conditions, text rendering, style control), and Designer Level prompts. Critically, it introduces prompt length robustness testing with paired short/long versions, revealing that current models struggle substantially with verbose instructions---a gap that matters for real-world use where users often write detailed, paragraph-length prompts. Complementing TIIF-Bench at the dense-prompt end, DPG-Bench~\citep{hu2024ella} contributes 1{,}000 long descriptive prompts scored across Global, Entity, Attribute, Relation, and Other categories; it has become a default reporting target across recent industrial tech reports~\citep{wu2025qwen, team2025longcat, cai2025z, team2026longcat}, where adherence scores cluster in the 84--89 range and fine-grained category breakdowns reveal that Relation and Other dimensions remain the discriminating axes among frontier models.

For editing, the benchmark landscape has evolved through three generations, each addressing limitations of its predecessor. The first generation, represented by MagicBrush~\citep{UltraEdit, AnyEdit} (10,388 annotated pairs), suffered from \emph{tool bias}---annotators used DALL-E 2's editing interface, causing models trained on it to overfit to that system's capabilities and generalize poorly elsewhere. The second generation shifted to \emph{real-world user distributions}: GEdit-Bench~\citep{Liu2025Step1XEditAP} collects 606 genuine editing instructions from platforms like Reddit, covering 11 editing categories with bilingual support and dual evaluators (GPT-4o and Qwen2.5-VL-72B) to ensure reproducibility. Its VIEScore evaluation---which uses a VLM to rate Semantic Consistency (SC) between the edit result and the instruction, alongside Perceptual Quality (PQ) of the output image---reveals these two axes as critical and largely independent dimensions. The third generation introduces \emph{difficulty stratification}: ImgEdit-Bench~\citep{ImgEdit} decomposes editing into Basic-Edit (734 cases), Understanding-Grounding-Editing (47 complex cases requiring simultaneous comprehension, localization, and manipulation), and Multi-Turn (30 cases testing content memory and version backtracking). Its three-dimensional evaluation---instruction adherence, editing quality, and detail preservation---captures the multi-faceted nature of editing fidelity, though its fine-tuned evaluator (ImgEdit-Judge) achieves only approximately 70\% alignment with human judgments, highlighting the need for better automated evaluation.

Beyond instruction-based editing, EditWorld~\citep{EditWorld} tests instruction following for world-aware edits, categorizing instructions into seven types including Long-Term temporal changes, Story-Type narratives, and Exaggeration, and evaluates through both CLIP score and a VLM-based MLLM Score. ByteMorph-Bench~\citep{ByteMorph} introduces CLIP-Dimg---which measures the cosine similarity between the directional change in CLIP image embeddings (source$\to$target) and a reference direction, complementing the standard text-based CLIP-Dtxt---to evaluate non-rigid motion editing across 613 manually curated samples.

A fourth generation has emerged in the past year as industrial tech reports scale the GEdit-Bench / ImgEdit-Bench formula to bilingual coverage and broader task taxonomies. LongCat-Image's CEdit-Bench~\citep{team2025longcat} extends GEdit-Bench with reference-image generation, structural modification, and viewpoint transformation tasks that the earlier benchmark omits, providing 1{,}464 bilingual editing pairs across 15 categories. FireRed's REDEdit-Bench~\citep{team2026firered} pushes further to 1{,}673 bilingual pairs with 15 categories that include beautification and low-level enhancement, and---uniquely---introduces a dual OCR + VLM-Judge metric for the text-editing subset that scores OCR fidelity (Levenshtein distance + word accuracy) alongside VLM judgments on SuccessEdit, OverEdit, Style, and Consistency. The latter design directly addresses a reward-hacking failure mode in earlier text-editing evaluation, where models could inflate OCR scores by rendering oversized characters that disrupted layout but parsed cleanly. Both benchmarks position themselves as complementary to GEdit-Bench rather than replacements, with shared evaluators (GPT-4o or Gemini-class VLMs) enabling direct cross-benchmark comparison.

\subsubsection{Identity Preservation and Personalization}

Subject-driven generation and editing require the model to maintain visual identity---preserving the distinctive appearance of a specific person, object, or character across different contexts, poses, and prompts. This ability is central to personalization applications such as customized product imagery, character-consistent storytelling, and portrait editing.

DreamBench~\citep{ruiz2023dreambooth} established the foundational evaluation protocol using DINO scores for subject fidelity and CLIP-T scores for text alignment, measuring whether generated images preserve the reference subject's identity while following new textual instructions. DreamBench++~\citep{peng2024dreambench++} extends this with broader concept coverage and more challenging composition scenarios that stress the editability-fidelity trade-off. MultiRef-bench~\citep{MultiRef} addresses a critical gap by evaluating multi-reference generation across 10 reference types (bounding box, mask, pose, caption, subject, semantic map, depth, canny, sketch, art style) spanning 33 combinations. Even the best unified model (OmniGen) achieves only 66.6\% accuracy on synthetic samples, exposing the difficulty of handling multiple conditional inputs simultaneously.

In editing, identity preservation manifests as the tension between faithfully executing an edit instruction and maintaining unedited regions. ByteMorph-Bench~\citep{ByteMorph} directly quantifies this trade-off across 613 non-rigid motion editing samples, finding that proprietary models (GPT-4o, Gemini-2.0-flash) excel in instruction-following but struggle with identity preservation, while models like SeedEdit~\citep{wang2025seededit} maintain stronger visual consistency at the cost of reduced instruction fidelity. Recent industrial human-evaluation protocols converge on the same separation, treating instruction-following and consistency as orthogonal dimensions reported side by side rather than fused into a single score~\citep{seedream2025seedream, team2025longcat, team2026firered, mao2026wan}; their cross-paper findings echo the ByteMorph picture, with proprietary models (GPT-Image-1, Nano Banana Pro) leading on instruction-following and open-source models trailing on the same axis but often tying or leading on consistency. This \textbf{instruction fidelity vs. identity preservation trade-off} is a defining tension for the field, and current benchmarks suggest no model has satisfactorily resolved it (see Dimension~IV in \Cref{sec:stress_test} for stress tests on temporal consistency and identity preservation).

\subsubsection{Compositional Understanding and Spatial Reasoning}

Compositional understanding---the ability to correctly bind attributes to objects, count entities, and arrange them in specified spatial relationships---is arguably the most fundamental cognitive ability for controllable generation. A model that cannot reliably place ``a red cube to the left of a blue sphere'' fails at the most basic level of instruction following, regardless of its aesthetic quality.

DrawBench~\citep{saharia2022imagegen}, introduced alongside Imagen, was among the first benchmarks to probe this ability through curated challenging prompts spanning attribute binding, counting, spatial relations, and unusual concept combinations. GenEval~\citep{GenEval} formalized compositional evaluation into six structured dimensions (single objects, two objects, counting, colors, spatial positioning, and attribute binding), though even the best models at the time achieved only 0.61 overall accuracy. T2I-CompBench~\citep{huang2023t2i} extended this line with emphasis on 2D/3D spatial relationships, but suffered from high semantic redundancy (less than 30\% unique prompts after deduplication) and fixed short prompt lengths that fail to capture real-world prompt diversity.

On the editing side, EditWorld~\citep{EditWorld} evaluates spatial reasoning through its Spatial-Trans category, which requires models to perform viewpoint shifts while maintaining scene consistency. ByteMorph-Bench~\citep{ByteMorph} further probes spatial understanding through camera zoom and camera motion editing, where the model must infer how spatial configurations transform under viewpoint changes. Pushing further into geometrically verifiable supervision, JoyAI's SpatialEdit-Bench~\citep{xiao2026spatialedit} decomposes spatial editing into object-level manipulation (Moving Score, Rotation Score) and camera-level control (Viewpoint Error, Framing Error), with ground-truth transformations rendered through Blender so that errors can be measured against known 3-DOF camera or object poses rather than judged perceptually. The reported separation is sizable---JoyAI-Image-Edit attains a Camera Overall Error of 0.429 versus LongCat-Image-Edit's 0.743~\citep{jdjoyaiimage}, evidence that geometry-aware editing is now a distinct capability from general edit fidelity rather than a special case of it. Across both generation and editing, a consistent finding emerges: \textbf{counting and precise spatial positioning remain the weakest dimensions} even for state-of-the-art models, suggesting that current architectures lack robust mechanisms for discrete, relational reasoning (see Dimension~I in \Cref{sec:stress_test} for case studies).

\subsubsection{World Knowledge and Physical Reasoning}

Beyond compositional arrangement, a capable generation model must ground its outputs in real-world knowledge---producing images where shadows obey light sources, chemical structures follow valence rules, and cultural symbols carry correct meaning. This dimension tests whether models have internalized an implicit world model, or merely learned surface-level correlations from training data.

PhyBench~\citep{meng2024phybench} isolates physical commonsense across four domains---mechanics, optics, thermodynamics, and material properties---covering 31 physical scenarios. Using GPT-4o-based scoring calibrated against human judgments, it reveals persistently low performance across all tested models, indicating that physical reasoning remains a fundamental weakness. WISE~\citep{niu2025wise} broadens the scope to world knowledge more generally, spanning Cultural Common Sense, Spatio-Temporal Reasoning, and Natural Science across 25 fine-grained subdomains. Its WiScore metric---a weighted composite of knowledge-image consistency (assessed by a VLM verifying whether factual details are visually correct), visual realism, and aesthetic quality---reveals a clear knowledge gap: most of the 20 evaluated models fail to reach a satisfactory threshold of 0.6. The same gap holds at the frontier: LongCat-Image~\citep{team2025longcat} reports a WISE overall of 0.65, the strongest among open-source diffusion models, yet still trails Seedream~4.0's 0.78~\citep{seedream2025seedream}, with Chemistry as the weakest domain across both.

GenExam~\citep{GenExam} takes a unique exam-oriented approach, testing multidisciplinary understanding across 10 academic subjects (mathematics, physics, chemistry, biology, geography, computer science, music, linguistics, economics, architecture). By separating semantic correctness from visual plausibility, it reveals that open-source models score below 5\% under strict evaluation, compared to 72.7\% for the best closed-source model, suggesting fundamental rendering limitations beyond mere knowledge deficiencies. R2I-Bench~\citep{chen2025r2ibench} provides the most fine-grained decomposition with seven reasoning categories---commonsense, compositional, concept mixing, logical, numerical, mathematical, and causal---across 32 subcategories. Its R2I-Score employs a QA-style protocol: for each generated image, a set of instance-specific questions is posed to a VLM judge, and the score aggregates correctness across text-image alignment, reasoning accuracy, and image quality. Mathematical and causal reasoning prove universally the most challenging, with most open-source models scoring below 0.45 versus 0.77 for GPT-Image-1.

For editing, EditWorld~\citep{EditWorld} complements these generation benchmarks through its Physical-Trans (structural transformations obeying physical laws) and Implicit-Logic (edits requiring logical reasoning) categories, demonstrating that world knowledge is equally critical when modifying existing images. The consistent finding across all these benchmarks is a \textbf{substantial gap between closed-source and open-source models} on knowledge-intensive tasks, with physical and causal reasoning representing the frontier of difficulty (see Dimension~II in \Cref{sec:stress_test} for case studies on physical reasoning failures).

\subsubsection{Text Rendering}

Rendering legible, correctly spelled text within generated images has been one of the most persistent challenges in visual generation. Unlike other visual elements that tolerate approximate representations, text demands character-level precision---a single wrong letter makes the output unacceptable. This dimension is critical for practical applications ranging from poster design to document generation.

TextAtlasEval~\citep{TextAtlas5M} provides the most comprehensive assessment with 4,000 test samples spanning four domains of increasing difficulty: CleanTextSynth (text on white backgrounds), StyledTextSynth (text in complex scenes), TextScenesHQ (diverse real-world scenarios), and TextVisionBlend (interleaved image-text content). Character-level evaluation reveals large performance gaps: GPT-4o achieves 60.69--82.88\% word-level accuracy with 0.32--0.36 Character Error Rate (CER), while open-source baselines struggle at 0.11--2.98\% accuracy with 0.83--0.99 CER---an order-of-magnitude gulf that underscores text rendering as a key differentiator between model tiers. LongText-Bench~\citep{X-Omni} further stresses this ability with prompts containing 30--50 English words or 60+ Chinese characters, exposing limitations of even autoregressive architectures in maintaining coherent long text.

Two finer-grained text-rendering benchmarks have emerged from the industrial frontier and now serve as default reporting targets across recent tech reports. CVTG-2K~\citep{tai2025investigating} probes \emph{multi-region} English text rendering with 2{,}000 prompts each requiring 2--5 distinct text regions in the same image, scored jointly by Word Accuracy, Normalized Edit Distance, and CLIPScore; reported numbers from frontier reports cluster in the 0.83--0.87 word-accuracy range~\citep{wu2025qwen, cai2025z, team2025longcat, jdjoyaiimage}, with all systems degrading as the region count grows. ChineseWord, introduced by Qwen-Image~\citep{wu2025qwen} with 8{,}105 prompts grouped into three frequency tiers (3{,}500 / 3{,}000 / 1{,}605 characters drawn from the Ministry of Education's General Standard Chinese Characters), evaluates character-level accuracy on a single character per image; LongCat-Image~\citep{team2025longcat} extends the protocol to cover the full 8{,}105-character set and switches the evaluator from MLLM judging to PPOCRv5, on the explicit observation that VLM judges fail on rare characters. Together with TextAtlasEval and LongText-Bench, these benchmarks span the practical text-rendering matrix: short bilingual single characters (ChineseWord), long multi-line blocks (LongText-Bench, TextAtlasEval), and multi-region typography (CVTG-2K). For design-oriented bilingual layouts, GlyphDraw2~\citep{ma2025glyphdraw2} contributes a Poster-Set of 200 prompts in design contexts and a Complex-Set drawing from 2{,}000 frequent Chinese characters in random combinations, isolating the glyph-rendering axis from broader scene composition.

On the editing side, GEdit-Bench~\citep{Liu2025Step1XEditAP} includes text modification as one of its 11 editing categories, testing whether models can alter existing text in images while preserving surrounding visual context. Across generation and editing, \textbf{text rendering remains the weakest single dimension} for nearly all evaluated models~\citep{OneIG-Bench}, making it a critical bottleneck for real-world deployment (see Dimension~III in \Cref{sec:stress_test} for case studies on visual-textual integration).

\subsubsection{Visual Quality and Resolution}

As generation models move toward high-resolution outputs for professional use, traditional quality metrics designed for 256$\times$256 or 512$\times$512 images become inadequate. Ultra-high-resolution synthesis introduces new failure modes---texture degradation, repetitive patterns, and loss of fine details---that require dedicated evaluation frameworks.

Diffusion-4K~\citep{Diffusion-4K} pioneers 4K-resolution evaluation through two novel metrics: GLCM Score, which measures texture richness via the Gray Level Co-occurrence Matrix (SRCC 0.75 with human preferences), and Compression Ratio, which assesses fine detail preservation via JPEG compression resistance (SRCC 0.53). These resolution-aware metrics substantially outperform conventional quality metrics such as MUSIQ (SRCC 0.36) and MANIQA (SRCC 0.20) at 4K resolution, validating that standard metrics fail to capture what humans care about in high-resolution outputs. For editing quality assessment, EditReward-Bench~\citep{wu2025editreward} introduces multi-way preference comparison (2-way, 3-way, 4-way) beyond traditional pairwise evaluation, with 500 groups annotated by 3 independent experts. The strict all-or-nothing evaluation reveals that even the best evaluator models achieve only 38.42\% overall accuracy, demonstrating that \textbf{fine-grained quality discrimination remains an unsolved problem} for both generative models and their evaluators.

\subsubsection{Holistic and Cross-Dimensional Benchmarks}

While dimension-specific benchmarks diagnose individual abilities, several recent efforts aim to provide unified evaluation across multiple cognitive dimensions simultaneously. These holistic benchmarks are valuable for overall model comparison and for identifying unexpected correlations between abilities.

Two recent benchmarks represent contrasting design philosophies for multi-dimensional evaluation. OneIG-Bench~\citep{OneIG-Bench} covers six evaluation tracks with 1,120 prompts (semantic alignment, text rendering, knowledge and reasoning, stylization, diversity, and multilingualism) and adopts a \emph{conditional QA} approach: GPT-4o generates a dependency graph of evaluation questions for each image, and Qwen2.5-VL-7B answers them with conditional validation---if a prerequisite question fails, dependent questions are automatically marked incorrect. This fine-grained protocol shows superior correlation with human preferences compared to coarse-grained scoring, revealing that even state-of-the-art closed-source models achieve only 52--66\% accuracy on semantic alignment. PRISM-Bench~\citep{FLUX-Reason-6M_PRISM-Bench} takes a different approach with seven specialized tracks (Imagination, Entity, Text Rendering, Style, Affection, Composition, and Long Text), employing \emph{dual-axis scoring} that evaluates each track along both prompt-image alignment and aesthetic quality independently---capturing the insight that a well-aligned but ugly image and a beautiful but misaligned one represent different failure modes. Prompts are selected via K-Means clustering with human-in-the-loop refinement to ensure representative coverage. Notably, closed-source models (GPT-Image-1: 86.3, Gemini2.5-Flash-Image: 85.3) substantially outperform open-source alternatives (Qwen-Image: 79.9, SEEDream 3.0: 79.6), with the Long Text track proving universally challenging.

ImagenWorld~\citep{ImagenWorld} provides the first truly unified framework spanning both generation and editing, covering six task types: Text-guided Image Generation/Editing and Single/Multiple Reference Image Generation/Editing. Across 3.6K condition sets, it evaluates Prompt Relevance, Aesthetic Quality, Content Coherence, and Artifacts, with additional object-level error attribution via Set-of-Mark segmentation. VLM evaluation (Gemini-2.5-Flash) achieves Spearman correlation of 0.57--0.70 with human judgments, though object-level explainability remains a capability exclusive to human annotators. A key finding is that \textbf{editing tasks are consistently harder than generation} by approximately 0.1 points across all models, and closed-source systems (GPT-Image-1: 0.91) maintain a substantial lead over open-source alternatives.

In parallel, industrial labs have built their own holistic benchmarks to address three gaps left by the public ones: lag behind frontier capabilities, English-only coverage, and absence of productivity scenarios. HunyuanImage~3.0's SSAE~\citep{cao2025hunyuanimage} decomposes 500 prompts into 3{,}500 manually verified key points across 12 fine-grained semantic fields and locks the key-point set across all model comparisons, ensuring that cross-model differences reflect generation quality rather than evaluator drift. Seedream~4.0's MagicBench~4.0~\citep{seedream2025seedream} explicitly bundles three task families---T2I (325 prompts), single-image editing (300 prompts), and multi-image editing (100 prompts)---with bilingual coverage, while its companion DreamEval introduces \emph{tiered difficulty} (Easy / Medium / Hard) over 1{,}600 prompts and 128 sub-tasks, scored through fine-grained VQA so that interpretability and headline numbers come from the same protocol. Seedream~3.0's earlier Bench-377~\citep{gao2025seedream} contributes the productivity dimension explicitly, partitioning its 377 prompts across cinematic, arts, entertainment, aesthetic-design, and \emph{practical-design} (Prac-Design) scenarios---the last of which targets icon arrangements in slides and illustration design in handwritten newspapers, content typically absent from research-oriented benchmarks. Although these suites are internally curated and not always publicly released, their methodological design choices---bilingual prompts, multi-task bundling, productivity scenarios, fixed key-point sets---are increasingly visible in the public benchmarks discussed above.

\begin{highlightbox}{Community Message: Proxy Capabilities Are the Default Training-Progress Indicators}

Direct measurement of overall gen/edit quality is increasingly unreliable: FID and CLIP-score have effectively saturated at the frontier, and pairwise human evaluation does not scale to weekly training-progress checks. The community has therefore implicitly converged on a set of \emph{proxy capabilities} that are easy to measure, hard to game, and correlate well with overall capability lift: text rendering (English and Chinese)\footnote{The Nano Banana team explicitly describes text rendering as a training-progress indicator in their public interview: \url{https://youtu.be/H6ZXujE1qBA?si=SdeAnvCwRTTY90oq}.}, long-form text rendering, and structured-content rendering (flowcharts, chemical formulas, electrical schematics, mathematical layouts). The signal is concrete: JoyAI-Image's~\citep{jdjoyaiimage} LongText-Bench EN/ZH score of 0.963 and LongCat-Image's~\citep{team2025longcat} 90.7 on ChineseWord versus Seedream~4.0's~\citep{seedream2025seedream} 58.5 (a 30-plus-point gap in a single year) track real underlying-capability shifts that bulk perceptual metrics miss. The takeaway: when designing internal evaluation harnesses for a frontier image-generation training run, instrument these proxies first---they catch regressions and improvements weeks before bulk preference studies do.
\end{highlightbox}

\subsubsection{Domain-Specific Benchmarks}

Beyond general-purpose evaluation, several benchmarks target specialized application domains where visual generation serves as a component in a larger pipeline.

\paragraph{Robotics and Manipulation.} Generated and edited images serve as data augmentation for robot learning. RoboSeg~\citep{RoboEngine} provides 3,800 annotations for robot scene segmentation, enabling physics-aware augmentation that delivers 210\% improvement over no-augmentation baselines. Gen2Sim~\citep{Gen2Sim} evaluates image-to-3D asset generation for robot simulation, demonstrating successful sim-to-real policy transfer. NICE~\citep{NICE} introduces manipulation robustness evaluation across three clutter levels (864 real-world trials), while RoboPearls~\citep{RoboPearls} and Masquerade~\citep{lepert2025masquerade} benchmark scene editing for task generalization, with Masquerade showing that edited human video pre-training achieves 74\% success versus 12\% for standard pre-training.

\paragraph{Creative and Artistic Generation.} Layer-Bench~\citep{ART_plus_multi-layer} evaluates transparent image generation across 1,500 prompts using TIPS and HPSv2 metrics, with user studies (20+ participants) revealing 57--59\% win rates for the best methods over baselines.

\paragraph{Synthetic Media Detection.} As generated images become more realistic, detecting them grows critical. DRAGON~\citep{DRAGON} benchmarks this with 150K images across 25 generative models, revealing robustness challenges: JPEG compression degrades DIRE accuracy from 0.916 to 0.745 at quality factor 30. REALEDIT~\citep{REALEDIT} focuses on edited image detection in the wild, achieving 4.87\% F1 improvement through training on real editing data from Reddit.

\paragraph{Industrial Productivity Scenarios.} A distinct domain has crystallized as image-generation models target professional design workflows. LongCat-Image's Poster\&SceneBench~\citep{team2025longcat} evaluates poster typography and natural scene text on 500 prompts that span signage on textured surfaces and shop fronts under complex lighting. Wan-Image~\citep{mao2026wan} introduces an Image Series Generation benchmark covering 13 application categories (fashion, e-commerce, slide presentations, film storyboarding, architecture, and others), evaluating up to 12 still images forming a coherent series rather than video, alongside an Interactive Image Editing benchmark targeting precise visual-cue-driven local edits. The shared methodological shift is from relative win-rate scoring to absolute Pass Rate---the proportion of outputs deemed satisfactory in isolation---reflecting deployment requirements where a model must clear an absolute usability bar regardless of how it ranks against competitors.

\subsubsection{Emerging Evaluation Frontiers}
\label{subsubsec:emerging_eval}

Despite rapid progress, the current benchmark ecosystem has significant blind spots that fail to capture several emerging capabilities of frontier models.

\paragraph{Structured Visual Content.} State-of-the-art models can now generate flowcharts, circuit diagrams, chemical structural formulas, and mathematical notation---content that demands both domain knowledge and precise spatial layout. Yet no established benchmark systematically evaluates the correctness of such outputs. Evaluating a generated chemical structure requires verifying valence constraints and bond angles; evaluating a flowchart requires checking logical consistency of node connections. These tasks demand domain-specific evaluation protocols that go well beyond prompt-image alignment.

\paragraph{Evaluation Methodology Gaps.} Current VLM-as-judge pipelines, while scalable, show only moderate correlation with human preferences (typically 0.57--0.75 Spearman). More concerning, VLM judges can exhibit systematic biases: PSR~\citep{taesiri2025understanding} finds that when comparing AI-generated edits against human-made edits, VLM evaluators achieve only 0.14--0.25 Cohen's $\kappa$ agreement with human raters, and show a consistent bias toward preferring AI outputs---likely because VLMs share similar visual priors with the generative models they evaluate. The field lacks standardized protocols for when VLM evaluation is sufficient versus when human evaluation is necessary, and for how to calibrate automated metrics against human judgments across different cognitive dimensions.

\paragraph{Bilingual Coverage as a First-Class Dimension.} Frontier industrial reports now routinely report Chinese-track results (ChineseWord, GEdit-Bench-CN, OneIG-ZH, LongText-Bench-ZH, CEdit-Bench-CN, REDEdit-Bench-CN), and the cross-language gap is itself diagnostic---models that lead on English alignment can lag substantially on Chinese~\citep{wu2025qwen, team2025longcat, team2026firered}. Yet most academic benchmarks remain English-only, and there is no widely adopted protocol for how to weight bilingual performance into a composite score, leaving a systematic blind spot for any model whose deployment markets include Chinese-speaking users.

We discuss potential solutions to these evaluation challenges---including domain-expert-in-the-loop evaluation and compositional correctness verification---in \Cref{sec:frontier}.

\subsubsection{Arena-Based Human Preference Evaluation}
\label{subsubsec:arena}

All benchmarks discussed above evaluate specific cognitive dimensions through automated pipelines, but the ultimate measure of a generative model is whether real users prefer its outputs. \textbf{Arena-based evaluation} addresses this by collecting large-scale anonymous human preferences, providing the closest proxy to real-world user satisfaction.

\paragraph{Mechanism.} Arena platforms present users with a blind side-by-side comparison: given a prompt, two models generate images whose identities are hidden, and the user votes for the preferred output (or declares a tie). The accumulated votes are aggregated into a global ranking via the \textbf{Bradley-Terry model}~\citep{ChatbotArena}, which estimates a latent strength parameter $r_i$ for each model $i$ by maximizing the likelihood of all observed pairwise outcomes:
\begin{equation}
    P(i \succ j) = \frac{\exp(r_i)}{\exp(r_i) + \exp(r_j)},
\end{equation}
where the resulting scores are typically reported as Elo ratings. Bootstrap resampling provides confidence intervals to ensure that ranking differences are statistically significant rather than artifacts of sampling noise.

\paragraph{Major Platforms.} LM Arena~\citep{ChatbotArena} extends the widely-used Chatbot Arena infrastructure to text-to-image and image editing leaderboards, with millions of votes across 50+ models, making it the largest-scale visual generation arena. GenAI Arena~\citep{GenAI-Arena} provides an open evaluation platform covering text-to-image generation and image editing with Elo-based rankings across 35+ open-source models. Two further platforms have become standard reporting targets in industrial tech reports. Artificial Analysis Image Arena~\citep{aaarena2025} runs an independent third-party leaderboard for both text-to-image and image-editing tracks and---unlike the others---publishes \textbf{inference cost} (USD per 1{,}000 generated images) alongside Elo, exposing the cost--quality Pareto frontier directly~\citep{seedream2025seedream, gao2025seedream, cai2025z}. Alibaba AI Arena~\citep{wu2025qwen}, introduced together with Qwen-Image, requires each model to participate in at least 10{,}000 anonymous pairwise comparisons before reporting and draws votes from over 200 evaluators across professional backgrounds, with explicit dimensions for general image quality, subject accuracy, style adherence, and photographic perspective diversity.

\paragraph{Structured Pairwise Human Evaluation.} Beyond public arenas, frontier tech reports increasingly run their own structured pairwise or trinary protocols that complement arenas at smaller scale but with controls public platforms cannot enforce. The shared protocol is \emph{Good--Same--Bad} (GSB) or its scoring variants---Win Score = (Wins + 0.5$\times$BothGood) / Total~\citep{mao2026wan}, and absolute Pass Rate---computed over a fixed prompt set (typically 100--1{,}000 prompts) with single-run inference per model to forbid cherry-picking, professional evaluators (HunyuanImage~3.0 reports more than 100), and explicit dimension separation along instruction following, consistency, structural integrity, and text-editing performance~\citep{cao2025hunyuanimage, seedream2025seedream, team2025longcat, team2026firered, mao2026wan}. The trade-offs are the inverse of public arenas: smaller scale and self-conducted evaluation introduce risk of selection bias, but controlled inference conditions and pre-registered dimensions yield diagnostic information that public Elo scores conceal---HunyuanImage~3.0's reported $+1.17\%$ relative win rate over Seedream~4.0 and Seedream~4.0's $\sim$$20\%$ GSB advantage over GPT-Image-1 in multi-image editing illustrate the resolution this protocol can achieve. The two evaluation modes are now best read as complementary: public arenas establish overall standing, while structured pairwise protocols isolate specific capability differences.

\paragraph{Value and Limitations.} Arena rankings serve as the de facto \textbf{gold standard} for overall model comparison---they directly reflect aggregated user preferences without relying on proxy metrics or VLM judges. However, they have fundamental limitations that make them complementary to, rather than a replacement for, dimension-specific benchmarks. First, arenas \textbf{cannot scale} to serve as development-time evaluation: each data point requires a human vote, making rapid iteration impossible. Second, arenas provide \textbf{no diagnostic information}---a model may rank highly overall while failing silently on specific abilities such as text rendering or spatial reasoning. Third, arena rankings can be \textbf{biased by prompt distribution}: the mix of easy and hard prompts submitted by users may not uniformly cover all cognitive dimensions, potentially masking systematic weaknesses. For these reasons, the dimension-specific benchmarks discussed in the preceding sections remain essential for understanding \emph{where} and \emph{why} models succeed or fail.

\subsection{Infrastructure and Ecosystem: Systematizing Visual Generation}
\label{subsec:infrastructure_systems}

As visual generation models transition from traditional U-Nets to Diffusion Transformers (DiTs) and unified autoregressive architectures, the computational bottleneck has fundamentally shifted. The challenge is no longer merely optimizing localized convolutional operators, but managing memory walls, communication overhead, and token scheduling at an unprecedented scale. Modern visual generation has evolved into a rigorous systems engineering problem. In this section, we review the underlying infrastructure primitives driving this shift, followed by the open-source ecosystem that democratizes these capabilities.

\subsubsection{Scaling the Context: Sequence Parallelism and Load Balancing}
\label{subsubsec:infra_scaling_context}
In the era of DiTs and unified multimodal models (e.g., Transfusion, OmniGen), a single high-resolution image or short video is patchified into massive sequences ranging from $10^5$ to over $10^6$ tokens. At this scale, traditional Data Parallelism (DP) and Tensor Parallelism (TP) quickly exhaust GPU memory. Modern infrastructure must incorporate \textbf{Sequence Parallelism (SP)}, leveraging communication-efficient mechanisms such as DeepSpeed-Ulysses~\citep{jacobs2023deepspeed} or Ring-Attention~\citep{liu2023ring} to distribute the extreme sequence context across multiple devices.

Furthermore, unlike the linear nature of text LLMs, visual models frequently process inputs with dynamic resolutions (e.g., jointly training on $512^2$ and $1024^2$ images). Naive padding strategies can result in up to 50\% computational waste. To address this, state-of-the-art visual infrastructure adopts \textbf{Sequence Packing} (or \emph{Patch n' Pack})~\citep{dehghani2023patch} techniques, dynamically binning and routing patches of varying resolutions into a unified batch to maintain near-optimal Model Flops Utilization (MFU) without compromising spatial integrity.

\subsubsection{Orchestrating Visual RL: The Memory Wall in Alignment}
\label{subsubsec:infra_visual_rl}
Post-training alignment via algorithms like PPO~\citep{schulman2017proximal} or GRPO~\citep{shao2024deepseekmath} introduces a severe memory wall, which is significantly more taxing in the visual domain than in text. Visual RL requires maintaining massive Actor and Reference models in memory while simultaneously caching gradients across multi-step denoising trajectories.

To overcome this, infrastructure designs are moving toward highly decoupled architectures. Key system-level solutions include efficient weight synchronization between dedicated training and inference nodes, aggressive CPU offloading for optimizer states, and utilizing high-throughput inference backends (such as vLLM~\citep{kwon2023efficient} or SGLang~\citep{zheng2024sglang}) to dramatically accelerate the Actor's generation \emph{rollout} phase. This separation of concerns ensures that GPU compute is not left idle during the computationally expensive reward-scoring and generation steps.

Beyond decoupling the generation and scoring stages, a recent line of work pushes the boundary of what is jointly optimized even further. \textbf{PromptRL}~\citep{wang2026promptrl} inserts a trainable language model as a prompt-refinement module inside a flow-based RL loop, so that prompt adaptation and generator alignment are co-optimized rather than treated as separate stages. This design illustrates a broader trend in visual RL infrastructure: as the system-level memory and throughput bottlenecks are progressively tamed, the scope of end-to-end optimization keeps expanding beyond the generator itself.

\subsubsection{Production-Grade Serving and Acceleration}
\label{subsubsec:infra_serving}
Transitioning these foundation models to production shifts the focus toward high-throughput and low-latency serving. Modern visual serving engines are beginning to adopt optimizations originally pioneered for LLMs, but tailored for spatial-temporal data. This includes \textbf{Continuous Batching}~\citep{yu2022orca} for asynchronous request handling, advanced \textbf{KV-Cache management} (e.g., adapting PagedAttention~\citep{kwon2023efficient} to handle cross-modal prefixes in unified generation models), and hardware-aware \textbf{Operator Fusion} via Triton~\citep{tillet2019triton} or custom CUDA kernels to minimize memory fragmentation and I/O overhead. A concrete realization of this direction is \textbf{SGLang Diffusion}~\citep{sglang-diffusion-2026}, which extends the SGLang LLM engine with Unified Sequence Parallelism (USP)---a composition of Ulysses-SP and Ring-Attention---alongside CFG-parallelism and Tensor Parallelism, so that AR and diffusion workloads share a single optimized runtime. Reported results show $1.2\times$--$5.9\times$ speedups over standard baselines on representative image and video backbones (Flux, Hunyuan, Wan), delivered behind an OpenAI-compatible API that makes the engine drop-in for production deployments.

\subsubsection{The Open-Source Ecosystem and Toolchains}
\label{subsubsec:infra_ecosystem}
Built atop these underlying system primitives is a rich ecosystem of modular toolchains that democratizes visual generation research. Rather than writing monolithic training scripts, researchers now rely on specialized frameworks categorized by their operational focus:
\begin{itemize}
    \item \textbf{Omni-modal Training Frameworks:} Systems like \textbf{VeOmni}~\citep{ma2025veomni} provide model-centric distributed recipes, allowing researchers to easily compose heterogeneous encoders and language backbones across parallel execution environments.
    \item \textbf{RL and Post-Training Toolboxes:} \textbf{UniRL-Zero}~\citep{wang2025unirl} extends RL beyond single-generator optimization to \emph{unified models} with joint language-model and diffusion-model experts, and integrates a broad reward stack---aesthetic scorers, fidelity models (ImageReward~\citep{xu2023imagereward}, PickScore~\citep{kirstain2023pick}, HPS~\citep{wu2023human,ma2025hpsv3widespectrumhumanpreference}), and utility metrics (PaddleOCR~\citep{cui2025paddleocr30technicalreport} for text rendering, EditReward~\citep{wu2025editreward} for instruction following)---as pluggable backends so that reward signals can be swapped without touching the training loop. \textbf{Flow-Factory}~\citep{ping2026flowfactory} complements this with a registry-based design that cleanly decouples algorithms, models, and reward functions, supporting methods such as GRPO, DiffusionNFT, and AWM on flow-matching backbones. Pushing this composability toward full generality, Tencent-Hunyuan's \textbf{UniRL} framework~\citep{unirl_github} applies a single RL post-training loop---generate, score, estimate advantages, update, and sync weights back to rollout workers---uniformly across diffusion, autoregressive, prompt-enhancer, and unified AR+diffusion models, treating model families and RL algorithms (e.g., GRPO, DiffusionNFT, DanceGRPO) as independent, composable dimensions atop a shared distributed runtime (Ray, FSDP, and high-throughput rollout via vLLM/SGLang).
    \item \textbf{Base Libraries:} At the foundational layer, \textbf{Hugging Face Diffusers}~\citep{von-platen-etal-2022-diffusers} and \textbf{MMagic}~\citep{mmagic2023} remain the standard modular toolboxes, exposing schedulers, pipelines, and pretrained building blocks in a highly recombinable format---the essential substrate on which custom research and higher-level frameworks are built. Complementing these, \textbf{DiffSynth-Studio}~\citep{diffsynth-studio} redesigns the training and inference pipelines for mainstream open-weight diffusion model families (FLUX~\citep{labs2025flux1kontextflowmatching}, Qwen-Image~\citep{wu2025qwen}, Z-Image~\citep{cai2025z}), adding a richer set of training modalities such as FP8 training, end-to-end distillation, two-stage split training, and differential LoRA---targeted at the engineering needs of recent open releases rather than generic schedulers.
\end{itemize}

\section{Applications and Evolving Frontiers}
\label{sec:applications}

\begin{figure}[!t]
  \centering
  \resizebox{0.93\linewidth}{!}{\hbadness=10000\hfuzz=200pt%
\begin{forest}
for tree={
  grow'=east,
  forked edges,
  draw=black!35,
  rounded corners=2pt,
  align=left,
  font=\sffamily\small,
  inner xsep=4pt,
  inner ysep=3pt,
  l sep=8pt,
  s sep=6pt,
  edge={thick, gray!60},
  parent anchor=east,
  child anchor=west,
  anchor=west,
  base=left,
  edge path={
    \noexpand\path [\forestoption{edge}]
      (!u.parent anchor) -- +(4pt,0) |- (.child anchor)\forestoption{edge label};
  },
},
root style/.style={
  fill=gray!15,
  draw=gray!60,
  text=black,
  font=\sffamily\bfseries\normalsize,
  minimum width=3.3cm,
  align=center,
  inner xsep=6pt,
  inner ysep=5pt,
},
level 1/.style={
  font=\sffamily\bfseries\small,
  minimum width=3.2cm,
  align=center,
  inner xsep=5pt,
  inner ysep=4pt,
},
level 2/.style={
  fill=white,
  draw=black!25,
  text=black,
  line width=0.9pt,
  font=\sffamily\bfseries\footnotesize,
  minimum width=2.5cm,
  align=center,
  inner xsep=4pt,
  inner ysep=3pt,
},
level 3/.style={
  fill=white,
  draw=black!25,
  text=black,
  font=\fontsize{6.2}{7.0}\selectfont,
  line width=0.6pt,
  edge={black!35},
  align=left,
  text width=8.2cm,
  inner xsep=6pt,
  inner ysep=6pt,
  anchor=west,
}
    [Applications \\ (\Cref{sec:applications}), root style
      [Fine-grained Controllability \\ (Structural Scaffolding),
        level 1, fill=BlueGreen!15, draw=BlueGreen!70, edge={thick, BlueGreen!70}
          [Spatial \& Relational, level 2
              [{\textbf{Layout:} ReCon~\citep{ReCon}, CreatiLayout~\citep{CreatiLayout} \\
              MIGLoRA~\citep{MIGLoRA}, MOSAIC~\citep{MOSAIC} \\
              FROSS~\citep{FROSS}, HybridLayout~\citep{wu2025hybrid} \\
              \textbf{Geometric Primitives:} ControlNet++~\citep{controlnet++}, \\
              OmniRefiner~\citep{OmniRefiner}, RichControl~\citep{RichControl}}, level 3]
          ]
          [Disentangled Synthesis, level 2
              [{\textbf{Concept-centric personalization:} \\
              Chimera~\citep{Chimera}, DreamBooth~\citep{ruiz2023dreambooth} \\
              \textbf{Style-content separation:} \\
              StyleShot~\citep{StyleShot}, AIComposer~\citep{AIComposer} \\
              OminiControl~\citep{OminiControl}, Easycontrol~\citep{Easycontrol} \\
              \textbf{Multi-view/3D consistency:} \\
              Mv-adapter~\citep{Mv-adapter}, FlexGen~\citep{FlexGen} \\
              PRISM~\citep{PRISM}}, level 3]
          ]
          [Identity \& Character \\ Customization, level 2
              [{\textbf{Optimization-based personalization:} \\
              Textual Inversion~\citep{gal2022image}, DreamBooth~\citep{ruiz2023dreambooth} \\
              LoRA~\citep{hu2022lora} \\
              \textbf{Reusable identity injectors:} \\
              E4T~\citep{gal2023encoder}, IP-Adapter~\citep{ye2023ip} \\
              InstantID~\citep{wang2024instantid}, PuLID~\citep{guo2024pulid} \\
              PhotoMaker~\citep{li2024photomaker} \\
              \textbf{Character/full-body consistency:} \\
              PortraitBooth~\citep{peng2024portraitbooth}, StoryMaker~\citep{zhou2024storymaker} \\
              Visual Persona~\citep{nam2025visual}, InstantCharacter~\citep{InstantCharacter} \\
              UniPortrait~\citep{UniPortrait}}, level 3]
          ]
      ]
      [Constructionist Paradigms \\ (Domain-Specific Adaptation),
        level 1, fill=Periwinkle!20, draw=Periwinkle!70, edge={thick, Periwinkle!70}
          [Modular Workflows, level 2
              [{\textbf{Multi-agent planning systems:} \\
              PosterVerseAF~\citep{Liu2026PosterVerseAF}, PosterGenAP~\citep{Zhang2025PosterGenAP} \\
              \textbf{Reward modeling \& distillation:} \\
              PosterOmniGA~\citep{Chen2026PosterOmniGA}, POSTAAG~\citep{Chen2025POSTAAG} \\
              PosterCraftRH~\citep{Chen2025PosterCraftRH} \\
              \textbf{In-context domain alignment:} \\
              ICCustomDI~\citep{Li2025ICCustomDI}}, level 3]
          ]
          [Semantic Typography, level 2
              [{\textbf{Visual approximation to semantic precision:} \\
              GlyphByT5v2AS~\citep{Liu2024GlyphByT5v2AS} \\
              \textbf{Decoupled typographic control:} \\
              ControlTextUC~\citep{Jiang2025ControlTextUC}, EasyTextCD~\citep{Lu2025EasyTextCD} \\
              UniGlyphUS~\citep{Wang2025UniGlyphUS}, ReChar~\citep{Rechar} \\
              \textbf{3D-aware semantic rendering:} \\
              CharGenHA~\citep{Ma2024CharGenHA}, UMTextAU~\citep{Ma2026UMTextAU} \\
              WordConWT~\citep{Shi2025WordConWT}, BeyondFT~\citep{Luo2025BeyondFT}}, level 3]
          ]
      ]
      [Reasoning-Driven Editing \\ (State Transition Modeling),
        level 1, fill=Melon!20, draw=Melon!70, edge={thick, Melon!70}
          [Agentic Decomposition, level 2
              [{\textbf{Multi-step logical planning:} \\
              BeyondSE~\citep{Yeh2025BeyondSE}, MIRAMI~\citep{Zeng2025MIRAMI} \\
              \textbf{Symbolic executable programs:} \\
              ReasonEditTR~\citep{Yin2025ReasonEditTR}, LegoEditAG~\citep{Jia2025LegoEditAG} \\
              ImageEA~\citep{Hu2025ImageEA}}, level 3]
          ]
          [Manifold Preservation, level 2
              [{\textbf{Global-local invariance:} \\
              IMAGHarmonyCI~\citep{Shen2025IMAGHarmonyCI}, \\ TBStarEditFI~\citep{Fang2025TBStarEditFI} \\
              SpotEditEV~\citep{Ghazanfari2025SpotEditEV} \\
              \textbf{Physically grounded layer decomposition:} \\
              ControllableLI~\citep{Yang2026ControllableLI}}, level 3]
          ]
          [Deployment \& Efficiency, level 2
              [{\textbf{Weight-free/In-context editing:} \\
              InContextEE~\citep{Zhang2025InContextEE}, VisualAM~\citep{Mao2025VisualAM} \\
              X2EditRA~\citep{Ma2025X2EditRA} \\
              \textbf{Real-time production pipelines:} \\
              Step1XEditAP~\citep{Liu2025Step1XEditAP}, ImplementationFF~\citep{Yao2025ImplementationFF}}, level 3]
          ]
      ]
      [Embodied Domain \\ (Physical Interaction \\ \& Agent Learning),
        level 1, fill=gray!12, draw=gray!75, edge={thick, gray!75}
          [Data Generation \& Augmentation, level 2
              [{\textbf{Embodied data foundations:} \\
              Droid~\citep{khazatsky2024droid}, BridgeData V2~\citep{walke2023bridgedatav2} \\
              Open X-Embodiment~\citep{openx2024rtx} \\
              \textbf{Cross-morphology editing:} \\
              Rovi-aug~\citep{chen2024roviaug}, Phantom~\citep{lepert2025phantom} \\
              H2R~\citep{li2025h2r}, Masquerade~\citep{lepert2025masquerade} \\
              RoboPaint~\citep{fan2026robopaint}, MITTY~\citep{song2025mitty} \\
              \textbf{Trajectory diversification:} \\
              IRASim~\citep{zhu2024irasim}, RoboMaster~\citep{fu2025robomaster} \\
              RoboDreamer~\citep{zhou2024robodreamer}, DreMa~\citep{barcellona2024dreamtomanipulate}}, level 3]
          ]
          [Visual Prediction for \\ Embodied Interaction, level 2
              [{\textbf{Future visual prediction for planning:} \\
              IRASim~\citep{zhu2024irasim}, VPP~\citep{hu2024vpp} \\
              U-VLA~\citep{li2025uva}, UWM~\citep{zhu2025uwm} \\
              \textbf{Predict-then-act:} \\
              CoT-VLA~\citep{zhao2025cotvla}, Seer~\citep{tian2024predictive} \\
              UniPi~\citep{du2023unipi}, AVDC~\citep{ko2024actionlessvideos} \\
              LVP~\citep{chen2025largevideoplanner}, RoboDreamer~\citep{zhou2024robodreamer} \\
              \textbf{Act-first with visual prediction regularization:} \\
              U-VLA~\citep{li2025uva}, UWM~\citep{zhu2025uwm} \\
              VideoVLA~\citep{shen2025videovla}, VPP~\citep{hu2024vpp}}, level 3]
          ]
      ]
    ]
  \end{forest}
  }
  \caption{Taxonomy of Applications and Evolving Frontiers in Visual Intelligence. The four branches correspond to the four subsections of this section: conditional image generation, custom domain adaptation, conditional image editing, and embodied-domain generation.}
  \label{fig:app_taxonomy}
\end{figure}
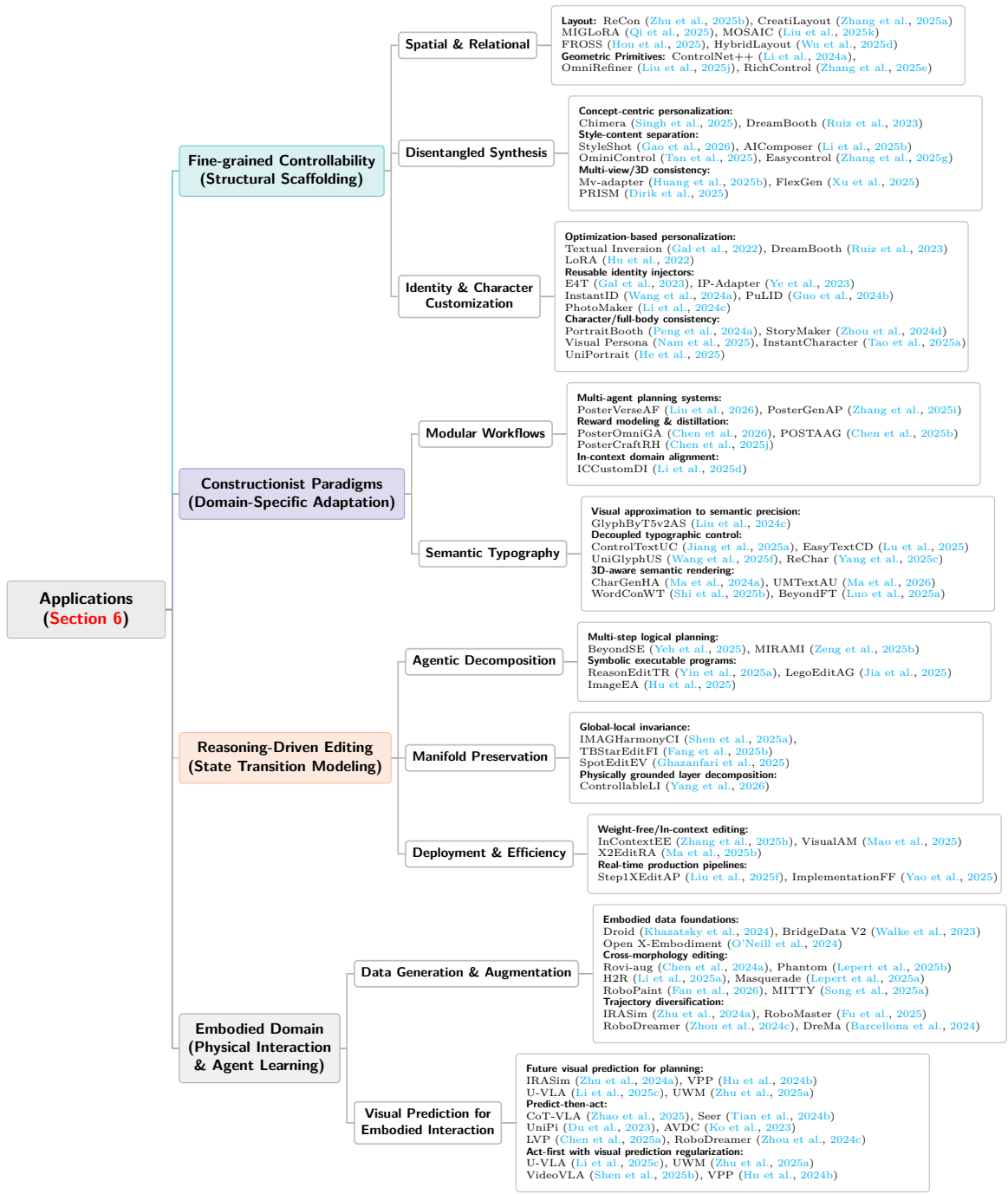

\Cref{fig:app_taxonomy} summarizes the application landscape discussed in this section. Rather than treating applications as a flat list of tasks, the figure organizes them by the type of constraint they impose on visual generation. \emph{Fine-grained controllability} corresponds to conditional image generation, where auxiliary conditions such as layouts, references, identities, and 3D cues reduce the ambiguity of text prompts. \emph{Constructionist paradigms} correspond to custom domain adaptation, where generation must obey domain-specific rules such as layout grammar, layered composition, typography, and design conventions. \emph{Reasoning-driven editing} corresponds to conditional image editing, where the model performs controlled state transitions while preserving unedited content. Finally, the \emph{embodied domain} covers settings where generation supports physical agents through data augmentation or future-state prediction. This organization makes explicit the common thread across the section: applications are increasingly defined by verifiable structure, memory, and interaction rather than by visual realism alone.

The evolution of visual intelligence, viewed through our hierarchical taxonomy, reflects a broader shift from optimizing \textit{statistical plausibility} toward satisfying increasingly \textit{explicit constraints}. Early text-to-image synthesis centered on perceptual realism, focusing on reconstructing fine visual details from compressed latent representations. As perceptual fidelity becomes less of the primary bottleneck, recent work has increasingly prioritized controllability, requiring generation to remain faithful not only to natural image statistics but also to structured user intent. Under this view, the generative model is no longer treated solely as a probabilistic sampler, but as a controllable neural rendering system whose intermediate representations must remain open to intervention.

This transition becomes more visible as visual generation moves beyond open-ended synthesis toward settings where spatial layout, temporal coherence, and semantic persistence must be maintained without drifting away from the pretrained image manifold. Following our taxonomy, we organize this progression into three increasingly constrained regimes: inference-time intervention for conditional synthesis, adaptation to domains governed by explicit design rules, and reasoning-driven modification of existing visual content. Together, these regimes reveal a common trend: visual generation is gradually evolving from implicit appearance modeling toward structured, programmable control.

\subsection{Conditional Image Generation}
\label{subsec:conditional_gen}

Conditional image generation arises from the persistent under-specification of natural language, whose semantics are often insufficient to determine concrete visual outcomes. To reduce this ambiguity, recent methods augment textual prompts with auxiliary control signals from complementary modalities, ranging from spatial maps and reference images to non-visual signals such as brain activity~\citep{wang2024mindbridge}. Early concept personalization methods such as DreamBooth~\citep{ruiz2023dreambooth} grounded novel subjects through explicit weight adaptation, establishing a strong yet optimization-heavy baseline. Subsequent adapter-based methods, including StyleShot~\citep{StyleShot}, improved inference-time flexibility by injecting stylistic cues directly into decoupled cross-attention layers rather than relying only on text. As diffusion backbones scale toward Diffusion Transformers, unified pipelines such as OminiControl~\citep{OminiControl} further extend this trend by routing appearance-specific priors through separated normalization or attention pathways, improving the disentanglement between semantic content and visual presentation.

Building on this line of control, identity customization focuses on preserving subject-specific appearance while maintaining text-driven editability. Optimization-based methods, including Textual Inversion~\citep{gal2022image} and LoRA~\citep{hu2022lora}, showed that personalized concepts can be incorporated effectively into pretrained generators, but their dependence on per-subject tuning often weakens compositional flexibility, editability, and deployment efficiency. This has motivated a gradual shift toward tuning-free or near tuning-free personalization. Representative frameworks such as IP-Adapter~\citep{ye2023ip}, PhotoVerse~\citep{chen2023photoverse}, and FastComposer~\citep{xiao2025fastcomposer} mitigate identity loss caused by compressed CLIP features by allocating dedicated cross-attention pathways to visual prompts and aligning multi-scale visual representations with text-conditioned generation. Few-shot methods such as PhotoMaker~\citep{li2024photomaker} further improve robustness by aggregating multiple references into a unified identity representation, balancing facial fidelity with semantic controllability. Related efforts that incorporate facial priors or orthogonal projection strategies~\citep{wang2024instantid, guo2024pulid, shi2024instantbooth, zhou2024storymaker, mou2025dreamo} reinforce persistent identity cues while reducing interference with the model's native prompt-following ability.

Despite this progress, face-centric personalization transfers only partially to full-body humans and open-domain characters, where large pose changes, self-occlusion, and non-rigid deformation expose the limits of identity representations dominated by facial appearance. In these settings, preserving clothing patterns, body proportions, and cross-view consistency requires stronger spatial correspondence and broader supervision than portrait-focused generation typically provides. Visual Persona~\citep{nam2025visual} addresses this gap by leveraging Multi-modal Large Language Models to curate dense multi-pose supervision, improving coverage of holistic human structure beyond facial identity. CompleteMe~\citep{tsai2025completeme} further highlights the importance of reference-guided detail transfer in human image completion, especially for recovering fine-grained clothing patterns and accessories under partial observations. In parallel, InstantCharacter~\citep{InstantCharacter} introduces reference-aligned spatial attention to track fine-grained traits across diverse articulations. Together, these efforts move conditional generation from rigid subject preservation toward articulated character consistency at the whole-body level.

Once subject consistency becomes more reliable, a related challenge emerges at the scene level: generation must preserve not only who appears in the image, but also where each semantic element should be placed. This has driven the development of structural control mechanisms that inject explicit geometric cues into the generative backbone. Frameworks such as \textit{ControlNet++}~\citep{controlnet++} improve geometric faithfulness through reward-based cycle consistency, while layout-aware methods such as ReCon~\citep{ReCon} rely on visual Chain-of-Thought reasoning to parse compositional instructions and bind semantic entities to designated regions via localized cross-attention masking. When these constraints must remain stable across viewpoints, multi-view frameworks~\citep{FlexGen, PRISM} incorporate 3D-aware priors to maintain epipolar consistency, which is especially important for downstream 3D reconstruction and interactive generation.

As identity preservation and spatial control improve, the dominant failure mode shifts again in multi-entity compositions and long-horizon visual narratives, where independent control signals often interfere with one another and cause feature bleeding across subjects or frames. Recent methods address this challenge by introducing more explicit state management during generation. For example, UniPortrait~\citep{UniPortrait} routes identity tokens to region-specific masks to reduce cross-subject interference, while SpinMeRound~\citep{SpinMeRound} disentangles persistent appearance cues from geometric variation during localized manipulation. These developments extend conditional generation beyond static customization toward more persistent visual state modeling. At the same time, most existing methods still address individual forms of control in isolation, leaving open the broader challenge of unifying identity, structure, and temporal persistence within a more compositional control framework.

\subsection{Custom Domain Adaptation}

\label{subsec:custom_domain}

A recurring challenge in domain adaptation is that the continuous visual priors learned by foundation diffusion models do not naturally align with the discrete structural rules imposed by many specialized applications. Although these models are highly effective at synthesizing naturalistic textures and broad semantic content, they often underrepresent explicit layout grammar, topological regularity, and coordinate-sensitive design constraints. As a result, domains that require precise spatial organization expose a persistent mismatch between generic visual realism and task-specific correctness.

Layout-to-image generation addresses this mismatch by conditioning image generation on explicit spatial specifications such as bounding boxes or segmentation masks. GLIGEN~\citep{li2023gligen} introduces a layout-conditioned diffusion paradigm that combines textual descriptions with designated spatial regions via gated self-attention, enabling fine-grained control over entity position and attributes. InstanceDiffusion~\citep{wang2024instancediffusion} broadens the spatial interface to support diverse formats---points, scribbles, boxes, and masks---within a unified fusion architecture. MIGC~\citep{zhou2024migc} enhances multi-instance control by decomposing generation into per-entity subtasks and leveraging instance-specific attention to better regulate quantity, position, and attributes. CreatiLayout~\citep{CreatiLayout} strengthens region-entity alignment through a siamese multimodal mechanism that jointly encodes layout and visual features with a Diffusion Transformer. HybridLayout~\citep{wu2025hybrid} further questions the assumption that strong layout adherence for modern Diffusion Transformers requires millions of densely annotated regional prompts. Its key insight is to decompose layout control into two sequential stages: first grounding anonymous layouts so the model learns \emph{where} objects should appear, and then applying lightweight semantic refinement so the model learns \emph{what} each designated region should contain. By combining this hybrid supervision strategy with low-resolution regional tokens and a region-wise diffusion loss, the method achieves strong spatial fidelity with dramatically fewer semantic layout annotations while preserving better visual aesthetics. DreamRenderer~\citep{zhou2025dreamrenderer} achieves training-free multi-instance attribute control by selectively applying hard attribute binding at critical transformer layers. Collectively, these advances transform foundation diffusion models into controllable generative systems capable of precise, instance-level semantic and positional regulation under increasingly data-efficient supervision.

Beyond explicit layout conditioning, an increasingly relevant direction is layered image generation, which represents an image as a set of compositional layers rather than a single flattened output. Compared with bounding boxes or segmentation masks alone, layered representations expose richer structural information, including foreground-background decomposition, occlusion ordering, transparency, and layer-wise editability. This makes them particularly suitable for domains where visual elements must remain independently controllable after synthesis, such as graphic design, advertising creatives, interface assets, and document-like media. Existing approaches generally follow two paradigms. Simultaneous generation methods generate multiple layers jointly within a unified model, improving global coherence and cross-layer consistency during synthesis, as seen in Text2Layer~\citep{zhang2023text2layer}, LayerDiff~\citep{huang2024layerdiff}, ART~\citep{pu2025art}, PrismLayer~\citep{chen2025prismlayers}, and Qwen-Image-Layered~\citep{yin2025qwenimagelayered}. In contrast, sequential generation methods decompose the process into progressive layer prediction or transparent-image reconstruction, which offers greater flexibility for iterative composition and editing, as explored by LayerDiffuse~\citep{zhang2024transparent}, COLE~\citep{jia2023cole}, OpenCOLE~\citep{inoue2024opencole}, and LayerD~\citep{suzuki2025layerd}. From the perspective of domain adaptation, layered generation can be viewed as a stronger structural abstraction than conventional layout control: it specifies not only where content should appear, but also how visual elements are separated, ordered, and recombined. As such, it extends controllable generation from spatial alignment toward editable compositional reasoning.

Despite these spatial control capabilities, many specialized domains impose additional constraints beyond layout arrangement. In commercial media and poster generation, for instance, generation must jointly satisfy typography placement, visual balance, and compositional aesthetics. Recent frameworks such as PosterVerse~\citep{Liu2026PosterVerseAF} and PosterGen~\citep{Zhang2025PosterGenAP} address this by reformulating generation as a modular planning problem, using Large Language Models to translate design intent into explicit layout variables and refine them through iterative visual feedback. Going further, CreatiDesign~\citep{zhang2025creatidesign} unifies heterogeneous design conditions--reference images, layouts, and rendered text--within a single model through a multimodal attention mask mechanism, enabling fine-grained control over each sub-condition while maintaining overall compositional harmony.

As these pipelines become more structured, a second line of work focuses on aligning generated outputs with domain-specific quality measures and content regimes. BizGen~\citep{peng2025bizgen}, for example, extends visual generation from sentence-level text rendering to article-level business content such as infographics and slides, combining scalable dense-layout data construction with layout-guided cross-attention for ultra-dense region-wise rendering. PosterOmni~\citep{Chen2026PosterOmniGA}, PosterCraft~\citep{Chen2025PosterCraftRH}, and POSTAAG~\citep{Chen2025POSTAAG} further combine task-aware distillation, reward modeling, and feedback-driven refinement to discourage failures such as text-background occlusion, hierarchy violations, or spatial imbalance. Rather than treating background synthesis and foreground arrangement as separate problems, these frameworks optimize them jointly so that visual content and layout evolve consistently. In contrast to these training-heavy solutions, ICCustom~\citep{Li2025ICCustomDI} shows that part of this domain alignment can also be induced at inference time through multimodal in-context learning, where interleaved visual and textual exemplars steer generation without requiring domain-specific parameter updates.

Beyond scene-level composition, specialized domains also demand precise rendering of local symbolic content, where approximate visual resemblance is no longer sufficient. This requirement is especially evident in text generation, where the granularity of language tokenization does not align well with the stroke-level geometry needed for accurate glyph synthesis. As a result, standard diffusion pipelines often produce visually plausible yet semantically incorrect characters. Although dedicated text-rendering architectures such as GlyphByT5-v2~\citep{Liu2024GlyphByT5v2AS} have substantially improved typographic quality, more recent efforts increasingly emphasize localized and decoupled control over character structure.

Segmentation-conditioned methods address this issue by injecting explicit typographical constraints into spatially bounded regions. ControlText~\citep{Jiang2025ControlTextUC} and ReChar~\citep{Rechar}, for example, modulate region-specific cross-attention so that glyph generation is anchored to designated masks and local structural priors~\citep{Lu2025EasyTextCD, PaCaNet}. In parallel, methods such as CharGen~\citep{Ma2024CharGenHA} and UMText~\citep{Ma2026UMTextAU} revisit the representation problem itself by replacing generic language tokenizers with character-aware encoders, thereby strengthening the alignment between symbolic input and visual realization. Together, these approaches shift the problem from coarse text-conditioned synthesis toward more faithful character-level rendering.

Recent work further extends text rendering beyond planar overlays toward physically grounded generation in spatially structured environments. Frameworks such as WordCon~\citep{Shi2025WordConWT} and Beyond~\citep{Luo2025BeyondFT} introduce self-guidance mechanisms that extract intermediate geometric signals, including depth and surface normals, from the generative process and feed them back as structural conditions. This design enables text to inherit explicit 3D geometry and material-aware appearance, making it more consistent with scene lighting, viewpoint, and physical support surfaces. Overall, these developments show that domain adaptation is moving from broad stylistic transfer toward rule-aware generation under both global and local constraints. Yet much of this progress remains tied to narrowly defined domains, with limited support for more general, reusable control abstractions across tasks.

\subsection{Conditional Image Editing}
\label{subsec:conditional_editing}

Conditional image editing reframes generative modeling from open-ended synthesis to a constrained state transition problem, where the desired output must satisfy editing instructions while preserving the source image wherever no change is required. This creates a persistent tension between \textit{plasticity}, namely the ability to modify targeted attributes, and \textit{stability}, namely the preservation of identity, structure, and context outside the edited region. Standard diffusion models often struggle with this balance because globally shared attention can entangle edited and unedited content, causing local instructions to propagate unexpectedly across the image. Recent work therefore places increasing emphasis on structured control, decomposition, and consistency preservation to regulate how edits are introduced into the latent trajectory.

A prominent direction addresses complex and multi-step edits by decomposing high-level instructions into more explicit intermediate operations. Frameworks such as \textit{X-Planner}~\citep{Yeh2025BeyondSE} and \textit{MIRA}~\citep{Zeng2025MIRAMI} employ Multi-modal Large Language Models to parse abstract user intent into sequentially executable subgoals, making the editing process more transparent and controllable across multiple stages. Related methods, including \textit{ReasonEdit}~\citep{Yin2025ReasonEditTR} and \textit{LegoEdit}~\citep{Jia2025LegoEditAG}, further separate semantic reasoning from low-level rendering, thereby reducing ambiguity in how instructions are grounded into pixel-space transformations. At a more symbolic level, program-like editing formulations~\citep{Hu2025ImageEA} translate natural language into executable operations, allowing image transitions to be governed by explicit logical structure rather than by diffusion dynamics alone.

While instruction decomposition improves semantic clarity, edit quality still depends critically on whether the original visual manifold can be preserved outside the target region. To reduce unwanted structural drift, recent methods introduce explicit consistency modules that regularize the denoising trajectory under spatial and quantitative constraints. \textit{IMAGHarmony}~\citep{Shen2025IMAGHarmonyCI} and \textit{TBStarEdit}~\citep{Fang2025TBStarEditFI}, for example, strengthen local-global consistency so that modifications remain concentrated around the intended edit scope. For scenarios requiring even finer control, physically grounded methods such as \textit{PhysicEdit}~\citep{PhysicEdit} and layered editing frameworks~\citep{Yang2026ControllableLI} decompose the image into semantically separable strata, enabling foreground manipulation while preserving the surrounding context more explicitly. The precision of such methods is increasingly examined by dedicated benchmarks~\citep{Ghazanfari2025SpotEditEV}, which evaluate not only whether the intended content is changed correctly, but also whether the remaining image stays stable.

As editing systems move closer to interactive use, a different bottleneck emerges at deployment time: per-image optimization becomes increasingly difficult to reconcile with responsive applications. This has motivated recent work on weight-free or feed-forward editing paradigms that retain instruction following while reducing adaptation cost. Methods based on visual in-context learning~\citep{Zhang2025InContextEE} and task-aware functional representations~\citep{Mao2025VisualAM, Ma2025X2EditRA} show that robust editing behavior can be induced without expensive gradient updates. Complementary architectural efforts further streamline the inference path toward high-fidelity, near single-step editing~\citep{Liu2025Step1XEditAP, Yao2025ImplementationFF}. Together, these advances move conditional image editing from offline optimization toward more deployable visual manipulation. However, most existing methods still optimize individual aspects of editing, such as planning, preservation, or efficiency, separately, rather than treating editing as part of a unified control process that jointly reasons about intent, locality, and state consistency.

\subsection{Embodied Domain}
Embodied applications of visual generation and editing extend beyond static synthesis to settings where visual outputs interact with or support physical agents and environments. In these contexts, generative models serve as tools for augmenting data and for modeling visual state transitions under control, enabling downstream tasks in robotics, simulation, and agent learning~\citep{meietal2026videoroboticsurvey,li2025uva,zhu2025uwm}. 
We categorize embodied applications into two primary modalities: 1) Visual Data Generation and Augmentation for Agents, and 2) Visual Prediction for Embodied Interaction.

\begin{figure}[!htbp]
\centering
\begin{tikzpicture}[
    font=\normalsize,
    box/.style={
        rounded corners=2mm,
        draw=black!12,
        fill=white,
        inner sep=7pt
    },
    title/.style={
        rounded corners=2mm,
        draw=black!12,
        fill=black!4,
        inner sep=6pt,
        font=\normalsize\bfseries,
        align=center,
        minimum width=9.5cm
    },
    headL/.style={
        rounded corners=1.5mm,
        fill=blue!15,
        inner sep=5pt,
        font=\normalsize\bfseries,
        align=center,
        text width=4.4cm
    },
    headR/.style={
        rounded corners=1.5mm,
        fill=teal!15,
        inner sep=5pt,
        font=\normalsize\bfseries,
        align=center,
        text width=4.4cm
    },
    item/.style={
        rounded corners=1.2mm,
        draw=black!10,
        fill=white,
        inner sep=5pt,
        text width=4.2cm,
        align=left,
        font=\small
    },
    roleL/.style={
        rounded corners=1.2mm,
        draw=black!10,
        fill=blue!8,
        inner sep=5pt,
        text width=4.2cm,
        align=left,
        font=\small
    },
    roleR/.style={
        rounded corners=1.2mm,
        draw=black!10,
        fill=teal!8,
        inner sep=5pt,
        text width=4.2cm,
        align=left,
        font=\small
    }
]

\node[title] (T) at (0,0) {Embodied Applications of Visual Generation};

\node[headL] (LH) at (-2.6,-1.3) {Data Generation \& Augmentation};
\node[headR] (RH) at ( 2.6,-1.3) {Prediction for Interaction};

\node[item, anchor=north] (L1) at ($(LH.south)+(0,-0.3)$)
{\textbf{Data construction}\\[-1pt]\footnotesize Datasets, simulation};

\node[item, anchor=north] (L2) at ($(L1.south)+(0,-0.22)$)
{\textbf{Transfer \& editing}\\[-1pt]\footnotesize Cross-embodiment adaptation};

\node[item, anchor=north] (L3) at ($(L2.south)+(0,-0.22)$)
{\textbf{Diversification}\\[-1pt]\footnotesize Expanded behaviors};

\node[roleL, anchor=north] (L4) at ($(L3.south)+(0,-0.28)$)
{\textbf{Role}\\[-1pt]\footnotesize Coverage \& generalization};

\node[item, anchor=north] (R1) at ($(RH.south)+(0,-0.3)$)
{\textbf{Predict-then-act}\\[-1pt]\footnotesize Future prediction};

\node[item, anchor=north] (R2) at ($(R1.south)+(0,-0.22)$)
{\textbf{Prediction-regularized}\\[-1pt]\footnotesize Auxiliary signal};

\node[item, anchor=north] (R3) at ($(R2.south)+(0,-0.22)$)
{\textbf{Hybrid}\\[-1pt]\footnotesize Joint modeling};

\node[roleR, anchor=north] (R4) at ($(R3.south)+(0,-0.28)$)
{\textbf{Role}\\[-1pt]\footnotesize Dynamics \& temporal structure};

\draw[black!15, line width=0.5pt] (0,-1.1) -- (0,-4.3);

\begin{scope}[on background layer]
\node[box, fit=(LH)(L1)(L2)(L3)(L4)(RH)(R1)(R2)(R3)(R4)] {};
\end{scope}

\end{tikzpicture}
\caption{Embodied applications of visual generation can be organized into two complementary roles. 
The first uses generation and augmentation to expand embodied training data through construction, transfer, and diversification, thereby improving coverage and generalization. 
The second uses visual prediction to support embodied interaction by modeling future observations, regularizing action learning, and capturing environment dynamics.}
\label{fig:embodied_visual_generation}
\end{figure}
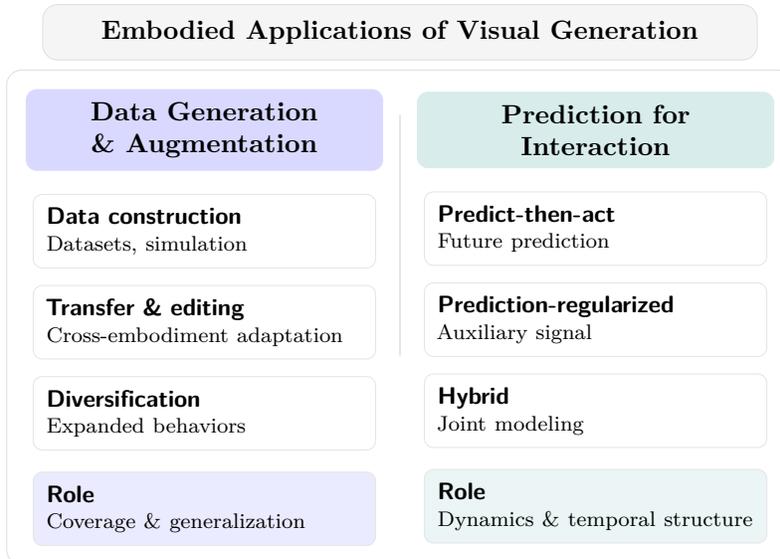

\subsubsection{Visual Data Generation and Augmentation for Agents}

In many embodied systems, the limited availability of diverse interaction data inhibits both learning and generalization~\citep{khazatsky2024droid,openx2024rtx,walke2023bridgedatav2}. Visual generation can mitigate this by synthesizing additional data that preserves both visual realism and physically plausible state diversity~\citep{chen2024roviaug,lepert2025phantom,li2025h2r,lepert2025masquerade,fan2026robopaint,song2025mitty}. Representative uses include:

\begin{itemize}
    \item \textbf{Trajectory Diversification:} Given an initial state (e.g., an image depicting an object configuration or scene), generate multiple plausible future visual sequences that correspond to different action sequences or environmental perturbations~\citep{zhu2024irasim,fu2025robomaster,zhou2024robodreamer,barcellona2024dreamtomanipulate}. This expanded dataset enhances robustness during policy learning or control refinement~\citep{zhu2024irasim,zhou2024robodreamer}.
    
    \item \textbf{Cross-Morphology Editing:} Edit existing perception data to reflect alternative agent embodiments (e.g., replacing a human arm with a robotic manipulator or substituting object appearances) without collecting new real-world samples~\citep{chen2024roviaug,lepert2025phantom,li2025h2r,lepert2025masquerade,fan2026robopaint,song2025mitty}. Such editing supports transfer learning and evaluation under varied agent morphologies~\citep{chen2024roviaug,li2025h2r}.
\end{itemize}

These applications treat visual generation and editing as data augmentation mechanisms that enrich interaction datasets with minimal real-world collection cost, while preserving semantic and physical coherence.

\subsubsection{Visual Prediction for Embodied Interaction}

A prominent embodied application of visual generation is to support decision-making through predictive modeling of future observations under interaction~\citep{zhu2024irasim,hu2024vpp,li2025uva,zhu2025uwm}. In this setting, visual generation is used to anticipate how the environment may evolve from the current observation, providing informative context for downstream action prediction~\citep{zhu2024irasim,hu2024vpp}.

A widely adopted approach first predicts multiple plausible future visual states given the current observation, and then infers executable actions for the agent based on these predicted futures~\citep{zhao2025cotvla,tian2024predictive,du2023unipi,ko2024actionlessvideos,chen2025largevideoplanner}. By explicitly modeling prospective environment evolution, this formulation enables the model to reason over alternative outcomes and long-term consequences before committing to actions~\citep{du2023unipi,chen2025largevideoplanner,zhou2024robodreamer}. Learning accurate visual dynamics is typically the more challenging component, while action inference is comparatively simpler once future states are available~\citep{zhu2024irasim,tian2024predictive,hu2024vpp}.

An alternative formulation reverses this order: models first predict actions directly from current visual inputs, and subsequently generate future visual observations as an auxiliary objective~\citep{li2025uva,zhu2025uwm,liang2025videopolicy,shen2025videovla,hu2024vpp}. In this case, visual prediction does not serve as an explicit intermediate for planning, but instead provides additional supervision that encourages temporal consistency, physical plausibility, and structured representations aligned with environment dynamics~\citep{li2025uva,zhu2025uwm}.

These two formulations reflect different design choices rather than fundamentally different objectives. Both couple visual generation with action reasoning, differing mainly in whether future visual prediction is treated as a primary decision-support signal or as auxiliary regularization~\citep{tian2024predictive,li2025uva,zhu2025uwm}. In practice, hybrid designs that interleave or jointly optimize visual prediction and action inference are also common~\citep{li2025uva,zhu2025uwm,liang2025videopolicy,shen2025videovla}.

Overall, visual prediction in embodied interaction functions as a mechanism for encoding environment dynamics and temporal structure, enabling agents to make informed decisions in sequential, interactive settings rather than producing visually accurate frames in isolation~\citep{meietal2026videoroboticsurvey,hu2024vpp,li2025uva}.

\section{Assessment for the Future: Stress Testing the Limits}
\label{sec:stress_test}

The five-level taxonomy introduced in \Cref{sec:evolution} claims that visual intelligence progresses from atomic generation (L1) through conditional generation (L2), in-context generation (L3), and agentic generation (L4) toward world-modeling generation (L5). But how far have current frontier models actually advanced along this trajectory? Standard benchmarks---FID, CLIP-Score, isolated prompt-following metrics---predominantly evaluate L1 and L2 capabilities: whether the generated image is realistic and whether it matches the text. They tell us little about whether models can solve spatial puzzles (L2 boundary), maintain identity across multi-turn edits (L3), self-correct through reasoning (L4), or simulate physical causality (L5).

To probe these higher-level capabilities, we complement benchmark-based evaluation with a series of \emph{in-the-wild stress tests}: carefully designed tasks that target the boundary conditions of each level. Each dimension below is explicitly mapped to the level it primarily challenges, as summarized in \Cref{tab:stress_test_level_map}. This design ensures that the stress tests serve not merely as anecdotal failure cases, but as structured evidence for where the current frontier sits within our taxonomy.

\subsection{Methodology: From Benchmarks to "In-the-Wild" Scenarios}

\paragraph{Generator and outputs.}
Each figure caption indicates the generator used (\textbf{Nano Banana} or \textbf{GPT-Image-2}). The two systems are evaluated under identical prompts and settings, enabling direct cross-system comparison without changing the task definitions.

\paragraph{Paper figures as embedded probes.}
Several explanatory figures in this work are themselves produced with frontier image-generation models under expert-authored descriptions. We do not treat these illustrations as formal benchmarks, because their prompts, selection process, and revision decisions are coupled to the authors' communicative goals. Nevertheless, they provide a complementary form of real-world case study: when a model is asked to generate a semantically correct roadmap figure, taxonomy diagram, system schematic, or scientifically constrained visual explanation from a dense expert specification, it must simultaneously follow domain semantics, visual grammar, layout constraints, and readability requirements. Success therefore offers evidence of generation capability beyond photorealism, while failure exposes the same issues studied in the stress tests below---missing relations, visually plausible but scientifically inaccurate structure, weak symbolic grounding, or unstable text rendering. For this reason, AI-assisted figures used in the paper should be read as \emph{embedded probes} of structured visual generation, with final correctness still requiring expert verification.

\subsection{Dimension I: Spatial Structuring \& Layout Precision}

\noindent\textit{Primary Level Tested: L2 (Conditional Generation) --- Can the model faithfully translate spatial constraints into precise geometric arrangements?}

\paragraph{Core Ability: Can the model follow strict geometric constraints? (Count, Ratio, Position).}

\subsubsection{Case Study I: The Jigsaw Puzzle Challenge (Geometric Rigidity vs. Generative Hallucination)}
\label{subsubsec:jigsaw_case}

To strictly evaluate the model's capability in \textit{Spatial Structuring} and \textit{Conservation of Content}, we designed the "Jigsaw Reconstruction" task. Unlike standard image generation where creativity is encouraged, this task requires the model to act as a rigid solver: it must reassemble visual fragments into a coherent whole based strictly on edge matching and texture continuity, without adding or removing visual information.

\paragraph{Experimental Setup.}
We employed a "Visual Reconstruction Specialist" system prompt (see Appendix for full prompt), instructing the model to strictly adhere to "Seamless Integrity" and "Strict Edge Matching."
\begin{itemize}
    \item \textbf{Input:} A fragmented view of hot air balloons (`puzzle\_input.jpg`), containing disjointed puzzle pieces with significant spatial gaps and randomized overlaps.
    \item \textbf{Ground Truth:} A high-resolution photograph of multiple hot air balloons against a clear blue sky (`puzzle\_gt.jpg`).
    \item \textbf{Challenge:} The model must recognize the discrete nature of the pieces and spatially rearrange them to restore the original topology.
\end{itemize}

\begin{figure}[!htbp]
    \centering
    \begin{subfigure}[b]{0.3\textwidth}
        \includegraphics[width=\linewidth,height=3.2cm,keepaspectratio]{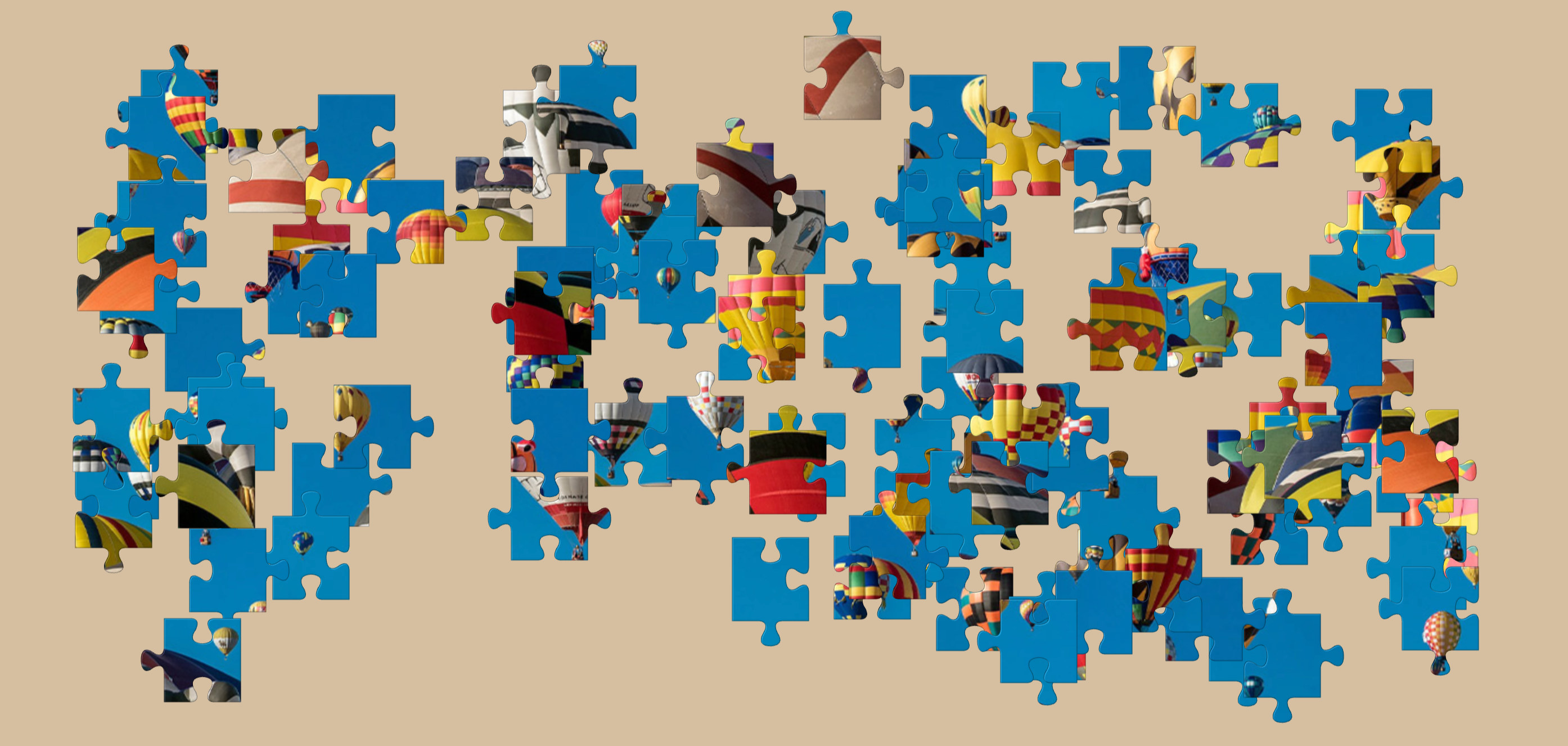}
        \caption{Input: Scrambled Fragments}
    \end{subfigure}
    \hfill
    \begin{subfigure}[b]{0.3\textwidth}
        \includegraphics[width=\linewidth,height=3.2cm,keepaspectratio]{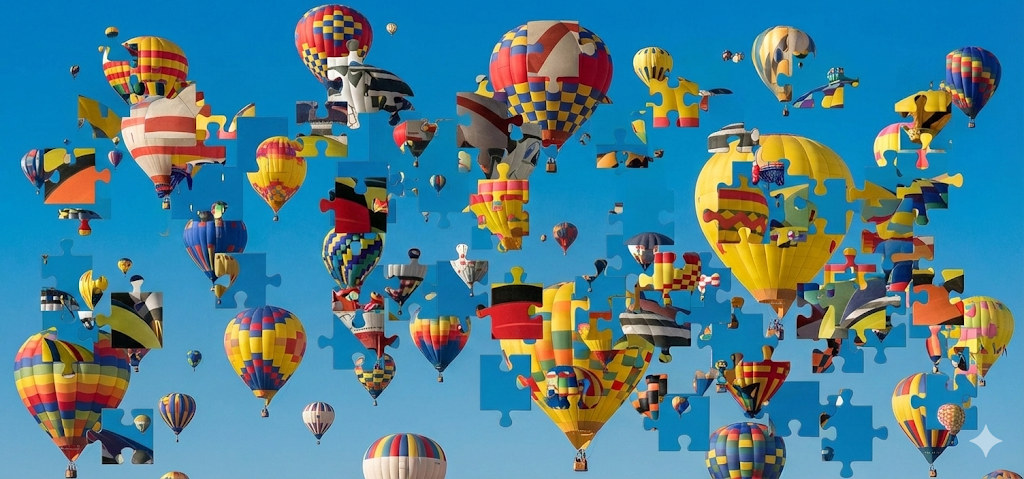}
        \caption{Model Output}
    \end{subfigure}
    \hfill
    \begin{subfigure}[b]{0.3\textwidth}
        \includegraphics[width=\linewidth,height=3.2cm,keepaspectratio]{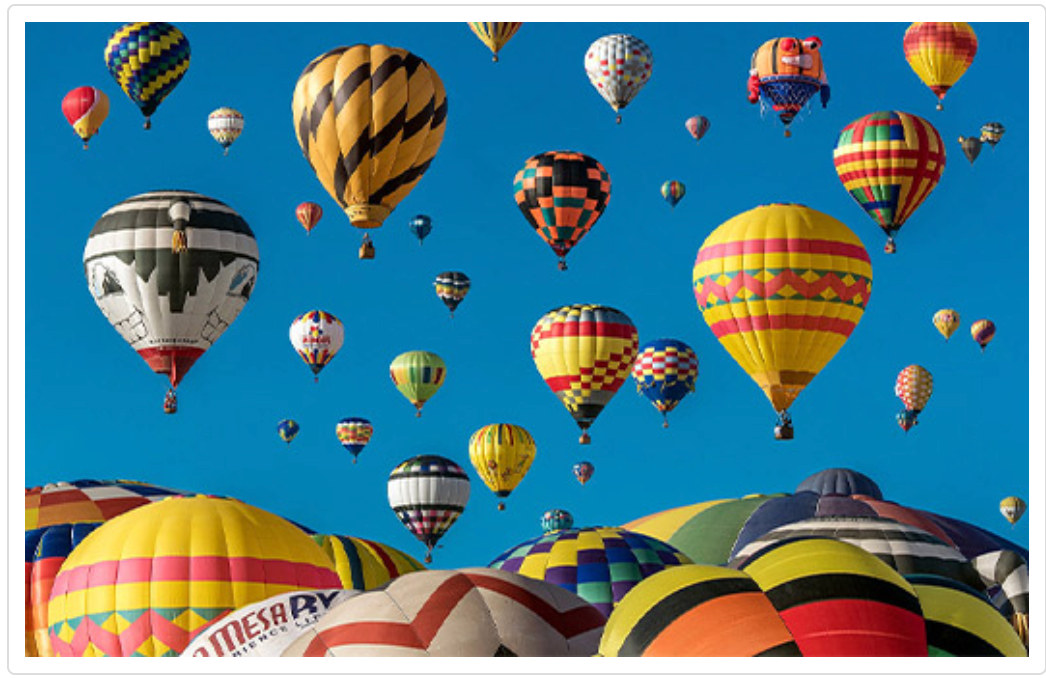}
        \caption{Ground Truth}
    \end{subfigure}
    \caption{\textbf{Failure Mode in Spatial Reconstruction.} While the model successfully recovers the \textit{semantic theme} (balloons in the sky), it fails the \textit{geometric test}. Instead of solving the puzzle logic (matching tabs and slots), it "hallucinates" a plausible-looking but structurally incoherent image, blending distinct pieces into amassed artifacts. Generated by Nano Banana.}
    \label{fig:jigsaw_case}
\end{figure}

\paragraph{Analysis of Results.}
As shown in \Cref{fig:jigsaw_case}, the result highlights a critical limitation in current foundation models:
\begin{itemize}
    \item \textbf{Semantic Success vs. Geometric Failure:} The model correctly identifies the semantic context (colorful balloons, blue sky). It attempts to group similar textures (blue pieces together, striped pieces together).
    \item \textbf{The "Hallucination" Trap:} Instead of performing rigid transformations (rotation/translation) on the pixel fragments, the model utilizes its generative prior to "dream" a solution. It generates \textit{new} balloon-like textures to fill the gaps rather than stitching the existing edges.
    \item \textbf{Lack of Object Permanence:} The input constraints explicitly forbade "hallucinating" content. However, the output shows significant artifacts where multiple pieces are fused into non-existent geometries, and original edge details (the "tabs" and "blanks" of puzzle pieces) are dissolved rather than utilized for alignment.
\end{itemize}

\paragraph{Insight.}
This case study demonstrates that even with a strong "Agentic" prompt, current visual models operate primarily on \textbf{Probabilistic Correlation (Level 1/2)} rather than \textbf{Causal/Physical Logic (Level 5)}. They prioritize making the image \textit{look} like a completed puzzle (semantic consistency) over actually \textit{solving} the puzzle (spatial consistency). True spatial intelligence requires the ability to treat image patches as rigid bodies with permanent properties, a capability currently lacking in standard diffusion pipelines.

\begin{figure}[!htbp]
    \centering
    \begin{subfigure}[b]{0.48\textwidth}
        \includegraphics[width=\linewidth]{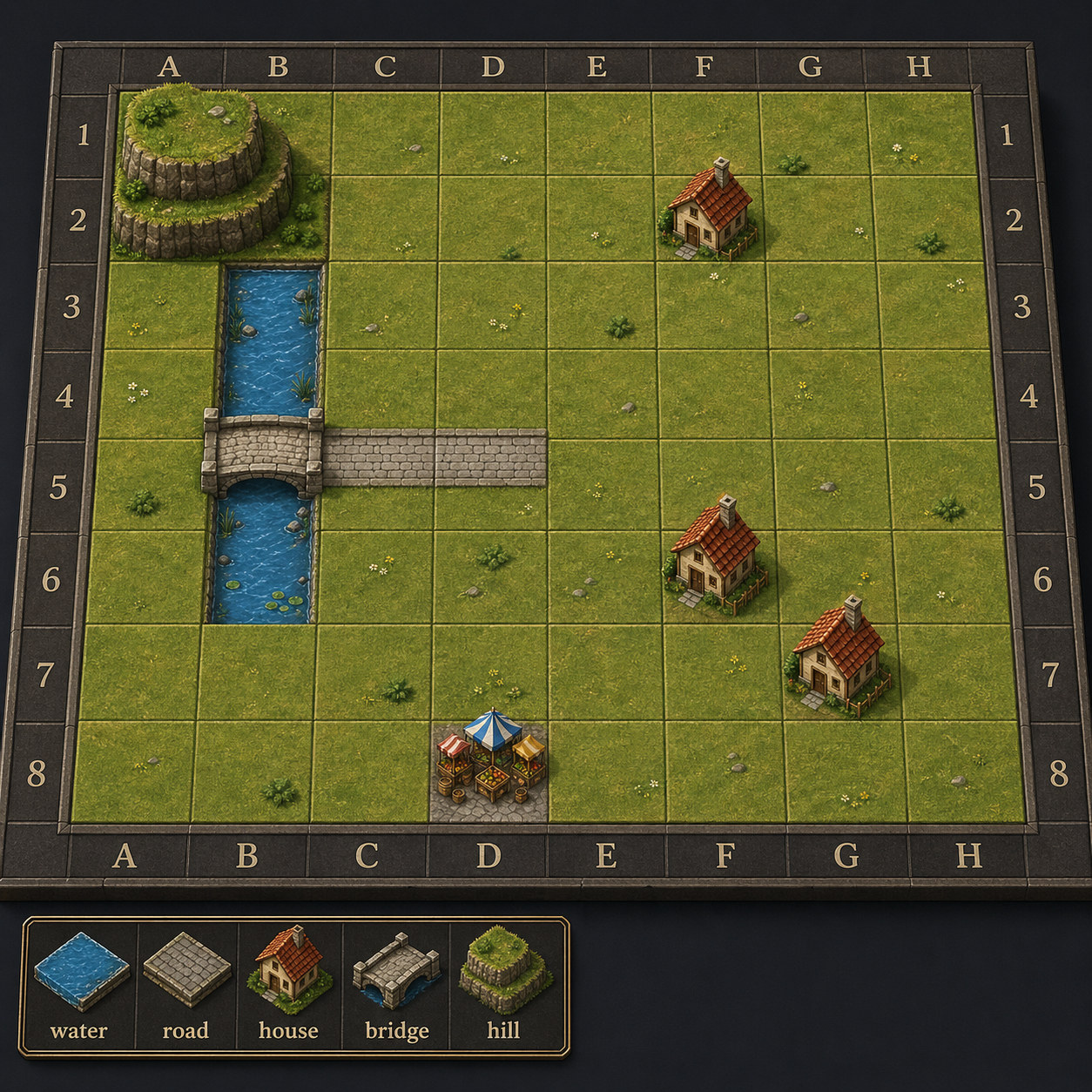}
        \caption{Isometric tile map}
        \label{fig:tile_map_case}
    \end{subfigure}
    \hfill
    \begin{subfigure}[b]{0.48\textwidth}
        \includegraphics[width=\linewidth]{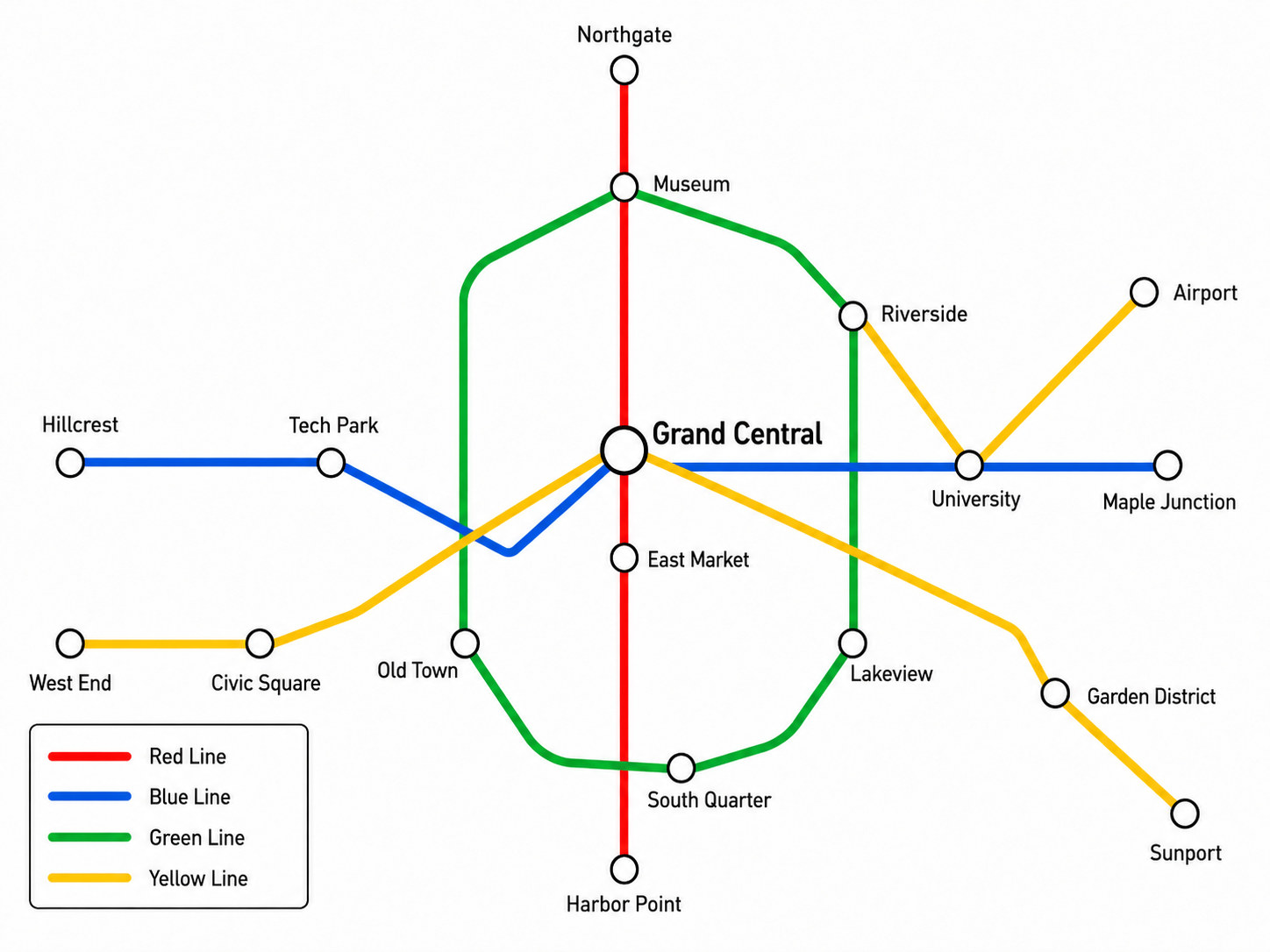}
        \caption{Metro-system map}
        \label{fig:metro_map_case}
    \end{subfigure}
    \caption{\textbf{Spatial structuring under explicit layout constraints.}
    Both examples are generated by \textbf{GPT-Image-2} from prompts with discrete spatial or topological requirements.
    The outputs are visually polished and closely match the intended design styles: a professional isometric tactical map in \Cref{fig:tile_map_case} and a clean vector-style metro map in \Cref{fig:metro_map_case}.
    However, both cases reveal subtle but important failures in constraint execution, including coordinate-level object shifts in the tile map and graph-topology violations in the metro map.}
    \label{fig:structured_layout_cases}
\end{figure}

\subsubsection{Case Study II: The Metro Map Challenge (Topological Plausibility vs. Constraint Verification)}
\label{subsubsec:metro_map_case}

\paragraph{Prompt.}
Create a professional metro-system map for a fictional city. It must contain four colored lines: Red, Blue, Green, and Yellow. There are exactly 18 station names. Three stations are transfer stations shared by exactly two lines, and one central station is shared by all four lines. The Red and Blue lines cross twice but share only one station; the Green line forms a loop; the Yellow line branches once after its third station. Use clean vector-map aesthetics, readable station labels, a legend, and no geographic background. The topology must be internally consistent.

\paragraph{Result.}
As shown in \Cref{fig:metro_map_case}, the generated map is visually polished and satisfies several high-level requirements. It correctly contains \textbf{18 stations} in total, the \textbf{Green Line forms a loop}, and there are \textbf{three transfer stations shared by exactly two lines}. Nevertheless, several key topological constraints are violated. Most notably, the prompt requires \textbf{one central station shared by all four lines}, but the generated central station (\textit{Grand Central}) is shared by only \textbf{three} lines---Red, Blue, and Yellow---while the Green Line does not pass through it. In addition, the constraint that the \textbf{Red and Blue lines should cross twice but share only one station} is not satisfied, since the generated map shows only \textbf{one} such crossing. Finally, the \textbf{Yellow Line branching rule} is also incorrect: the branch is not placed \emph{after its third station} as specified.

\paragraph{Insight.}
This case highlights a central limitation of current text-to-image systems in \emph{spatially structured generation}: they can often reproduce the \emph{visual grammar} of a structured diagram, but fail to enforce the underlying \emph{discrete relational constraints}. The model captures the appearance of a professional metro map---color-coded routes, readable station labels, loop-like geometry, transfer stations, and a legend---yet it does not faithfully preserve the graph-level topology required by the prompt. In other words, the output is visually plausible but not topologically valid.

A particularly revealing aspect of this example is the contrast between \emph{generation} and \emph{post-hoc verification}. In this case, GPT-Image-2 spent \textbf{13m15s} ``thinking'' before generation, yet still produced a map that violates multiple explicit constraints. By contrast, when the generated image and the original prompt were provided jointly to GPT 5.5 with the simple query \textit{``What parts of the image do not match the prompt?''}, the model identified all major mismatches in only \textbf{9s}. This suggests that constructing a constraint-satisfying visual artifact is substantially harder than checking an already rendered artifact against symbolic requirements. More broadly, long deliberation during image generation does not necessarily translate into reliable constraint satisfaction.

\subsubsection{Case Study III: The Isometric Tile Map Challenge (Coordinate Grounding vs. Visual Plausibility)}
\label{subsubsec:isometric_tile_map_case}

\paragraph{Prompt.}
Generate a clean isometric game map made of exactly 8$\times$8 square tiles. Use a strict isometric angle and make every tile edge align perfectly. Place a river occupying tiles B2--B6, a stone bridge crossing the river at B4--D4, three houses at F2, F5, and G6, one marketplace at D7, and a 2-tile-high hill in the northwest corner. Add a small legend showing ``water,'' ``road,'' ``house,'' ``bridge,'' and ``hill.'' The result should look like a professional tactical board-game map, not a fantasy painting.

\paragraph{Result.}
As shown in \Cref{fig:tile_map_case}, the generated map is visually convincing and successfully captures the intended style of a professional isometric tactical board-game map. The overall grid structure is clean, the isometric viewpoint is mostly consistent, tile boundaries are well aligned, and the terrain assets are rendered with coherent lighting and texture. The legend is also correctly included, covering the requested categories: \textit{water}, \textit{road}, \textit{house}, \textit{bridge}, and \textit{hill}. Several high-level spatial constraints are also broadly satisfied: the river, bridge, northwest hill, and one of the houses are placed in a way that is largely consistent with the prompt.

However, the result still fails on precise coordinate grounding. The houses specified at \textbf{F5} and \textbf{G6} are shifted to nearby but incorrect cells, approximately \textbf{F6} and \textbf{G7}. Similarly, the marketplace requested at \textbf{D7} is placed lower than expected, closer to \textbf{D8}. These errors are visually subtle because the map remains coherent and game-like, but they violate the rule-based placement constraints that define the task.

\paragraph{Insight.}
This case illustrates a recurring failure mode in spatially structured generation: the model can reproduce the \emph{visual grammar} of a tile-based map, but does not reliably execute the underlying \emph{coordinate program}. Isometric grids impose a discrete symbolic structure, where each object must be bound to a specific cell. Yet the model appears to treat coordinates such as \textit{F5} and \textit{G6} as soft layout cues rather than exact addresses. As a result, it produces a perceptually plausible map while committing off-by-one placement errors.

More broadly, this suggests that current text-to-image models still lack a robust internal mechanism for translating symbolic spatial specifications into deterministic geometric layouts. Their strength lies in synthesizing coherent visual patterns, not in maintaining a verifiable grid state throughout generation. Similar to the metro-map case in \Cref{fig:metro_map_case}, the failure is not aesthetic but procedural: the image looks correct at a glance, yet the discrete spatial relations are not fully satisfied. This again reveals a gap between generation and verification. A multimodal reasoning model such as GPT 5.5 can readily identify these coordinate mismatches after seeing the image and prompt, while the image generator itself struggles to prevent them during generation.

\begin{figure}[t]
\centering
\begin{subfigure}[b]{0.16\textwidth}
\includegraphics[width=\linewidth]{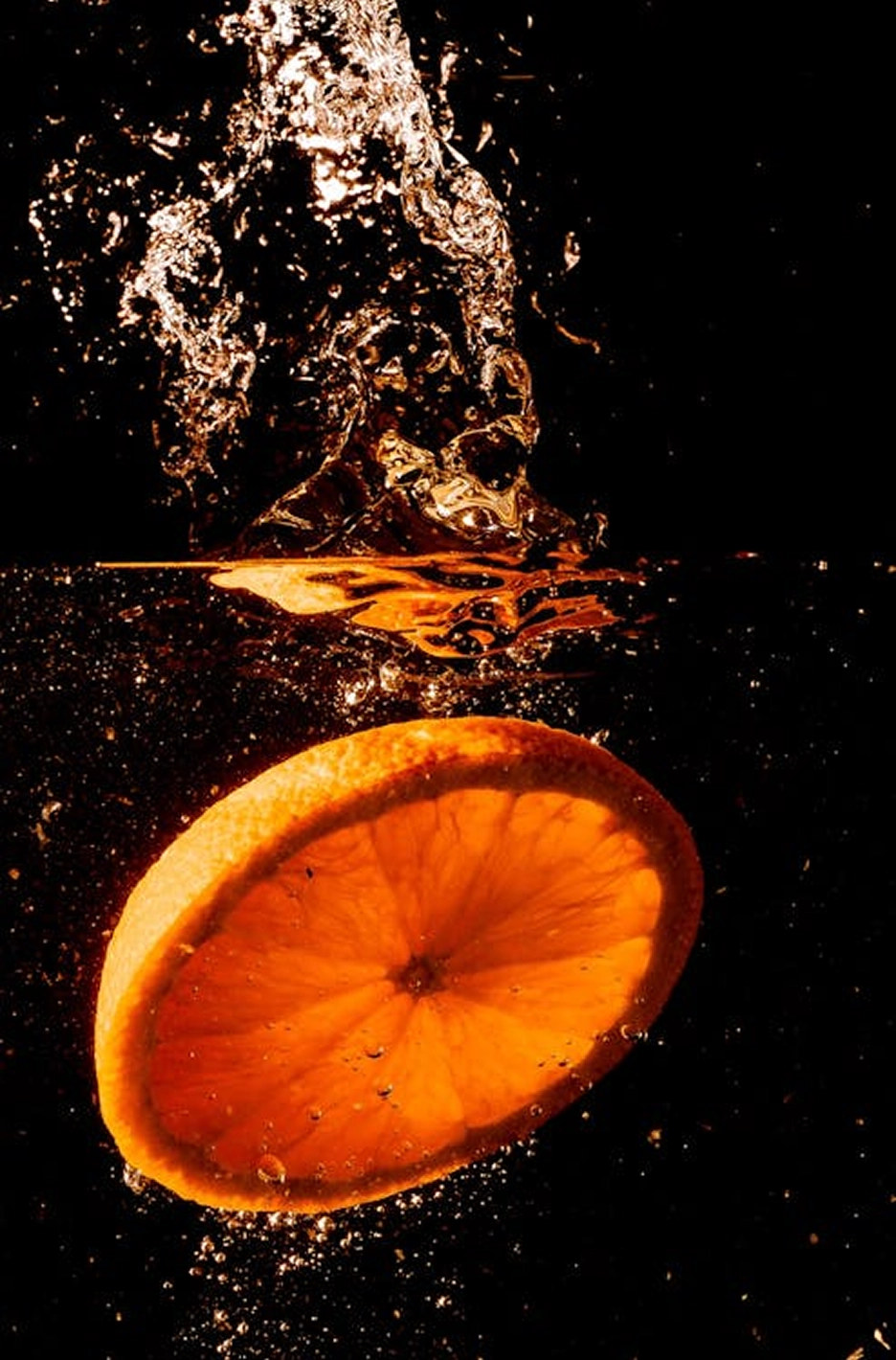}
\caption{Input: Floating State}
\vspace{-3mm}
\end{subfigure}
\hfill
\begin{subfigure}[b]{0.41\textwidth}
\includegraphics[width=\linewidth]{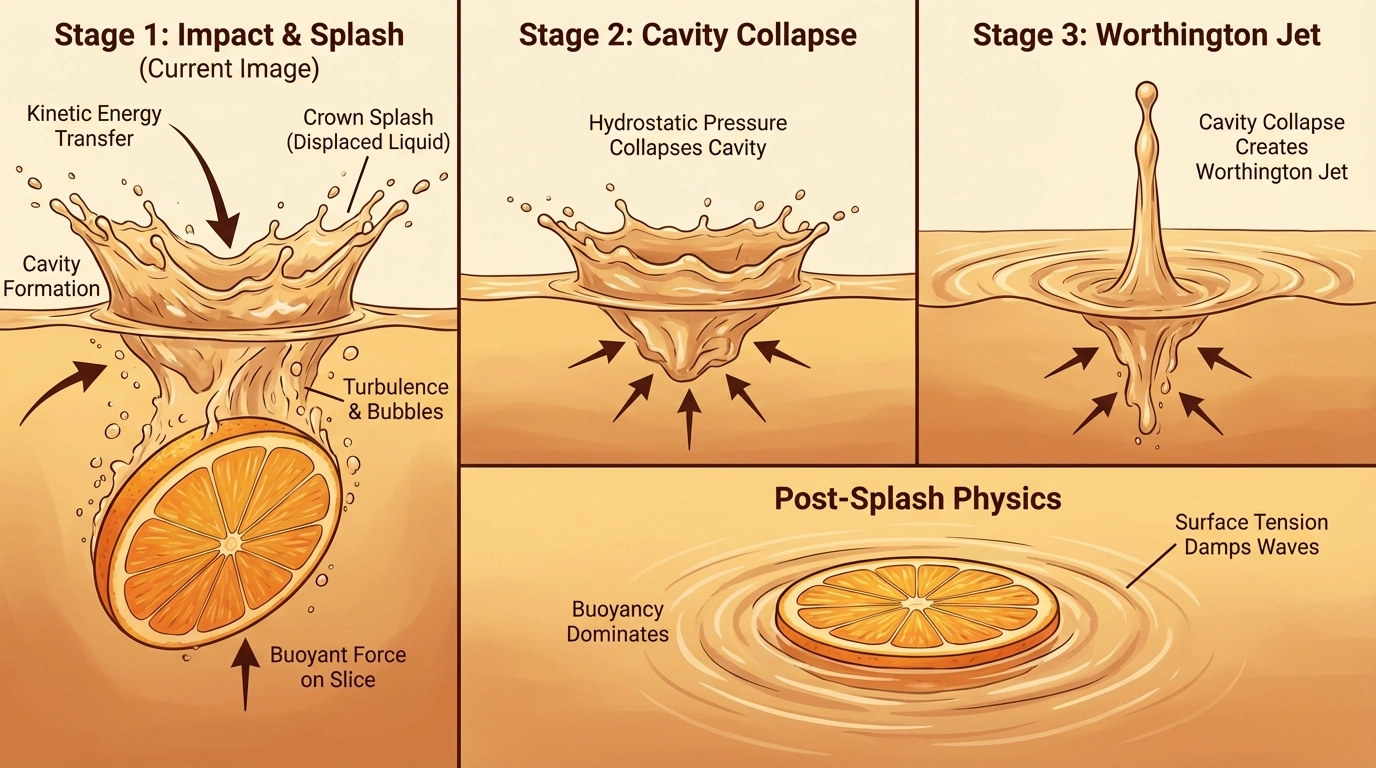}
\caption{Analytical Explainer}
\end{subfigure}
\hfill
\begin{subfigure}[b]{0.41\textwidth}
\includegraphics[width=\linewidth]{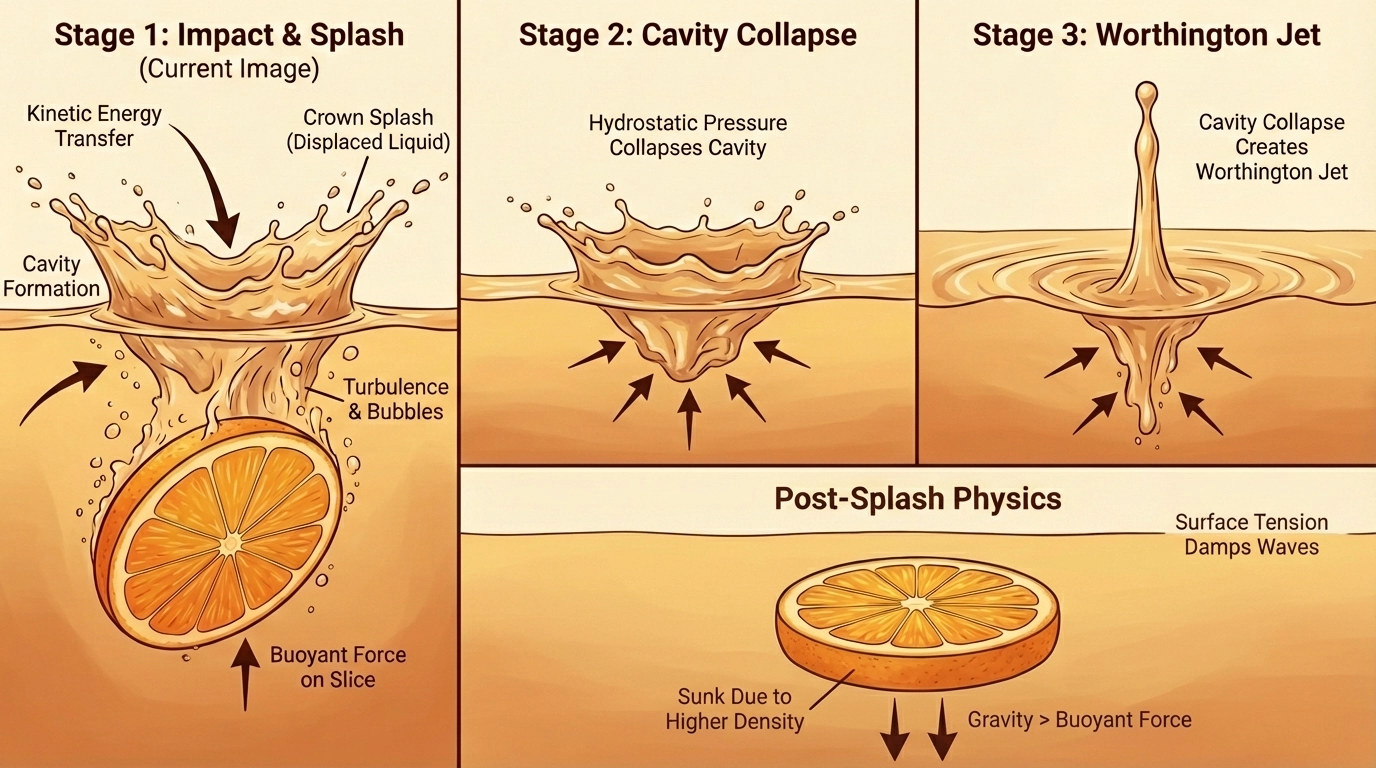}
\caption{Counterfactual: Sinking}
\end{subfigure}
\caption{\textbf{Causal Transition in Fluid Dynamics.} The model successfully transitions from a surface-level state (a) to a submerged state (c) upon a counterfactual prompt. Note the addition of micro-bubbles and shifted refractive caustics, indicating an implicit understanding of fluid interaction. Generated by Nano Banana.}
\label{fig:fluid_case}
\end{figure}

\subsection{Dimension II: Physical Reasoning \& Causal Fidelity}

\noindent\textit{Primary Level Tested: L5 (World-Modeling Generation) --- Can the model predict what would physically happen under intervention, not merely what looks plausible?}

\paragraph{Core Ability: Does the model understand "How the world works" (Gravity, Lighting, Crowd Dynamics)?}
Pure physical reasoning, however, is not always evaluated in a purely textual form. Some of the most revealing cases arise when the model must \emph{read} a visually presented physics problem, \emph{solve} it, and then \emph{write the derivation back} onto the same image. Such tasks lie at the boundary between \textit{Physical Reasoning \& Causal Fidelity} and \textit{Visual-Textual Integration \& Logic}. In this section, we place our physics-exam case primarily under Dimension III, because the decisive challenge is not only whether the model can infer the correct physical law, but whether it can close the full OCR$\rightarrow$reasoning$\rightarrow$rendering loop without breaking the visual structure of the page.

\subsubsection{Case Study I: Fluid Dynamics and Counterfactual Buoyancy}
\label{subsubsec:fluid_dynamics_case}

To probe the model's grasp of \textit{World-Modeling Generation}, we designed a two-stage stress test involving a floating object and a counterfactual physical intervention.

\paragraph{Experimental Setup.}
We utilized a high-resolution image of a citrus slice floating on a water surface as the anchor. The evaluation was split into two instructions:
\begin{itemize}
\item \textbf{Instruction A (Analytical Rendering):} "Create an illustrated explainer detailing the physics of the fluid dynamics caught in this image." This tests \textit{Visual-Textual Integration} (Dimension III) and the ability to map physical concepts (e.g., surface tension, Archimedes' principle) onto a visual substrate.
\item \textbf{Instruction B (Counterfactual State Transition):} "What if the orange piece sinks down the water?" This requires the model to move beyond "Stochastic Plausibility" toward "Causal Simulation," predicting the resulting change in lighting, bubbles, and displacement.
\end{itemize}

\paragraph{Observed Behavior.}
As shown in \Cref{fig:fluid_case}, the model demonstrates a sophisticated "VLM-first, renderer-second" pipeline. In the analytical explainer, it successfully generates vector-like annotations and labels that align with the spatial coordinates of the object. More critically, in the sinking scenario, the model does not merely lower the object's vertical position; it introduces \textbf{causal artifacts} such as trailing air bubbles and altered light refraction consistent with a submerged environment.

\paragraph{Insight.}
This case suggests that frontier models are beginning to bridge the gap between \textbf{L2 (Conditional Generation)} and \textbf{L5 (World-Modeling Generation)}. While the "Thinking Trace" often reveals a redundant search for consistency, the final output indicates that "Reasoning-conditioned document editing" is a viable path toward agentic visual systems that understand "how the world works".

\subsubsection{Case Study II: Action-Conditioned Navigation and Predictive World Modeling}
\label{subsubsec:driving_collision_case}

We further evaluate the model's capacity for \textbf{predictive world modeling} within an embodied driving context. This requires the model to predict future visual states conditioned on specific agent actions or high-kinetic counterfactuals.

\paragraph{Experimental Setup.}
The model is presented with two navigation-based challenges:
\begin{itemize}
    \item \textbf{Scenario A (Intersection Turning):} From a first-person view at a signalized intersection, the model is asked if a turn is possible and to generate the resulting view. This tests the preservation of landmarks and dynamic actors (pedestrians) across a 3D coordinate shift.
    \item \textbf{Scenario B (High-Speed Collision):} Given a highway scene, the model is prompted with a severe counterfactual: "What if I further speed up and crash into the car ahead?".
\end{itemize}

\begin{figure}[t]
    \centering
    \begin{subfigure}[b]{0.48\textwidth}
        \includegraphics[width=\linewidth,height=4cm,keepaspectratio]{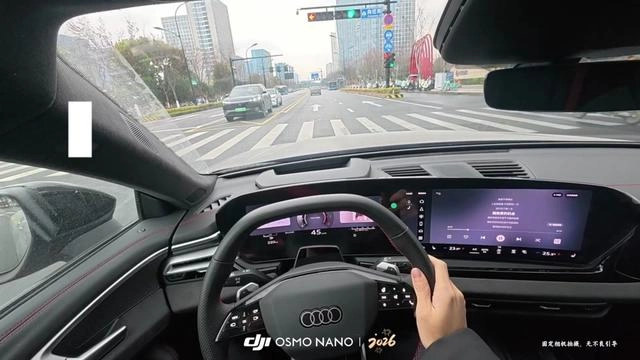}
        \caption{Scenario A: Initial Intersection State}
    \end{subfigure}
    \hfill
    \begin{subfigure}[b]{0.48\textwidth}
        \includegraphics[width=\linewidth,height=4cm,keepaspectratio]{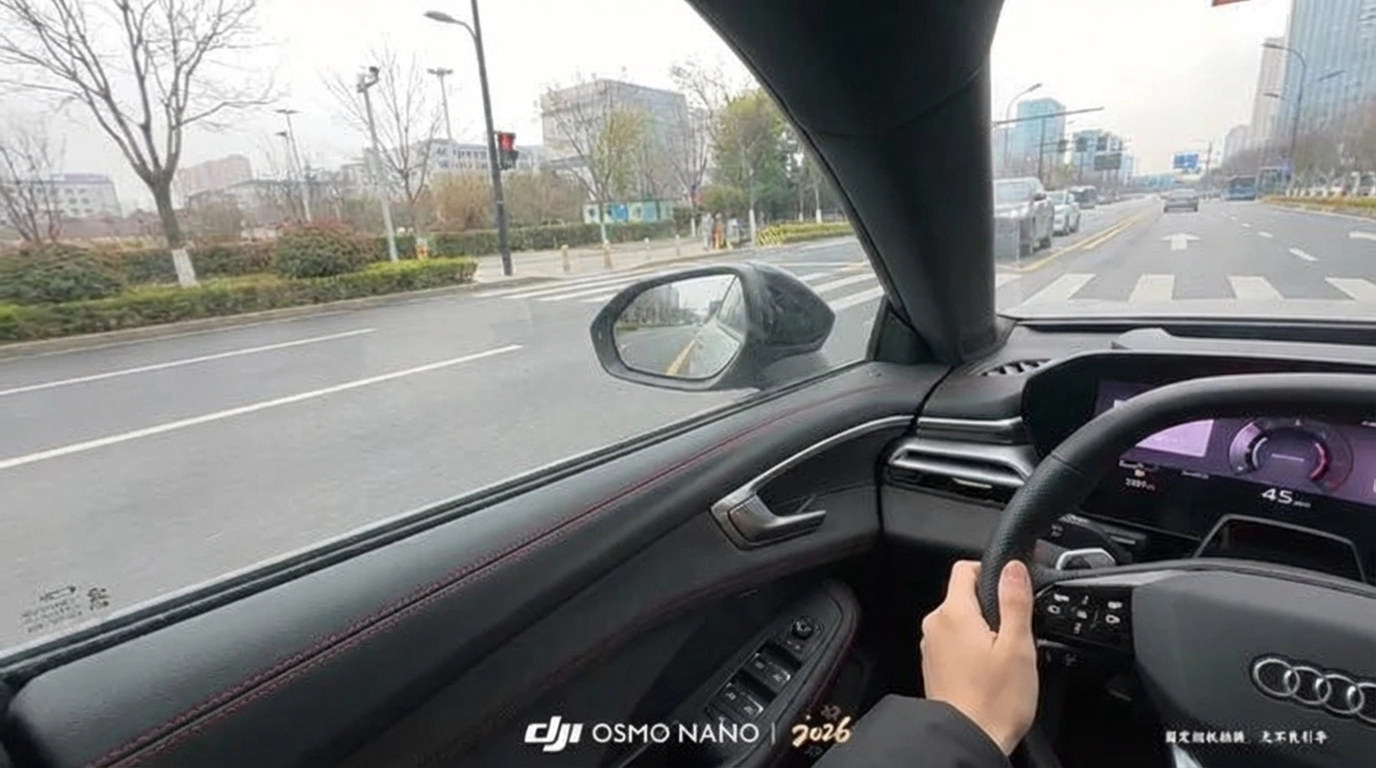}
        \caption{Predicted State: Perspective Shift}
    \end{subfigure}
    
    \vspace{1em}
    
    \begin{subfigure}[b]{0.31\textwidth}
        \includegraphics[width=\linewidth,height=4cm,keepaspectratio]{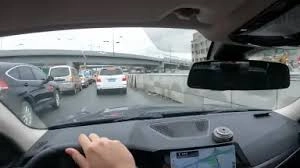}
        \caption{Scenario B: Highway Anchor}
    \end{subfigure}
    \hfill
    \begin{subfigure}[b]{0.31\textwidth}
        \includegraphics[width=\linewidth,height=4cm,keepaspectratio]{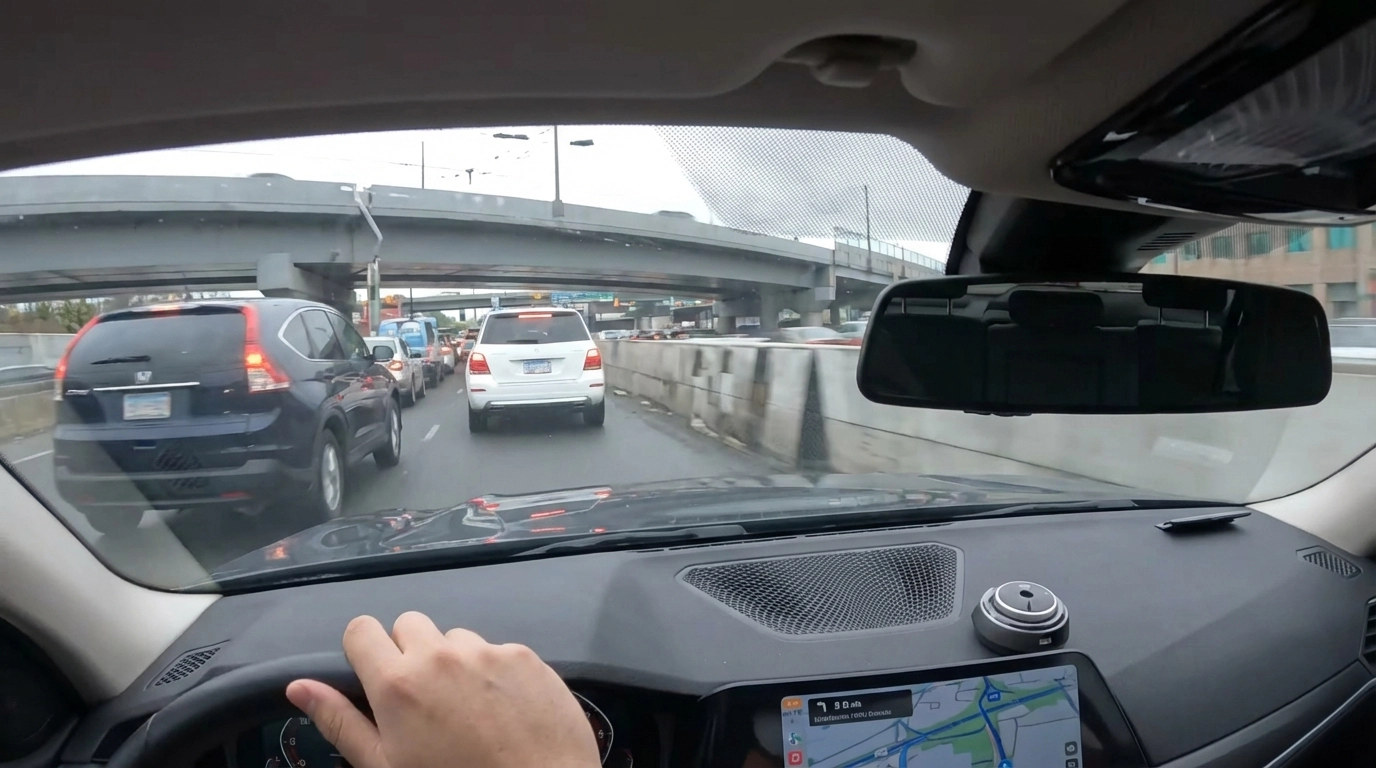}
        \caption{Predicted State: High Speed}
    \end{subfigure}
    \hfill
    \begin{subfigure}[b]{0.31\textwidth}
        \includegraphics[width=\linewidth,height=4cm,keepaspectratio]{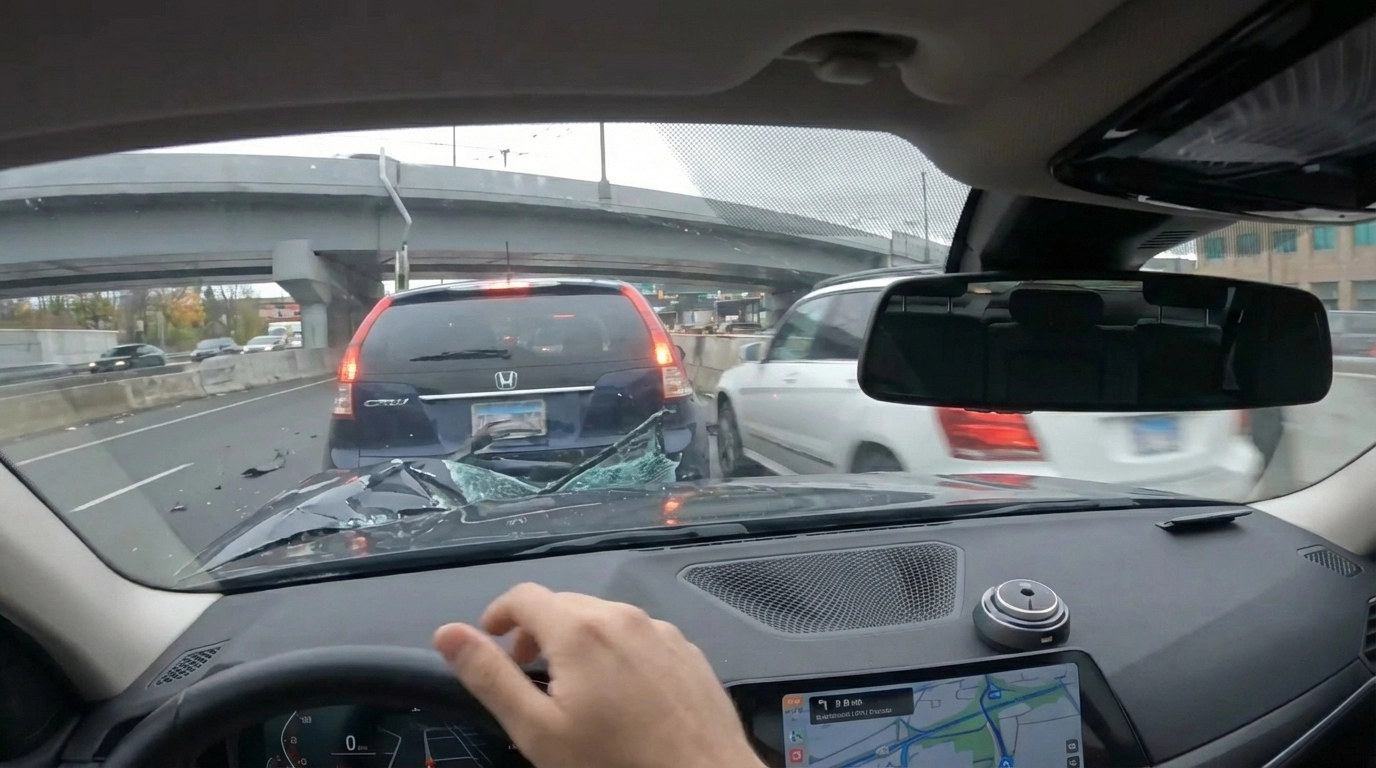}
        \caption{Counterfactual: Collision}
    \end{subfigure}
    \caption{\textbf{Action-Conditioned Causal Prediction.} The model successfully simulates visual markers of high-speed travel (motion blur) and the structural consequences of a physical crash, including metal deformation and fragmentation, indicating significant progress toward Level 5 World Simulation. Generated by Nano Banana.}
    \label{fig:driving_causal_test}
\end{figure}

\paragraph{Insight.}
While the model demonstrates high structural accuracy in the collision test, its failure to explicitly flag immediate safety constraints (e.g., the pedestrian currently in the crosswalk during Scenario A) highlights a critical gap between "Visual Prediction" and "Grounded Causal Reasoning". This reinforces the finding that current systems often prioritize semantic plausibility over robust causal competence.

\subsubsection{Case Study III: Task-Oriented Action Grounding and Robotic Manipulation}
\label{subsubsec:robot_grasp_case}

To move from passive physical simulation toward proactive, goal-directed behavior, we evaluate the model's capacity for \textbf{Generative Action Grounding}. This task requires the model to translate a high-level functional intent (e.g., ``grasp the cup'') into a visually and physically coherent action plan.

\paragraph{Experimental Setup.}
The model is presented with a robot workspace featuring a robotic end-effector and a target object (a white cup) on a flat surface. We issue the prompt: \textit{``If I have a robot arm with gripper and I need to grasp the cup, how should the gripper grasp the cup on the table. Visualize it.''} The model's response is evaluated based on its ability to propose a gripper orientation that accounts for the object's geometry and the physics of contact.

\begin{figure}[t]
    \centering
    \begin{subfigure}[b]{0.31\textwidth}
        \includegraphics[width=\linewidth]{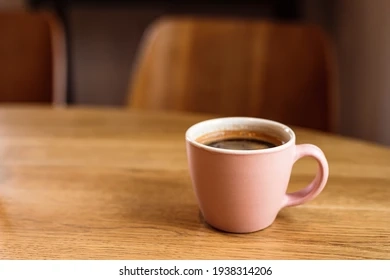}
        \caption{Input: Mug to grasp}
    \end{subfigure}
    \hfill
    \begin{subfigure}[b]{0.31\textwidth}
        \includegraphics[width=\linewidth]{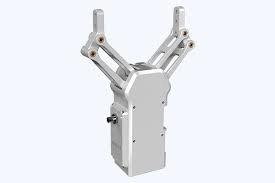}
        \caption{Input: Robot gripper}
    \end{subfigure}
    \hfill
    \begin{subfigure}[b]{0.31\textwidth}
        \includegraphics[width=\linewidth]{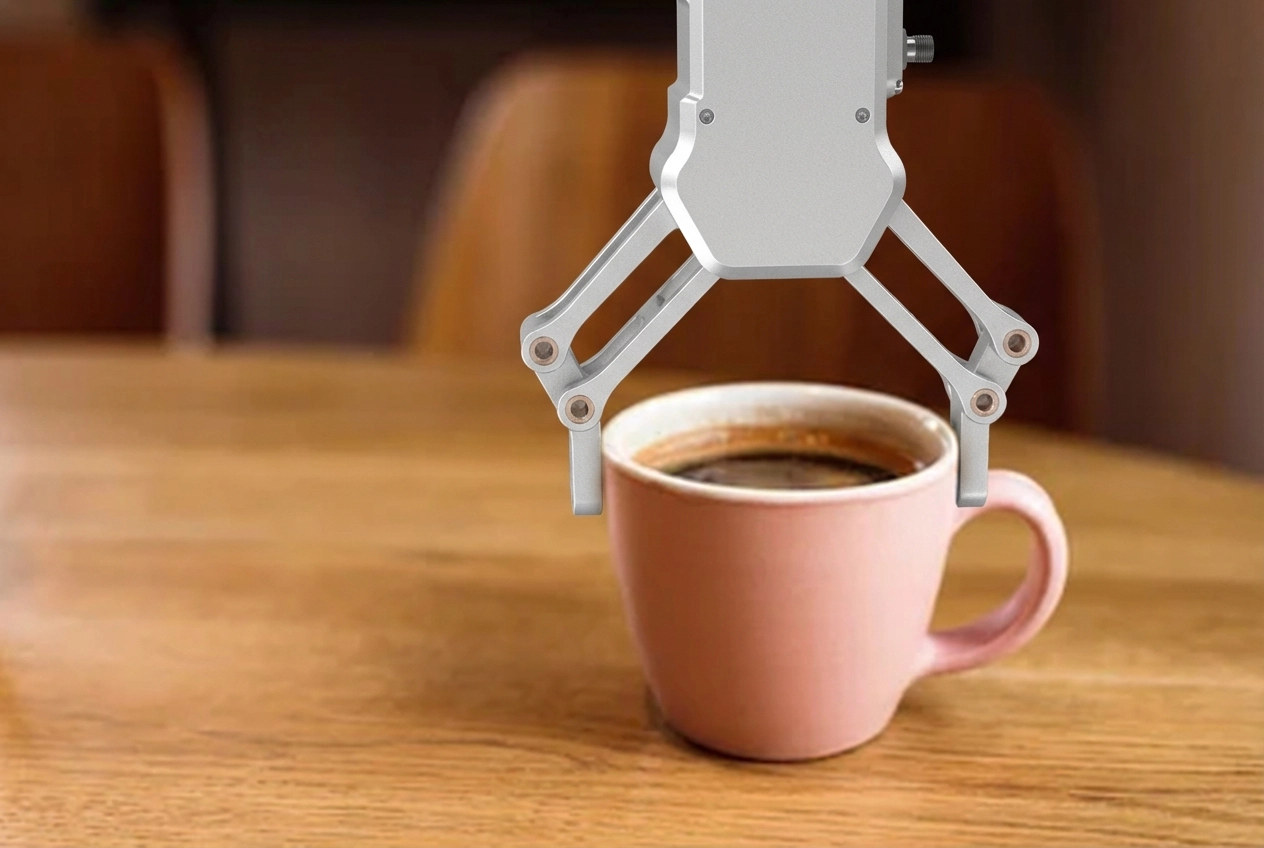}
        \caption{Photorealistic Execution}
    \end{subfigure}
    \caption{\textbf{Generative Action Grounding for Robotic Manipulation.} (a-b) The input mug to grasp and the robot gripper. (c) The model successfully synthesizes a high-fidelity anthropomorphic grasp, demonstrating an implicit understanding of \textbf{Contact Manifold} dynamics and force-closure. Generated by Nano Banana.}
    \label{fig:robot_grasp}
\end{figure}

\paragraph{Analysis of Physical Reasoning.}
The results in \Cref{fig:robot_grasp} reveal that the model possesses an emergent ``World Model'' for manipulation. In the schematic examples (a-b), the model maintains \textbf{Object Permanence} and correctly identifies the cylindrical affordance of the cup. In the photorealistic rendering (c), the model avoids the common failure of interpenetrating the mesh; instead, it wraps the robotic fingers around the mug in a way that suggests an understanding of friction and surface normals.

\paragraph{Insight.}
This case study highlights that advanced VLMs can act as \textbf{Visual Policy Proposals}. While the model is not outputting raw joint torques, its ability to generate a visually valid future state—conditioned on a functional goal—marks a significant step toward \textit{Agentic Visual Intelligence}. It demonstrates that the model is not just a stochastic parrot of pixels, but a system capable of running a ``mental simulation'' of physical contact to guide its generative process.

\subsubsection{Case Study IV: Spatiotemporal Trajectory Synthesis for Multi-Step Tasks}
\label{subsubsec:trajectory_synthesis}

To evaluate the model's capacity for long-horizon physical reasoning, we move beyond static pose estimation to \textbf{Spatiotemporal Trajectory Synthesis}. In this task, the model must visualize a multi-step sequence of actions that transition a scene from a "source" state to a "target" state, maintaining causal logic throughout the motion.

\paragraph{Experimental Setup.}
The model is provided with a workspace containing a green spoon and a wooden bowl (\texttt{spoon\_input.png}). We issue a trajectory-based prompt: \textit{``Generate the trajectory to put the green spoon into the wooden bowl as a sequence of images from the same camera.''} This requires the model to simulate a continuous path that respects the physical constraints of both the robotic end-effector and the environment.

\begin{figure}[!htbp]
    \centering
    \begin{subfigure}[b]{0.48\textwidth}
        \includegraphics[width=\linewidth]{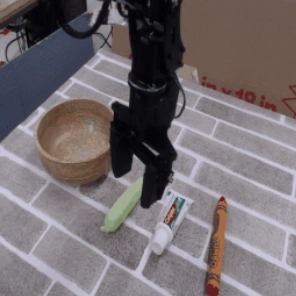}
        \caption{Input: Initial State}
    \end{subfigure}
    \hfill
    \begin{subfigure}[b]{0.48\textwidth}
        \includegraphics[width=\linewidth]{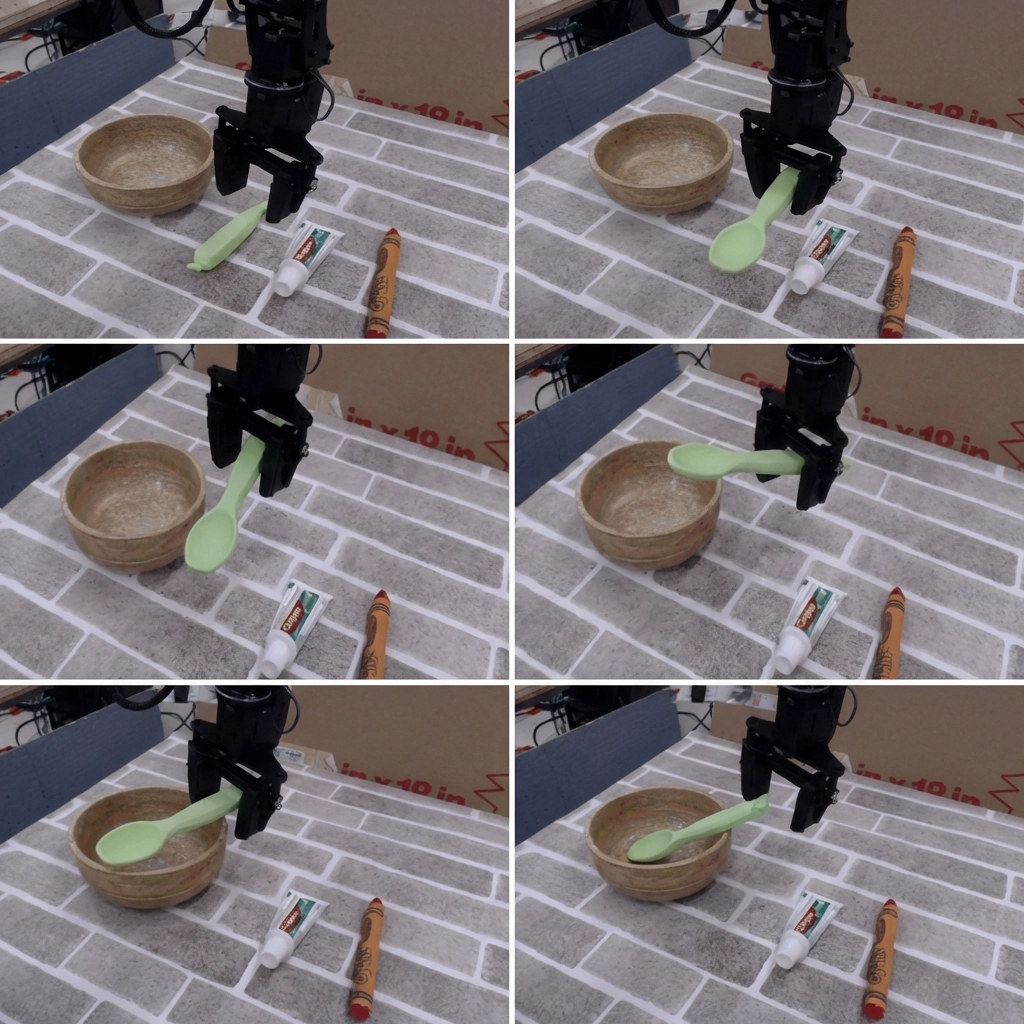}
        \caption{Output: Generated trajectory}
    \end{subfigure}
    \caption{\textbf{Sequential Trajectory Generation.} The model demonstrates an ability to synthesize a physically plausible path through time. It maintains strict \textbf{Identity Persistence} for the spoon and correctly predicts the spatial containment relationship required for the task. This illustrates a transition toward Level 5 intelligence, where visual generation serves as a proxy for physical planning. Generated by Nano Banana.}
    \label{fig:spoon_trajectory}
\end{figure}

\paragraph{Analysis of Spatiotemporal Consistency.}
The generated sequence in \Cref{fig:spoon_trajectory} reveals two critical competencies:
\begin{enumerate}
    \item \textbf{Kinematic Logic:} The robotic arm's movement follows a natural arc, indicating an implicit understanding of joint constraints and end-effector orientation.
    \item \textbf{Containment Awareness:} In the final frame, the spoon is correctly occluded by the rim of the bowl, signifying that the model understands the 3D volume of the container and the physics of the ``inside'' relationship.
\end{enumerate}
However, there is an obvious failure in the first two frames of the generated trajectory, where the spoon orientation is wrong, indicating that more research is needed for visual detail consistencies. 

\begin{figure}[t]
    \centering
    \begin{subfigure}[b]{\textwidth}
        \centering
        \includegraphics[width=0.95\linewidth]{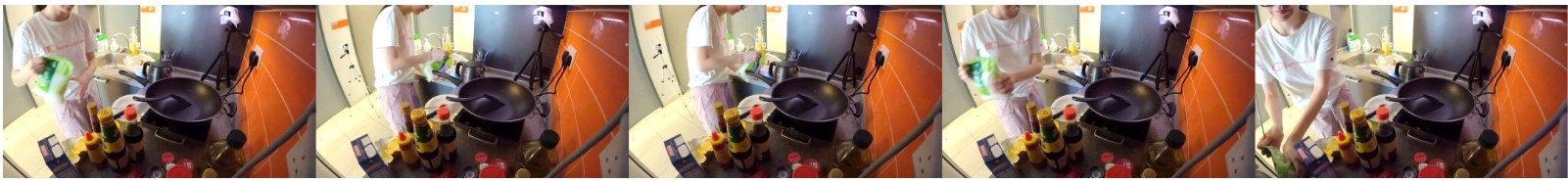}
        \caption{Original Human Sequence: Multi-Step Causal Tasks (e.g., Pouring Action in Frame 2)}
    \end{subfigure}
    
    \vspace{1em}
    
    \begin{subfigure}[b]{\textwidth}
        \centering
        \includegraphics[width=0.95\linewidth,keepaspectratio]{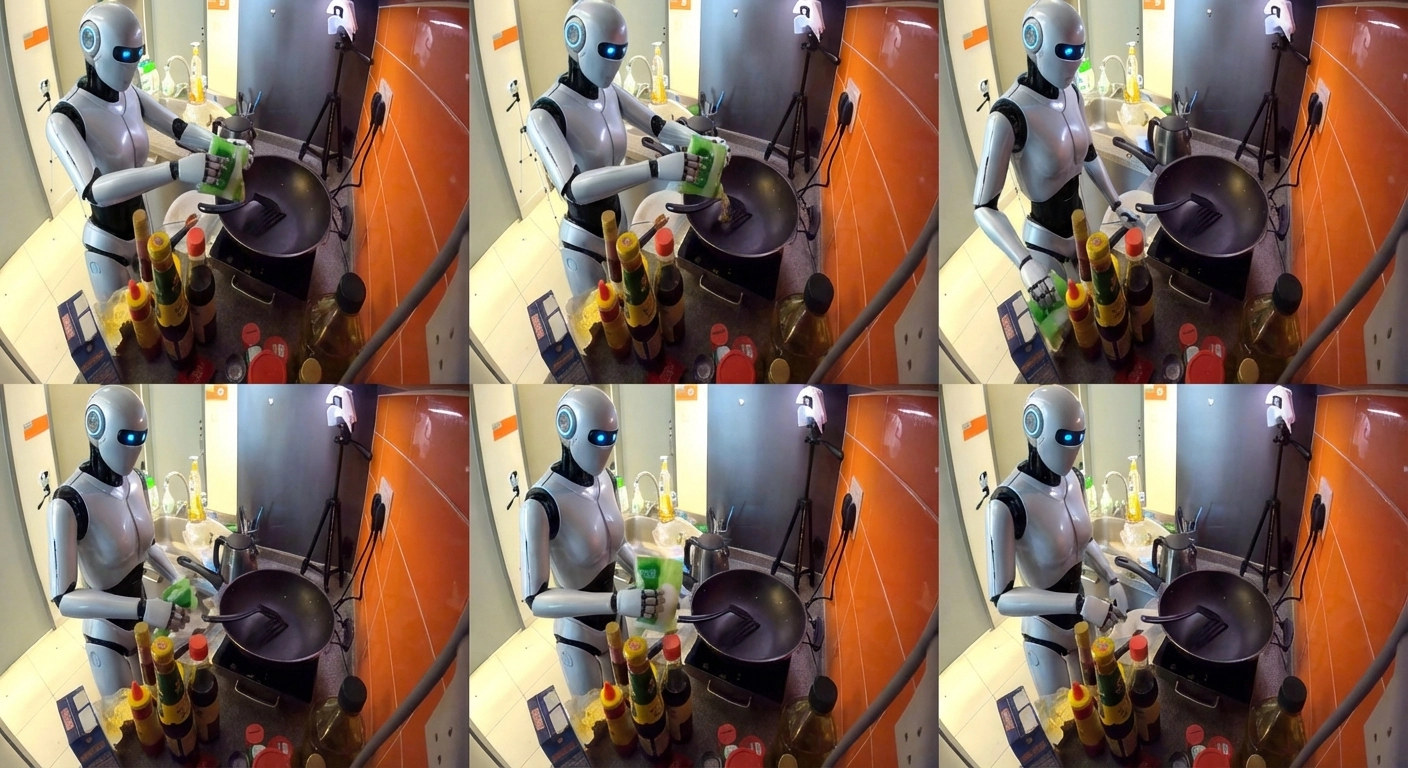}
        \caption{Edited Humanoid Sequence: Plausible Identity but Causal Action Failure (Missing Pouring Simulation)}
    \end{subfigure}
    
    \caption{\textbf{Video Re-rendering and Causal Divergence.} The model achieves high controllable creativity (Level 2) by consistently rendering a single humanoid robot along the original spatiotemporal path. However, it fails to simulate the functional grounding of the task; the pouring action present in the source (a, Frame 2) disappears and has no causal consequence in the edited sequence (b), demonstrating a decoupling of visual fidelity and causal world simulation (Level 5). Generated by Nano Banana.}
    \label{fig:video_rerendering_cases}
\end{figure}

\paragraph{Insight.}
This case study highlights the emergence of \textbf{Visual Action Planning}. By successfully rendering a temporal sequence that adheres to Newtonian intuition, the model proves it can act as a simulator for high-level robotic tasks. However, a remaining challenge for ``Agentic Intelligence'' is ensuring these trajectories are not just visually plausible but also kinematically executable in real-world control systems.

\subsubsection{Case Study V: Video Re-rendering with Functional Causal Failure}
\label{subsubsec:video_re-rendering_case}

To evaluate the limits of high-fidelity spatiotemporal consistency in video-to-video editing, we conducted a real-world multi-frame stress test requiring complex identity replacement: transforming a human performing a sequential kitchen task into a consistent humanoid robot while preserving the trajectory and functional intent.

\paragraph{Experimental Setup.}
The model is presented with a consecutive four-frame sequence from a kitchen camera (\texttt{human\_input.jpg}), featuring a human performing complex manipulation tasks around a stove and sink. We issue a task-conditioned visual editing prompt: \textit{``Replace the human in those sequential frames with a humanoid. The robot must be consistent along frames.''} The evaluation criteria are (a) identity persistence of the humanoid, and (b) preservation of the original functional action and its causal consequences.

\paragraph{Observation and Analysis.}
As depicted in \Cref{fig:video_rerendering_cases}, the model successfully replaces the human with a single, visually detailed, and physically plausible humanoid that maintains identity persistence across all frames. However, a significant divergence occurs regarding \textbf{Functional Causal Persistence}. In frame 2 of the source sequence (\Cref{fig:video_rerendering_cases}a, top right), the human is clearly pouring from a container into a pot. The model, while pathing the robot to the same location, fails to render the \textit{act of pouring} or simulate its consequences (e.g., fluid dynamics, maintained pose). The action is reduced to a semantic gesture of interaction, rather than a simulation.

\paragraph{Insight.}
This real-world example exposes the gap between ``Pixel-space Control'' and ``Physical Simulation''. The model excels at L2/L3 (Conditional and In-Context Generation) but remains far from L5 (World-Modeling Generation). It treats a sequential visual action as a chain of static texturally consistent states rather than a continuous event with logical consequences. The failure of the pouring action highlights that while visual generation is reaching high maturity, grounding these generations in robust physical and logical world models is a persistent bottleneck for agentic visual systems that must operate safely and functionally in dynamic environments.

\subsubsection{Case Study VI: Irreversible State Transitions and Internal Material Consistency}
\label{subsubsec:material_transformation_case}

To push the limits of \textit{World-Modeling Generation}, we evaluate the model's ability to represent irreversible physical transformations—specifically \textbf{Material Removal} and \textbf{Subdivision}. These tasks require the model to predict the internal structure of an object that is hidden in the initial state, testing the depth of its volumetric world knowledge.

\paragraph{Experimental Setup.}
The model is provided with a whole zucchini and a rainbow carrot (\texttt{veg\_input.jpg}). We issued two goal-conditioned instructions: (1) \textit{``Generate the goal image of cutting a zucchini,''} and (2) \textit{``Generate the goal image of peeling a rainbow carrot.''} A successful simulation must preserve the object's identity while correctly rendering the internal flesh and resulting debris (slices/shavings).

\begin{figure}[!htp]
    \centering
    \begin{subfigure}[b]{0.45\textwidth}
        \includegraphics[width=\linewidth]{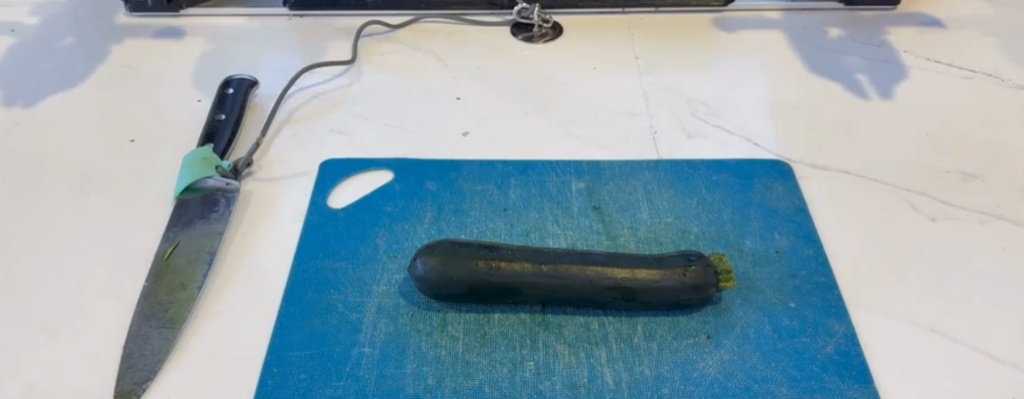}
        \caption{Initial State: Original Zucchini}
    \end{subfigure}
    \hfill
    \begin{subfigure}[b]{0.45\textwidth}
        \includegraphics[width=\linewidth]{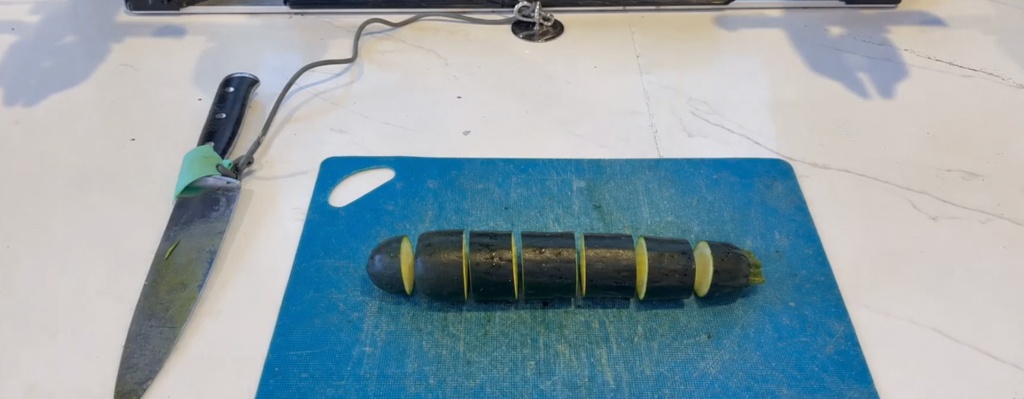}
        \caption{Goal State: Cut Zucchini}
    \end{subfigure}
    
    \vspace{1em}
    
    \begin{subfigure}[b]{0.45\textwidth}
        \includegraphics[width=\linewidth]{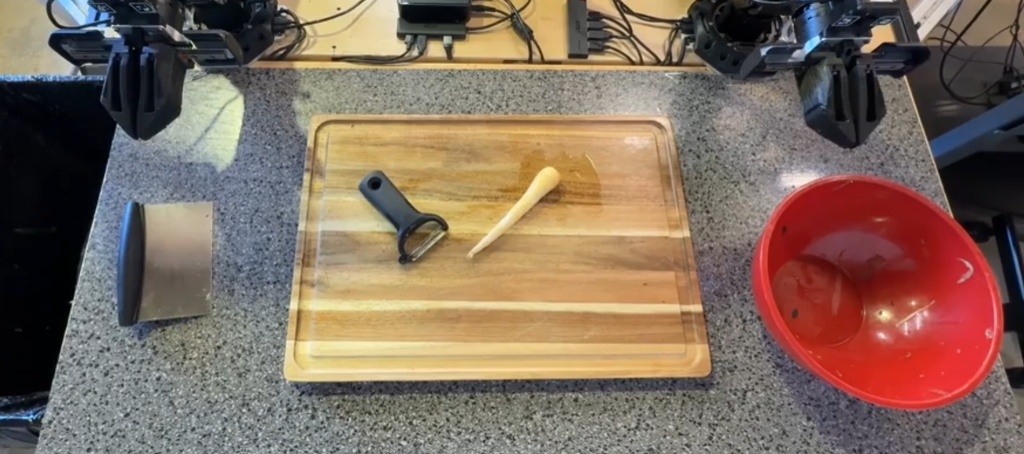}
        \caption{Initial State: Original Carrot}
    \end{subfigure}
    \hfill
    \begin{subfigure}[b]{0.45\textwidth}
        \includegraphics[width=\linewidth]{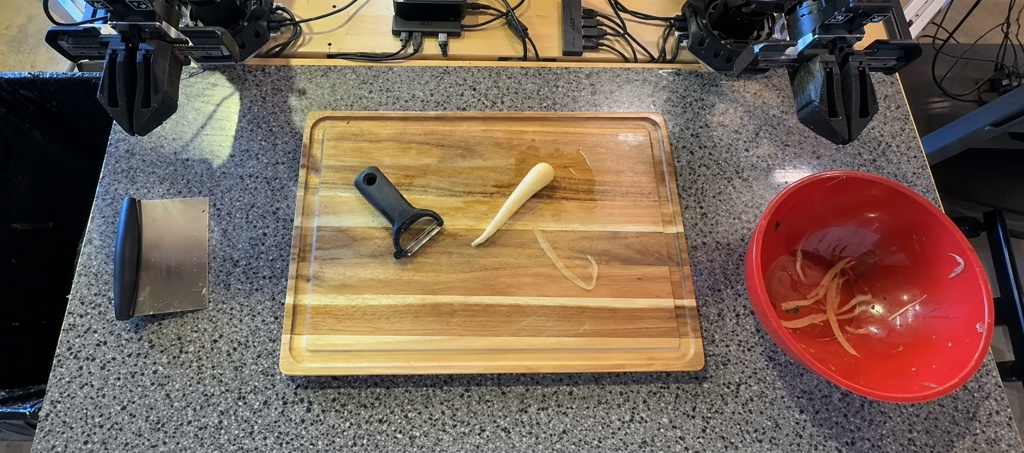}
        \caption{Goal State: Peeled Rainbow Carrot}
    \end{subfigure}
    \caption{\textbf{Irreversible Material Transformation.} The model demonstrates an advanced understanding of object interiority. In (b), it correctly renders the internal seeds and texture of the zucchini. In (d), it maintains the specific color gradient of the rainbow carrot in the resulting shavings, showcasing high \textbf{Causal Identity Persistence} across state changes. Generated by Nano Banana.}
    \label{fig:transformation_cases}
\end{figure}

\paragraph{Analysis of Volumetric Reasoning.}
The generated images in \Cref{fig:transformation_cases} reveal that the model treats the vegetables as 3D volumes rather than 2D sprites. The transition from (a) to (b) and (c) to (d) involves a significant change in geometry. The model accurately simulates the ``Internal Material Properties'': the zucchini slices have a logically consistent diameter and interior color, while the carrot shavings exhibit the correct thickness and curvature expected from a manual peeler.

\paragraph{Insight.}
This case study highlights the model's capacity for \textbf{Counterfactual State Synthesis}. By accurately imagining the "inside" of an object based on its "outside," the model moves toward Level 5 intelligence. However, the exact arrangement of the slices in (b) remains somewhat stochastic; a truly agentic model would need to ground these transformations in a rigid kinematic path (e.g., the specific movement of a knife), bridging the gap between visual plausibility and physical execution.

\subsection{Dimension III: Visual-Textual Integration \& Logic}

\noindent\textit{Primary Level Tested: L4 (Agentic Generation) --- Can the model read, reason, and write back onto the same visual substrate in a coherent closed-loop workflow?}

\paragraph{Core Ability: The fusion of OCR (Reading), Reasoning (Solving), and Rendering (Writing).}

\subsubsection{Case Study I: Solving a Physics Exam Directly on the Image}
\label{subsubsec:physics_solver_case}

To probe whether a frontier model can truly \emph{read, think, and write back} on the same visual substrate, we constructed a physics-tutoring case in which the input is a Chinese Gaokao-style electromagnetism problem and the output must be an \emph{annotated solution image} rather than plain text. The user prompt explicitly casts the model as a ``Visual Physics Solver \& Annotator'' and requires an end-to-end workflow: parse the question and diagram, identify the governing physical laws, derive the equations, perform the calculations, and overlay the step-by-step solution directly onto the image or an extended whitespace region. The output must preserve the original question, use legible high-contrast handwriting or clean typesetting, remain in Chinese, and present a tutor-like structure including \emph{Given}, \emph{Analysis}, \emph{Derivation}, and \emph{Result}. This transforms a seemingly standard physics problem into a much more demanding multimodal task.

\paragraph{Experimental Setup.}
The case revolves around a classic inclined-plane electromagnetic induction problem: a conducting rod and a U-shaped frame slide on a smooth slope, enter and leave a magnetic field at different moments, and are coupled through induced current, Ampere force, gravity, and friction. Unlike text-only physics QA, the model must jointly process three types of information:
\begin{itemize}
    \item \textbf{Dense document OCR:} The problem statement is embedded in a photographed or scanned exam page with Chinese text, inline formulas, and a geometric diagram.
    \item \textbf{Diagram grounding:} Variables such as $CD$, $EF$, $L$, $s_0$, $s_1$, $\alpha$, and the direction of the magnetic field must be aligned with the figure rather than treated as independent text tokens.
    \item \textbf{Layout-aware response generation:} The derivation must be inserted without obscuring the original question, and the rendered equations must remain readable and pedagogically structured.
\end{itemize}
The target output therefore is not merely a correct scalar answer. It is a visually integrated worked solution.

\begin{figure}[!htbp]
    \centering
    \begin{subfigure}[b]{0.48\textwidth}
        \centering
        \IfFileExists{img/stress_test/math/input.jpg}{%
            \includegraphics[width=\linewidth,height=7.2cm,keepaspectratio]{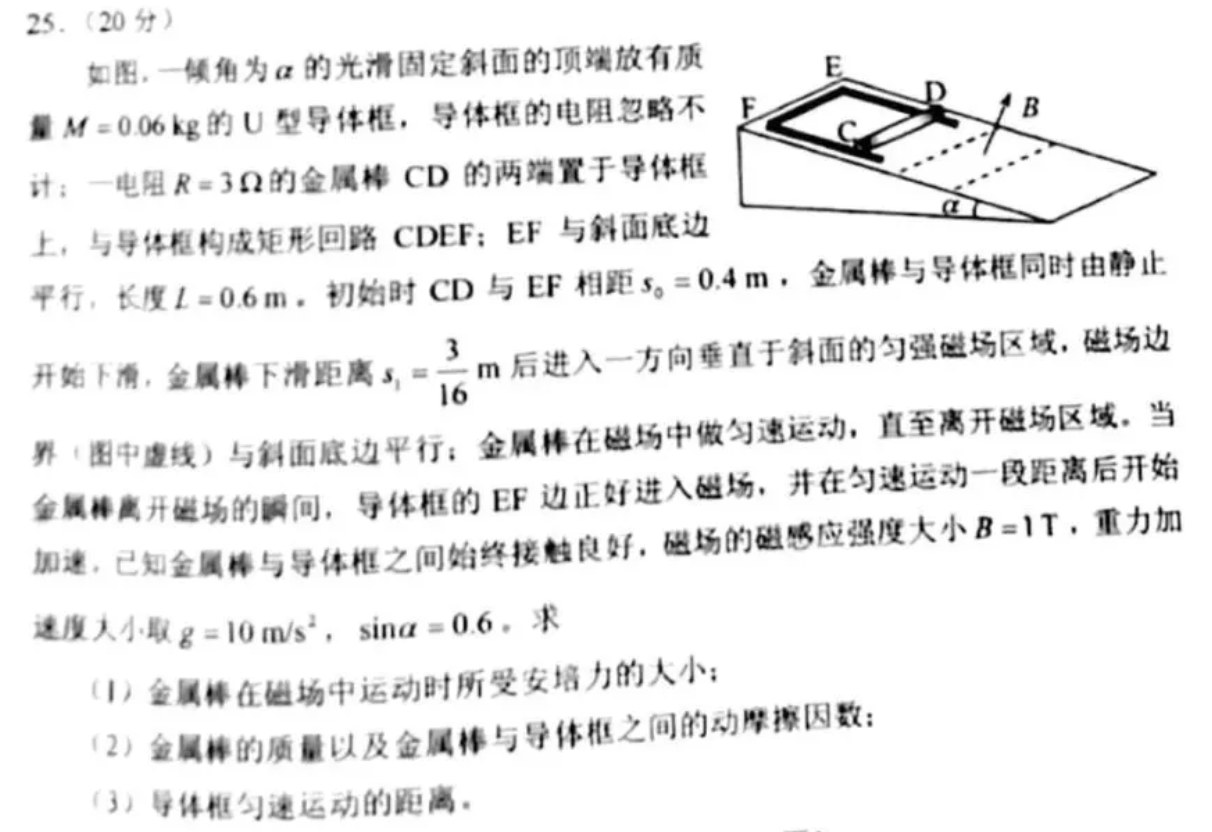}%
        }{%
            \fbox{\parbox[c][7.2cm][c]{0.92\linewidth}{\centering
            \textbf{Input Image Placeholder}\\[0.5em]
            Put the source image at\\
            \texttt{img/stress\_test/math/input.jpg}}}%
        }
        \caption{Input: exam-style physics problem}
    \end{subfigure}
    \hfill
    \begin{subfigure}[b]{0.48\textwidth}
        \centering
        \IfFileExists{img/stress_test/math/output.jpg}{%
            \includegraphics[width=\linewidth,height=7.2cm,keepaspectratio]{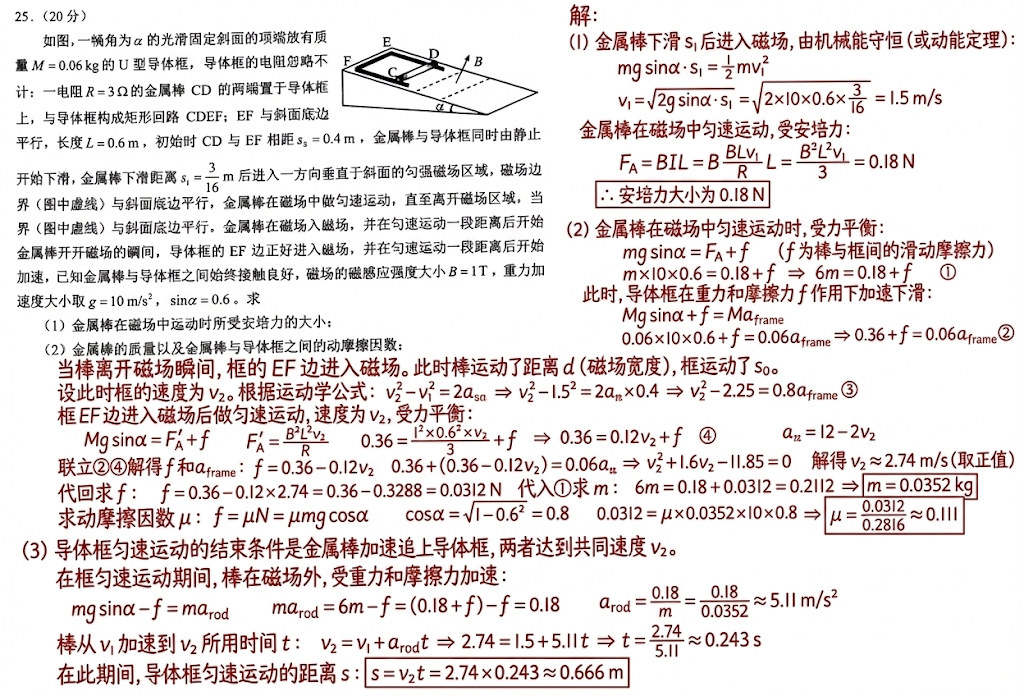}%
        }{%
            \fbox{\parbox[c][7.2cm][c]{0.92\linewidth}{\centering
            \textbf{Output Image Placeholder}\\[0.5em]
            Put the solved image at\\
            \texttt{img/stress\_test/math/physics\_exam\_output.png}}}%
        }
        \caption{Model output: solution rendered onto the image}
    \end{subfigure}
    \caption{\textbf{Physics problem solving as reasoning-conditioned document editing.} The model is asked not only to solve the problem, but to write the derivation back onto the original visual artifact without obscuring the source text or diagram. This case therefore jointly tests OCR, diagram grounding, symbolic reasoning, layout planning, and text rendering. Generated by Nano Banana.}
    \label{fig:physics_solver_case}
\end{figure}

\paragraph{Observed Behavior.}
As shown in \Cref{fig:physics_solver_case}, the generated result is striking because it does more than decorate the image with red formulas. It appears to reconstruct a plausible expert workflow: first deriving the rod's entry velocity from energy conservation, then computing the induced current and Ampere force during the constant-velocity stage in the magnetic field, and finally solving for the rod mass, kinetic-friction coefficient, and the frame's constant-velocity distance through a coupled kinematic and force-balance analysis. The final image also demonstrates strong document-level awareness. It preserves the original question and diagram, writes the derivation into unused white space, keeps the explanation language consistent with the Chinese prompt, and uses contrasting red annotations to separate solution content from source content. Importantly, the final numerical results are not merely stylistic artifacts; they are mutually consistent under a standard derivation, yielding $F_A=0.18\,\mathrm{N}$, $m\approx 0.0352\,\mathrm{kg}$, $\mu \approx 0.111$, and $s\approx 0.666\,\mathrm{m}$.

\paragraph{What the Thinking Trace Reveals.}
The accompanying reasoning trace reveals a much less smooth internal process than the polished output image suggests. Rather than following a compact symbolic plan, the model repeatedly revisits the same subproblems: it re-analyzes the rod-frame velocity relation, reconsiders whether the relevant friction should be treated as kinetic or static, and repeatedly rebuilds the force-balance equations around the transition points at which the rod exits the magnetic field and the frame edge $EF$ enters it. This repetitive self-correction is highly informative. It suggests that the model can eventually converge to a coherent solution, but it does not maintain a stable symbolic workspace in the way a dedicated theorem prover or physics engine would. Instead, it appears to perform a broad and somewhat redundant search over partially grounded hypotheses until a numerically self-consistent chain of reasoning emerges. In other words, the final image demonstrates \emph{productive reasoning}, but the trace indicates that the reasoning process remains fragile, expensive, and only partially structured.

\paragraph{Why This Case Matters.}
This case is substantially richer than a standard OCR benchmark, text-rendering benchmark, or multimodal math QA task. A model can have high OCR accuracy yet still fail to bind symbols to the correct entities in the figure. It can solve the physics problem in text yet fail to produce a legible annotated image. It can render beautiful formulas in red ink yet still hallucinate the derivation or place annotations in semantically inappropriate regions. The benchmark value of this case lies precisely in forcing these subskills to compose. It requires the model to treat the image not only as an object to perceive, but also as an external workspace that must be updated in a logically and spatially coherent way.

\paragraph{Does This Capability Truly Require Image Generation?}
This question is especially important because the task superficially looks like an image-generation benchmark, but its cognitive core is different. The hardest part of the task is not natural-image synthesis. It is the multimodal chain of \textbf{OCR $\rightarrow$ diagram grounding $\rightarrow$ symbolic reasoning $\rightarrow$ structured visual re-rendering}. Under this decomposition, the first three stages are much closer to VLM understanding and symbolic problem solving than to diffusion-style generation. A strong modular system---for example, a \emph{VLM parser}, a \emph{symbolic physics solver}, a \emph{layout planner}, and a \emph{text renderer}---could likely solve this task more reliably than a monolithic end-to-end image generator. From that perspective, the task does \emph{not} primarily measure whether a model can synthesize realistic pixels from scratch.

At the same time, image generation or image editing is not irrelevant. Once the required output format is ``return only the solved image,'' the model must make document-level visual decisions that a plain VLM or text-only solver does not naturally solve: where to place the derivation, whether to extend the canvas, how to preserve the original diagram, how to draw auxiliary arrows or free-body diagrams, how to maintain contrast, and how to imitate a human tutor's annotation style. These are not classical text-to-image problems, but they are still genuine visual generation and editing problems. The key distinction is that the model is not generating an image \emph{for its own sake}; it is using visual generation as a \emph{structured output interface} for reasoning. We therefore argue that this ability is best described as \textbf{reasoning-conditioned document editing}, not standard image generation.

\paragraph{A Likely Mechanism: VLM-First, Renderer-Second.}
The most plausible account of the observed behavior is a \textbf{VLM-first, renderer-second} pipeline, whether implemented explicitly or implicitly within a unified multimodal foundation model. The system likely first solves the problem in a latent or textual reasoning space, and only then projects the solution back into the image through text rendering or controllable image editing. The final output supports this interpretation: the annotations behave more like a well-controlled document renderer than like unconstrained image synthesis. Meanwhile, the lengthy and repetitive thinking trace suggests that the upstream reasoning stage remains the true bottleneck. This asymmetry is revealing. It implies that recent models may already be surprisingly competent at \emph{presenting} a solution visually once a derivation has been found, while still being much weaker at maintaining a concise, robust, and fully grounded symbolic chain of thought.

\paragraph{Insight.}
This case exposes a capability frontier that standard visual-generation benchmarks largely ignore: the ability to use images as \emph{interactive reasoning canvases}. The task is not well captured by OCR accuracy, math QA accuracy, or text rendering fidelity in isolation. Instead, it tests whether a model can compose all three into a coherent end-to-end behavior. More broadly, it suggests that the next stage of visual intelligence may not be defined only by prettier or more photorealistic images, but by whether models can read a complex visual artifact, reason over it, and then write back into it in a structured, pedagogically useful manner. That is a meaningful step toward agentic visual systems. Yet the thinking trace also warns us that much of the apparent competence may currently arise from a loose coupling between VLM reasoning and a strong rendering layer, rather than from a truly unified, world-grounded model.

\subsection{Dimension IV: Multi-Turn Editing --- Markovian Chaining and Silent Drift}

\noindent\textit{Primary Level(s) Tested: L3 (In-Context Generation) as floor, L4 (Agentic Generation) as ceiling --- Mechanically, each round of multi-turn editing is a single forward pass over the cumulative input $\{I_0, p_1, I_1, \ldots, p_t\}$, the L3 method signature. What we probe, however, is whether the system also exhibits L4 capability: does it genuinely use the full edit history and selectively re-anchor on relevant past turns, or does it collapse to the Markovian shortcut $f(I_{t-1}, p_t)$ whose observable signature is silent, cumulative drift across rounds?}

\paragraph{Core Ability: Beyond carrying multi-turn context across rounds (L3) and selecting which past turns to consult for the current edit (L4), resisting the silent drift that compounds even when every turn is locally faithful---drift is the observable surface of the Markovian shortcut, since with no early-turn anchor each round amplifies the previous round's reconstruction error.}

\begingroup

Multi-turn editing is the natural stress test that separates L3 from L4. Mechanically each round is a single forward pass over the growing input---the L3 method signature; the L4 capability we then probe is whether the model can (i)~remember the full edit history, (ii)~selectively retrieve the past turns that bear on the current instruction, and (iii)~hold cross-turn invariants stable as the chain grows. In practice, however, many systems implement multi-turn as a chain of single-turn calls $f(I_{t-1}, p_t)$, retaining only the last image and the current prompt---a Markovian shortcut whose observable signature is silent, cumulative drift: each individual turn looks locally acceptable, yet small per-turn error accumulates so that earlier-panel detail decays beyond recovery and the trajectory across turns no longer closes back to the original. We probe two complementary layers at which this drift surfaces: \emph{representational} drift inside the model that compounds because every panel is decoded from the previous round's lossy reconstruction (Case~I), and \emph{semantic} drift visible to the user---identity, size, and object persistence shifting silently across turns---which surfaces sharply when the model is asked to ``restore to original'' (Case~II). Together they reveal a failure mode that no single-turn benchmark can surface: each turn is correct in isolation, yet the trajectory across turns is not. Unless a caption states otherwise, outputs in this dimension are from \textbf{Nano Banana}.

\subsubsection{Case Study I: Cumulative Visual Quality Degradation in Multi-Panel Sequential Editing}

\begin{figure}[!htbp]
    \centering
    \begin{subfigure}[b]{0.235\textwidth}
        \centering
        \includegraphics[width=\linewidth,height=4cm,keepaspectratio]{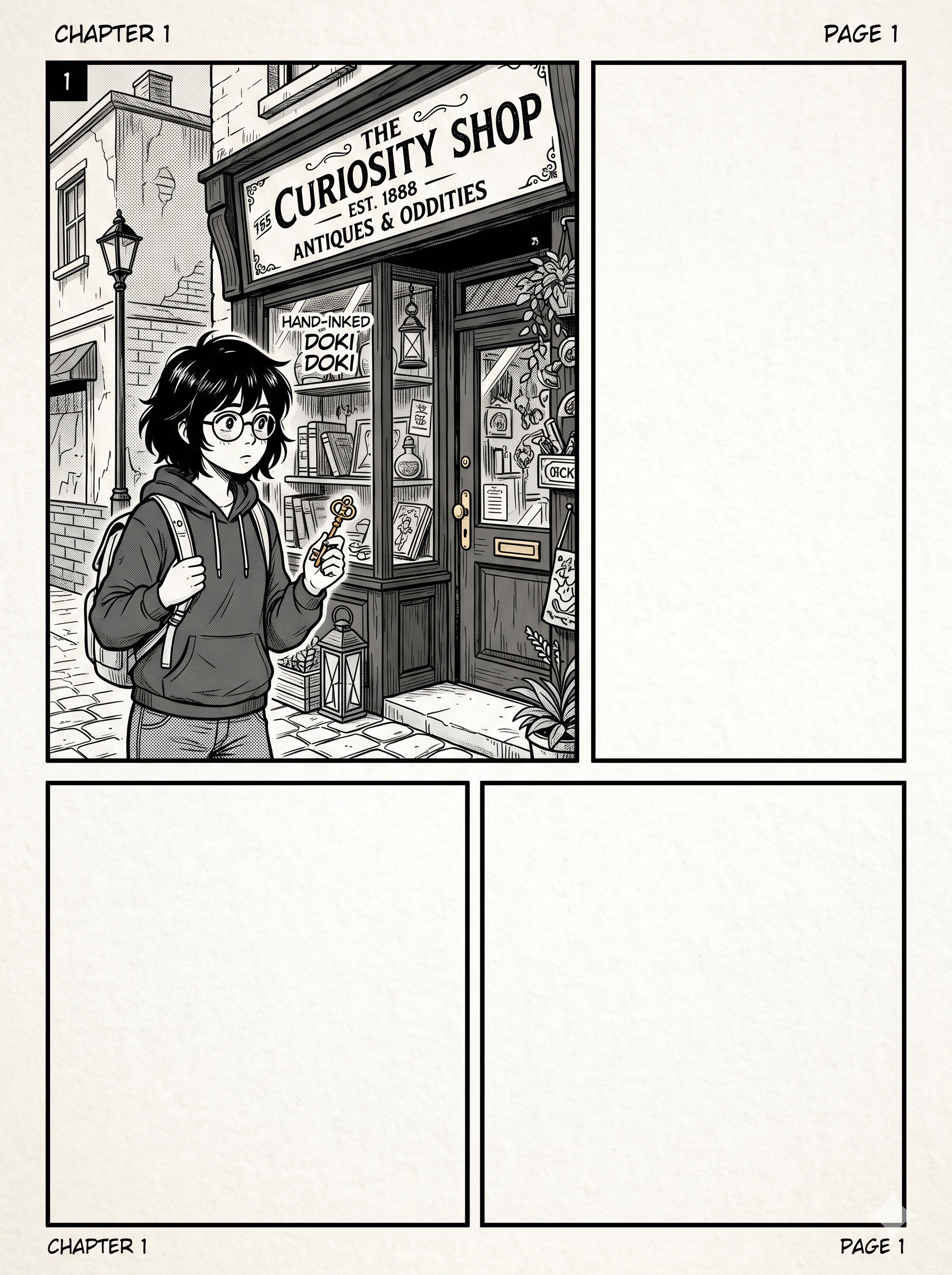}
        \caption{Turn 0}
    \end{subfigure}\hfill
    \begin{subfigure}[b]{0.235\textwidth}
        \centering
        \includegraphics[width=\linewidth,height=4cm,keepaspectratio]{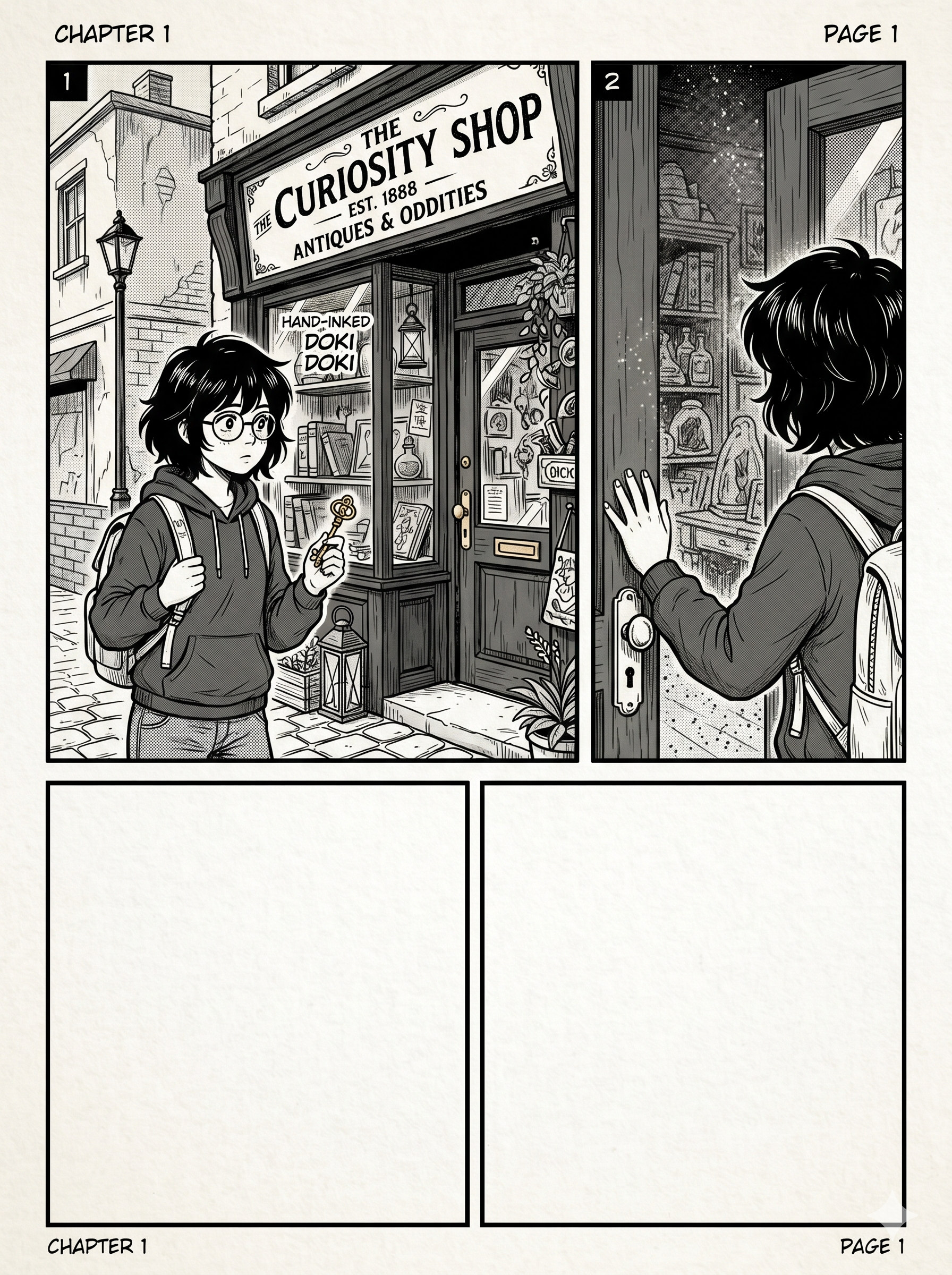}
        \caption{Turn 1}
    \end{subfigure}\hfill
    \begin{subfigure}[b]{0.235\textwidth}
        \centering
        \includegraphics[width=\linewidth,height=4cm,keepaspectratio]{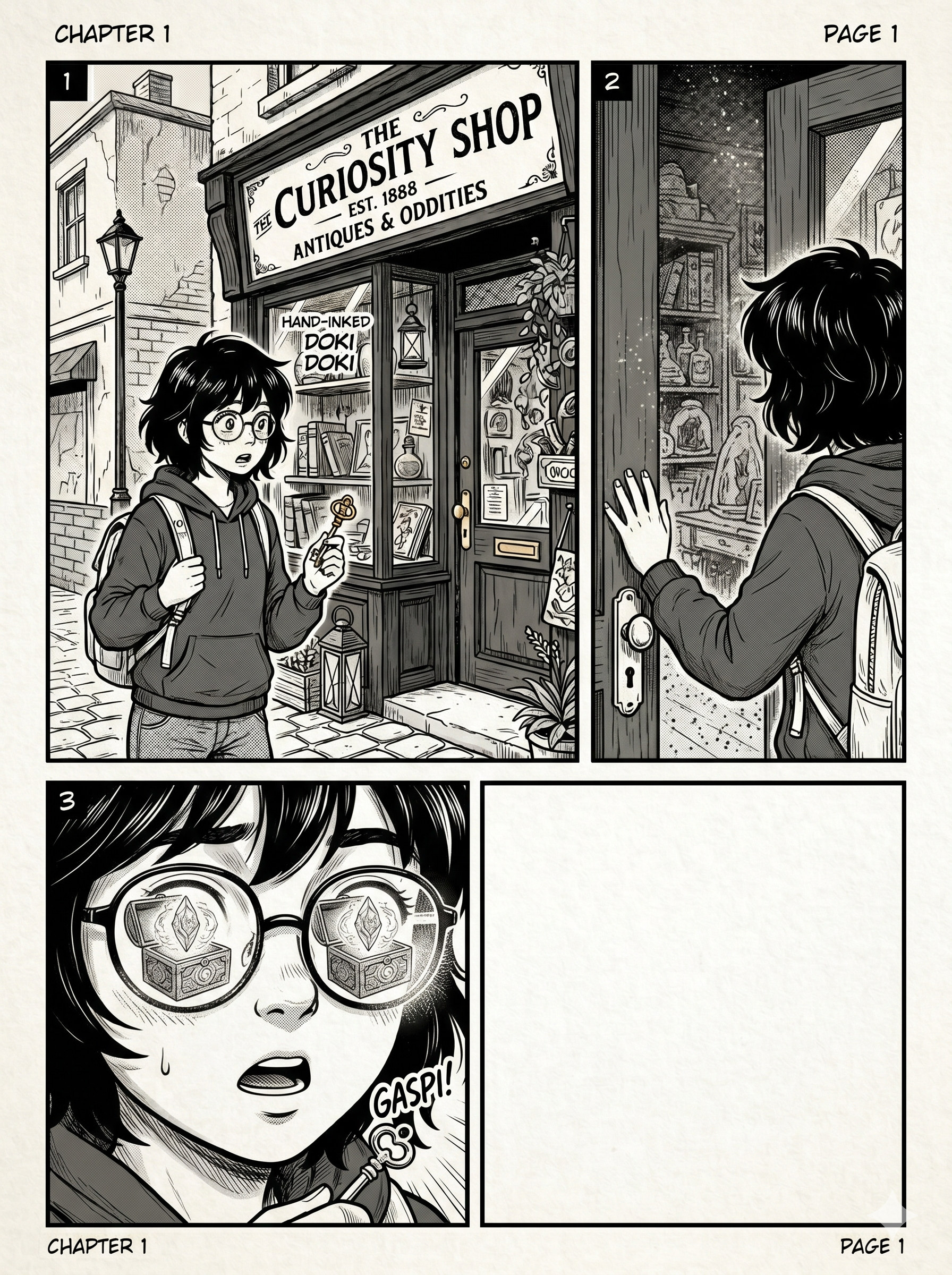}
        \caption{Turn 2}
    \end{subfigure}\hfill
    \begin{subfigure}[b]{0.235\textwidth}
        \centering
        \includegraphics[width=\linewidth,height=4cm,keepaspectratio]{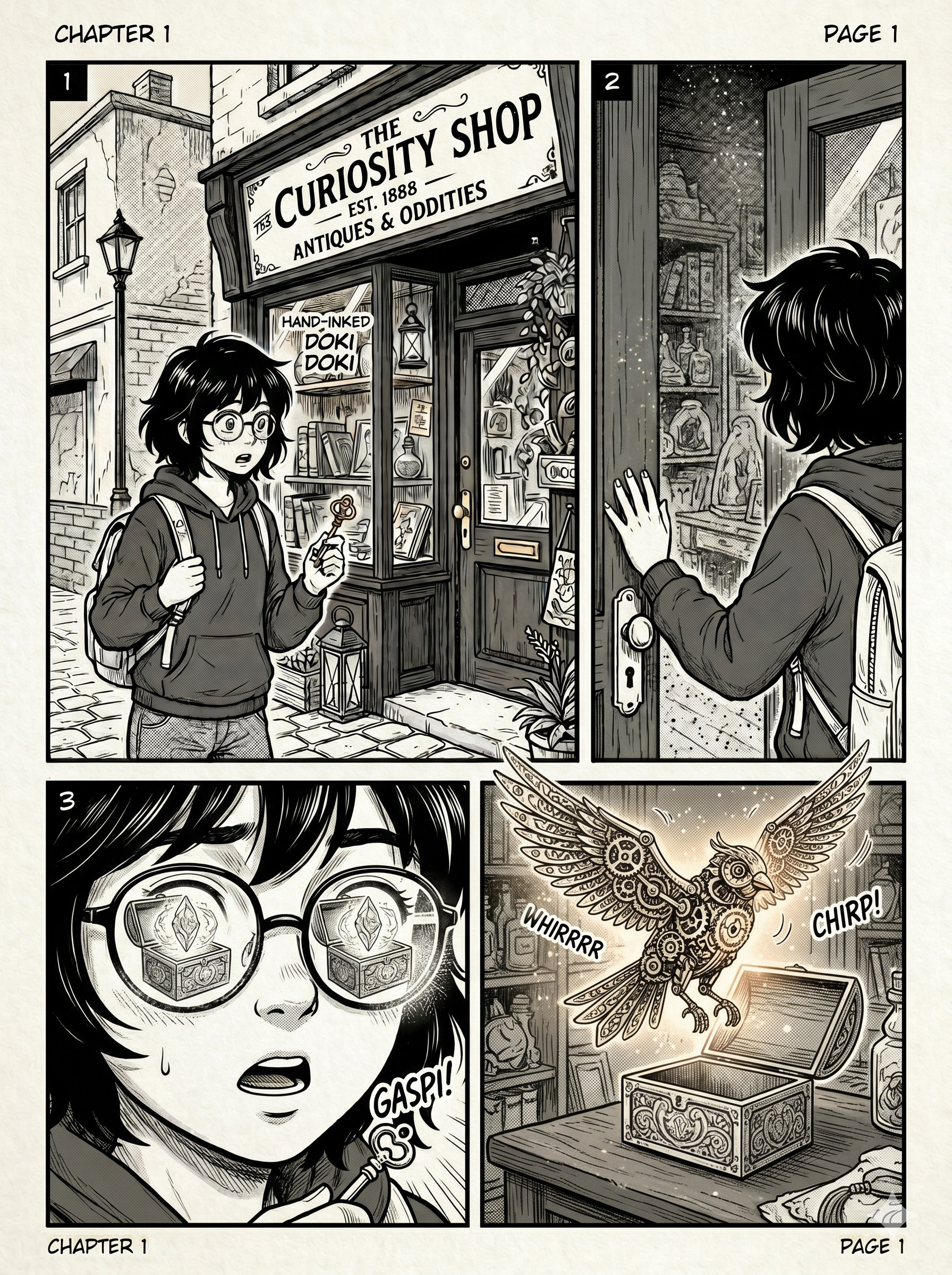}
        \caption{Turn 3}
    \end{subfigure}
    \caption{\textbf{Cumulative quality degradation in sequential multi-panel editing (Nano Banana).} Each turn fills only one empty panel of a four-panel manga page (T0: top-left; T1: top-right; T2: bottom-left; T3: bottom-right), yet because the model must re-encode and re-render the full image every round, JPEG-like compression artifacts accumulate and earlier panels drift even when no instruction targets them---for instance, at T2 the top-left girl's facial expression has silently shifted from her T0 rendering. We recommend zooming in to inspect the cumulative image-quality degradation across turns. This pattern reveals a structural cost of single-forward in-context generation that long context alone cannot resolve.}
    \label{fig:multi_turn_cumulative_degradation}
\end{figure}

\paragraph{Setup.}
We stress the pixel-level fidelity axis of multi-turn editing, deliberately decoupled from semantic identity. The input is a blank four-panel manga page outline, and the four-turn sequence fills the panels one at a time: (T0) top-left panel---a short-haired girl with round glasses outside an antique shop, holding a brass key, ink-line greyscale; (T1) top-right panel---the same girl pushing the door open from behind, dust motes in the air; (T2) bottom-left panel---a close-up of her surprised face, her glasses reflecting a glowing object; (T3) bottom-right panel---the reveal of what she sees, a floating gear bird with faint glow. Crucially, at each of T1--T3 the instruction targets a single empty panel; the previously-drawn panels should remain pixel-identical. The full prompt set is recorded in \texttt{img/stress\_test/multi\_turn/prompts.md}.

\paragraph{Observed Behavior.}
Even though each turn names exactly one panel as the target, Nano Banana visibly re-renders the full page on every round, with three failure modes accumulating across turns. (i) \emph{Compression-like artifacts}: subtle JPEG-style noise appears in turn~1 and intensifies turn by turn, most visible in flat shaded regions and near sharp edges. (ii) \emph{Text and fine-detail drift}: lettering and small symbols established in earlier panels become blurred or shift in content by later turns, despite no instruction asking the model to touch them. (iii) \emph{Non-edited regions are not strictly stable}: turn-by-turn micro-shifts in line weight, shading density, and panel alignment accumulate---most visibly, the top-left girl's facial expression at T2 no longer matches her T0 rendering, even though T2 only targeted the bottom-left panel.

\paragraph{Insight.}
This case isolates a structural cost of L3 in-context generation. Because a four-panel layout cannot be edited as four independently rendered crops, every turn must run the entire image through the encoder, the backbone, and the decoder. Each round therefore incurs a VAE round-trip with non-zero quantization loss, and these losses accumulate monotonically across turns. Long context solves \emph{capacity} (the model knows what should not change) but does not solve \emph{pixel-level fidelity} (the encode/decode cycle still degrades it). This is a structural property of the architecture, not an engineering bug: current single-forward in-context systems cannot guarantee strict freezing of non-edited regions. Read through the Markovian lens, this is the \emph{pixel-level drift} signature of $f(I_{t-1}, p_t)$ chaining: with no agentic decision to re-anchor on $I_0$ at any point in the chain, every panel survives only as the lossy reconstruction of the previous round's lossy reconstruction, and quality drifts monotonically downward with depth. The implication is that production-grade multi-turn editing tools likely need an explicit non-edit mask plus a copy-paste fallback for unmodified regions---or, more generally, a controller that selects \emph{which} past turn to anchor each new edit on---rather than relying on the model's in-context capability alone to preserve untouched content.

\subsubsection{Case Study II: Restore-to-Original --- Long-Range Recall under Cascading Drift}

\begin{figure}[!htbp]
    \centering
    \begin{subfigure}[b]{0.235\textwidth}
        \centering
        \includegraphics[width=\linewidth,height=6cm,keepaspectratio]{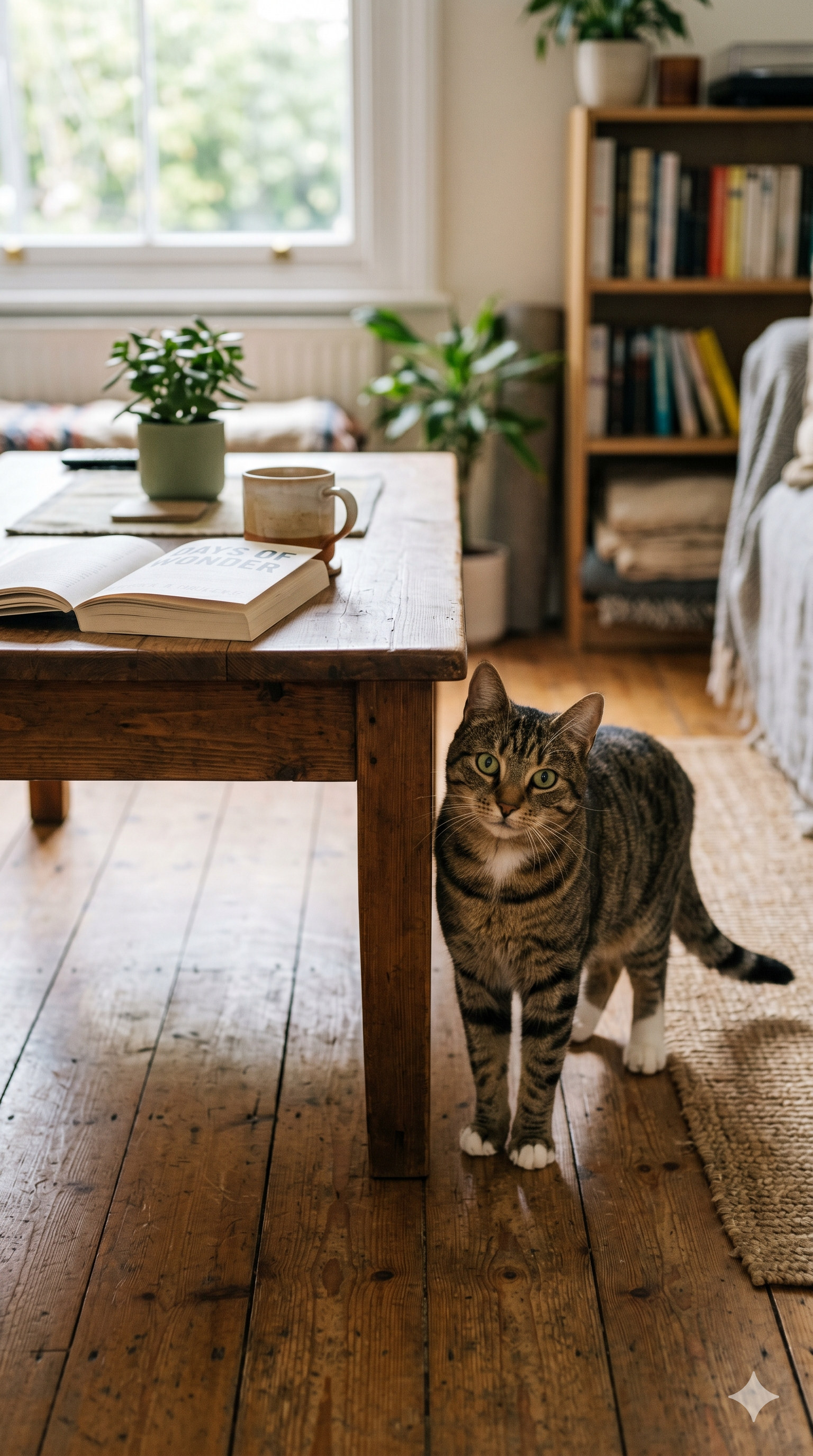}
        \caption{Turn 0}
    \end{subfigure}\hfill
    \begin{subfigure}[b]{0.235\textwidth}
        \centering
        \includegraphics[width=\linewidth,height=6cm,keepaspectratio]{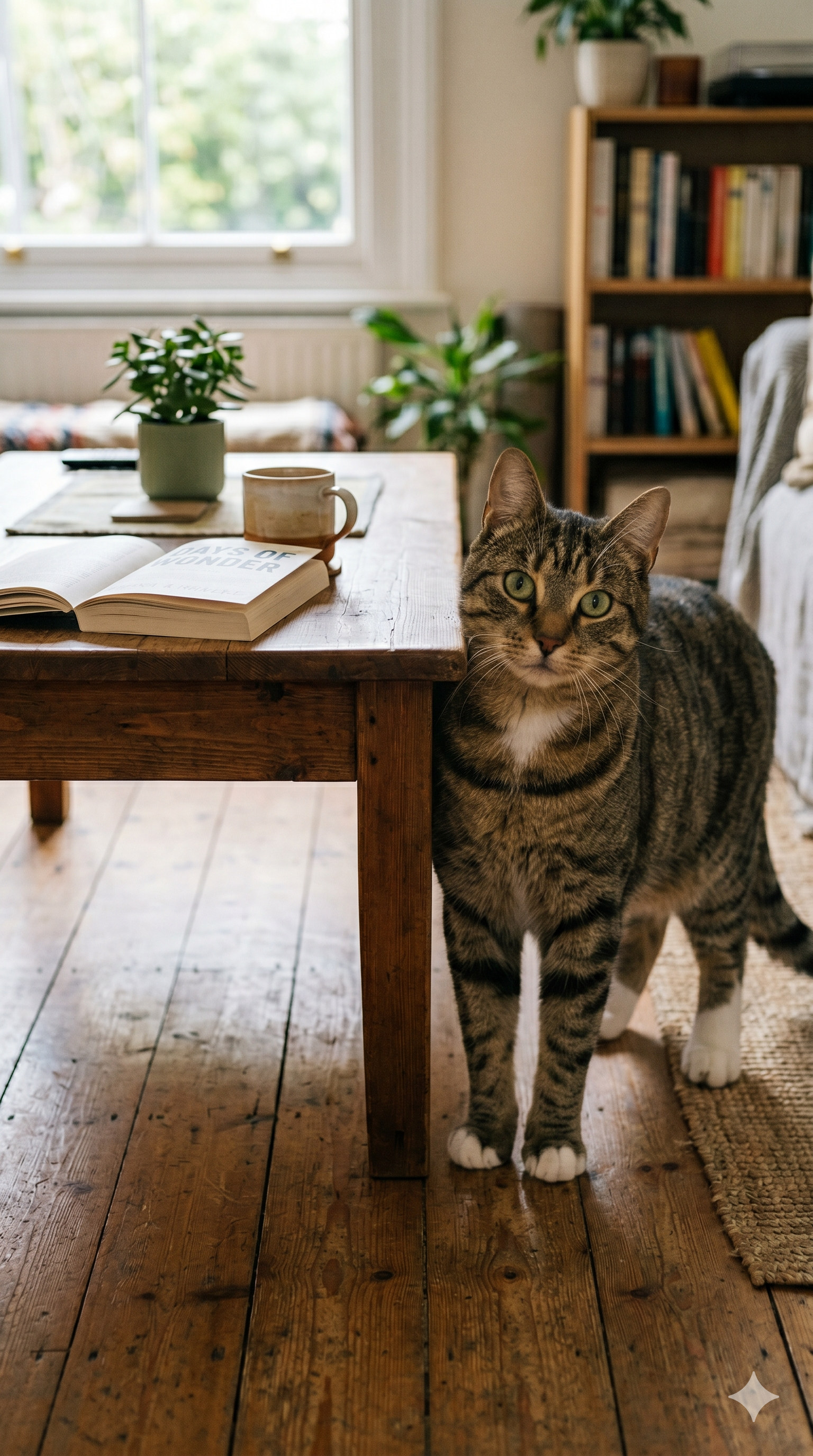}
        \caption{Turn 1}
    \end{subfigure}\hfill
    \begin{subfigure}[b]{0.235\textwidth}
        \centering
        \includegraphics[width=\linewidth,height=6cm,keepaspectratio]{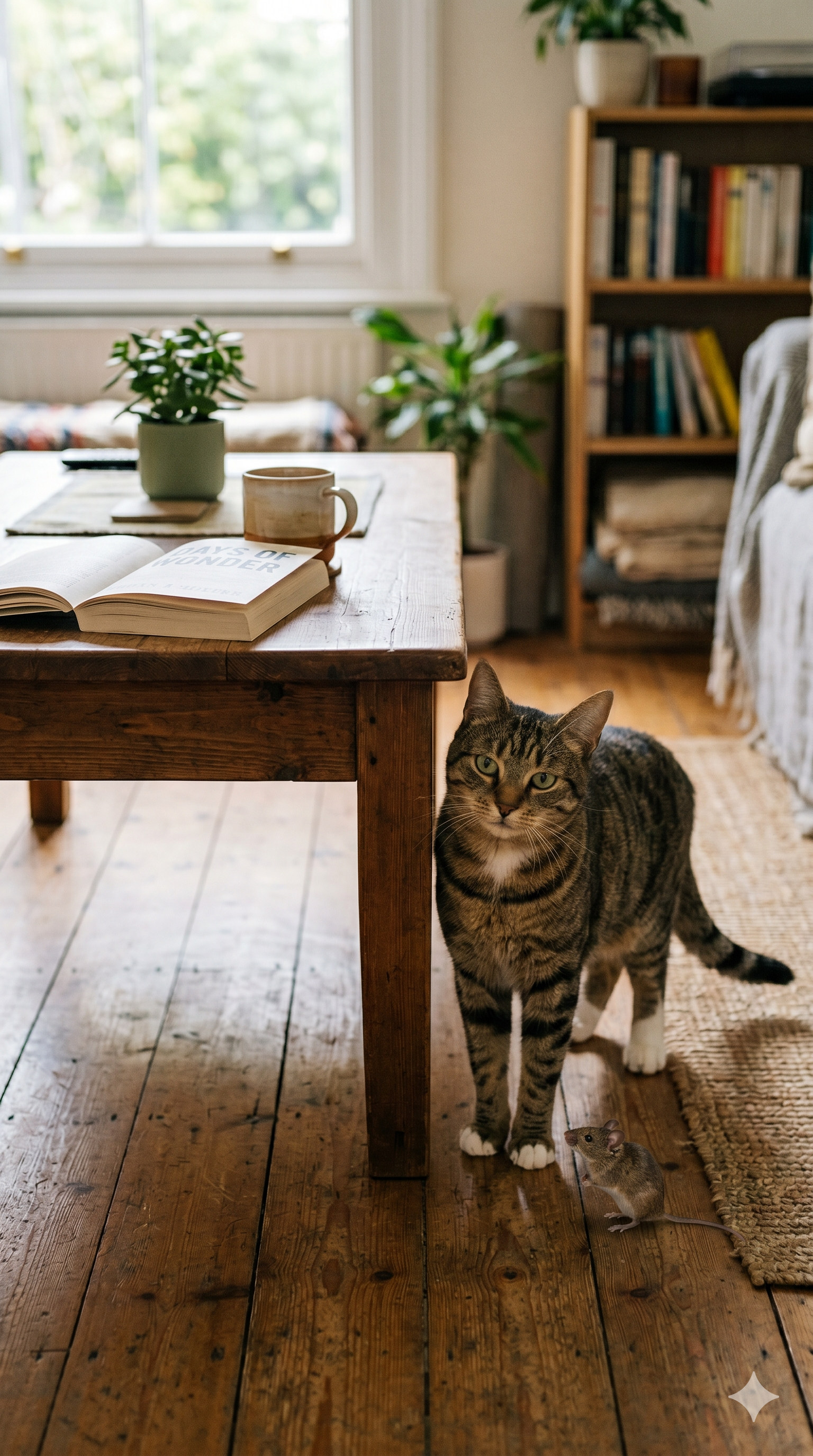}
        \caption{Turn 2}
    \end{subfigure}\hfill
    \begin{subfigure}[b]{0.235\textwidth}
        \centering
        \includegraphics[width=\linewidth,height=6cm,keepaspectratio]{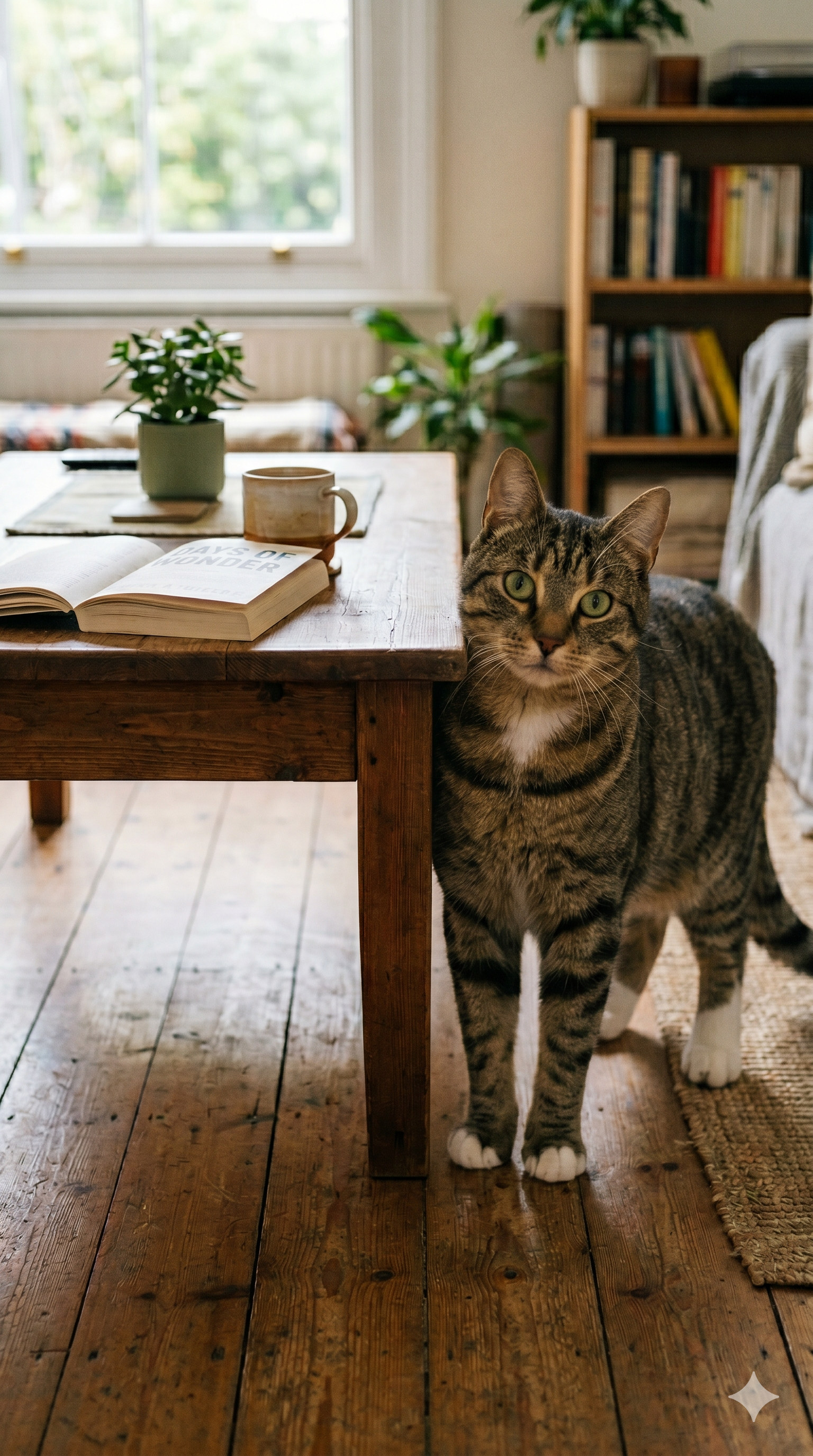}
        \caption{Turn 3}
    \end{subfigure}
    \caption{\textbf{Sequential edits applied to a photograph of a cat (Nano Banana).} Each panel shows the model's output after the corresponding turn. Prompts---Turn 0: ``A cat standing next to a table.''; Turn 1: ``Make the cat 2x larger.''; Turn 2: ``Add a mouse beside the cat.''; Turn 3: ``Restore the cat to its original size.''}
    \label{fig:multi_turn_cat_recall}
\end{figure}

\paragraph{Setup.}
We probe semantic-level multi-turn drift with a four-turn sequence whose final turn explicitly triggers long-range recall. Starting from a photograph of a tabby cat standing next to a wooden coffee table ($I_0$), we issue: (T1) ``Make the cat 2x larger''; (T2) ``Add a mouse beside the cat''; (T3) ``Restore the cat to its original size.'' Each turn is conditioned on the previous turn's output. The decisive instruction is T3: ``original size'' refers unambiguously to $I_0$, but in a Markovian regime $f(I_2, p_3)$ the model has access only to $I_2$, where the cat has already drifted from both $I_0$ and $I_1$. Successful execution therefore requires the model to either retain the full image history or carry an explicit memory of $I_0$'s appearance.

\paragraph{Observed Behavior.}
Turns~0--1 succeed: the cat in $I_1$ is visibly larger than in $I_0$ while preserving identity, gaze, and pose, indicating that the model handles a single explicit edit cleanly. The first failure surfaces at T2. Although ``add a mouse'' is the only requested change, the cat is silently re-rendered---it shrinks back to roughly its $I_0$ size and its gaze tightens into a different expression---even though no instruction touched the cat at this turn. The second and more decisive failure surfaces at T3. The prompt asks the model to revert the cat to its $I_0$ size; instead, the cat is enlarged further to a size comparable to or larger than $I_1$, and the mouse introduced at T2 silently disappears. T3 therefore exhibits two simultaneous failures: a long-range recall failure (the model cannot reach back to $I_0$ for the reference size) and an object-persistence failure (the mouse is dropped despite no instruction to remove it).

\paragraph{Insight.}
T3's instruction ``restore to original'' makes long-range recall a hard requirement: it can only be satisfied by consulting $I_0$ via image history or an explicit textual memory of T0's state. The observed behavior---enlarging the cat in the wrong direction and silently dropping the mouse---is consistent with $f(I_2, p_3)$, conditioning on a single drifted frame and reinterpreting ``original'' against whatever appearance happens to be in $I_2$. T2's silent drift is the upstream cause: each unrequested change at turn $t$ shifts the reference frame against which turn $t{+}1$ must reason, so drift compounds and recall degrades together. Read alongside Case~I, this completes the picture of Markovian multi-turn editing: Case~I exposes \emph{representational} drift inside the model (pixel-level, model-internal), while Case~II exposes \emph{semantic} drift visible to the user (identity, size, object persistence). Both are surface symptoms of the same shortcut, and motivate the agentic, history-aware architectural directions discussed in \Cref{sec:frontier}.

\endgroup

\subsection{Dimension V: Human-Centric Heredity \& Aesthetic Editing}

\noindent\textit{Primary Level(s) Tested: L2 (Conditional Generation) / L4 (Agentic Generation) / L5 (World-Modeling Generation) --- Can the model reason about human faces across heredity, cosmetic modification, and stylistic editing, translating subjective, biological, or cultural constraints into coherent visual outputs while preserving identity?}

\paragraph{Core Ability: Reasoning about human faces---their genetic plausibility, surgically modifiable structure, and stylistic presentation---while decomposing vague aesthetic prompts into coordinated, identity-preserving edits.}

\subsubsection{Case Study I: Predicting Children's Appearance}

\begin{figure}[!htbp]
    \centering
    \begin{subfigure}[b]{0.2275\textwidth}
        \centering
        \includegraphics[width=\linewidth,height=8cm,keepaspectratio]{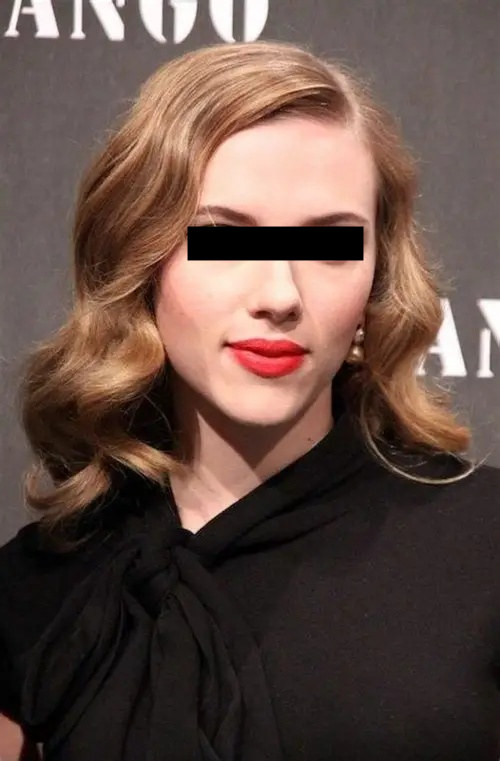}
        \caption{Input: female}
    \end{subfigure}\hfill
    \begin{subfigure}[b]{0.23\textwidth}
        \centering
        \includegraphics[width=\linewidth,height=8cm,keepaspectratio]{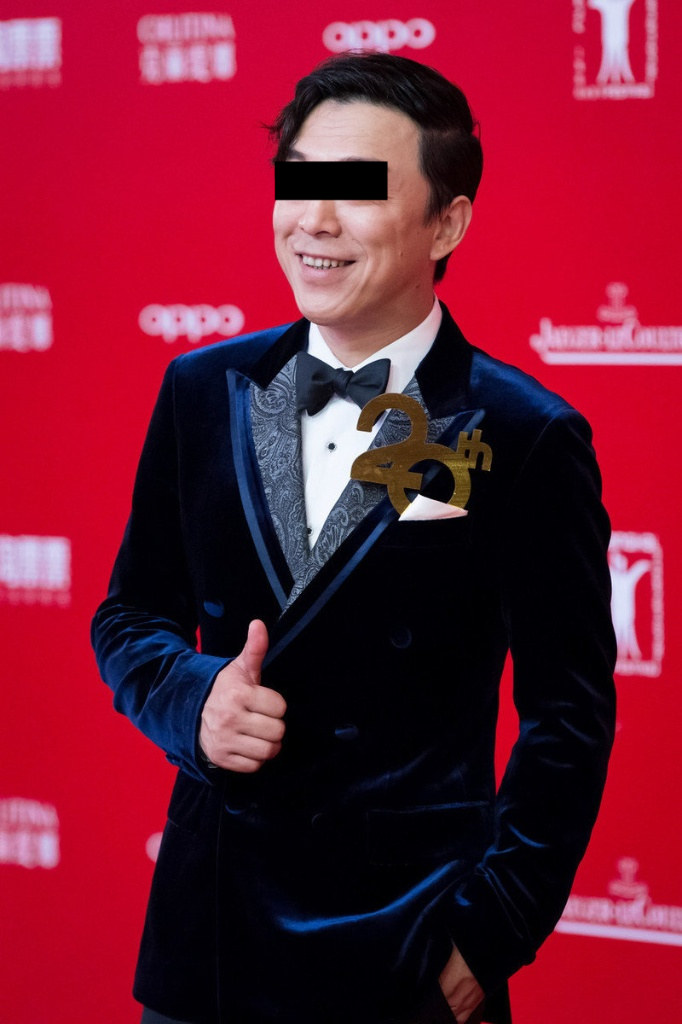}
        \caption{Input: male}
    \end{subfigure}\hfill
    \begin{subfigure}[b]{0.23\textwidth}
        \centering
        \includegraphics[width=\linewidth,height=8cm,keepaspectratio]{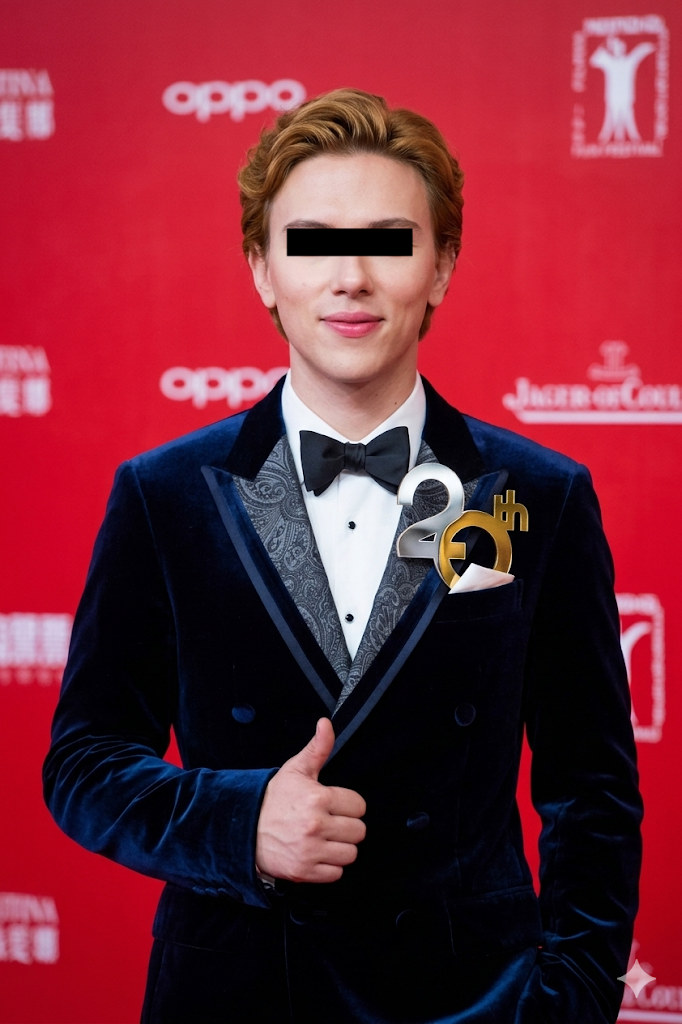}
        \caption{Attempt 1}
    \end{subfigure}\hfill
    \begin{subfigure}[b]{0.225\textwidth}
        \centering
        \includegraphics[width=\linewidth,height=8cm,keepaspectratio]{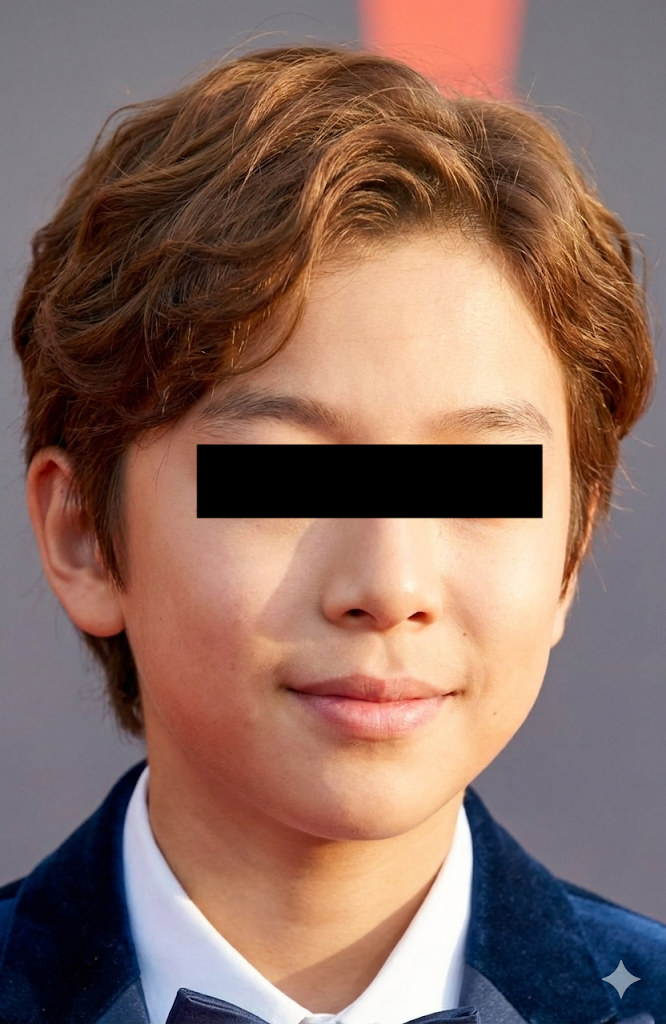}
        \caption{Attempt 2}
    \end{subfigure}
    \caption{\textbf{Predicting Children's Appearance.} Attempt~1 (minimal prompt) yields a compositional shortcut; Attempt~2 (with an explicit blending cue) yields a genuine trait blend. Generated by Nano Banana.}
    \label{fig:children}
\end{figure}

\paragraph{Setup and Objectives.}
We test whether the model can synthesize a plausible child's face by blending the traits of two parents. The input is two portraits of unrelated adults---a white woman and an East Asian man---and we probe prompt sensitivity by issuing two variants. Attempt~1 uses the minimal prompt: ``Please predict what the child of these two people would look like and generate an image of the future child.'' Attempt~2 adds an explicit blending cue: ``\dots combining the facial features of both.''

\paragraph{Observed Behavior.}
The two prompts produce qualitatively different outputs (\Cref{fig:children}). Under Attempt~1, the model takes a \emph{compositional shortcut}: the generated face is essentially a copy of the mother's appearance, while the clothing, gesture, and background are lifted from the father's photograph. The image quality is high, but this is a superficial recombination of source regions rather than true hereditary blending. Under Attempt~2, the output exhibits a genuine blend---dark hair carrying a soft wave from the mother, a larger-than-baseline East Asian eye shape, a relatively high nose bridge, a sharper chin, and skin tone interpolating between the two parents---consistent with elementary genetic reasoning where dominant traits suppress recessive ones while morphology interpolates.

\paragraph{Insight.}
The model's prior over ``child given two parents'' \emph{does} contain the structure needed for real trait blending, and the Attempt~2 result is readily accepted as a possible child of this couple. However, whether this capability is invoked depends sensitively on prompt wording: without an explicit blending cue, the model falls back to a shallower strategy that mixes source regions rather than source features. The underlying skill exists, but surfacing it reliably still requires the user to spell out the intended reasoning---a gap in deep instruction understanding.

\subsubsection{Case Study II: Plastic Surgery Simulation}

\begin{figure}[!htbp]
    \centering
    \begin{subfigure}[b]{0.24\textwidth}
        \centering
        \includegraphics[width=\linewidth,height=5cm,keepaspectratio]{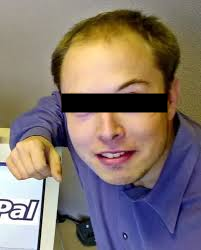}
        \caption{Ex.~1 Input}
    \end{subfigure}\hfill
    \begin{subfigure}[b]{0.24\textwidth}
        \centering
        \includegraphics[width=\linewidth,height=5cm,keepaspectratio]{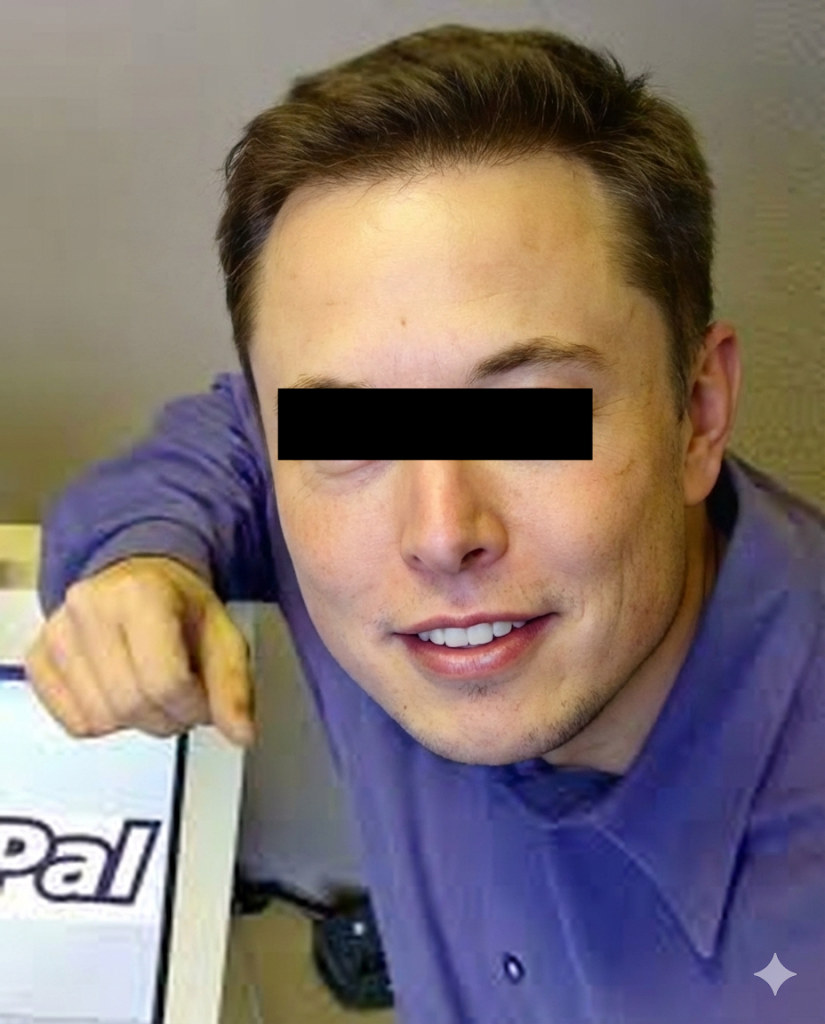}
        \caption{Ex.~1 Output}
    \end{subfigure}\hfill
    \begin{subfigure}[b]{0.24\textwidth}
        \centering
        \includegraphics[width=\linewidth,height=5cm,keepaspectratio]{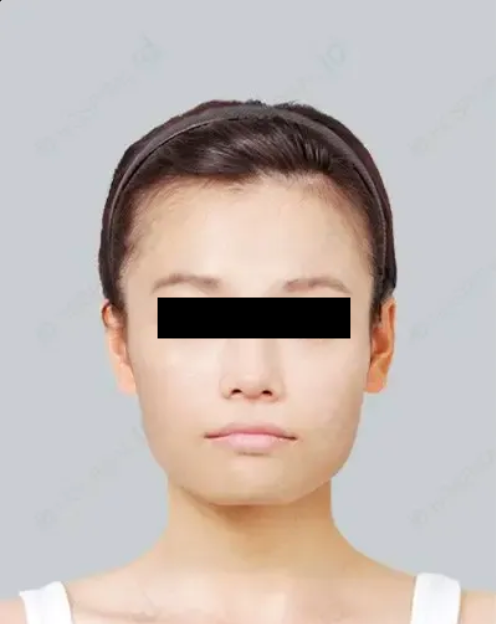}
        \caption{Ex.~2 Input}
    \end{subfigure}\hfill
    \begin{subfigure}[b]{0.24\textwidth}
        \centering
        \includegraphics[width=\linewidth,height=5cm,keepaspectratio]{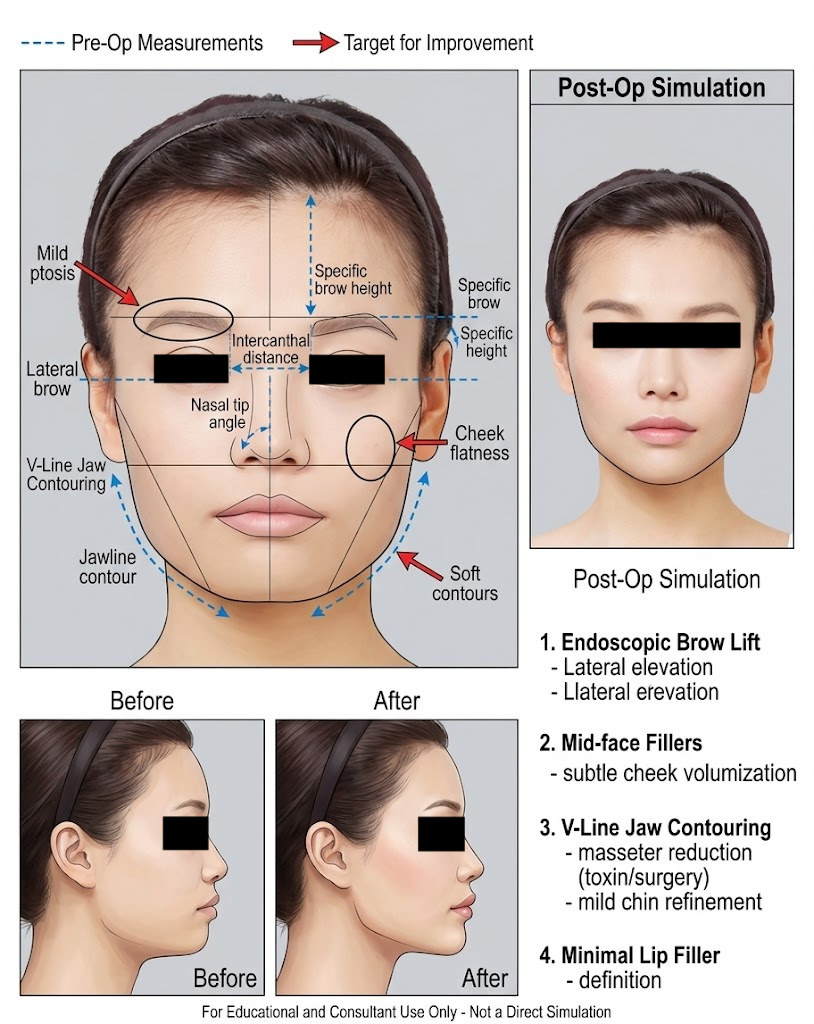}
        \caption{Ex.~2 Output}
    \end{subfigure}
    \caption{\textbf{Plastic Surgery Simulation.} Example~1 (vague ``make him handsome'' prompt) yields a coordinated multi-axis edit of the input portrait; Example~2 (``generate an analysis chart'' prompt) yields a full surgical-consultation document with annotated measurements, itemized procedures, and a clinical-style disclaimer. Generated by Nano Banana.}
    \label{fig:plastic_surgery}
\end{figure}

\paragraph{Setup and Objectives.}
The second case probes the model's ability to translate a vague aesthetic intent into concrete surgical output. We probe prompt sensitivity with two variants that differ in both the subject and the requested output modality. Example~1 (male subject) uses the prompt: ``Please analyze this person's facial structure and perform a plastic surgery that makes him more handsome''---an underspecified beautification request with no procedural detail. Example~2 (female subject) uses the prompt: ``Please analyze this woman's facial structure and generate a plastic surgery analysis chart''---a specification that explicitly requests an analysis document rather than a modified portrait. Together the two variants test whether the model can (i) decompose an underspecified goal into an implicit edit plan, and (ii) adapt its output modality to the prompt's intended format.

\paragraph{Observed Behavior.}
The two prompts trigger qualitatively different outputs (\Cref{fig:plastic_surgery}). In Example~1, the output is a single modified portrait in which the subject is changed along multiple coherent axes at once---advanced hairline, slimmer face with a more defined jawline, enlarged eye region, and smoothed skin---with identity preserved. In Example~2, the output is \emph{not} a single portrait but a complete surgical-consultation sheet: annotated measurement arrows on the main headshot labeling anatomical landmarks (mid philtrum, brow height, jawline contour, cheek profile, soft contours), side-view before/after thumbnails, and a numbered procedure plan listing specific terms used in clinical practice (Endoscopic Brow Lift, Mid-face Fillers, V-Line Jaw Contouring, Minimal Lip Filler). The chart even appends a disclaimer (``For Educational and Consultation Use Only---Not a Direct Simulation''), suggesting that the model has learned the surrounding conventions of such documents, not just the main visual.

\paragraph{Insight.}
The two examples together reveal two layers of capability. First, under a vague beautification prompt (Example~1), the model implicitly decomposes ``handsome'' into a coordinated set of edit dimensions (hair, contour, eyes, skin) and executes them harmoniously. Second, and more strikingly, the model \emph{adapts its output modality} to the prompt (Example~2): asked for an ``analysis chart,'' it produces a document that follows the layout, terminology, and even educational disclaimers of real clinical consultation sheets, rather than returning a modified portrait. This indicates a non-trivial depth of world knowledge in the plastic-surgery domain, well beyond the level required for surface-level beautification.

\subsubsection{Case Study III: Hairstyle Generation}

\begin{figure}[!htbp]
    \centering
    \begin{subfigure}[b]{0.48\textwidth}
        \centering
        \includegraphics[width=\linewidth,height=6cm,keepaspectratio]{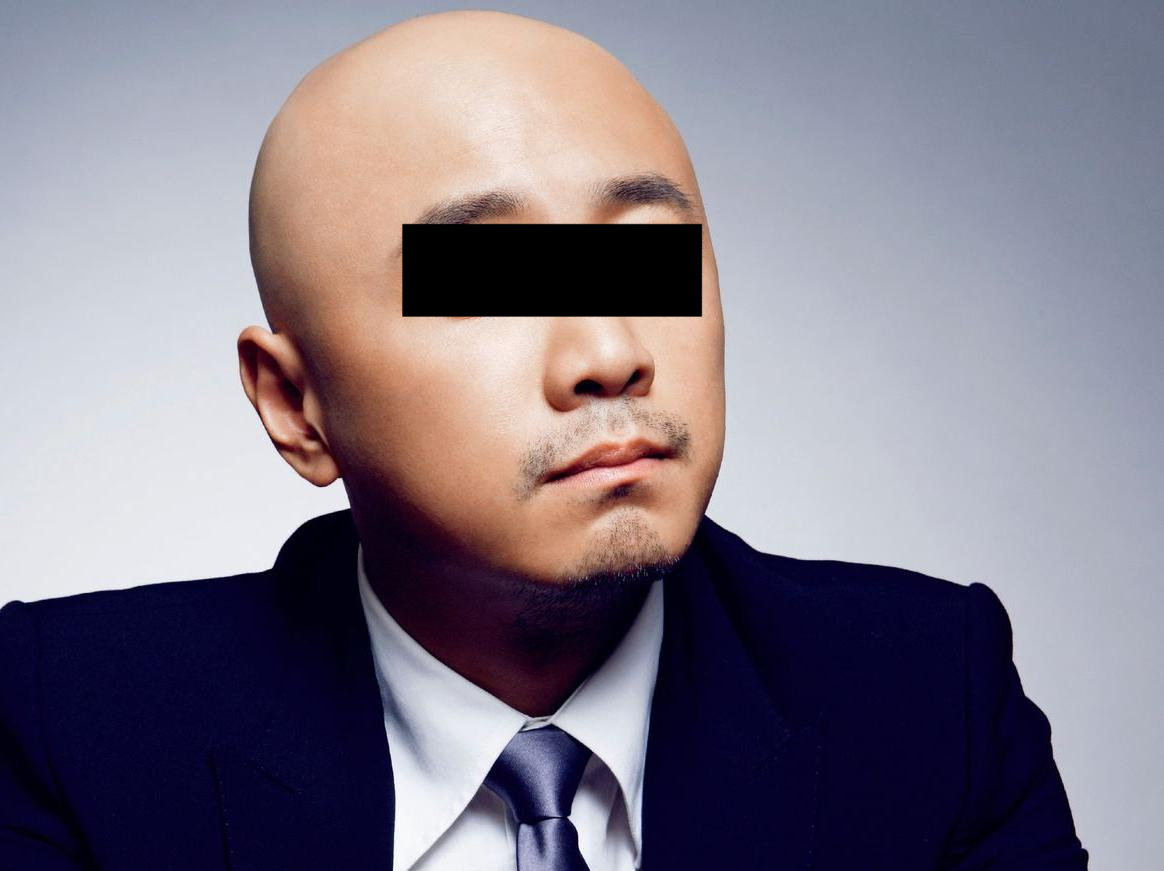}
        \caption{Input}
    \end{subfigure}
    \hfill
    \begin{subfigure}[b]{0.48\textwidth}
        \centering
        \includegraphics[width=\linewidth,height=6cm,keepaspectratio]{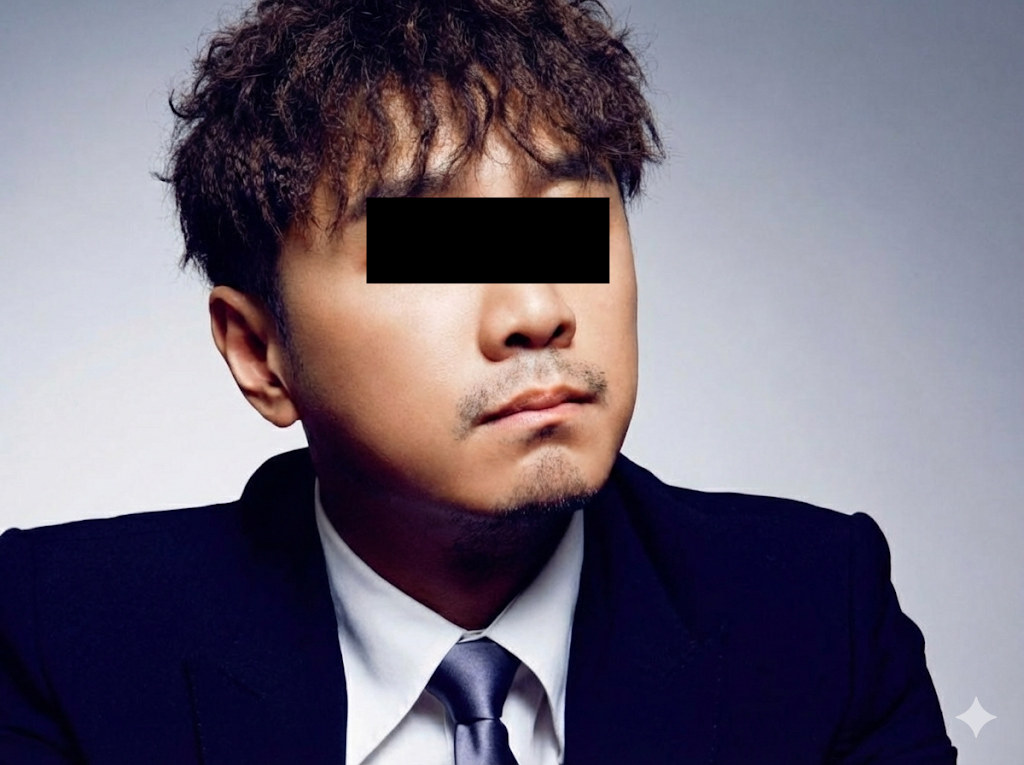}
        \caption{Model Output}
    \end{subfigure}
    \caption{\textbf{Hairstyle Generation.} The model correctly resolves a culturally specific Chinese styling term (``foil perm'') and composites the hairstyle onto the subject while preserving identity. Generated by Nano Banana.}
    \label{fig:hair_case}
\end{figure}

\paragraph{Setup and Objectives.}
We evaluate the model on a concrete, culturally grounded styling request. The input is a photograph of a bald male subject, and the prompt asks: ``Please generate an image of this person with a foil perm hairstyle.'' The foil perm is a styling term popular in contemporary East Asian contexts, characterized by short, tightly coiled curls produced with foil wrapping during setting. The case simultaneously probes whether the model carries sufficient \emph{culturally specific world knowledge} to recognize the target style from its local name, and whether it can composite that style onto an existing face without disturbing identity.

\paragraph{Observed Behavior.}
As shown in \Cref{fig:hair_case}, the output renders a hairstyle visually faithful to the foil perm as a category: short, tight, lofted curls rather than a generic wavy default. The hair integrates naturally with the original head shape, with plausible lighting and proportions and no visible ``pasted-on'' artifact at the hair--face boundary. Identity is well preserved: facial landmarks, expression, and clothing remain largely unchanged.

\paragraph{Insight.}
Two properties make this case noteworthy. First, the model correctly resolved a culturally specific term supplied in a non-English prompt---an outcome that relies on substantial region-specific styling knowledge absorbed during training. Second, the result demonstrates clean \emph{localized editing}: the hairstyle is changed while the rest of the portrait remains stable, which is the operational property that makes virtual try-on useful. Together these point to a practical sweet spot for current models: concrete, visually specified styling requests anchored by strong world knowledge and reliable identity preservation.

\subsection{Dimension VI: Low-level Vision Tasks}

\noindent\textit{Primary Level Tested: L1 (Atomic Generation) / L2 (Conditional Generation) --- Can the model perform pixel-level restoration and geometric estimation tasks that require precise signal recovery rather than semantic plausibility?}
\subsubsection{Case Study I: Out-of-Distribution Depth Estimation}
\paragraph{Setup and Objectives.} 
In this case study, we test the model on a real-world image that it has never seen during training (an out-of-distribution test). The image contains a large number of objects in a complex setting. We have two main goals for this setup: first, to see if the model can successfully identify the different objects and their outlines; second, to find out if it can correctly estimate the depth of these objects, meaning how far away they are from the camera.

\paragraph{Success in Object Recognition.} 
When given this entirely new image, the model shows that it can follow instructions very well. It successfully finds the outlines of most objects in the foreground. This proves that the model has a strong ability to recognize and separate items from each other, even when placed in real-world situations that it is not used to. It clearly understands what the objects are in the picture.

\begin{figure}[!htbp]
    \centering
    \begin{subfigure}[b]{0.48\textwidth}
        \centering
        \includegraphics[width=\linewidth,height=7.2cm,keepaspectratio]{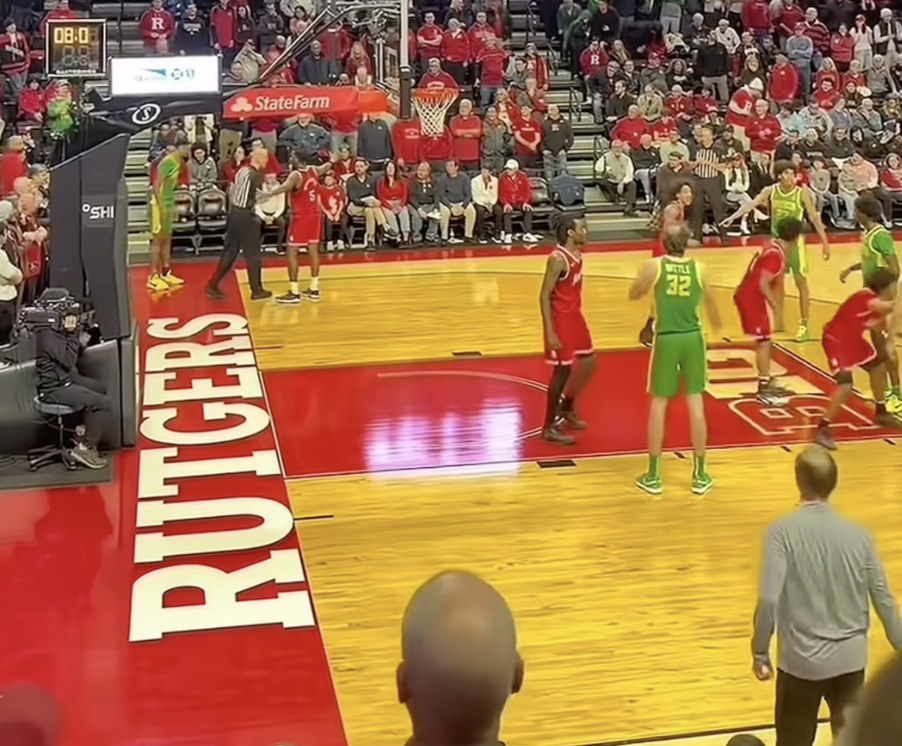}
        \caption{Input: Real-world image}
    \end{subfigure}
    \hfill
    \begin{subfigure}[b]{0.48\textwidth}
        \centering
        \includegraphics[width=\linewidth,height=7.2cm,keepaspectratio]{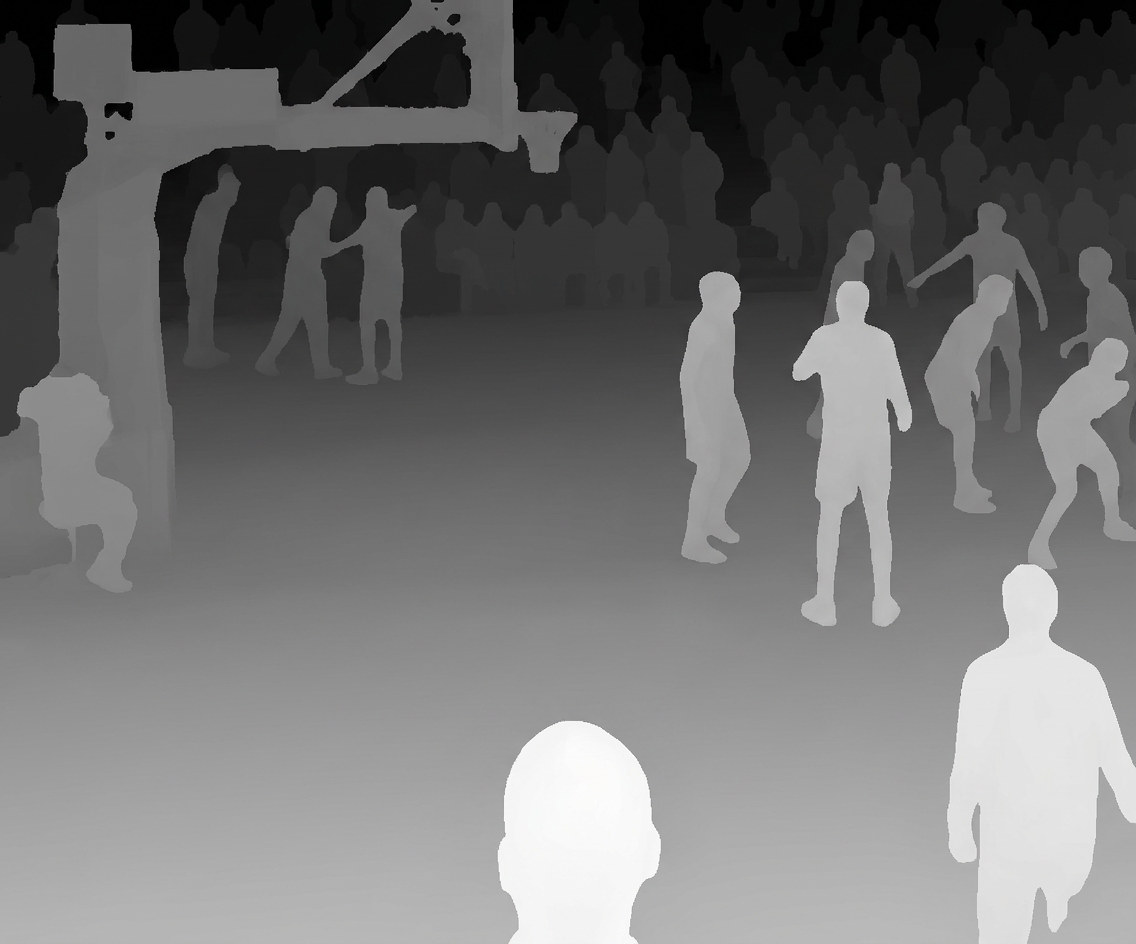}
        \caption{Model output: grey depth estimation}
    \end{subfigure}
    \caption{\textbf{Depth Estimation.} While the model performs well in object recognition, it fails to provide accurate depth estimation. Many regions that should share the same depth are not generated with consistent shades. Generated by Nano Banana.}
    \label{fig:depth_estimation}
\end{figure}

\paragraph{Failure in Depth Estimation.} 
However, the model runs into major problems when it tries to estimate depth. As we can see in \Cref{fig:depth_estimation}, the model colors many different objects with the exact same shade. Because color represents depth here, the model is incorrectly guessing that all these objects are at the same distance from the camera, which is not true. On top of this error, the model completely misses the outlines of several people standing in the background. These mistakes reveal a clear weakness: while the model is very good at recognizing objects and following text commands, it still has a poor understanding of 3D space and the true physical relationships between objects in the real world.

\subsubsection{Case Study II: Low-Level Visual Restoration Across Heterogeneous Degradations}

\paragraph{Setup and Objectives.}
To probe the model's capability in low-level vision, we evaluate it on a small but diverse set of restoration tasks covering both synthetic and real-image degradations. Specifically, we randomly select one degraded image from each of Set5~\citep{bevilacqua2012set5}, LOL~\citep{wei2018lol}, BSD68~\citep{martin2001bsd68}, Rain100H~\citep{yang2017rain100h}, and GoPro~\citep{nah2017gopro}, corresponding to super-resolution, low-light enhancement, denoising, deraining, and deblurring, respectively. In addition, we include one real-world image to examine out-of-distribution (OOD) generalization. Unlike higher-level multimodal tasks that primarily test semantic understanding or reasoning, these examples test whether the model can restore visual quality while preserving scene structure under a range of heterogeneous degradations.

\paragraph{Overall Behavior.}
Across these examples, the model shows a surprisingly unified restoration capability. It can follow task instructions and produce outputs that are substantially cleaner, sharper, and more visually pleasing than the degraded inputs. Super-resolution recovers crisper edges, low-light enhancement reveals scene content hidden in darkness, denoising removes strong corruption, deraining suppresses dominant streak artifacts, and deblurring reconstructs a much clearer scene from severely blurred input. At a coarse perceptual level, the outputs are often strong enough to be immediately useful, suggesting that the model has learned a robust prior over what a ``restored'' image should look like.

\paragraph{Task-by-Task Observations.}
In the super-resolution example, the model produces a sharper butterfly wing with clearer boundary structure and richer internal texture than the low-resolution input. In the low-light example, it dramatically lifts brightness and reveals room content that is barely visible in the original dark image. In the denoising example, heavy noise is removed while the global layout of the racing scene is preserved. In the deraining example, the dominant rain streaks are largely eliminated and the bear scene becomes much easier to read visually. In the deblurring example, the model transforms an extremely blurred street image into a recognizable urban scene with plausible object boundaries and textures. Taken together, these results indicate that the model can handle multiple restoration tasks within a single general-purpose image-editing framework, rather than relying on a separate specialized pipeline for each degradation type.

\paragraph{A Key Limitation: Restoration by Detail Hallucination.}
However, a closer comparison with the ground truth reveals that the model's outputs are not simply faithful reconstructions of the underlying clean images. The main issue is not over-smoothing; rather, the model frequently \emph{adds}, \emph{alters}, or \emph{reimagines} details, effectively producing a more detailed version of the scene rather than strictly recovering the original signal. This tendency is especially evident in the low-light enhancement case, where the model does not merely brighten the dark image but appears to rewrite textures, local structures, and illumination patterns according to a learned natural-image prior. Similar behavior can also be seen in super-resolution and deblurring: the outputs are often sharper and richer, but some of the recovered details are better interpreted as plausible inventions than as verifiable restorations of the true image content.

\paragraph{What This Suggests About the Model.}
This pattern suggests that the model is not performing low-level restoration as a classical inverse problem solver. Instead, it behaves more like a generative image corrector guided by degraded input and task instructions. In other words, it uses the corrupted image as a constraint, but much of the final quality appears to come from prior-driven image synthesis rather than strict signal recovery. This is a meaningful strength from the perspective of perceptual quality: many outputs look subjectively good and align well with user expectations of what the restored scene should resemble. But it is also a nontrivial limitation, because visually appealing restoration does not necessarily imply fidelity to the original scene details.

\paragraph{Insight.}
These examples suggest that the model already possesses a fairly general low-level editing capability across heterogeneous degradations, including enhancement, suppression of structured corruption, and perceptual detail recovery. Yet the mechanism behind this capability appears closer to \emph{prior-guided image rewriting} than to exact physical restoration. That is, the model is often very good at making a degraded image look clean, sharp, and natural, but less clearly optimized for reconstructing the precise underlying image that generated the degraded observation. This distinction is especially important in low-level vision, where perceptual success and reconstruction fidelity are not always the same thing.

\begin{figure}[t]
    \centering
    \begin{subfigure}[t]{0.275\textwidth}
        \centering
        \includegraphics[width=\linewidth]{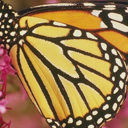}
        \caption{Input}
    \end{subfigure}\hfill
    \begin{subfigure}[t]{0.275\textwidth}
        \centering
        \includegraphics[width=\linewidth]{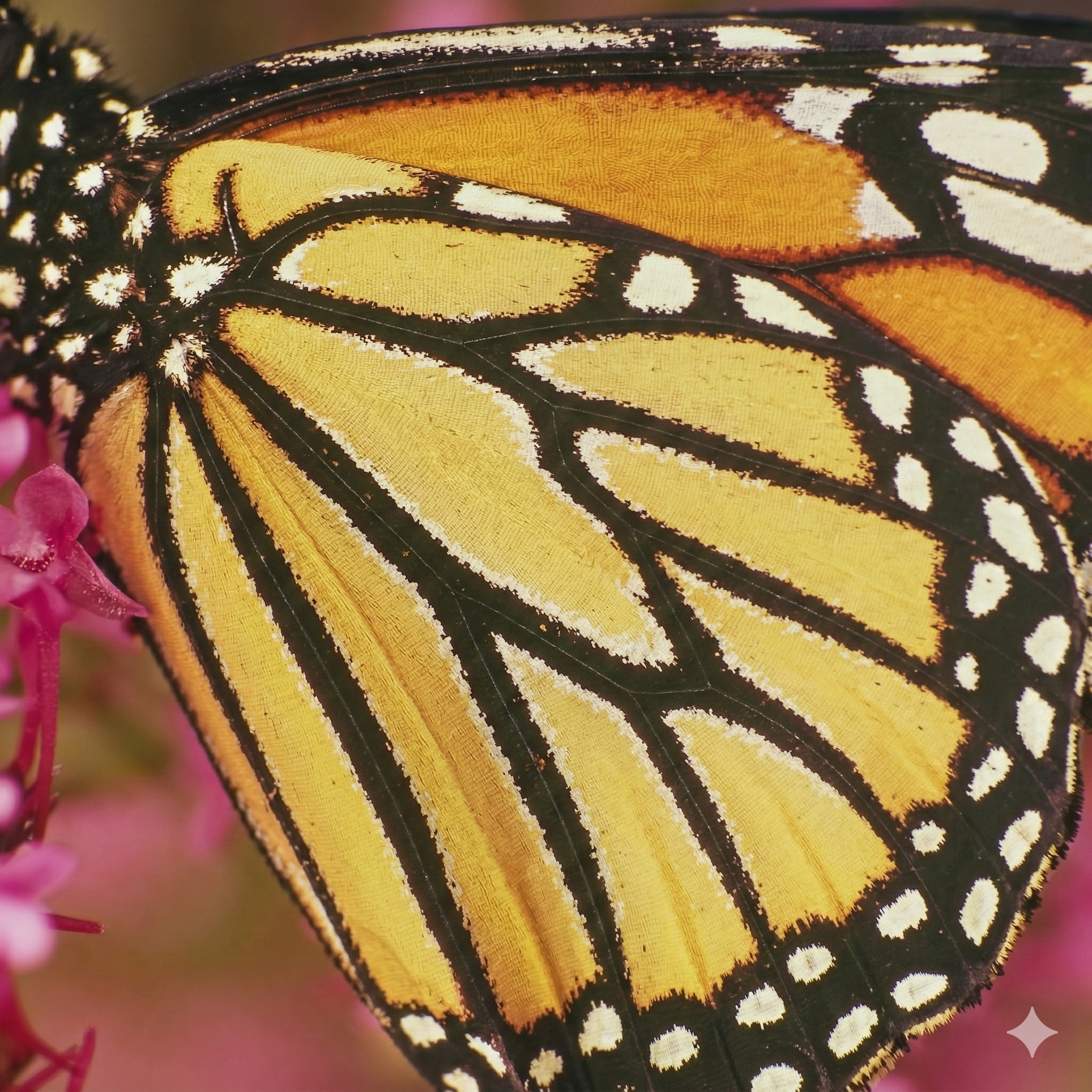}
        \caption{Model Output}
    \end{subfigure}\hfill
    \begin{subfigure}[t]{0.275\textwidth}
        \centering
        \includegraphics[width=\linewidth]{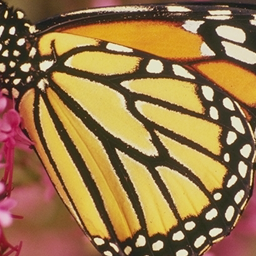}
        \caption{Ground Truth}
    \end{subfigure}
    \caption{\textbf{Super-resolution $\times$2.} The model recovers sharper edges and richer texture from the low-resolution input, but some restored texture details appear to be plausibly synthesized rather than faithfully reconstructed from the original signal. Generated by Nano Banana.}
\end{figure}

\begin{figure}[!htbp]
    \centering
    \begin{subfigure}[t]{0.275\textwidth}
        \centering
        \includegraphics[width=\linewidth]{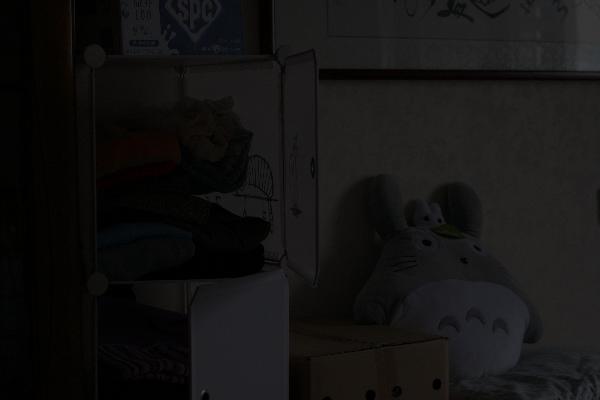}
        \caption{Input}
    \end{subfigure}\hfill
    \begin{subfigure}[t]{0.275\textwidth}
        \centering
        \includegraphics[width=\linewidth]{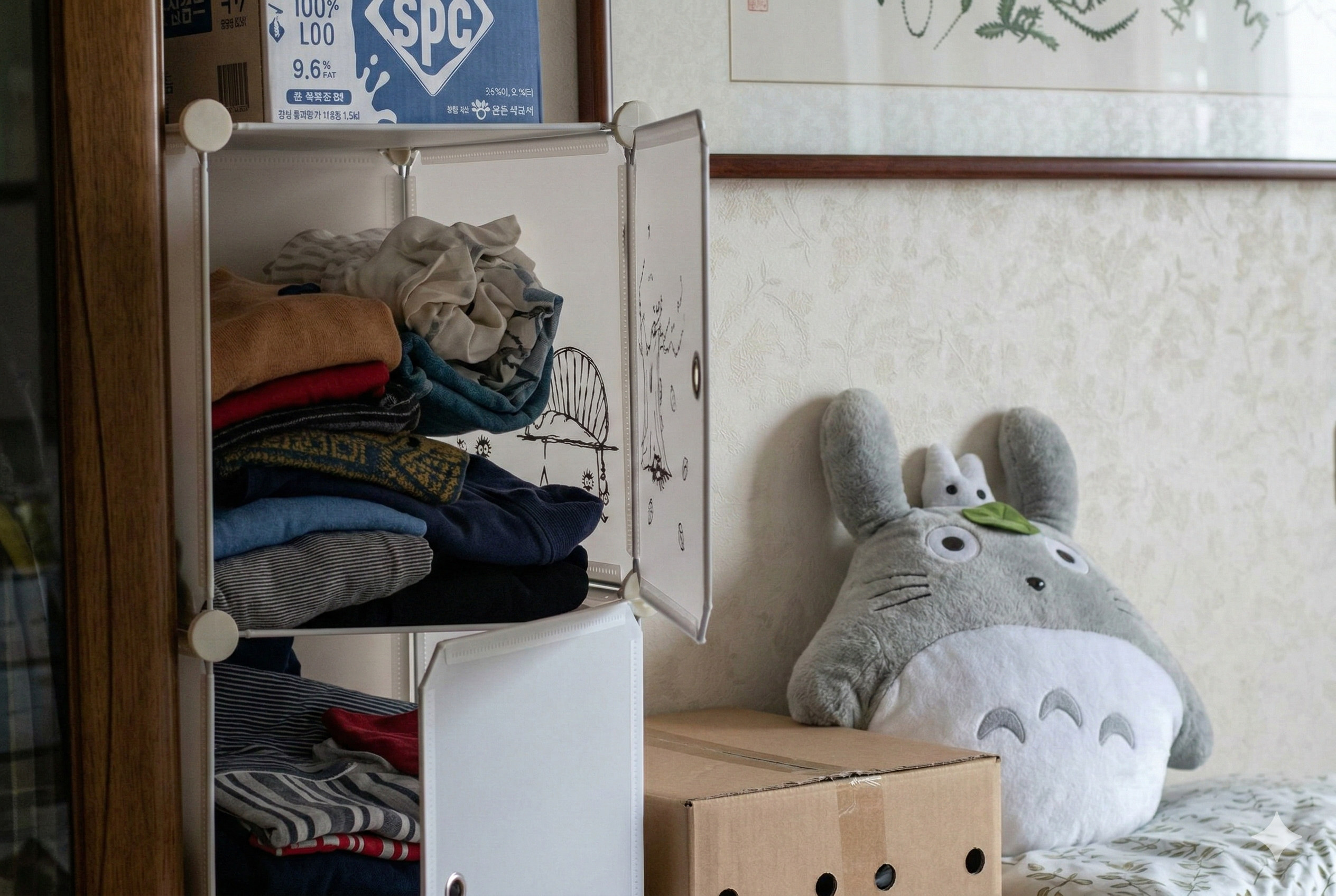}
        \caption{Model Output}
    \end{subfigure}\hfill
    \begin{subfigure}[t]{0.275\textwidth}
        \centering
        \includegraphics[width=\linewidth]{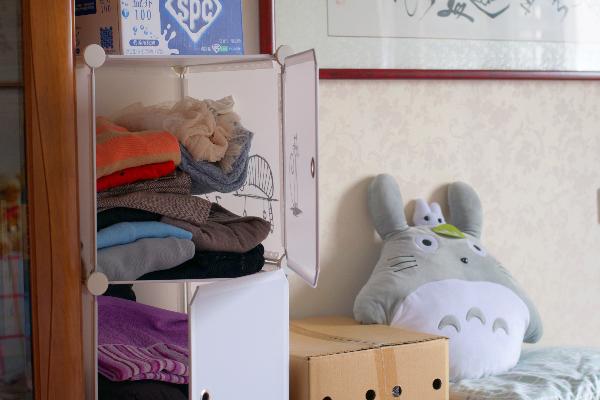}
        \caption{Ground Truth}
    \end{subfigure}
    \caption{\textbf{Low-light Enhancement.} The model substantially brightens the dark input and reveals scene content, but the result also shows clear detail rewriting, suggesting prior-guided enhancement rather than strictly faithful recovery. Generated by Nano Banana.}
\end{figure}

\begin{figure}[!htbp]
    \centering
    \begin{subfigure}[t]{0.275\textwidth}
        \centering
        \includegraphics[width=\linewidth]{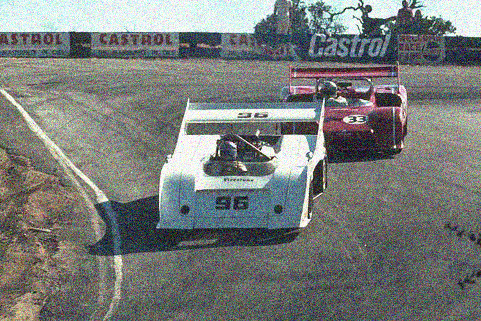}
        \caption{Input}
    \end{subfigure}\hfill
    \begin{subfigure}[t]{0.275\textwidth}
        \centering
        \includegraphics[width=\linewidth]{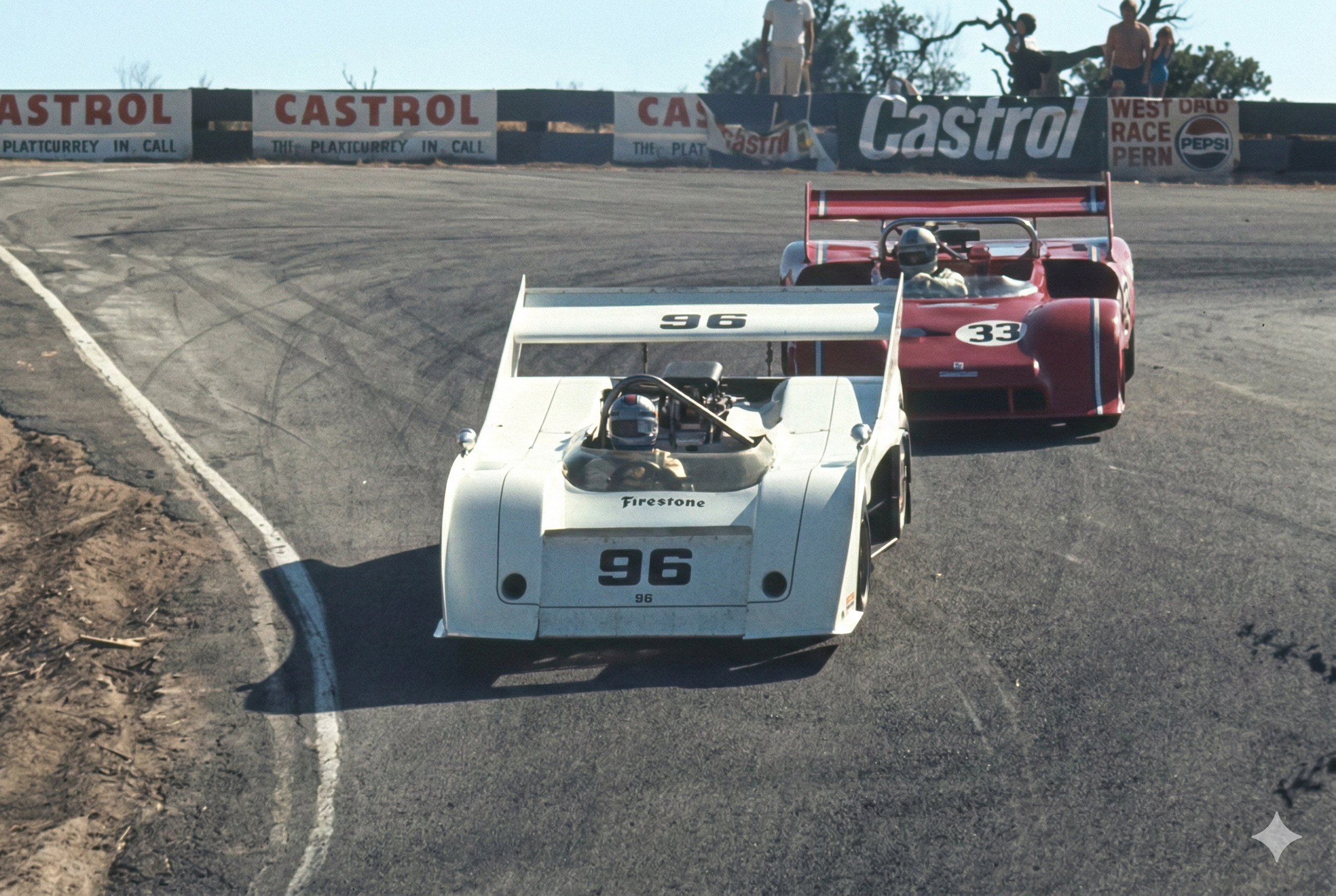}
        \caption{Model Output}
    \end{subfigure}\hfill
    \begin{subfigure}[t]{0.275\textwidth}
        \centering
        \includegraphics[width=\linewidth]{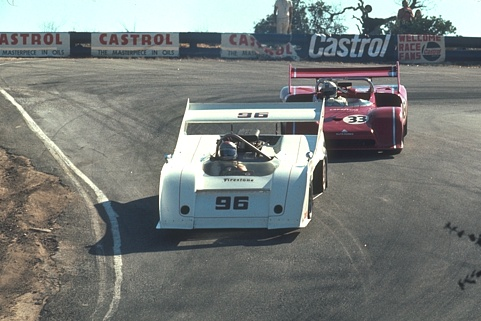}
        \caption{Ground Truth}
    \end{subfigure}
    \caption{\textbf{Denoising.} The model removes heavy noise and restores a visually clean image while preserving the global scene structure. Generated by Nano Banana.}
\end{figure}

\begin{figure}[!htbp]
    \centering
    \begin{subfigure}[t]{0.275\textwidth}
        \centering
        \includegraphics[width=\linewidth]{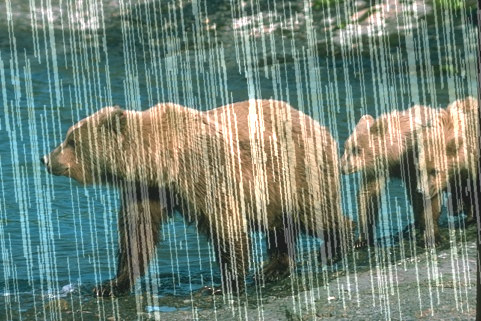}
        \caption{Input}
    \end{subfigure}\hfill
    \begin{subfigure}[t]{0.275\textwidth}
        \centering
        \includegraphics[width=\linewidth]{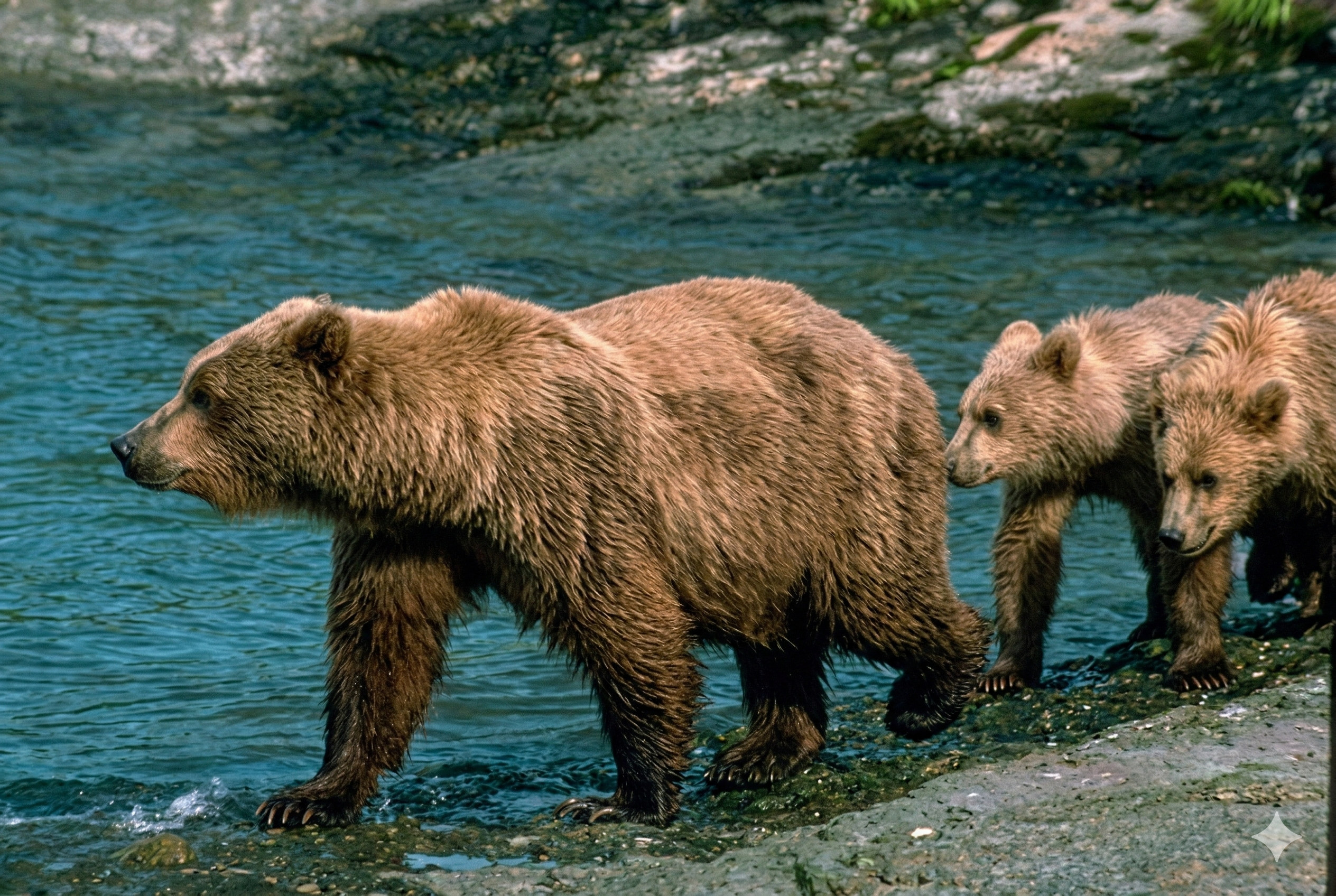}
        \caption{Model Output}
    \end{subfigure}\hfill
    \begin{subfigure}[t]{0.275\textwidth}
        \centering
        \includegraphics[width=\linewidth]{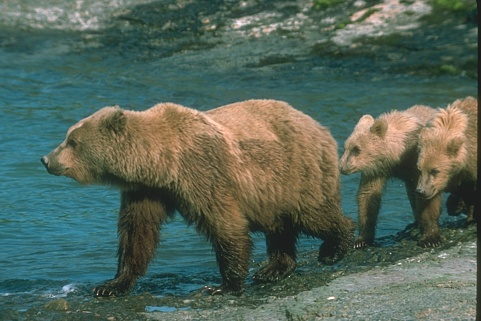}
        \caption{Ground Truth}
    \end{subfigure}
    \caption{\textbf{Deraining.} The model effectively suppresses dominant rain streaks and improves perceptual clarity, but the restored textures are partly regenerated under a learned natural-image prior. Generated by Nano Banana.}
\end{figure}

\begin{figure}[t]
    \centering
    \begin{subfigure}[t]{0.48\textwidth}
        \centering
        \includegraphics[width=\linewidth]{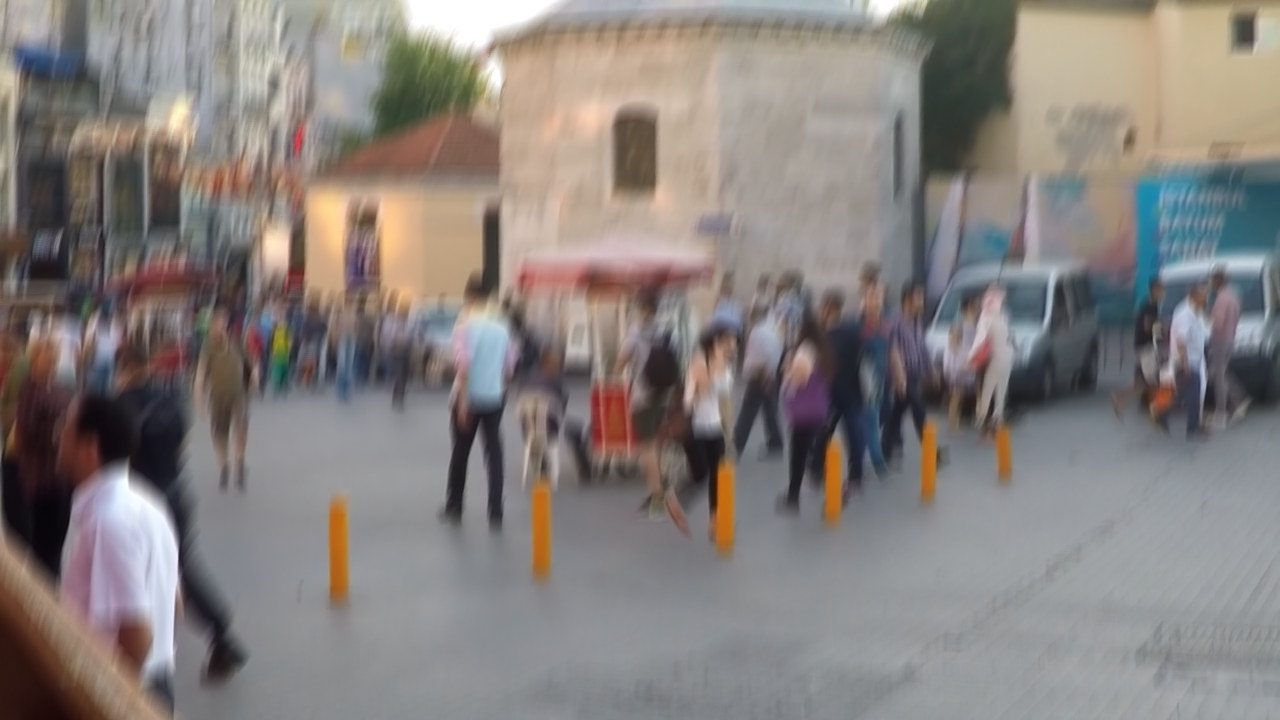}
        \caption{Input}
    \end{subfigure}\hfill
    \begin{subfigure}[t]{0.48\textwidth}
        \centering
        \includegraphics[width=\linewidth]{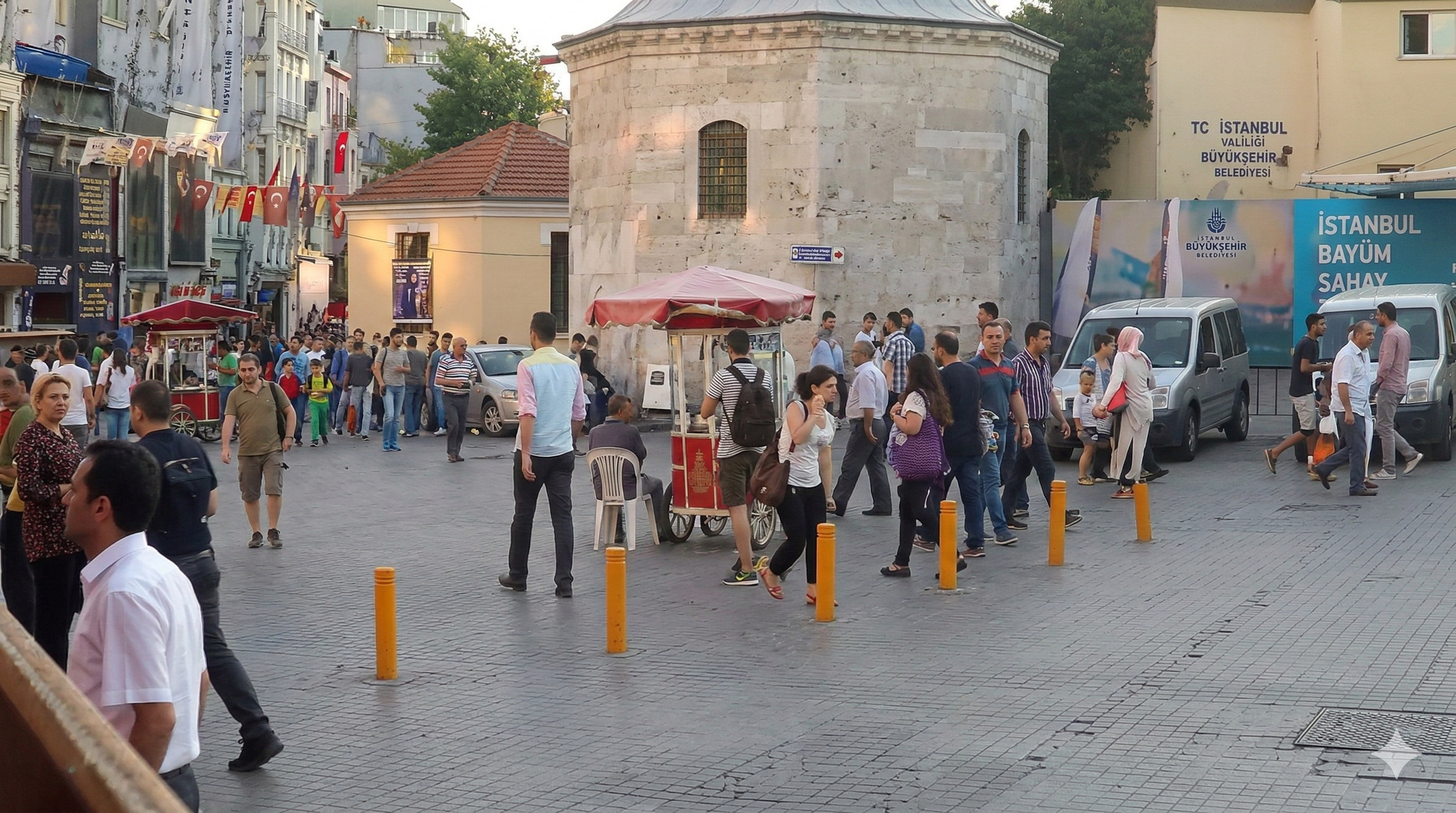}
        \caption{Model Output}
    \end{subfigure}\hfill
    \caption{\textbf{Deblurring.} The model converts a heavily blurred image into a much clearer scene with plausible structures and boundaries, indicating perceptual restoration that may rely on detail synthesis rather than exact inversion of blur. Generated by Nano Banana.}
\end{figure}

\subsection{Dimension VII: Cross-Disciplinary Real-World Applications}

\noindent\textit{Primary Level Tested: L4 (Agentic Generation) / L5 (World-Modeling Generation) --- Can the model serve as a practical tool in professional workflows requiring world knowledge, logical consistency, and domain-specific accuracy?}

To evaluate the model's practical utility beyond standard academic benchmarks, we explore its performance across a wide range of real-world scenarios. In this dimension, we test whether the model can serve as a tool in diverse professional fields. We categorize these real-life applications into three main areas: (1) \textbf{Design and Engineering}, including architectural drafting, UI/UX, product design, and fashion; (2) \textbf{Scientific and Medical Analysis}, covering tasks like X-ray interpretation, meteorology forecasting, and natural sciences; and (3) \textbf{Complex Reasoning and Society}, which involves legal and forensic scene reconstruction, coding, psychology, and educational content generation. By simulating these complex tasks, we aim to uncover the model's true potential and limitations when applied to actual human workflows.

\begin{figure}[!htbp]
    \centering
    \includegraphics[width=0.9\linewidth]{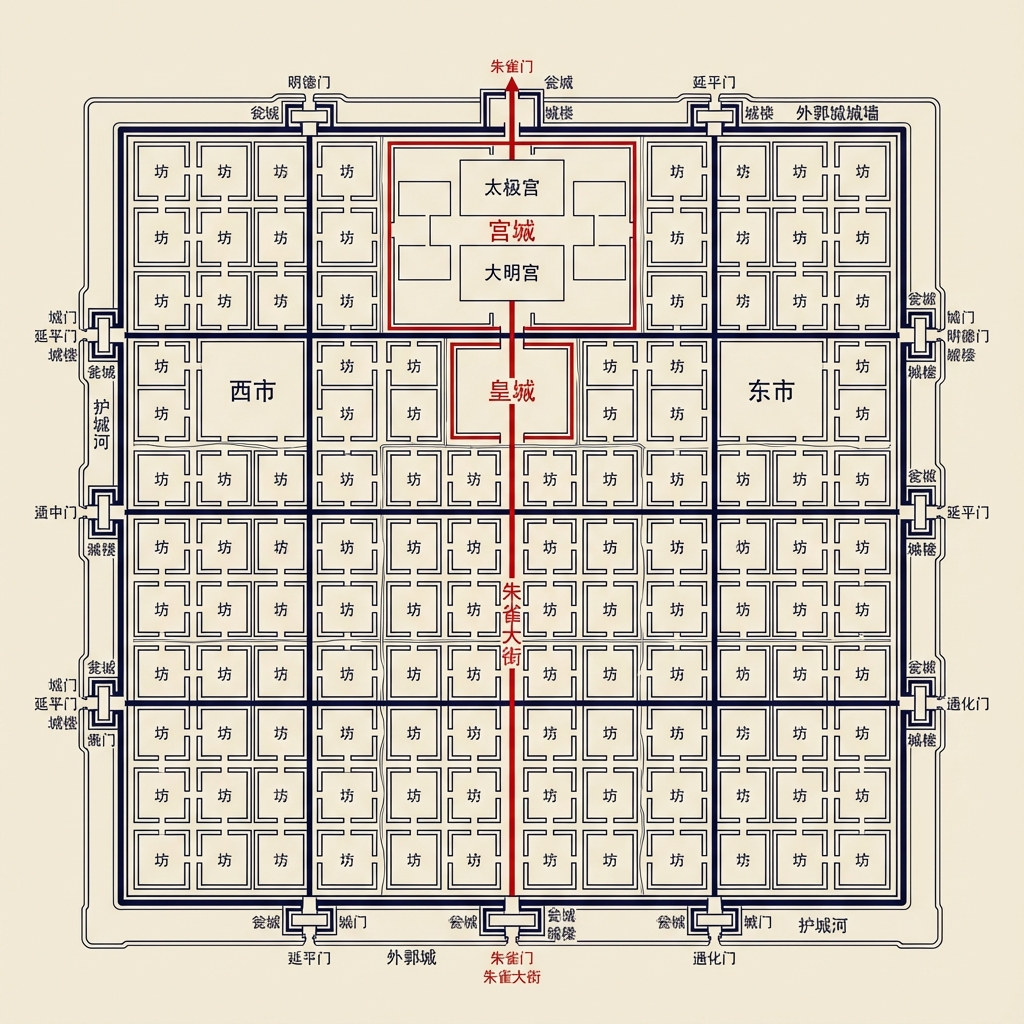}
    \caption{\textbf{Historical Urban Planning Design.} The model successfully generates a historically accurate layout of the Tang Dynasty Chang'an city, but struggles with rendering clear and logically unique Chinese text labels. Generated by Nano Banana.}
    \label{fig:city_design}
\end{figure}

\subsubsection{Case Study I: Historical Urban Planning Design}

\paragraph{Setup and Objectives.} 
In this case study, we test the model's capability in professional architectural and urban design. We prompted the model to generate a top-down floor plan of an ancient Chinese city, specifically based on the Tang Dynasty capital, Chang'an. The instructions were highly detailed: it required a solid background, clear structural logic, specific historical zones (such as the outer city, imperial city, palace, and ward/market areas), and a complete road and defense system. Furthermore, the model was asked to label these functional areas clearly in Chinese.

\paragraph{Success in Structural Accuracy.} 
The model performed exceptionally well in following the overall design instructions. As shown in \Cref{fig:city_design}, the generated image perfectly matches the requested orthographic (top-down) view. More importantly, the spatial layout of the city walls and the grid-like ward system (Fang) are highly accurate and fit the historical background perfectly. The image successfully captures the complex planning logic and authentic feel of an ancient capital.

\paragraph{Limitations in Text Rendering.} 
Despite the excellent spatial and structural design, we observe clear limitations in the finer details. First, the generated Chinese characters are somewhat blurry, making them difficult to read. Second, there are logical errors in the text itself, such as using the exact same name for different city gates. This indicates that while the model has a strong grasp of visual layout and historical styles, it still struggles with generating sharp, precise text and maintaining logical consistency in small, detailed elements.

\begin{figure}[!htbp]
    \centering
    \includegraphics[width=\linewidth]{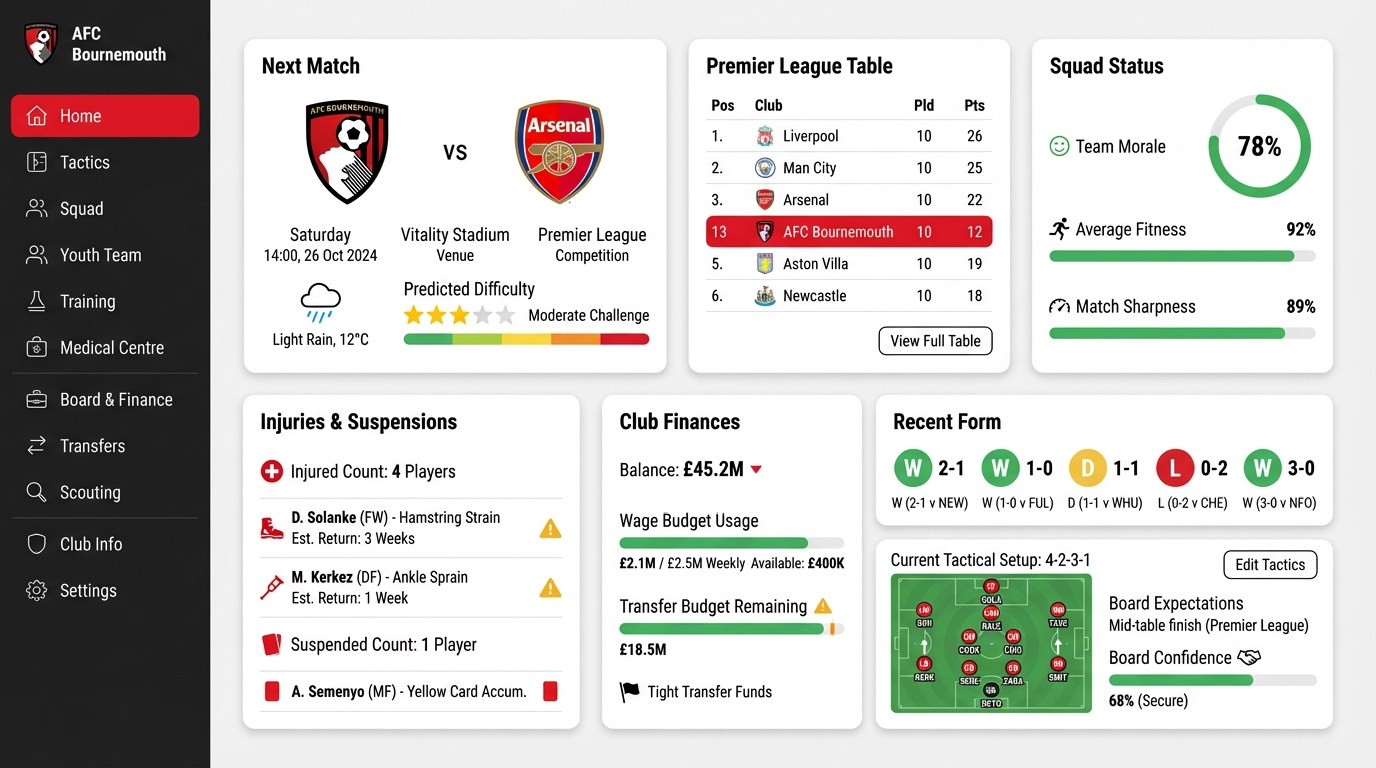}
    \caption{\textbf{Professional UI Dashboard Design.} The model successfully generates a highly realistic and visually appealing football manager interface with accurate world knowledge, though it exhibits minor logical flaws in fine-grained details. Generated by Nano Banana.}
    \label{fig:ui_design}
\end{figure}

\subsubsection{Case Study II: Professional UI Dashboard Design}
\paragraph{Setup and Objectives.} 
In this second case study, we test the model's ability to create a complex user interface (UI). We prompted the model to design a professional football manager dashboard for the English club AFC Bournemouth. The detailed prompt required a modern, data-driven layout containing multiple specific panels, such as the next match, league table, squad status, injuries, club finances, and a tactical pitch. 
\paragraph{Success in Visuals and World Knowledge.} 
The model performs exceptionally well in overall design and utilizes world knowledge. The generated UI is beautiful, clean, and very easy to understand. It correctly includes real-world details from its training data, such as actual Bournemouth players (e.g., Semenyo, Kerkez) and club-specific elements. Notably, unlike the previous city planning task, the model successfully generates perfectly clear and readable English text with no blurring or spelling issues.
\paragraph{Limitations in Fine-Grained Logic.} 
Despite the excellent overall appearance, we observe several minor logical flaws in the small details. First, the tactical pitch incorrectly displays four goals instead of two. Second, in the league table, Bournemouth is ranked 13th but is visually placed near the very top of the list. Third, Semenyo, who is suspended due to yellow card accumulation, is incorrectly given a red card icon. Fourth, the player position is labeled as a generic ``DF'' (Defender), whereas real football UI typically uses specific roles like LB (Left Back) or RB (Right Back). Finally, the remaining transfer budget progress bar is poorly drawn. These small mistakes show that while the model excels at broad visual design and semantic knowledge, it can still struggle with the strict logical consistency of fine-grained elements.

\begin{figure}[!htbp]
    \centering
    \begin{subfigure}[t]{0.48\linewidth}
        \centering
        \includegraphics[width=\linewidth]{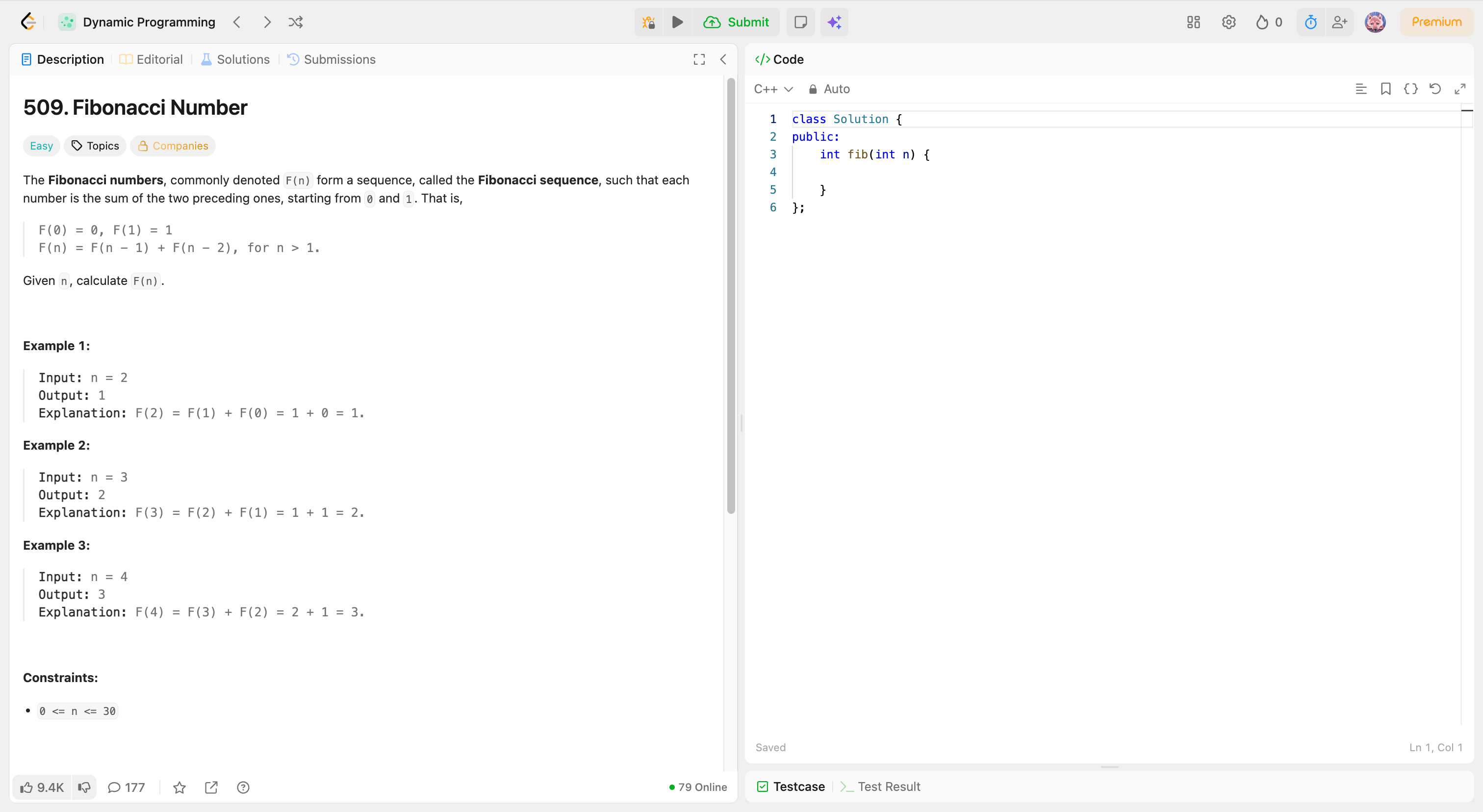}
        \caption{Easy input: Fibonacci Number}
    \end{subfigure}\hfill
    \begin{subfigure}[t]{0.48\linewidth}
        \centering
        \includegraphics[width=\linewidth]{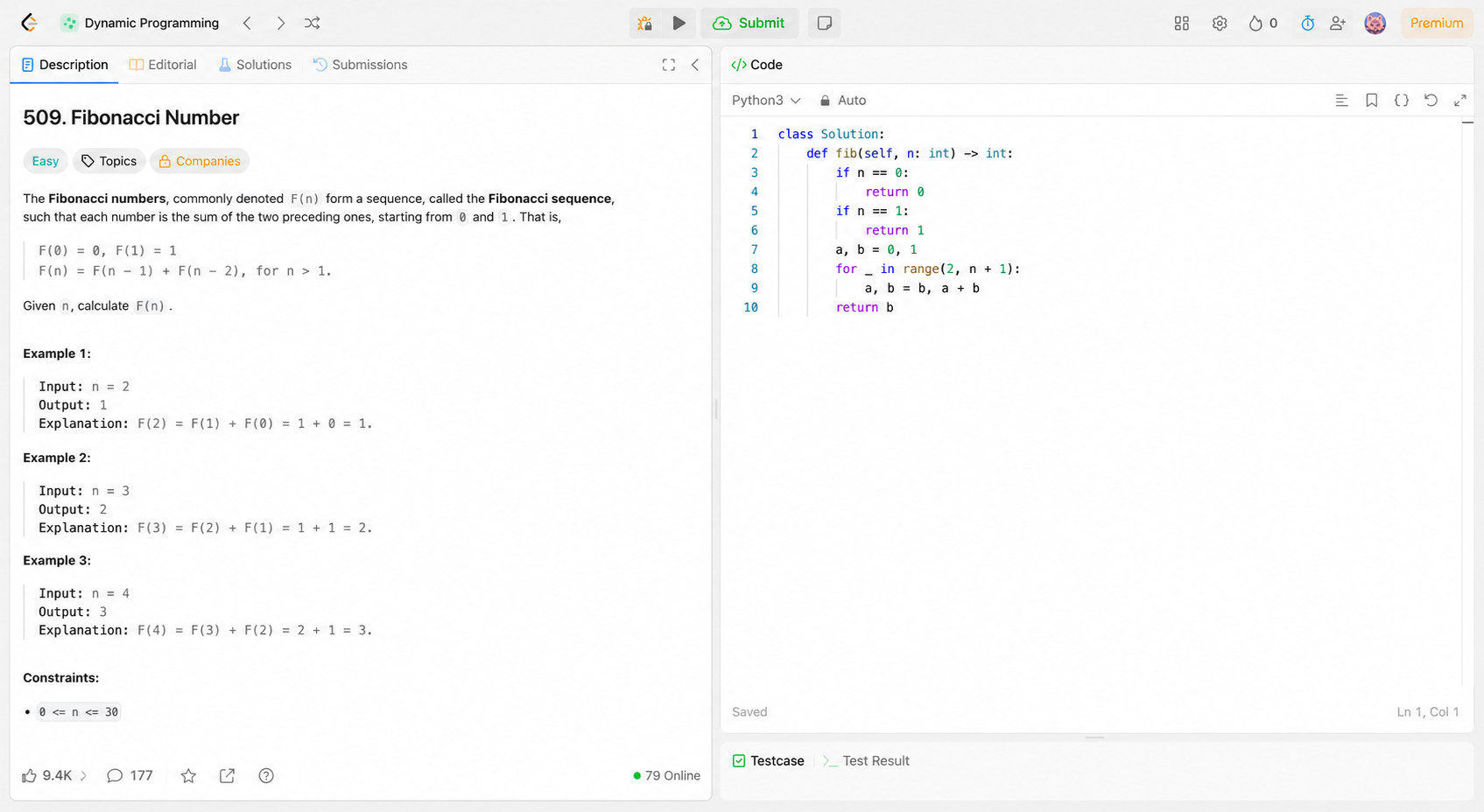}
        \caption{GPT-Image-2 output}
    \end{subfigure}

    \vspace{1mm}
    \begin{subfigure}[t]{0.48\linewidth}
        \centering
        \includegraphics[width=\linewidth]{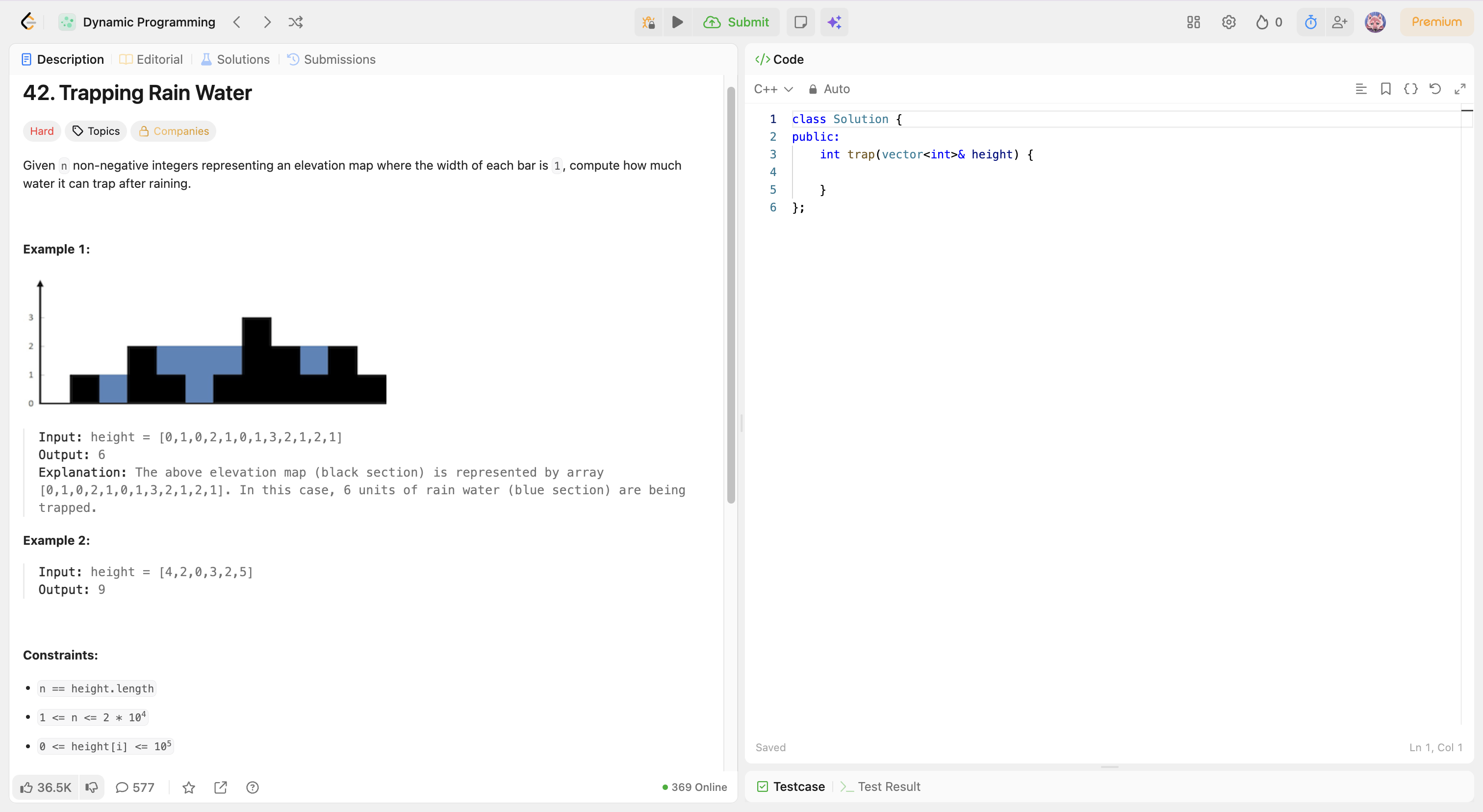}
        \caption{Hard input: Trapping Rain Water}
    \end{subfigure}\hfill
    \begin{subfigure}[t]{0.48\linewidth}
        \centering
        \includegraphics[width=\linewidth]{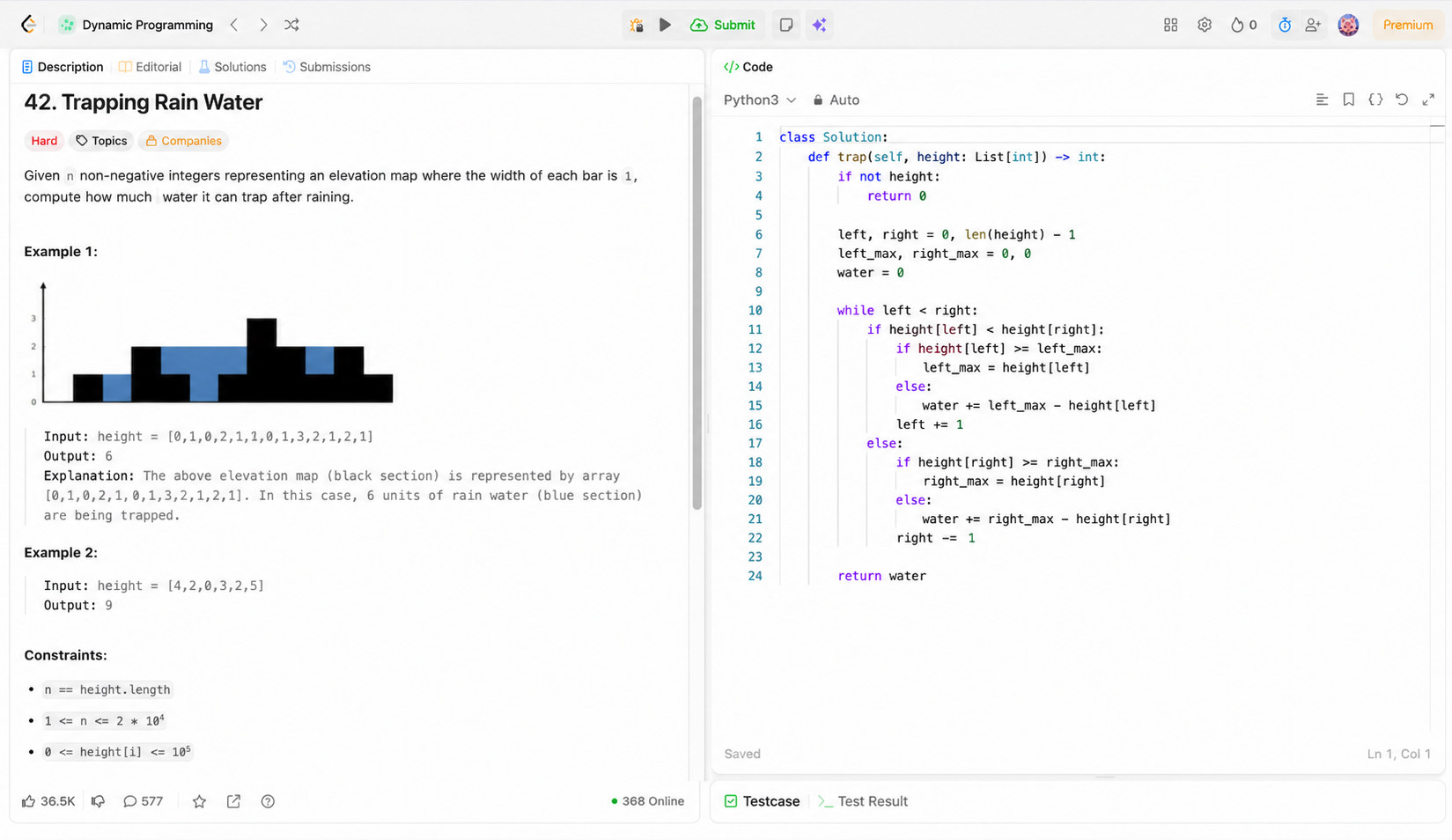}
        \caption{GPT-Image-2 output}
    \end{subfigure}
    \caption{\textbf{Code generation as a read-solve-render workflow with GPT-Image-2.} Given a screenshot of a LeetCode problem interface and a target programming language, the model must read the problem, infer the correct algorithm, and generate a new LeetCode-style image containing the original problem on the left and executable solution code on the right. Both the easy Fibonacci case and the hard trapping-rain-water case are rendered with clear UI structure, readable code, and correct Python solutions.}
    \label{fig:code_generation_case}
\end{figure}

\subsubsection{Case Study III: Coding Problem Solving as Visual UI Generation}

\paragraph{Setup and Objectives.}
We further test whether a frontier image generation model can support programming workflows rather than only visual design. The input is a screenshot of a LeetCode problem interface, and the prompt asks the model to read the problem, solve it, and output a \emph{single new image} that mimics a LeetCode UI. The generated image must contain two panels: the left panel reproduces the problem title, description, examples, and constraints, while the right panel renders complete solution code in a specified target language. This task combines OCR, algorithmic reasoning, code synthesis, and high-fidelity UI rendering in one closed-loop visual output.

\paragraph{Evidence from Easy and Hard Problems.}
As shown in \Cref{fig:code_generation_case}, GPT-Image-2 succeeds on both an easy dynamic-programming problem, \emph{Fibonacci Number}, and a harder two-pointer problem, \emph{Trapping Rain Water}. In the easy case, the model converts the blank C++ editor state into a Python LeetCode solution with the correct function signature and an iterative recurrence. In the hard case, it produces the standard linear-time two-pointer algorithm, maintaining left/right maxima and accumulating trapped water correctly. In both examples, the output also preserves the two-panel LeetCode layout, syntax highlighting, indentation, and readable high-resolution text.

\paragraph{Insight.}
This result suggests that GPT-Image-2 can already act as a practical \textbf{read-solve-render} interface for coding tasks: it does not merely redraw the input screenshot, but transforms visual problem understanding into executable symbolic content and then renders that content back into a professional UI. The capability is especially notable because success requires alignment across several layers: exact problem comprehension, algorithm selection, target-language formatting, and document-level visual composition. At the same time, this should be interpreted as a capability probe rather than a complete coding benchmark; broader coverage across languages, edge cases, hidden tests, and more adversarial screenshots will be needed to establish robustness.

\begin{figure}[!htbp]
    \centering
    \includegraphics[width=0.92\linewidth]{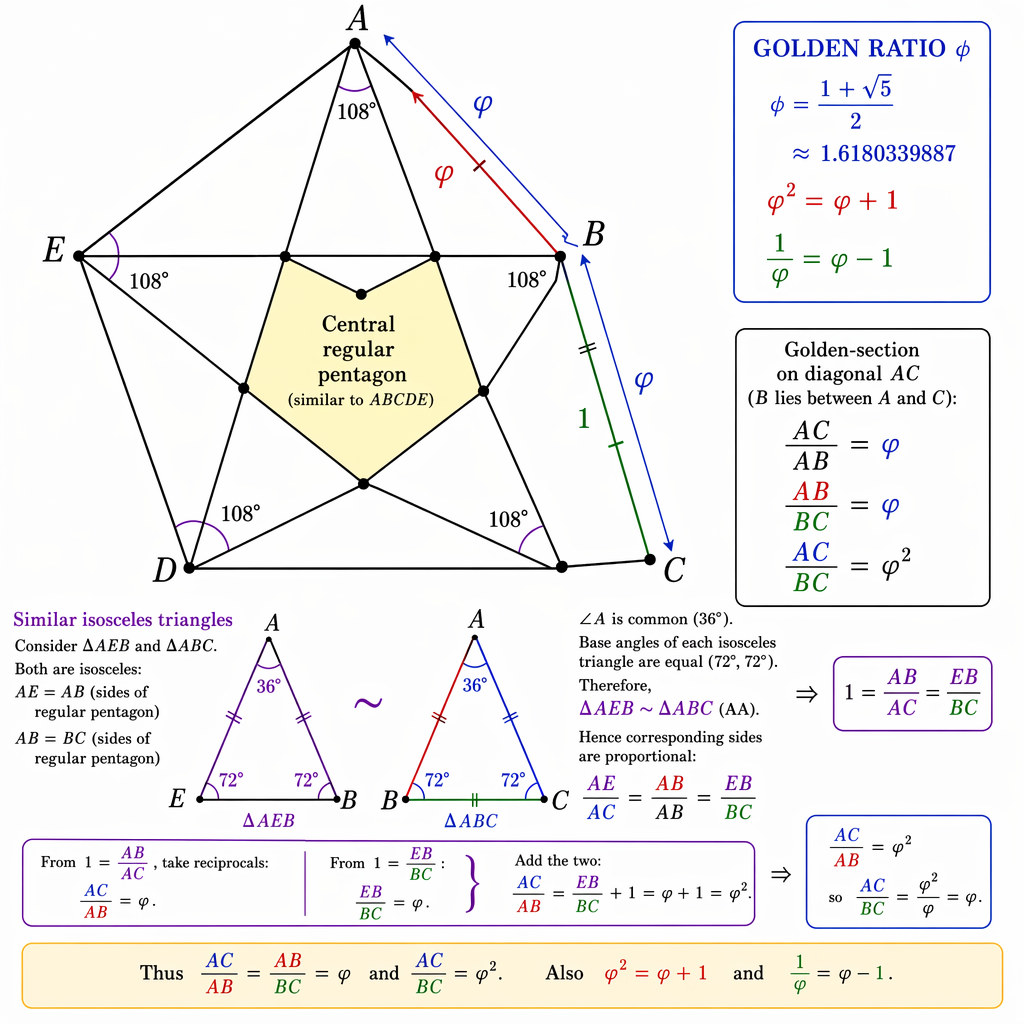}
    \caption{\textbf{Structured mathematical diagram generation with GPT-Image-2.} The prompt asks for a Euclidean proof diagram of the golden ratio in a regular pentagram, including a regular pentagon, all diagonals, a central similar pentagon, angle markings, side ticks, golden-section ratios, and algebraic identities involving $\phi$. The generated image organizes geometry, formulas, labels, and proof annotations into a readable educational diagram.}
    \label{fig:math_structure_case}
\end{figure}

\subsubsection{Case Study IV: Mathematical Proof Diagram Generation}

\paragraph{Setup and Objectives.}
We next test GPT-Image-2 on a structured mathematical visualization task. Unlike ordinary illustration, the prompt requires the model to construct a Euclidean proof diagram for the golden ratio in a regular pentagram. The output must combine multiple constraints: a regular pentagon $ABCDE$ with $108^\circ$ interior angles, all five diagonals forming a pentagram, a smaller central regular pentagon similar to the outer one, explicit golden-ratio relations such as $\phi=(1+\sqrt{5})/2$, $\phi^2=\phi+1$, and $1/\phi=\phi-1$, and at least one pair of similar isosceles triangles marked by angle arcs or side ticks. The task therefore probes whether the model can synthesize geometry, mathematical notation, and proof-oriented visual layout into one coherent image.

\paragraph{Observed Behavior.}
As shown in \Cref{fig:math_structure_case}, GPT-Image-2 produces a highly structured proof-style diagram rather than a decorative pentagram. The outer pentagon and internal diagonals are visually organized, the central similar pentagon is clearly highlighted, and the diagram includes repeated $108^\circ$ angle annotations, color-coded segment labels, matching side ticks, and boxed algebraic identities. The lower portion further converts the geometric construction into a proof narrative, explicitly comparing similar isosceles triangles and deriving proportional relations among $AC$, $AB$, $BC$, and $\phi$. Text, formulas, and arrows remain largely legible, which is crucial for an educational mathematical diagram.

\paragraph{Insight.}
This case suggests that GPT-Image-2 has a strong capacity for \textbf{structure-aware diagram synthesis}: it can follow a dense symbolic prompt, arrange mathematical objects in a pedagogically meaningful layout, and render both geometric and algebraic information with high visual clarity. Compared with free-form image generation, success here requires tighter coupling between spatial construction and symbolic semantics. The remaining limitation is that visual plausibility should not be mistaken for certified geometric correctness: exact regularity, proportional lengths, and proof validity still require external verification. Nevertheless, the result is a strong example of a model moving beyond aesthetic rendering toward usable mathematical visual communication.

\subsubsection{Case Study V: Multilingual Culinary Poster Design}

\paragraph{Setup and Objectives.}
In this case study, we evaluate whether GPT-Image-2 can generate a practical food-media design artifact that combines \textbf{culinary world knowledge}, \textbf{multilingual text rendering}, and \textbf{professional poster design}. The prompt asks for a premium food preparation poster for \emph{spicy crawfish}, featuring a beautiful hero dish, warm natural lighting, a cream background, and an elegant step-by-step recipe layout. The poster is also required to contain three functional sections---ingredients, cooking process, and serving suggestions---with headings and labels rendered in English, Chinese, Japanese, and Korean. This makes the task more complex than ordinary food photography, since the model must simultaneously produce an appetizing dish image, organize procedural recipe information, and maintain multilingual visual coherence.

\begin{figure}[t]
    \centering
    \begin{subfigure}[b]{0.48\textwidth}
        \includegraphics[width=\linewidth]{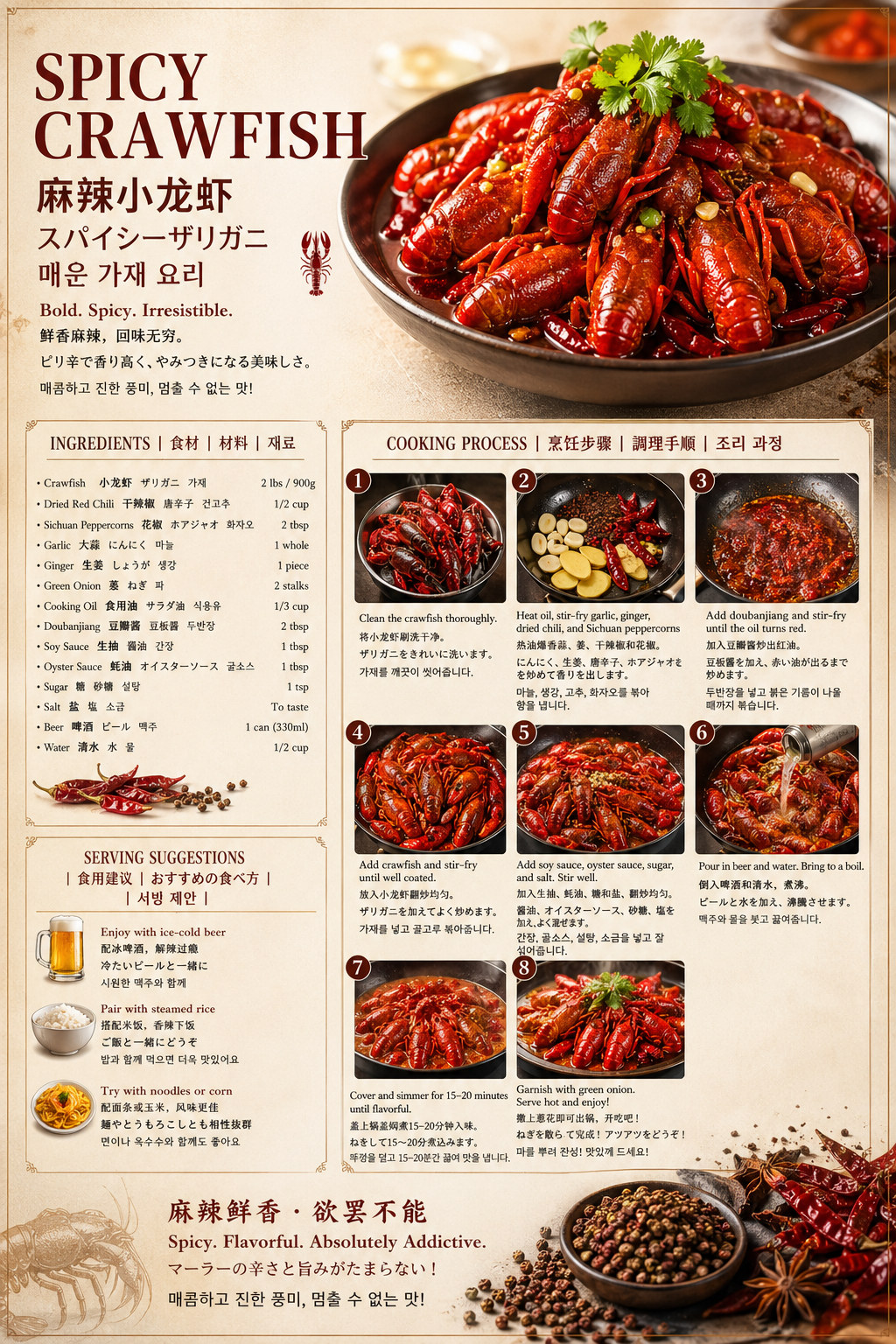}
        \caption{Multilingual culinary poster}
        \label{fig:food_poster_case}
    \end{subfigure}
    \hfill
    \begin{subfigure}[b]{0.45\textwidth}
        \includegraphics[width=\linewidth]{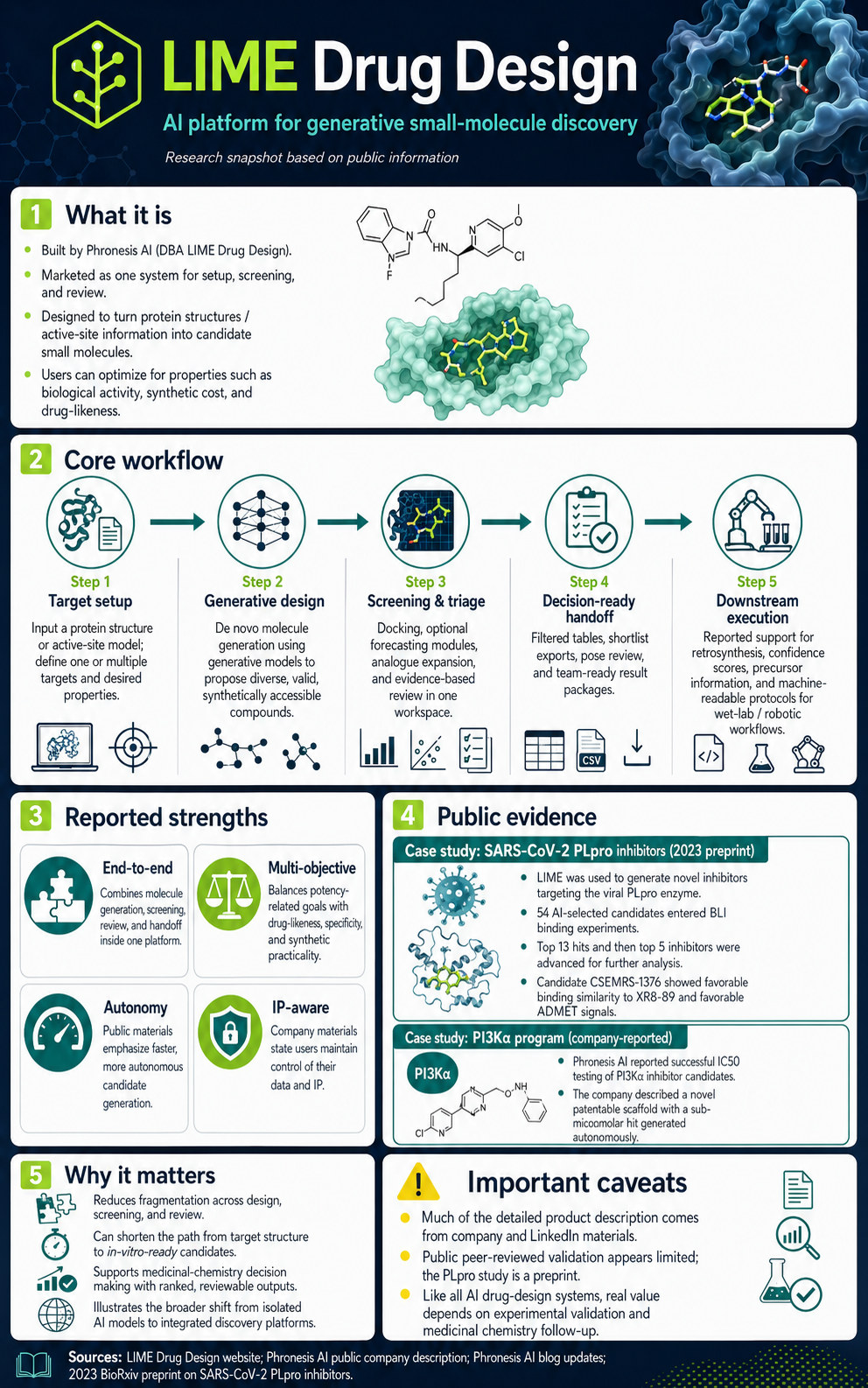}
        \caption{Biomedical infographic}
        \label{fig:lime_drug_design}
    \end{subfigure}
    \caption{\textbf{Cross-disciplinary poster and infographic generation with GPT-Image-2.}
    Both examples are generated by \textbf{GPT-Image-2} from open-ended prompts requiring world knowledge, dense text rendering, and professional information design.
    \Cref{fig:food_poster_case} shows a premium multilingual food preparation poster for spicy crawfish, combining culinary knowledge, English/Chinese/Japanese/Korean text rendering, and luxury restaurant-style recipe layout.
    \Cref{fig:lime_drug_design} shows a biomedical infographic on LIME Drug Design, integrating scientific content organization, domain-aware visual elements, hierarchical information design, and long-form English text rendering.}
    \label{fig:cross_domain_infographic_cases}
\end{figure}

\paragraph{Observed Behavior.}
As shown in \Cref{fig:food_poster_case}, GPT-Image-2 produces a visually polished poster that closely resembles a luxury restaurant advertisement or cookbook editorial page. The hero dish is highly appealing, with red crawfish, dried chilies, peppercorns, aromatics, glossy sauce, and warm lighting that together convey the sensory identity of spicy crawfish. The model also shows reasonable culinary knowledge: the ingredient list includes crawfish, dried chili, Sichuan peppercorns, garlic, ginger, green onion, doubanjiang, soy sauce, oyster sauce, beer, and water, while the cooking process follows a plausible workflow from cleaning and stir-frying to simmering and garnishing. The overall layout is clean and information-rich, with clear section division, numbered cooking steps, supporting food images, and decorative spice elements that reinforce the premium culinary theme.

\paragraph{Insight.}
A particularly important observation is that the model can render multiple languages in the same visual document with largely correct meanings. The English, Chinese, Japanese, and Korean headings and labels are mostly recognizable and semantically aligned with the corresponding recipe sections, demonstrating strong multilingual document-level rendering ability. However, the quality is not uniform across text scales: while large headings and major labels are generally clear, smaller non-English text, especially Chinese, Japanese, and Korean body text, shows a certain degree of blurriness and reduced legibility. This suggests that GPT-Image-2 already supports impressive cross-lingual poster generation, but fine-grained multilingual typography remains more challenging when the text becomes dense and small.

\subsubsection{Case Study VI: Biomedical Infographic Generation from Public Knowledge}

\paragraph{Setup and Objectives.}
In this case study, we test whether GPT-Image-2 can transform an open-ended domain-specific instruction into a professional scientific infographic. The prompt is intentionally brief: \emph{``Research LIME Drug Design and make a detailed infographic about it.''} Unlike tasks that provide explicit layout instructions, this prompt requires the model to determine what information should be included, how the content should be structured, and what visual form is appropriate for communicating a biomedical topic. The case therefore evaluates the model's \textbf{world knowledge}, \textbf{scientific information organization}, \textbf{infographic design ability}, and \textbf{text rendering quality}.

\paragraph{Observed Behavior.}
As shown in \Cref{fig:lime_drug_design}, GPT-Image-2 generates a polished multi-panel biomedical infographic rather than a generic scientific illustration. The output identifies and organizes multiple aspects of LIME Drug Design, including its definition, workflow, strengths, public evidence, broader significance, and caveats. The visual design is also highly structured: numbered modules, workflow arrows, molecular diagrams, protein-style illustrations, icons, and consistent color coding guide the viewer through the topic. The rendered English text is dense yet largely readable, with clear section headers, explanatory blocks, and scientific terminology arranged in a coherent hierarchy.

\paragraph{Insight.}
This case highlights GPT-Image-2's ability to function as a \textbf{knowledge-to-communication} model. Given only a short research-oriented prompt, it can select relevant conceptual components, impose an explanatory structure, and present the result as a visually coherent scientific communication artifact. The success of this example depends on more than image aesthetics: the model must combine topic understanding, information prioritization, visual hierarchy, scientific iconography, and long-form text rendering. This suggests that frontier text-to-image models are increasingly capable of supporting real-world infographic creation, especially when the task requires turning specialized knowledge into an accessible and professionally designed visual summary.

\subsection{Dimension VIII: High-level Vision Tasks}

\noindent\textit{Primary Level Tested: L2 (Conditional Generation) --- Can the model convert visual understanding into explicit, structured, and actionable predictions (bounding boxes, keypoints, masks, text)?}

Beyond open-ended image generation, a practically useful multimodal model is also expected to support visual tasks whose outputs are explicit, structured, and directly actionable. These settings are particularly informative because success is no longer measured by perceptual plausibility alone, but by whether visual understanding can be converted into symbolic or geometric predictions with clear operational meaning. We therefore examine four representative tasks, namely optical character recognition, keypoint estimation, semantic segmentation, and object detection. Taken together, they span text decoding, articulated pose modeling, dense region parsing, and instance-level localization, thus offering a compact yet revealing view of how the model transitions from holistic perception to structured visual prediction.

\begin{figure}[!htbp]
    \centering
    \begin{subfigure}[t]{0.48\linewidth}
        \centering
        \includegraphics[width=\linewidth]{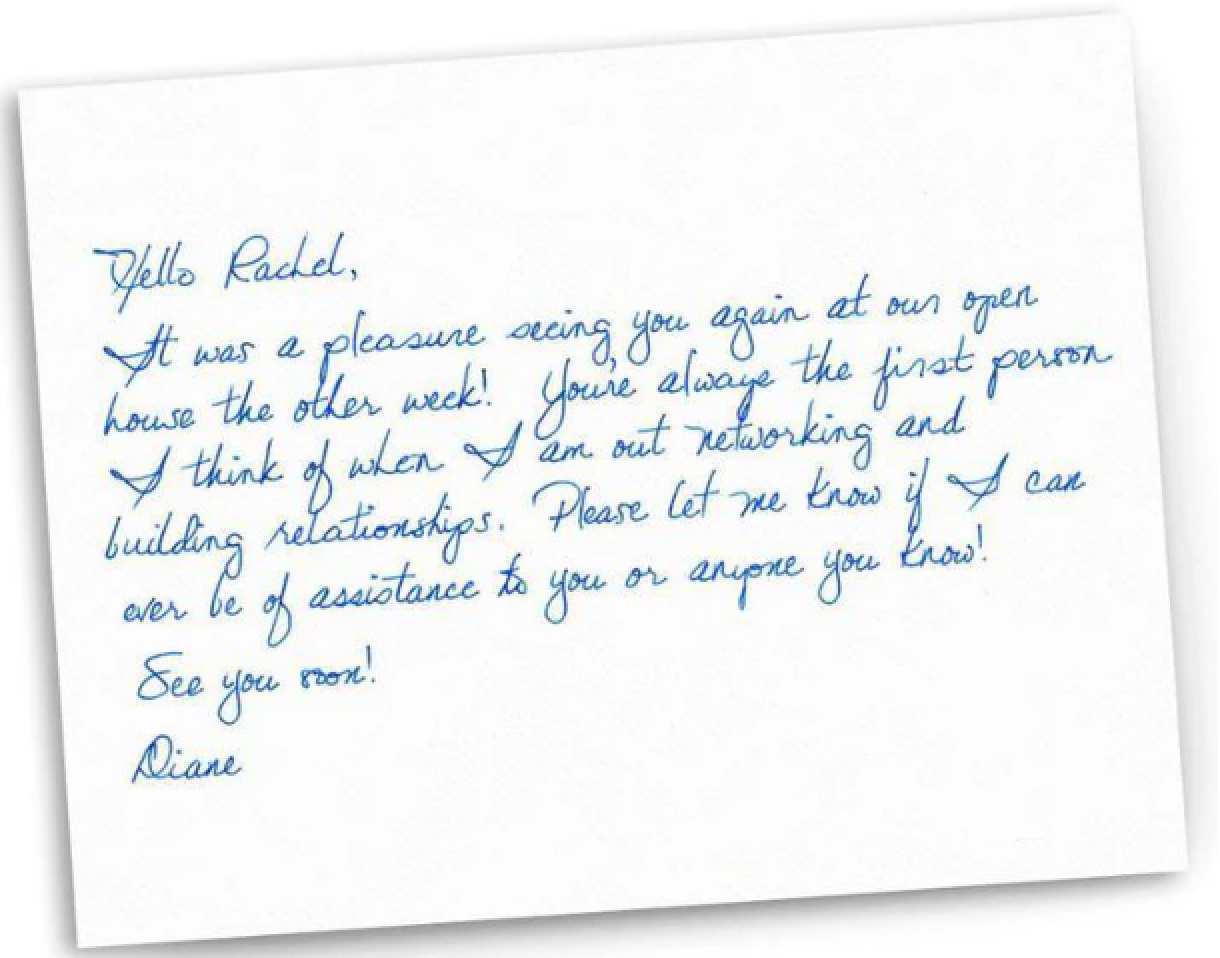}
        \caption{Input}
    \end{subfigure}\hfill
    \begin{subfigure}[t]{0.48\linewidth}
        \centering
        \includegraphics[width=\linewidth]{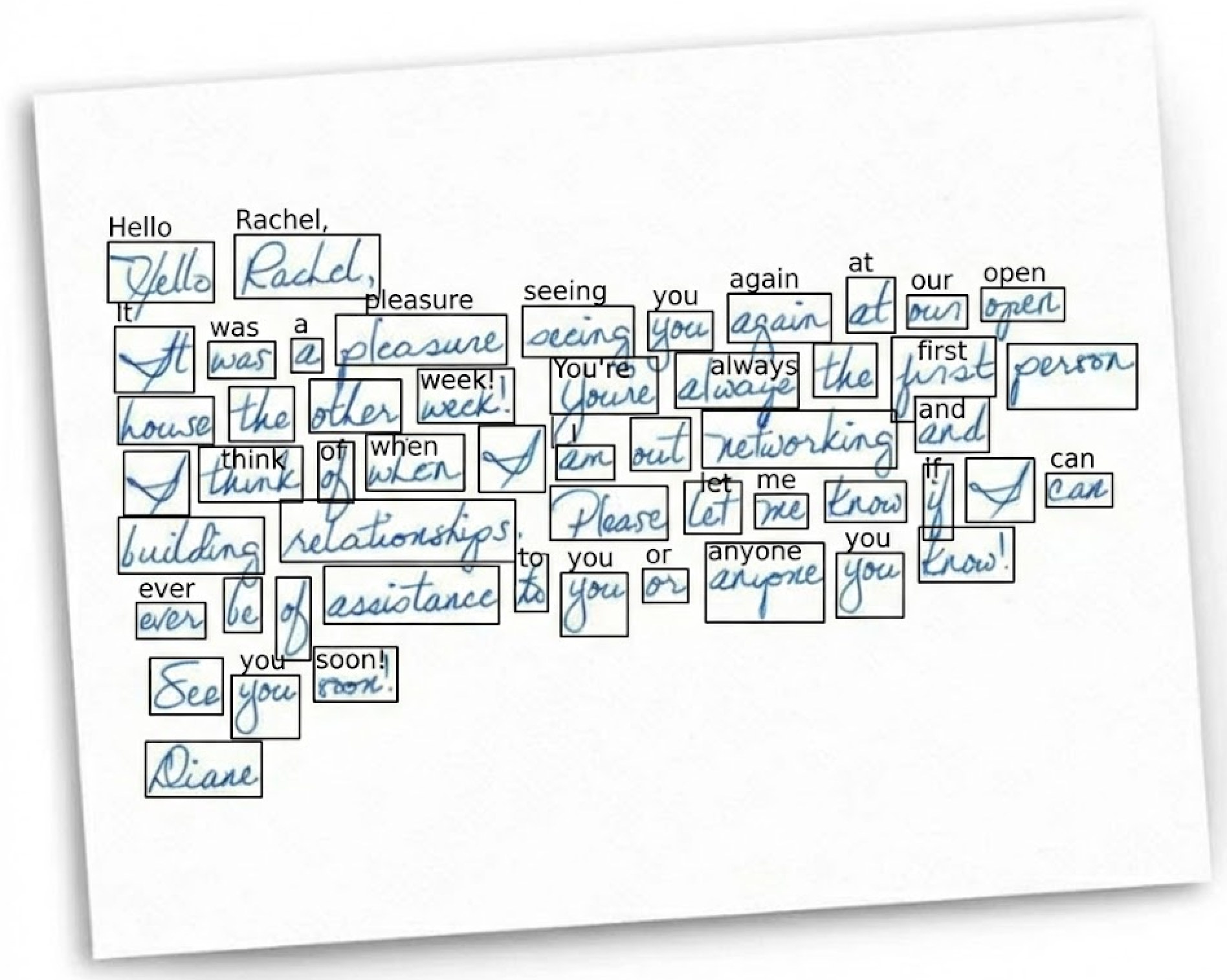}
        \caption{Model Output}
    \end{subfigure}
    \caption{\textbf{OCR.} The model recovers the dominant text regions and a substantial portion of the readable content, while character-level corruption and omissions remain visible for small or degraded text. Generated by Nano Banana.}
    \label{fig:ocr_case}
\end{figure}

\subsubsection{Case Study I: Optical Character Recognition}

\paragraph{Setup and Evaluation Focus.}
We begin with OCR, a task that requires the model to convert visual text into explicit symbolic output. Compared with open-ended generation, the tolerance for local error is much lower here, since minor deviations in character identity, spacing, or line structure can directly reduce usability. The task therefore provides a clean test of whether the model can move beyond recognizing the presence of text and preserve content in a form that remains legible and operationally meaningful.

\paragraph{Evidence of Global Competence.}
As shown in \Cref{fig:ocr_case}, the model captures the dominant text layout well and recovers a substantial portion of the salient content. At the level of paragraphs, lines, and major textual regions, the output remains broadly aligned with the input, suggesting that the model has acquired a meaningful representation of document structure rather than merely reproducing a superficial visual pattern. This result indicates that high-level text perception is already within reach.

\paragraph{Remaining Challenge in Precision.}
The main limitation appears when recognition must be resolved at the level of individual characters. Small, low-contrast, or degraded text is more prone to omission or corruption, and some local regions exhibit visible drops in readability. The overall pattern suggests that the model is better at preserving the existence and arrangement of text than its exact symbolic content. This gap is informative: it shows that structured prediction can already emerge from the model's visual representations, while dependable fine-grained transcription remains substantially harder.

\begin{figure}[!htbp]
    \centering
    \begin{subfigure}[t]{0.48\linewidth}
        \centering
        \includegraphics[width=\linewidth]{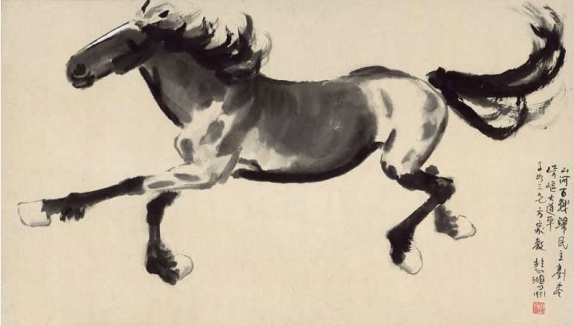}
        \caption{Input}
    \end{subfigure}\hfill
    \begin{subfigure}[t]{0.48\linewidth}
        \centering
        \includegraphics[width=\linewidth]{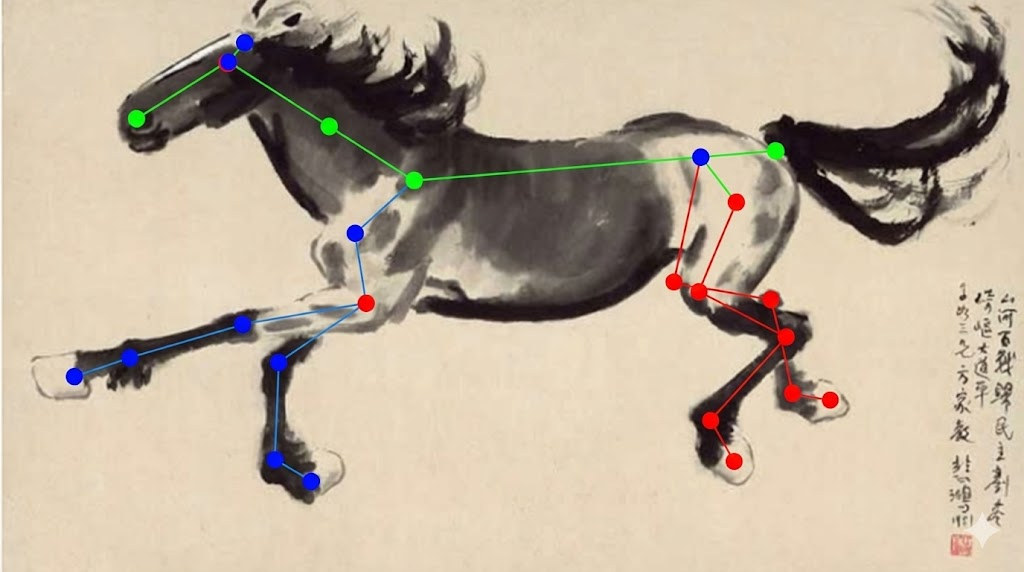}
        \caption{Model Output}
    \end{subfigure}
    \caption{\textbf{Keypoint Estimation.} The model produces a plausible keypoint configuration that captures the overall pose structure, although errors remain around occluded joints and locally ambiguous body parts. Generated by Nano Banana.}
    \label{fig:keypoint_case}
\end{figure}

\subsubsection{Case Study II: Keypoint Estimation}

\paragraph{Setup and Evaluation Focus.}
We next consider keypoint estimation, where the model must express body structure through a sparse yet semantically organized set of landmarks. Relative to OCR, the emphasis here shifts from symbolic decoding to geometric reasoning: success depends not only on recognizing the subject, but also on inferring how its parts are spatially related under pose variation, viewpoint change, and partial occlusion. This makes the task a natural probe of whether the model can impose structured spatial organization on top of raw appearance.

\paragraph{Evidence of Global Competence.}
The result in \Cref{fig:keypoint_case} shows that the model recovers the global pose configuration with considerable fidelity. Major joints are arranged in a manner consistent with the overall body layout, and the resulting skeleton preserves the dominant articulation pattern of the subject. This is a meaningful capability, as it suggests that the model can already move beyond holistic recognition and produce a structured geometric interpretation of the scene.

\paragraph{Remaining Challenge in Precision.}
The remaining errors are concentrated in the most delicate parts of the task, especially where limbs overlap, joints are partially hidden, or local evidence is inherently ambiguous. In such regions, predicted landmarks may drift toward nearby structures, reducing anatomical precision even when the global pose remains plausible. Much like the OCR case, the model appears stronger at recovering the correct structural form than at resolving every local detail with equal reliability, again pointing to a gap between high-level organization and fine-grained exactness.

\begin{figure}[!htbp]
    \centering
    \begin{subfigure}[t]{0.48\linewidth}
        \centering
        \includegraphics[width=\linewidth]{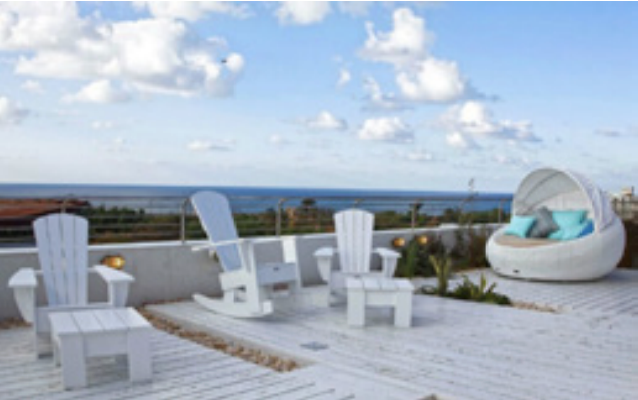}
        \caption{Input}
    \end{subfigure}\hfill
    \begin{subfigure}[t]{0.48\linewidth}
        \centering
        \includegraphics[width=\linewidth]{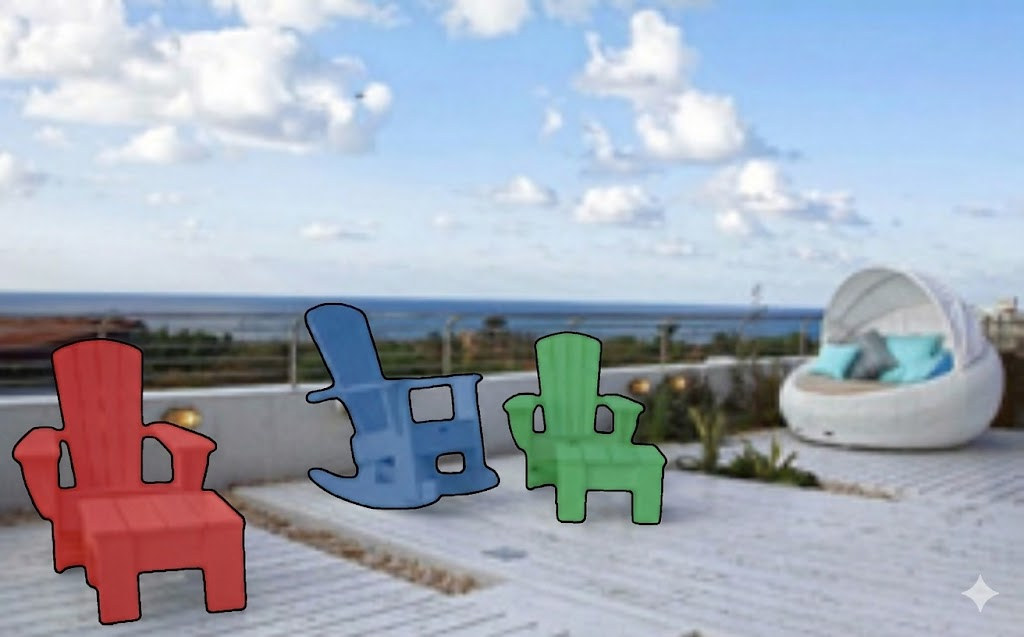}
        \caption{Model Output}
    \end{subfigure}
    \caption{\textbf{Segmentation.} The model separates major objects and semantic regions with broadly coherent masks, while boundary leakage and coarse delineation remain visible in cluttered areas. Generated by Nano Banana.}
    \label{fig:segment_case}
\end{figure}

\subsubsection{Case Study III: Semantic Segmentation}

\paragraph{Setup and Evaluation Focus.}
We then turn to semantic segmentation, which requires dense prediction across the entire image. Unlike sparse outputs such as keypoints or boxes, segmentation asks the model to assign coherent semantic labels over extended regions while preserving object boundaries. This makes it a useful test of whether the model can sustain structured region-level reasoning rather than relying primarily on a few salient cues.

\paragraph{Evidence of Global Competence.}
As illustrated in \Cref{fig:segment_case}, the model separates the principal semantic regions in a largely coherent manner. Foreground objects and dominant background areas are generally partitioned correctly, and the resulting masks preserve the broad scene composition. This suggests that the model already supports dense scene parsing at a meaningful level of abstraction.

\paragraph{Remaining Challenge in Precision.}
The main weakness appears near object contours and in cluttered regions, where boundaries become coarse and masks may leak across adjacent structures. Fine-grained delineation is therefore less stable than large-scale region assignment. In other words, semantic coherence is already reasonably strong, whereas geometric fidelity remains the more fragile component.
\begin{figure}[!htbp]
    \centering
    \begin{subfigure}[t]{0.48\linewidth}
        \centering
        \includegraphics[width=\linewidth]{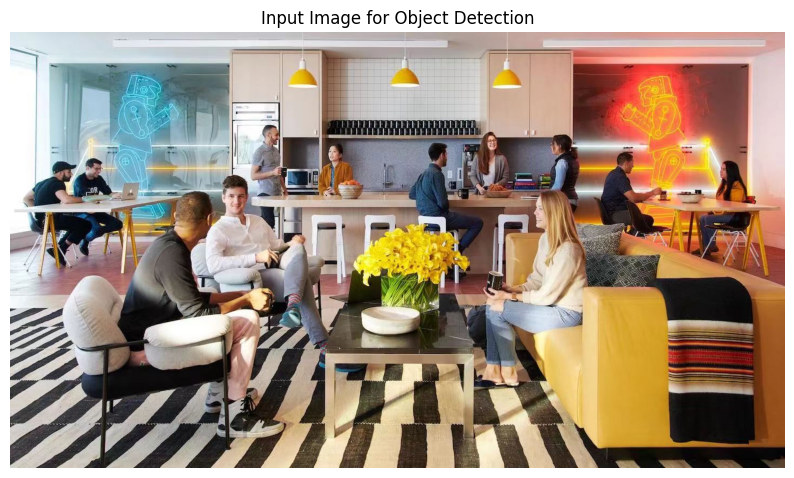}
        \caption{Input}
    \end{subfigure}\hfill
    \begin{subfigure}[t]{0.48\linewidth}
        \centering
        \includegraphics[width=\linewidth]{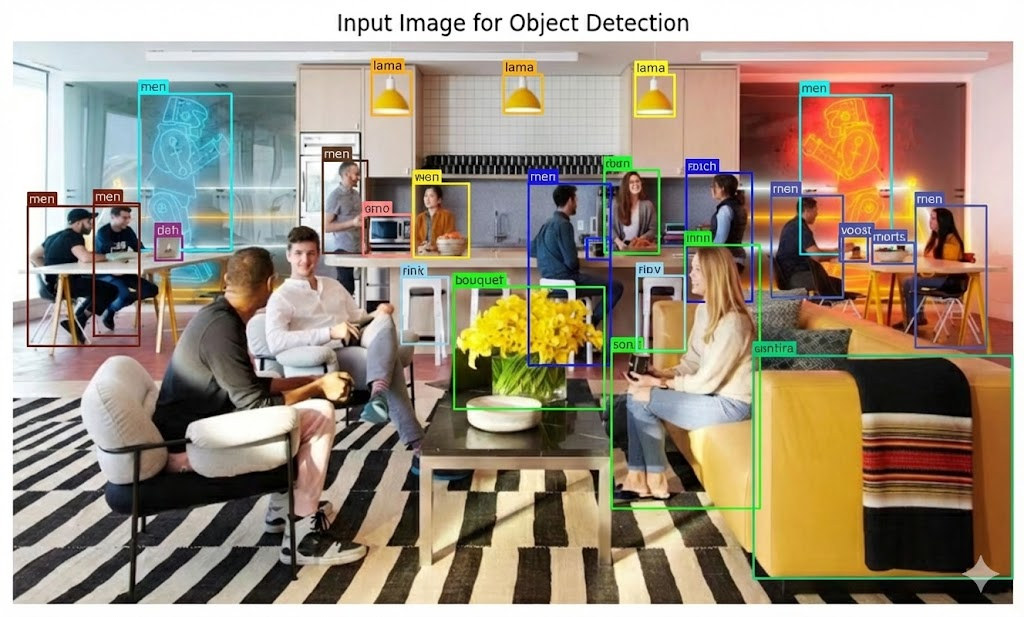}
        \caption{Model Output}
    \end{subfigure}
    \caption{\textbf{Detection.} The model localizes people and salient objects with broadly plausible bounding boxes, but still shows duplicate predictions and ambiguous assignments in crowded regions. Generated by Nano Banana.}
    \label{fig:detect_case}
\end{figure}

\subsubsection{Case Study IV: Object Detection}

\paragraph{Setup and Evaluation Focus.}
Finally, we examine object detection, where the model must identify distinct instances and localize them through explicit bounding boxes. Compared with segmentation, the task places less weight on pixel-level contours but introduces a different requirement: nearby objects must be separated into individually meaningful hypotheses with correct category assignments. It therefore offers a direct test of whether high-level scene understanding can be translated into instance-level structured prediction.

\paragraph{Evidence of Global Competence.}
The output in \Cref{fig:detect_case} shows that the model can identify many of the major entities in the scene and place boxes over the corresponding image regions. Human figures and salient objects are localized in a manner that is globally sensible, indicating that the model has a solid grasp of scene composition and object presence. This is already a nontrivial result, since the model must go beyond generic recognition and commit to explicit spatial decisions.

\paragraph{Remaining Challenge in Precision.}
The primary failure modes arise in crowded or visually entangled regions, where duplicate boxes, overlapping predictions, and ambiguous labels become more frequent. These errors suggest that the model is better at determining that an object should be present in a region than at resolving the exact extent and identity of each competing instance. Read together with the preceding cases, this result completes a consistent picture: across text, pose, region, and instance prediction, the model already exhibits credible structured visual understanding at the global level, yet its reliability still decreases when that understanding must be rendered with fine-grained, verifiable precision.

\subsection{Summary: Mapping Stress Tests to the Visual Intelligence Taxonomy}

\begin{table}[t]
\centering
\caption{\textbf{Mapping stress test dimensions to the visual intelligence taxonomy (\Cref{sec:evolution}).} Each dimension is designed to probe the boundary conditions of one or more levels, revealing where current frontier models sit within the five-level progression. The key finding column summarizes the dominant failure mode or capability gap observed.}
\label{tab:stress_test_level_map}
\scriptsize
\setlength{\tabcolsep}{3pt}
\renewcommand{\arraystretch}{1.15}
\resizebox{\textwidth}{!}{%
\begin{tabular}{@{}c l l l@{}}
\toprule[1pt]
\textbf{Dimension} & \textbf{Stress Test Focus} & \textbf{Primary Level(s)} & \textbf{Key Finding} \\ \midrule
I & Spatial Structuring \& Layout & L2 (Conditional Generation) & Semantic hallucination dominates over geometric reasoning \\ \midrule
II & Physical Reasoning \& Causality & L5 (World-Modeling Generation) & Emerging causal artifacts (bubbles, deformation) but no true physics \\ \midrule
III & Visual-Textual Integration & L4 (Agentic Generation) & VLM-first, renderer-second; fragile but productive reasoning \\ \midrule
IV & Multi-Turn Editing & L3 / L4 & Locally OK per turn, but Markovian chaining surfaces as cumulative drift \\ \midrule
V & Human-Centric Heredity \& Aesthetic Editing & L2 / L4 / L5 & Strong implicit human priors; room for deeper intent understanding \\ \midrule
VI & Low-level Vision Tasks & L1 / L2 & Prior-guided rewriting, not faithful signal recovery \\ \midrule
VII & Cross-Disciplinary Applications & L4 / L5 & Strong layout and world knowledge, but formal correctness remains task-dependent \\ \midrule
VIII & High-level Vision Tasks & L2 (Conditional Generation) & Global competence in structured prediction; precision degrades locally \\
\bottomrule[1pt]
\end{tabular}%
}
\end{table}
\section{Positions and Frontiers}
\label{sec:frontier}

The previous sections traced where visual generation \emph{is} today, level by level and pipeline by pipeline. This section turns to where the field is heading. Our organizing principle is the same five-level taxonomy of \Cref{sec:evolution}: each subsection picks an open frontier, names the unsolved problem at its center, and locates that problem within the L1--L5 progression. The intent is not exhaustive coverage but a small set of positions that, in our reading, will dominate the next two years of research on image generation and editing.

\subsection{Visual Chain-of-Thought (vCoT): Thinking Before Rendering}
\label{subsec:vcot}

The text-to-pixel mapping at the heart of every Level~1 generator is a single jump: it commits to a final image without any intermediate representation that the user, or the model itself, can inspect, debate, or revise. \emph{Visual Chain-of-Thought} (vCoT) attempts to break this jump into a sequence of intermediate states---textual analyses, layout sketches, regional decompositions, or coarse renderings---that the model produces before the final pixels are ever committed. The reason this matters mirrors why text CoT mattered for language models: complex visual outputs are easier to plan when the planning artefact is exposed to constraint checking, retrieval, or explicit revision. Several systems already realize restricted forms of vCoT---ReasonGen-R1~\citep{zhang2025reasongen} and T2I-R1~\citep{jiang2025t2i} let an autoregressive generator emit a reasoning trace before token-level synthesis, while X-Planner~\citep{Yeh2025BeyondSE} and MIRA~\citep{Zeng2025MIRAMI} translate user intent into stepwise edit sub-goals before any pixel changes. The frontier question is no longer whether reasoning helps, but how to design intermediate states that are simultaneously faithful to the eventual image and cheap enough to revise.

A first axis of design is the modality of the intermediate state. Pure-text reasoning is computationally cheap and inherits directly from the maturity of LMM-style chain-of-thought, but it is also lossy: many spatial relations and aesthetic judgements cannot be faithfully described in language without becoming verbose enough to be useless. ReasonEdit~\citep{Yin2025ReasonEditTR} and LegoEdit~\citep{Jia2025LegoEditAG} push toward typed symbolic edit programs that are richer than free-form text but still discrete; ImageEA~\citep{Hu2025ImageEA} goes further by translating natural language into executable visual operations whose effects can be verified step by step. A complementary line treats coarse visual sketches---layouts, depth proxies, or low-resolution drafts---as the intermediate representation: ReCon~\citep{ReCon} grounds compositional instructions through a visual reasoning trace, while CreatiLayout~\citep{CreatiLayout} and HybridLayout~\citep{wu2025hybrid} promote the layout itself to a first-class artefact that downstream rendering must respect, and Make Geometry Matter~\citep{zhang2026make} pushes this further by elevating explicit geometric structure as the planning substrate when 2D layouts cannot capture the spatial relations at stake. The trade-off mirrors the broader text-vs-image debate in multimodal reasoning: text traces are easier to learn and verify but information-poor; visual traces preserve geometry but are harder to score automatically.

Despite enthusiastic adoption, explicit vCoT does not always improve final image quality, and a healthy debate is emerging about when the additional latency is justified. Reasoning-augmented generation appears most useful in three regimes: (i)~when the prompt encodes hard compositional constraints---counts, spatial relations, attribute binding---that one-shot generators routinely fail, mirroring the Level-2 stress tests of \Cref{sec:stress_test}; (ii)~when the request requires external world knowledge that must be retrieved, verified, or combined, the case Gen-Searcher~\citep{feng2026gen} pushes to its logical extreme by training a search-augmented image-generation agent; and (iii)~when the user wants to inspect or steer the planning process itself, as in the agent-to-Lightroom protocol of JarvisArt~\citep{lin2025jarvisart}. Conversely, for purely aesthetic or familiar-distribution prompts, explicit reasoning often costs more than it saves and can amplify hallucinations in the trace, as the physics-exam case in \Cref{subsubsec:physics_solver_case} illustrates: the final solved image is correct, yet the trace repeatedly revisits the same sub-problems before converging.

The open challenges of vCoT therefore mirror those of text CoT but with sharper visual constraints. \emph{Faithfulness of the trace} is paramount---a sophisticated reasoning monologue that does not constrain the final pixels is worse than no reasoning, because it falsely advertises competence. \emph{Length control} matters too: many recent reasoning-augmented generators emit traces that revisit the same sub-problems repeatedly, suggesting that visual reasoning still lacks a stable internal workspace. Finally, \emph{verifiability} is largely unsolved for visual sketches: a textual reasoning step can be checked against a knowledge base, but a layout sketch can only be checked against the image it eventually produces, which makes debugging the planner difficult. Progress on vCoT will require richer intermediate representations that are simultaneously inspectable, executable, and grounded in the eventual visual output.

\subsection{Closed-Loop Visual Agents: Generation as Action}
\label{subsec:closed_loop_agents}

Across \Cref{sec:applications,subsec:agentic_visual_generation} we have seen visual generation increasingly embedded inside larger systems, but most of those systems still treat generation as an \emph{output}: the agent reasons, retrieves, plans, and only at the end commits to a final image. The next conceptual shift is to demote generation from output to \emph{action}---a primitive operation that an agent invokes mid-trajectory, observes, evaluates, and either commits or rolls back. Once generation is an action, the standard machinery of agent loops becomes available: planners decide when to generate, verifiers decide whether to keep the result, and policies learn over sequences of generate-and-revise steps rather than over single prompts. The intuition mirrors the move from one-shot LLM completion to ReAct-style tool-using assistants: the marginal gain comes not from a better generator, but from giving the model a closed loop in which to operate.

The most concrete instantiation of generation-as-action lives in the editing literature. In multi-turn editing, every operation---change the background, add an accessory, transfer a style---is an action whose precondition (current image) and post-condition (target image) are both visible to the agent. Frameworks such as X-Planner~\citep{Yeh2025BeyondSE}, MIRA~\citep{Zeng2025MIRAMI}, and ImageEA~\citep{Hu2025ImageEA} make this explicit by decomposing edit instructions into typed, executable steps. A complementary line studies \emph{transitions} themselves as a first-class action: VTG~\citep{yang2025vtg} learns to generate smooth, semantically coherent transitions between two given visual states, treating the intermediate motion as the unit of control rather than the final frame. Once transitions are addressable, an image agent can reason at the granularity of trajectories---how to morph, blend, or interpolate visual content under a high-level goal---instead of producing isolated end states.

A closed loop without verification quickly degenerates into one-shot generation in disguise. The defining capability of this family of agents is therefore not the generator, but the \emph{verifier} that decides whether an action's result is acceptable. GEMS~\citep{he2026gems} integrates a verifier and refiner inside the agent's persistent memory, allowing the system to back out of a faulty intermediate state without restarting from scratch. JarvisArt~\citep{lin2025jarvisart} exposes the same loop in a real creative application, where each Lightroom operation is reversible and inspectable before the next is committed. In embodied settings, CoT-VLA~\citep{zhao2025cotvla} and UniPi~\citep{du2023unipi} show that visual prediction can serve as a checked-out hypothesis space in which the agent simulates several plausible next states before committing to a physical action. The shared pattern is a clean separation between proposing (generation) and accepting (verification), with explicit fall-back state for the rejected path.

Even with explicit loops, three issues remain hard. \emph{Compounding error}: when an agent stitches many short generations into a long trajectory, small per-step inaccuracies accumulate, and identity drift or geometric inconsistency can become catastrophic---the same failure mode that current world models exhibit on long horizons (\Cref{subsec:world_models}) but with sharper visual signatures. \emph{Verifier reliability}: most current verifiers are themselves learned models, and a closed loop in which the verifier and the generator share a common bias is no better than open-loop generation. \emph{Latency and cost}: every additional generation in the loop multiplies inference cost, so practical systems must learn when to short-circuit a loop with a confident one-shot answer. Resolving these jointly is what separates a closed-loop visual agent from a slow open-loop generator with extra steps.

\subsection{Agentic Visual Generation: Tool-Augmented Rendering}
\label{subsec:agentic_visual_generation}

The recent rise of general-purpose agents in coding and computer-use domains points to a broader lesson for visual generation: capability is increasingly a property of the orchestration loop, not just of the underlying model. Systems such as Claude Code and Codex are powerful not because their next-token predictions are individually stronger, but because they plan, call tools, inspect external state, verify intermediate results, and iteratively refine outputs. A similar transition is now beginning in the visual domain. Instead of treating image generation as a one-shot mapping from prompt to pixels, an agentic visual system can decompose a request into retrieval, grounding, reasoning, and rendering substeps, invoking external tools whenever the answer depends on information that the base model should not be expected to memorise or hallucinate correctly.

We use the term \emph{Agentic Visual Generation} to denote this shift from passive rendering to \emph{tool-mediated visual production}. In this setting, generation is only one module inside a larger execution graph. For a single request, the agent may first query the web for real-time facts, retrieve structured data from databases, call OCR or document parsers, invoke charting or geometric engines for exact layout, use segmentation or tracking tools to localise edit regions, and finally hand the grounded plan to a visual renderer or editor. The output image is therefore not produced from latent priors alone; it is the result of a closed-loop pipeline that combines external state with learned visual priors.

Recent systems already instantiate different slices of this execution graph. GEMS~\citep{he2026gems} treats multimodal generation as an agent-native optimization loop with a planner, decomposer, verifier, refiner, persistent memory, and on-demand skill library, demonstrating that orchestration itself can unlock capability beyond the base generator. Gen-Searcher~\citep{feng2026gen} targets knowledge-intensive image requests by training a search-augmented agent that performs multi-hop web search and gathers textual evidence plus reference images before rendering, thereby grounding generation in external and potentially up-to-date state. JarvisArt~\citep{lin2025jarvisart} closes the loop on the editing side: rather than regenerating pixels from scratch, it maps multimodal user intent into executable Lightroom operations through a dedicated agent-to-software protocol, showing how visual agents can work \emph{inside} professional creative tools rather than outside them.

This tool-augmented setting is especially valuable for three failure regimes that recur throughout this work: \textbf{world knowledge}, \textbf{accuracy}, and \textbf{real-time information}. World knowledge matters when a model must generate historically correct city plans, scientifically faithful diagrams, or domain-specific dashboards. Accuracy matters when small symbolic mistakes---incorrect counts, wrong labels, misplaced overlays---destroy practical usefulness even if the image looks realistic. Real-time information matters when the requested output depends on current weather, sports standings, financial signals, software interfaces, or breaking events that cannot be reliably encoded in static model weights. In all of these cases, tool calls fundamentally change the problem formulation: instead of asking the visual model to \emph{know everything}, the agent only needs to know \emph{how to acquire, verify, and visualise} the right information.

The deeper implication is that future progress may depend less on scaling monolithic generators alone and more on building robust orchestration layers around them. The hard problem is no longer only how to render sharper pixels, but how to decide when retrieval is necessary, which tools are trustworthy, how to fuse tool outputs with visual priors, and how to verify the final image before returning it. Agentic Visual Generation is therefore not simply ``image generation with tool use.'' It is a transition from \emph{prompt-conditioned synthesis} to \emph{goal-conditioned visual task execution}. In the short term, external tools act as scaffolds that compensate for the model's missing factual, geometric, and temporal grounding; in the longer term, they may also shape the training targets for future unified systems that internalize part of this workflow. This makes Agentic Visual Generation a pragmatic bridge between today's strong but brittle renderers and tomorrow's truly grounded visual agents.

\subsection{Training with Synthetic Data and Visual Self-Play}
\label{subsec:synthetic_self_play}

A long-running thread of this work is data. \Cref{sec:resource_and_infra} documented how the field shifted from indiscriminate web scrape toward actively engineered corpora, with \Cref{sec:paradigm} tracing the rise of frontier-model distillation as the dominant data-construction recipe. Looking forward, the more provocative question is whether visual generation will increasingly close the loop and train on its own outputs---moving the field, as one early position paper put it, into the dawn of a synthetic era~\citep{yang2023aigs}. The motivation is structural rather than ideological: high-quality real images are bounded by what people choose to photograph and label, while synthetic pipelines can be steered toward exactly the failure modes the current generation of models reveals---the spatial, physical, and identity stress tests of \Cref{sec:stress_test}. The practical question is no longer whether to use synthetic data, but how to keep its distribution from collapsing under self-training.

Synthetic data in this roadmap spans three regimes whose boundaries are increasingly blurred. \emph{Frontier-model distillation} (\Cref{sec:paradigm,sec:generation}) uses a stronger generator to produce targets for a weaker one and remains the most reliable recipe when a frontier system genuinely outperforms the student. \emph{Reward-model-guided generation} pairs a generator with learned preference or correctness signals---DanceGRPO~\citep{xue2025dancegrpo}, RewardDance~\citep{wu2025rewarddance}, and EditReward~\citep{wu2025editreward} are recent examples---turning data construction into an RL-style search rather than a static teacher--student transfer. \emph{Visual self-play}, the most speculative of the three, lets the same generator both produce and critique, with retraining loops driven by self-rewarded preferences or external verifiers. Each regime trades a different axis of cost: distillation is expensive in API calls, reward-model RL is expensive in reward-model training, and self-play is expensive in the engineering required to avoid distribution collapse.

A useful complement to this high-level discussion is the concrete demand for \emph{domain-specific} synthesis where web data is sparse, culturally specific, or structurally constrained. Cross-domain transfer pipelines such as PaCaNet~\citep{PaCaNet} bridge stylistically distant Chinese painting and calligraphy through CycleGAN with transfer learning, illustrating that a small amount of paired data plus a bridging objective can manufacture training corpora for low-resource artistic domains long before diffusion-era distillation matured. More recent work in artistic character generation, exemplified by ReChar~\citep{Rechar}, focuses on \emph{structure-preserving} synthesis: each generated artefact must keep glyph integrity intact while admitting user-specified aesthetic enhancements, producing a constraint-aware synthetic-data pipeline that doubles as a benchmark for downstream training. These domain-targeted recipes are not just applications: they sketch what the next generation of synthetic-data engines must support natively---structurally constrained generation under sparse real supervision.

The frontier of this thread is visual self-play in the strict sense: a generator improved on its own filtered outputs across multiple rounds. In language modeling, similar loops have produced both impressive gains (when filtering is strict and external signals are used) and silent collapses (when the filter is itself the generator). The image-generation analogue is harder still, because visual quality is multi-dimensional and difficult to score with a single number. Promising signals come from preference-based fine-tuning---Diffusion-DPO~\citep{wallace2024diffusion}, dense-reward DPO~\citep{yang2024dense}, and video-DPO extensions~\citep{liu2025videodpo}---where a generator improves on top of its own samples filtered by human or VLM-derived preferences, but these methods still depend on an external preference signal rather than full self-play. Closed-loop self-play with no external grounding remains the open frontier of synthetic-data training, and the principal risk to manage is the one already named in the earliest synthetic-era position papers~\citep{yang2023aigs}: distribution narrowing toward whatever subset of the data manifold the generator and its self-judge already over-rate. As distillation, reward modeling, and self-play converge, the training pipeline of a frontier visual model is increasingly indistinguishable from a multi-agent system in which generators, judges, and curators continuously co-train.

\subsection{Visual Generation as World Simulation}
\label{subsec:world_models}

A recurring theme throughout this work is that visual generation is evolving from \emph{appearance synthesis} toward \emph{world modeling}: instead of producing a single static image or a passively unrolled video, the generative system must predict how a scene would evolve \emph{under intervention}. This ambition connects directly to Level~5 of our taxonomy (\Cref{subsec:level5}), where causal simulation rather than visual plausibility becomes the defining capability. We acknowledge upfront that the most vivid demonstrations in this area are video world models, which sit on the boundary of the roadmap's image-oriented scope; we include them here because they directly shape what the next generation of image generators must internalize. In this subsection, we examine the emerging frontier at which visual generators are repurposed---or purpose-built---as interactive world simulators, and discuss what this transition implies for the broader trajectory of visual intelligence.

\subsubsection{From Video Generation to Interactive Simulation}

The conceptual foundation for learning world models from visual observation dates back to the work of Ha and Schmidhuber~\citep{ha2018worldmodels}, who trained a compact latent model to predict environment dynamics and showed that policies could be learned entirely inside the model's ``dream.'' The Dreamer line of work~\citep{hafner2025dreamerv3} scaled this idea to diverse Atari and continuous-control domains, establishing that latent world models can support effective reinforcement learning with limited real-world interaction. LeCun's JEPA framework~\citep{lecun2022jepa} further articulated the theoretical case for predictive world models as the backbone of autonomous intelligence, arguing that learning to predict masked or future states in latent space is more scalable and robust than pixel-level generation.

What has changed in the past two years is the \emph{visual fidelity} and \emph{interaction bandwidth} of these models. Modern video generation backbones---DiT-based diffusion models, autoregressive transformers, and their hybrids---can now produce temporally coherent, photorealistic video at high resolution and long horizons. The natural question becomes: can these powerful visual generators be turned into \emph{playable} environments, where a user or an agent supplies actions at each step and the model renders the resulting next state in real time?

\subsubsection{Neural Game Engines and Playable Worlds}

The most vivid demonstrations of this idea come from neural game engines---systems that replace hand-coded simulators with learned generative models.

Genie~\citep{bruce2024genie} introduced a foundation world model trained on large-scale Internet video without action labels. By learning a latent action space from unlabeled video, Genie enables a user to ``play'' inside the generated environment by selecting from discovered actions, effectively creating an interactive 2D platformer from passive observation data alone. Genie~2~\citep{parker2024genie2} extended this paradigm to photorealistic 3D environments, supporting persistent memory, consistent object behavior across long horizons, and meaningful agent--world interaction at substantially higher visual fidelity than prior playable world models.

GameNGen~\citep{valevski2024gamenngen} demonstrated that a diffusion model can serve as a real-time neural game engine for the classic first-person shooter DOOM, achieving interactive frame rates while maintaining visual quality comparable to the original engine. The key insight is that once a generative model has internalized the game's dynamics, it can roll out new frames conditioned on player actions without access to any explicit game logic. DIAMOND~\citep{alonso2024diamond} pushed this direction further by showing that diffusion-based world models can match or exceed the visual fidelity of discrete-token approaches in Atari, while simultaneously serving as the environment for reinforcement learning agents.

Oasis~\citep{oasis2024} scaled the interactive world model to the domain of Minecraft, demonstrating that an autoregressive transformer can generate an open-world, first-person 3D environment in real time from keyboard and mouse inputs alone. GameGen-X~\citep{che2025gamegenx} extended interactive generation to open-world game video with support for diverse genres, multi-character scenarios, and fine-grained control over game events. Together, these systems establish a striking result: for environments with well-defined visual regularities, learned generative models can already replace hand-coded rendering and physics pipelines.

\subsubsection{World Models for Embodied and Autonomous Agents}

The same principle extends naturally to embodied settings, where visual world models serve not as entertainment engines but as simulation substrates for robot learning and autonomous driving. This line of work is closely related to the embodied applications discussed in \Cref{subsec:level5} and to the application section in \Cref{sec:applications}, but here we focus specifically on the \emph{simulation} role of world models---their capacity to stand in for the real environment and support policy optimization through imagined interaction.

In the robotics domain, UniSim~\citep{yang2024unisim} showed that a single generative model can serve as a universal simulator for visual experience, supporting both object manipulation and navigation through action-conditioned video generation. Cosmos Predict~2.5~\citep{CosmosPredict} and the broader Cosmos platform represent the industry push toward foundation-scale world models pretrained on internet-scale video and fine-tuned for embodied simulation, offering a reusable backbone for diverse downstream tasks including autonomous driving and robotic manipulation. In autonomous driving, GAIA-1~\citep{hu2024gaia1} demonstrated a generative world model that conditions on vehicle actions, text, and video to produce realistic driving scenarios, enabling controllable simulation of novel situations without real-world data collection.

A critical insight from this body of work, extensively analyzed in the companion world-model work~\citep{meietal2026videoroboticsurvey}, is that the value of a visual world model lies not in perceptual quality alone but in \emph{action faithfulness}: whether the generated future correctly reflects the causal consequences of the agent's intervention. A visually stunning rollout that ignores the commanded steering angle or misrepresents contact dynamics is worse than useless for policy learning---it actively corrupts the training signal. This requirement separates world simulation from video generation in a fundamental way: the former demands causal alignment, the latter only requires distributional plausibility. This distinction is increasingly being operationalized into evaluation: \textbf{WorldReasonBench}~\citep{wu2026worldreasonbench} reframes video-generation assessment as \emph{world-state prediction}---given an initial state and an action, it tests whether a model can generate a future whose evolution remains physically, causally, socially, and informationally consistent---and across modern video generators exposes a persistent gap between visual plausibility and genuine world reasoning, with a companion preference benchmark (WorldRewardBench) supporting reward modeling along the same reasoning axes.

\subsubsection{Open Challenges: From Rendering to Reasoning}

Despite the impressive demonstrations, the gap between current interactive world models and reliable world simulators remains substantial. We identify four principal challenges.

\paragraph{Long-horizon consistency.} Current models suffer from compounding errors over extended rollouts. Visual artefacts accumulate, object identities drift, and spatial layouts degrade as the generation horizon grows. Unlike traditional game engines or physics simulators that maintain explicit state, learned world models must reconstruct state implicitly from generated frames, making them vulnerable to hallucination and distributional drift.

\paragraph{Physical plausibility.} Visual world models learn correlations, not physical laws. They can approximate gravity, collision, and inertia in domains where training data is abundant, but they generalize poorly to novel physical scenarios. A model trained on platformer videos may fail on an unusual spring mechanism; a driving world model may produce implausible vehicle dynamics under extreme maneuvers. Bridging this gap likely requires integrating learned visual priors with structured physical reasoning---a direction that connects to the physics-aware generation challenges discussed in \Cref{sec:stress_test}.

\paragraph{Real-time performance.} Interactive simulation demands low-latency generation. While autoregressive models can approach real-time speeds, diffusion-based approaches typically require multiple denoising steps per frame, creating a tension between visual quality and interaction speed. The distillation and acceleration techniques discussed in \Cref{sec:method,sec:train} are directly relevant here, and further progress on few-step generation may be essential for practical deployment.

\paragraph{Compositional generalization.} Training a world model on one domain (e.g., a specific game or robot environment) does not automatically transfer to others. The most ambitious vision---a universal world simulator that can model arbitrary physical and social environments---requires compositional generalization far beyond what current models achieve. Foundation-scale approaches like Genie~2 and Cosmos represent steps in this direction, but the degree to which scale alone can solve the compositional challenge remains an open question.

\subsubsection{Implications for Visual Intelligence}

The transition from passive rendering to interactive world simulation marks a qualitative shift in what visual generation means. In a world model, an image is no longer the \emph{output}; it is a \emph{state} in a dynamic system. Generation is no longer a one-shot function call; it is one step in a closed-loop process where actions produce consequences and consequences inform future actions. This reframing dissolves the traditional boundary between ``generation'' and ``understanding'': a model that can simulate how a scene evolves under intervention necessarily possesses a form of visual understanding that goes beyond pattern matching.

From the standpoint of this roadmap's five-level taxonomy, world simulation sits at the apex because it subsumes lower levels as special cases. A world model that can faithfully simulate physical dynamics can also render controllable images (Level~2), maintain narrative consistency (Level~3), and operate in closed-loop interaction (Level~4). Whether the field will converge on monolithic world models that learn everything end-to-end, or on modular systems that compose learned visual generators with explicit physics engines and symbolic planners, remains the defining architectural question for the next stage of visual intelligence.

\subsection{Data-Centric Visual Intelligence}
\label{subsec:data_centric}

The visual generation systems discussed above are predominantly conditioned on text prompts or reference images. Conversely, a large and growing class of real-world applications requires \emph{understanding and visualizing} structured data: semi-structured tables with complex layouts, multi-page documents interleaving text, charts, and images, or analytical queries over relational databases and free-form text. As visual systems advance toward Agentic Generation (L4) and beyond, they will increasingly operate in closed-loop settings where an agent must read heterogeneous data, reason over its structure, and render faithful visual outputs such as charts, infographics, or data-driven documents for downstream consumption. This demands a capability that current generation pipelines largely lack: parsing and reasoning over structured input before generation.

Recent work from the data management community has made concrete progress on this front. \textbf{ST-Raptor}~\citep{STRaptor} introduces the Hierarchical Orthogonal Tree (HO-Tree) for question answering over semi-structured tables with merged cells, nested headers, and containment relationships, combining vision-language layout understanding with tree-based multi-step reasoning and improving answer accuracy by up to 20\% over prior methods. \textbf{MoDora}~\citep{MoDora} addresses the complementary problem at the document level, proposing a Component-Correlation Tree (CCTree) that organises pages containing interleaved text, tables, charts, and images, and supports hybrid retrieval combining LLM-guided navigation with embedding-based search, yielding accuracy gains of up to 61\%. \textbf{FDABench}~\citep{FDABench} scales to the multi-source setting, benchmarking data agents that must reason jointly over images, relational databases, and free-form text to answer a single analytical query, underscoring the difficulty of coordinating heterogeneous modalities within a unified pipeline.

A parallel line of work directly bridges structured-data understanding and visualization synthesis. \textbf{NL2VIS}~\citep{NL2VIS} formalizes the translation of natural-language queries over structured data into chart and graph specifications, connecting the well-studied NL2SQL pipeline to visual output. \textbf{DataVisT5}~\citep{DataVisT5} further trains a unified model that jointly understands text and data visualizations, supporting bidirectional tasks such as chart question answering and natural-language-to-visualization generation within a single architecture. These efforts echo the unified understanding-and-generation trend discussed throughout this work and demonstrate that faithful visualization demands deep comprehension of both data semantics and visual encoding conventions.

Together, these works point toward a data-centric vision of visual intelligence in which generation is grounded in real-world structured data rather than curated prompts alone. Tighter integration between data management and visual generation remains a largely unexplored yet promising frontier, particularly as visual agents (\Cref{subsec:closed_loop_agents,subsec:agentic_visual_generation}) begin to treat databases, spreadsheets, and document corpora as first-class inputs to the rendering loop.

\subsection{Evaluating New Generative Capabilities}
\label{subsec:eval_frontier}

A practical consequence of the new capabilities discussed above is that current evaluation protocols are no longer adequate. Frontier systems such as Nano Banana and GPT-Image now succeed routinely on tasks that earlier benchmarks did not anticipate---rendering scientifically faithful flow diagrams, drawing chemical structures with correct bond geometry, executing dense multi-line typography in multiple scripts, and producing tutor-style annotated worked solutions on top of an existing image (the physics-exam case in \Cref{subsubsec:physics_solver_case}). FID, CLIP-Score, and isolated prompt-following metrics tell us nothing about whether the molecular structure is correct, whether the flowchart's edges respect the underlying graph, or whether the rendered Chinese characters are stroke-faithful. \Cref{sec:resource_and_infra} already documents the move from heuristic perceptual scores to VLM-as-a-judge protocols; the next step is to define benchmarks that directly probe \emph{structural correctness} for the capability classes that matter.

Three concrete directions are emerging. First, \emph{symbolic-graph correctness} for diagrammatic outputs: a generated flowchart should be evaluated against the directed graph it claims to represent (nodes, edges, labels, decision branches), not against a perceptual similarity to a reference rendering. Second, \emph{domain-grounded factuality} for scientific and technical imagery: a chemistry diagram must be parsed back into a SMILES string and matched against the prompt's ground truth; an architectural floor plan must satisfy the geometric constraints implied by the requested layout. Third, \emph{glyph-level rendering audits} for typography: rendered text should be OCR'd and exact-matched against the prompt, with edit-distance scoring for partial credit and stroke-level analysis for non-Latin scripts. Each of these directions implies a different evaluator stack---a graph parser, a symbolic chemistry validator, an OCR + tokenizer pipeline---rather than a single similarity metric. Building these evaluators is itself a research agenda, and we expect it to be the bottleneck that forces visual benchmarks to mature in step with the capabilities of \Cref{sec:method,sec:applications}.

This evaluation gap is not isolated to image generation. World-simulation models (\Cref{subsec:world_models}) face the same problem on the temporal axis: how does one verify that a rolled-out physical trajectory is causally faithful, not merely visually plausible? The convergence point is that visual benchmarks of the next generation will need to look more like compilers and theorem provers than like image-similarity metrics---a shift that mirrors how NLP evaluation moved from BLEU toward execution-based and grounded metrics. Until that shift completes, headline progress on FID and CLIP-Score will continue to overstate the field's true competence on the higher rungs of \Cref{tab:5levels}.

\section{Conclusion}\label{sec:conclusion}

This work has argued that visual generation is moving beyond the goal of producing visually plausible images toward a broader problem of \emph{visual intelligence}. The central question is no longer only how realistic an output looks, but what capabilities a model acquires as it becomes controllable, persistent, interactive, and eventually world-aware. To make this transition explicit, we proposed a five-level taxonomy---\emph{Atomic Generation}, \emph{Conditional Generation}, \emph{In-Context Generation}, \emph{Agentic Generation}, and \emph{World-Modeling Generation}---that organizes progress from one-shot rendering to causal simulation under intervention.

We used this taxonomy as the organizing thread for the paper. First, we reviewed the modeling foundations and architectural components of modern visual generators, from GANs, diffusion, flow matching, and autoregressive models to hybrid and unified multimodal systems. Second, we analyzed the full training lifecycle, emphasizing that recent progress is shaped not only by model scale but also by data density, VLM-driven relabeling, continued training, SFT, preference-based alignment, reward modeling, and inference acceleration. Third, we examined the data and infrastructure layer, showing how the field has shifted from passive web scraping toward actively engineered and synthetic data pipelines. Fourth, we organized applications from conditional generation and domain adaptation to reasoning-driven editing and embodied visual prediction.

Our main contribution is therefore not a catalogue of methods, but a capability-centered framework for reading a rapidly expanding field. The five-level taxonomy clarifies what is being improved; the method, training, data, and application sections explain how those improvements are being produced; and the stress tests expose where frontier systems still fall short. Across spatial puzzles, physical reasoning, multi-turn editing, structured diagrams, coding interfaces, and embodied scenarios, we observed the same pattern: current models are increasingly strong at semantic and visual plausibility, but remain fragile when correctness depends on exact structure, persistent state, verification, or causal grounding.

This positioning also defines the next agenda. Visual generation will need stronger intermediate reasoning, closed-loop agents, tool-augmented rendering, synthetic-data feedback loops, and evaluation protocols that test structural and causal correctness rather than perceptual quality alone. In this sense, the path from atomic mapping to playable world models is not simply a scaling path; it is a shift from image synthesis to verifiable visual reasoning. We hope this roadmap provides a common language for that shift and a practical framework for locating future work along the trajectory from powerful generators to genuinely grounded visual intelligence.

\newpage
\bibliographystyle{assets/plainnat}
\bibliography{citation}

@inproceedings{NL2VIS,
    author    = {Luo, Yuyu and Tang, Nan and Li, Guoliang and Chai, Chengliang and Li, Wenbo and Qin, Xuedi},
    title     = {Synthesizing Natural Language to Visualization ({NL2VIS}) Benchmarks from {NL2SQL} Benchmarks},
    booktitle = {{SIGMOD} '21: International Conference on Management of Data},
    pages     = {1235--1247},
    publisher = {{ACM}},
    year      = {2021},
}

@inproceedings{DataVisT5,
    author    = {Wan, Zhuoyue and Song, Yuanfeng and Li, Shuaimin and Zhang, Chen Jason and Wong, Raymond Chi-Wing},
    title     = {{DataVisT5}: A Pre-Trained Language Model for Jointly Understanding Text and Data Visualization},
    booktitle = {41st {IEEE} International Conference on Data Engineering, {ICDE} 2025},
    pages     = {1704--1717},
    publisher = {{IEEE}},
    year      = {2025},
}

@inproceedings{zhang2017stackgan,
  title={Stackgan: Text to photo-realistic image synthesis with stacked generative adversarial networks},
  author={Zhang, Han and Xu, Tao and Li, Hongsheng and Zhang, Shaoting and Wang, Xiaogang and Huang, Xiaolei and Metaxas, Dimitris N},
  booktitle={Proceedings of the IEEE international conference on computer vision},
  pages={5907--5915},
  year={2017}
}

@inproceedings{ramesh2021zero,
  title={Zero-shot text-to-image generation},
  author={Ramesh, Aditya and Pavlov, Mikhail and Goh, Gabriel and Gray, Scott and Voss, Chelsea and Radford, Alec and Chen, Mark and Sutskever, Ilya},
  booktitle={International conference on machine learning},
  pages={8821--8831},
  year={2021},
  organization={Pmlr}
}

@article{FDABench,
    author    = {Wang, Ziting and Zhang, Shize and Yuan, Haitao and Zhu, Jinwei and Li, Shifu and Dong, Wei and Cong, Gao},
    title     = {{FDABench}: A Benchmark for Data Agents on Analytical Queries over Heterogeneous Data},
    journal   = {arXiv preprint arXiv:2509.02473},
    year      = {2025},
}

@inproceedings{STRaptor,
    author    = {Tang, Zirui and Niu, Boyu and Zhou, Xuanhe and Li, Boxiu and Zhou, Wei and Wang, Jiannan and Li, Guoliang and Zhang, Xinyi},
    title     = {{ST-Raptor}: {LLM}-Powered Semi-Structured Table Question Answering},
    booktitle = {Proceedings of the ACM SIGMOD International Conference on Management of Data},
    year      = {2026},
}

@article{peng2024dreambench++,
  title={Dreambench++: A human-aligned benchmark for personalized image generation},
  author={Peng, Yuang and Cui, Yuxin and Tang, Haomiao and Qi, Zekun and Dong, Runpei and Bai, Jing and Han, Chunrui and Ge, Zheng and Zhang, Xiangyu and Xia, Shu-Tao},
  journal={arXiv preprint arXiv:2406.16855},
  year={2024}
}

@inproceedings{mou2025dreamo,
  title={Dreamo: A unified framework for image customization},
  author={Mou, Chong and Wu, Yanze and Wu, Wenxu and Guo, Zinan and Zhang, Pengze and Cheng, Yufeng and Luo, Yiming and Ding, Fei and Zhang, Shiwen and Li, Xinghui and others},
  booktitle={Proceedings of the SIGGRAPH Asia 2025 Conference Papers},
  pages={1--12},
  year={2025}
}

@inproceedings{MoDora,
    author    = {Xu, Bangrui and Yao, Qihang and Tang, Zirui and Zhou, Xuanhe and He, Yeye and Yu, Shihan and Xu, Qianqian and Wang, Bin and Li, Guoliang},
    title     = {{MoDora}: Tree-Based Semi-Structured Document Analysis System},
    booktitle = {Proceedings of the ACM SIGMOD International Conference on Management of Data},
    year      = {2026},
}

@InProceedings{OminiControl,
    author    = {Tan, Zhenxiong and Liu, Songhua and Yang, Xingyi and Xue, Qiaochu and Wang, Xinchao},
    title     = {OminiControl: Minimal and Universal Control for Diffusion Transformer},
    booktitle = {Proceedings of the IEEE/CVF International Conference on Computer Vision (ICCV)},
    month     = {October},
    year      = {2025},
    pages     = {14940-14950}
}

@misc{Chimera,
      title={Chimera: Compositional Image Generation using Part-based Concepting}, 
      author={Shivam Singh and Yiming Chen and Agneet Chatterjee and Amit Raj and James Hays and Yezhou Yang and Chitta Baral},
      year={2025},
      eprint={2510.18083},
      archivePrefix={arXiv},
      primaryClass={cs.CV},
      url={https://arxiv.org/abs/2510.18083}, 
}

@InProceedings{FROSS,
    author    = {Hou, Hao-Yu and Lee, Chun-Yi and Sonogashira, Motoharu and Kawanishi, Yasutomo},
    title     = {FROSS: Faster-Than-Real-Time Online 3D Semantic Scene Graph Generation from RGB-D Images},
    booktitle = {Proceedings of the IEEE/CVF International Conference on Computer Vision (ICCV)},
    month     = {October},
    year      = {2025},
    pages     = {28818-28827}
}

@InProceedings{AIComposer,
    author    = {Li, Haowen and Fan, Zhenfeng and Wen, Zhang and Zhu, Zhengzhou and Li, Yunjin},
    title     = {AIComposer: Any Style and Content Image Composition via Feature Integration},
    booktitle = {Proceedings of the IEEE/CVF International Conference on Computer Vision (ICCV)},
    month     = {October},
    year      = {2025},
    pages     = {16840-16850}
}

@inproceedings{Easycontrol,
  title={Easycontrol: Adding efficient and flexible control for diffusion transformer},
  author={Zhang, Yuxuan and Yuan, Yirui and Song, Yiren and Wang, Haofan and Liu, Jiaming},
  booktitle={Proceedings of the IEEE/CVF International Conference on Computer Vision},
  pages={19513--19524},
  year={2025}
}

@inproceedings{Mv-adapter,
  title={Mv-adapter: Multi-view consistent image generation made easy},
  author={Huang, Zehuan and Guo, Yuan-Chen and Wang, Haoran and Yi, Ran and Ma, Lizhuang and Cao, Yan-Pei and Sheng, Lu},
  booktitle={Proceedings of the IEEE/CVF International Conference on Computer Vision},
  pages={16377--16387},
  year={2025}
}

@misc{PRISM,
      title={PRISM: A Unified Framework for Photorealistic Reconstruction and Intrinsic Scene Modeling}, 
      author={Alara Dirik and Tuanfeng Wang and Duygu Ceylan and Stefanos Zafeiriou and Anna Frühstück},
      year={2025},
      eprint={2504.14219},
      archivePrefix={arXiv},
      primaryClass={cs.GR},
      url={https://arxiv.org/abs/2504.14219} 
}

@misc{MIGLoRA,
      title={Efficient Multi-Instance Generation with Janus-Pro-Dirven Prompt Parsing}, 
      author={Fan Qi and Yu Duan and Changsheng Xu},
      year={2025},
      eprint={2503.21069},
      archivePrefix={arXiv},
      primaryClass={cs.CV},
      url={https://arxiv.org/abs/2503.21069}, 
}

@inproceedings{controlnet++,
    author    = {Ming Li and Taojiannan Yang and Huafeng Kuang and Jie Wu and Zhaoning Wang and Xuefeng Xiao and Chen Chen},
    title     = {ControlNet++: Improving Conditional Controls with Efficient Consistency Feedback},
    booktitle = {European Conference on Computer Vision (ECCV)},
    year      = {2024},
}

@InProceedings{FlexGen,
    author    = {Xu, Xinli and Ge, Wenhang and Lin, Jiantao and Feng, Jiawei and Xu, Lie and Zhao, Hanfeng and Zhang, Shunsi and Chen, Ying-Cong},
    title     = {FlexGen: Flexible Multi-View Generation from Text and Image Inputs},
    booktitle = {Proceedings of the IEEE/CVF International Conference on Computer Vision (ICCV)},
    month     = {October},
    year      = {2025},
    pages     = {18714-18724}
}

@InProceedings{SpinMeRound,
    author    = {Galanakis, Stathis and Lattas, Alexandros and Moschoglou, Stylianos and Kainz, Bernhard and Zafeiriou, Stefanos},
    title     = {SpinMeRound: Consistent Multi-View Identity Generation Using Diffusion Models},
    booktitle = {Proceedings of the IEEE/CVF International Conference on Computer Vision (ICCV)},
    month     = {October},
    year      = {2025},
    pages     = {14346-14356}
}

@InProceedings{MOSAIC,
    author    = {Liu, Zhixuan and Zhu, Haokun and Chen, Rui and Francis, Jonathan and Hwang, Soonmin and Zhang, Ji and Oh, Jean},
    title     = {MOSAIC: Generating Consistent, Privacy-Preserving Scenes from Multiple Depth Views in Multi-Room Environments},
    booktitle = {Proceedings of the IEEE/CVF International Conference on Computer Vision (ICCV)},
    month     = {October},
    year      = {2025},
    pages     = {27456-27465}
}

@InProceedings{UniPortrait,
    author    = {He, Junjie and Geng, Yifeng and Bo, Liefeng},
    title     = {UniPortrait: A Unified Framework for Identity-Preserving Single- and Multi-Human Image Personalization},
    booktitle = {Proceedings of the IEEE/CVF International Conference on Computer Vision (ICCV)},
    month     = {October},
    year      = {2025},
    pages     = {14399-14408}
}

@inproceedings{ruiz2023dreambooth,
  title={Dreambooth: Fine tuning text-to-image diffusion models for subject-driven generation},
  author={Ruiz, Nataniel and Li, Yuanzhen and Jampani, Varun and Pritch, Yael and Rubinstein, Michael and Aberman, Kfir},
  booktitle={Proceedings of the IEEE/CVF conference on computer vision and pattern recognition},
  pages={22500--22510},
  year={2023}
}

@article{gal2022image,
  title={An image is worth one word: Personalizing text-to-image generation using textual inversion},
  author={Gal, Rinon and Alaluf, Yuval and Atzmon, Yuval and Patashnik, Or and Bermano, Amit H and Chechik, Gal and Cohen-Or, Daniel},
  journal={arXiv preprint arXiv:2208.01618},
  year={2022}
}

@article{hu2022lora,
  title={Lora: Low-rank adaptation of large language models},
  author={Hu, Edward J and Shen, Yelong and Wallis, Phillip and Allen-Zhu, Zeyuan and Li, Yuanzhi and Wang, Shean and Wang, Liang and Chen, Weizhu and others},
  booktitle={International Conference on Learning Representations},
  volume={1},
  number={2},
  pages={3},
  year={2022}
}

@article{gal2023encoder,
  title={Encoder-based domain tuning for fast personalization of text-to-image models},
  author={Gal, Rinon and Arar, Moab and Atzmon, Yuval and Bermano, Amit H and Chechik, Gal and Cohen-Or, Daniel},
  journal={ACM Transactions on Graphics (TOG)},
  volume={42},
  number={4},
  pages={1--13},
  year={2023},
  publisher={ACM New York, NY, USA}
}

@article{ye2023ip,
  title={Ip-adapter: Text compatible image prompt adapter for text-to-image diffusion models},
  author={Ye, Hu and Zhang, Jun and Liu, Sibo and Han, Xiao and Yang, Wei},
  journal={arXiv preprint arXiv:2308.06721},
  year={2023}
}

@article{wang2024instantid,
  title={Instantid: Zero-shot identity-preserving generation in seconds},
  author={Wang, Qixun and Bai, Xu and Wang, Haofan and Qin, Zekui and Chen, Anthony},
  journal={arXiv preprint arXiv:2401.07519},
  year={2024}
}

@article{guo2024pulid,
  title={Pulid: Pure and lightning id customization via contrastive alignment},
  author={Guo, Zinan and Wu, Yanze and Zhuowei, Chen and Zhang, Peng and He, Qian and others},
  journal={Advances in neural information processing systems},
  volume={37},
  pages={36777--36804},
  year={2024}
}

@inproceedings{shi2024instantbooth,
  title={Instantbooth: Personalized text-to-image generation without test-time finetuning},
  author={Shi, Jing and Xiong, Wei and Lin, Zhe and Jung, Hyun Joon},
  booktitle={Proceedings of the IEEE/CVF conference on computer vision and pattern recognition},
  pages={8543--8552},
  year={2024}
}

@inproceedings{li2024photomaker,
  title={Photomaker: Customizing realistic human photos via stacked id embedding},
  author={Li, Zhen and Cao, Mingdeng and Wang, Xintao and Qi, Zhongang and Cheng, Ming-Ming and Shan, Ying},
  booktitle={Proceedings of the IEEE/CVF conference on computer vision and pattern recognition},
  pages={8640--8650},
  year={2024}
}

@inproceedings{peng2024portraitbooth,
  title={Portraitbooth: A versatile portrait model for fast identity-preserved personalization},
  author={Peng, Xu and Zhu, Junwei and Jiang, Boyuan and Tai, Ying and Luo, Donghao and Zhang, Jiangning and Lin, Wei and Jin, Taisong and Wang, Chengjie and Ji, Rongrong},
  booktitle={Proceedings of the IEEE/CVF Conference on Computer Vision and Pattern Recognition},
  pages={27080--27090},
  year={2024}
}

@article{zhou2024storymaker,
  title={Storymaker: Towards holistic consistent characters in text-to-image generation},
  author={Zhou, Zhengguang and Li, Jing and Li, Huaxia and Chen, Nemo and Tang, Xu},
  journal={arXiv preprint arXiv:2409.12576},
  year={2024}
}

@article{xiao2025fastcomposer,
  title={Fastcomposer: Tuning-free multi-subject image generation with localized attention},
  author={Xiao, Guangxuan and Yin, Tianwei and Freeman, William T and Durand, Fr{\'e}do and Han, Song},
  journal={International Journal of Computer Vision},
  volume={133},
  number={3},
  pages={1175--1194},
  year={2025},
  publisher={Springer}
}

@inproceedings{nam2025visual,
  title={Visual persona: Foundation model for full-body human customization},
  author={Nam, Jisu and Son, Soowon and Xu, Zhan and Shi, Jing and Liu, Difan and Liu, Feng and Kim, Seungryong and Zhou, Yang},
  booktitle={Proceedings of the Computer Vision and Pattern Recognition Conference},
  pages={18630--18641},
  year={2025}
}

@article{Li2025ICCustomDI,
  title={IC-Custom: Diverse Image Customization via In-Context Learning},
  author={Yaowei Li and Xiaoyu Li and Zhaoyang Zhang and Yuxuan Bian and Gan Liu and Xinyuan Li and Jiale Xu and Wenbo Hu and Yating Liu and Lingen Li and Jing Cai and Yuexian Zou and Yancheng He and Ying Shan},
  journal={ArXiv},
  year={2025},
  volume={abs/2507.01926},
  url={https://api.semanticscholar.org/CorpusID:280150169}
}

@article{Chen2026PosterOmniGA,
  title={PosterOmni: Generalized Artistic Poster Creation via Task Distillation and Unified Reward Feedback},
  author={Sixiang Chen and Jianyu Lai and Jialin Gao and Hengyu Shi and Zhongying Liu and Tian Ye and Junfeng Luo and Xiaoming Wei and Lei Zhu},
  journal={arXiv preprint},
  year={2026}
}

@article{Liu2026PosterVerseAF,
  title={PosterVerse: A Full-Workflow Framework for Commercial-Grade Poster Generation with HTML-Based Scalable Typography},
  author={Junle Liu and Peirong Zhang and Yuyi Zhang and Pengyu Yan and Hui Zhou and Xinyue Zhou and Fengjun Guo and Lianwen Jin},
  journal={ArXiv},
  year={2026},
  volume={abs/2601.03993},
  url={https://api.semanticscholar.org/CorpusID:284532482}
}

@article{Zhang2025PosterGenAP,
  title={PosterGen: Aesthetic-Aware Paper-to-Poster Generation via Multi-Agent LLMs},
  author={Zhilin Zhang and Xiang Zhang and Jiaqi Wei and Yiwei Xu and Chenyu You},
  journal={ArXiv},
  year={2025},
  volume={abs/2508.17188},
  url={https://api.semanticscholar.org/CorpusID:280710654}
}

@article{zhang2025creatidesign,
  title={Creatidesign: A unified multi-conditional diffusion transformer for creative graphic design},
  author={Zhang, Hui and Hong, Dexiang and Yang, Maoke and Cheng, Yutao and Zhang, Zhao and Shao, Jie and Wu, Xinglong and Wu, Zuxuan and Jiang, Yu-Gang},
  journal={arXiv preprint arXiv:2505.19114},
  year={2025}
}

@article{Shi2025WordConWT,
  title={WordCon: Word-level Typography Control in Scene Text Rendering},
  author={Wenda Shi and Yiren Song and Zihan Rao and Dengming Zhang and Jiaming Liu and Xingxing Zou},
  journal={ArXiv},
  year={2025},
  volume={abs/2506.21276},
  url={https://api.semanticscholar.org/CorpusID:280010723}
}

@article{Chen2025PosterCraftRH,
  title={PosterCraft: Rethinking High-Quality Aesthetic Poster Generation in a Unified Framework},
  author={Sixiang Chen and Jianyu Lai and Jialin Gao and Tian Ye and Haoyu Chen and Hengyu Shi and Shitong Shao and Yunlong Lin and Song Fei and Zhaohu Xing and Yeying Jin and Junfeng Luo and Xiaoming Wei and Lei Zhu},
  journal={ArXiv},
  year={2025},
  volume={abs/2506.10741},
  url={https://api.semanticscholar.org/CorpusID:279318662}
}

@article{Ma2026UMTextAU,
  title={UM-Text: A Unified Multimodal Model for Image Understanding and Visual Text Editing},
  author={Lichen Ma and Xiaolong Fu and Gaojing Zhou and Zipeng Guo and Ting Zhu and Yichun Liu and Yu Shi and Jason Li and Junshi Huang},
  journal={arXiv preprint},
  year={2026}
}

@article{Ma2024CharGenHA,
  title={CharGen: High Accurate Character-Level Visual Text Generation Model with MultiModal Encoder},
  author={Lichen Ma and Tiezhu Yue and Pei Fu and Yujie Zhong and Kai Zhou and Xiaoming Wei and Jie Hu},
  journal={ArXiv},
  year={2024},
  volume={abs/2412.17225},
  url={https://api.semanticscholar.org/CorpusID:274981660}
}

@article{Wang2025UniGlyphUS,
  title={UniGlyph: Unified Segmentation-Conditioned Diffusion for Precise Visual Text Synthesis},
  author={Yuanrui Wang and Cong Han and Yafei Li and Zhipeng Jin and Xiawei Li and Sinan Du and Wen Tao and Yi Yang and Shuanglong Li and Chun Yuan and Liu Lin},
  journal={ArXiv},
  year={2025},
  volume={abs/2507.00992},
  url={https://api.semanticscholar.org/CorpusID:280069916}
}

@article{Jiang2025ControlTextUC,
  title={ControlText: Unlocking Controllable Fonts in Multilingual Text Rendering without Font Annotations},
  author={Bowen Jiang and Yuan Yuan and Xinyi Bai and Zhuoqun Hao and Alyson Yin and Yaojie Hu and Wenyu Liao and Lyle Ungar and Camillo Jose Taylor},
  journal={ArXiv},
  year={2025},
  volume={abs/2502.10999},
  url={https://api.semanticscholar.org/CorpusID:276408159}
}

@article{Chen2025POSTAAG,
  title={POSTA: A Go-to Framework for Customized Artistic Poster Generation},
  author={Haoyu Chen and Xiaojie Xu and Wenbo Li and Jingjing Ren and Tian Ye and Songhua Liu and Ying-Cong Chen and Lei Zhu and Xinchao Wang},
  journal={2025 IEEE/CVF Conference on Computer Vision and Pattern Recognition (CVPR)},
  year={2025},
  pages={28694-28704},
  url={https://api.semanticscholar.org/CorpusID:277113604}
}

@article{Lu2025EasyTextCD,
  title={EasyText: Controllable Diffusion Transformer for Multilingual Text Rendering},
  author={Runnan Lu and Yuxuan Zhang and Jailing Liu and Haifa Wang and Yiren Song},
  journal={ArXiv},
  year={2025},
  volume={abs/2505.24417},
  url={https://api.semanticscholar.org/CorpusID:279070661}
}

@misc{PhysicEdit,
      title={From Statics to Dynamics: Physics-Aware Image Editing with Latent Transition Priors}, 
      author={Liangbing Zhao and Le Zhuo and Sayak Paul and Hongsheng Li and Mohamed Elhoseiny},
      year={2026},
      eprint={2602.21778},
      archivePrefix={arXiv},
      primaryClass={cs.CV},
      url={https://arxiv.org/abs/2602.21778}, 
}

@article{Liu2024GlyphByT5v2AS,
  title={Glyph-ByT5-v2: A Strong Aesthetic Baseline for Accurate Multilingual Visual Text Rendering},
  author={Zeyu Liu and Weicong Liang and Yiming Zhao and Bohan Chen and Ji Li and Yuhui Yuan},
  journal={ArXiv},
  year={2024},
  volume={abs/2406.10208},
  url={https://api.semanticscholar.org/CorpusID:270521692}
}

@article{Luo2025BeyondFT,
  title={Beyond Flat Text: Dual Self-inherited Guidance for Visual Text Generation},
  author={Minxing Luo and Zixun Xia and Liaojun Chen and Zhenhang Li and Weichao Zeng and Jianye Wang and Wentao Cheng and Yaxing Wang and Yu ZHOU and Jian Yang},
  journal={ArXiv},
  year={2025},
  volume={abs/2501.05892},
  url={https://api.semanticscholar.org/CorpusID:275458598}
}

@misc{CreatiLayout,
      title={CreatiLayout: Siamese Multimodal Diffusion Transformer for Creative Layout-to-Image Generation}, 
      author={Hui Zhang and Dexiang Hong and Yitong Wang and Jie Shao and Xinglong Wu and Zuxuan Wu and Yu-Gang Jiang},
      year={2025},
      eprint={2412.03859},
      archivePrefix={arXiv},
      primaryClass={cs.CV},
      url={https://arxiv.org/abs/2412.03859}, 
}

@misc{ReCon,
      title={ReCon: Region-Controllable Data Augmentation with Rectification and Alignment for Object Detection}, 
      author={Haowei Zhu and Tianxiang Pan and Rui Qin and Jun-Hai Yong and Bin Wang},
      year={2025},
      eprint={2510.15783},
      archivePrefix={arXiv},
      primaryClass={cs.CV},
      url={https://arxiv.org/abs/2510.15783}, 
}

@ARTICLE{StyleShot,
  author={Gao, Junyao and Sun, Yanan and Liu, Yanchen and Tang, Yinhao and Zeng, Yanhong and Qi, Ding and Chen, Kai and Zhao, Cairong},
  journal={IEEE Transactions on Pattern Analysis and Machine Intelligence}, 
  title={StyleShot: A Snapshot on Any Style}, 
  year={2026},
  volume={48},
  number={2},
  pages={1215-1228},
  doi={10.1109/TPAMI.2025.3610614}}

@article{InstantCharacter,
  title={InstantCharacter: Personalize Any Characters with a Scalable Diffusion Transformer Framework},
  author={Jiale Tao and Yanbing Zhang and Qixun Wang and Yiji Cheng and Haofan Wang and Xu Bai and Zhengguang Zhou and Ruihuang Li and Linqing Wang and Chunyu Wang and Qin Lin and Qinglin Lu},
  journal={ArXiv},
  year={2025},
  volume={abs/2504.12395},
  url={https://api.semanticscholar.org/CorpusID:277856764}
}

@article{RichControl,
  title={RichControl: Structure- and Appearance-Rich Training-Free Spatial Control for Text-to-Image Generation},
  author={Liheng Zhang and Lexi Pang and Hang Ye and Xiaoxuan Ma and Yizhou Wang},
  journal={ArXiv},
  year={2025},
  volume={abs/2507.02792},
  url={https://api.semanticscholar.org/CorpusID:280047690}
}

@article{OmniRefiner,
  title={OmniRefiner: Reinforcement-Guided Local Diffusion Refinement},
  author={Yaoli Liu and Zi-Juan Ouyang and Shengtao Lou and Yiren Song},
  journal={ArXiv},
  year={2025},
  volume={abs/2511.19990},
  url={https://api.semanticscholar.org/CorpusID:283250744}
}

@article{Zhang2025InContextEE,
  title={In-Context Edit: Enabling Instructional Image Editing with In-Context Generation in Large Scale Diffusion Transformer},
  author={Zechuan Zhang and Ji Xie and Yu Lu and Zongxin Yang and Yi Yang},
  journal={ArXiv},
  year={2025},
  volume={abs/2504.20690},
  url={https://api.semanticscholar.org/CorpusID:278171476}
}

@article{Yin2025ReasonEditTR,
  title={ReasonEdit: Towards Reasoning-Enhanced Image Editing Models},
  author={Fukun Yin and Shiyu Liu and Yucheng Han and Zhibo Wang and Peng Xing and Rui Wang and Wei Cheng and Yingming Wang and Aojie Li and Zixi Yin and Pengtao Chen and Xiangyu Zhang and Daxin Jiang and Xianfang Zeng and Gang Yu},
  journal={ArXiv},
  year={2025},
  volume={abs/2511.22625},
  url={https://api.semanticscholar.org/CorpusID:283439275}
}

@article{Ghazanfari2025SpotEditEV,
  title={SpotEdit: Evaluating Visually-Guided Image Editing Methods},
  author={Sara Ghazanfari and Wei-An Lin and Haitong Tian and Ersin Yumer},
  journal={ArXiv},
  year={2025},
  volume={abs/2508.18159},
  url={https://api.semanticscholar.org/CorpusID:280710822}
}

@article{Jia2025LegoEditAG,
  title={Lego-Edit: A General Image Editing Framework with Model-Level Bricks and MLLM Builder},
  author={Qifei Jia and Yu Liu and Yajie Chai and Xintong Yao and Qiming Lu and Yasen Zhang and Runyu Shi and Ying Huang and Guoquan Zhang},
  journal={ArXiv},
  year={2025},
  volume={abs/2509.12883},
  url={https://api.semanticscholar.org/CorpusID:281325583}
}

@misc{PaCaNet,
      title={PaCaNet: A Study on CycleGAN with Transfer Learning for Diversifying Fused Chinese Painting and Calligraphy}, 
      author={Zuhao Yang and Huajun Bai and Zhang Luo and Yang Xu and Wei Pang and Yue Wang and Yisheng Yuan and Yingfang Yuan},
      year={2023},
      eprint={2301.13082},
      archivePrefix={arXiv},
      primaryClass={cs.CV},
      url={https://arxiv.org/abs/2301.13082}, 
}

@inproceedings{Rechar,
author = {Yang, Zhongyu and Song, Junhao and Luo, Zhang and Yang, Zuhao and Xu, Yang and Lan, Jingfen and Zhang, Yonghan and Pang, Wei and Song, Siyang and Yuan, Yingfang},
title = {ReChar: Revitalising Characters with Structure Preserved and User-Specified Aesthetic Enhancements},
year = {2025},
isbn = {9798400721366},
publisher = {Association for Computing Machinery},
address = {New York, NY, USA},
url = {https://doi.org/10.1145/3757376.3771409},
doi = {10.1145/3757376.3771409},
booktitle = {Proceedings of the SIGGRAPH Asia 2025 Technical Communications},
articleno = {30},
numpages = {5},
location = {
},
series = {SA Technical Communications '25}
}

@article{Yeh2025BeyondSE,
  title={Beyond Simple Edits: X-Planner for Complex Instruction-Based Image Editing},
  author={Chun-Hsiao Yeh and Yilin Wang and Nanxuan Zhao and Richard Zhang and Yuheng Li and Yi Ma and Krishna Kumar Singh},
  journal={ArXiv},
  year={2025},
  volume={abs/2507.05259},
  url={https://api.semanticscholar.org/CorpusID:280150428}
}

@article{Zeng2025MIRAMI,
  title={MIRA: Multimodal Iterative Reasoning Agent for Image Editing},
  author={Ziyun Zeng and Hang Hua and Jiebo Luo},
  journal={ArXiv},
  year={2025},
  volume={abs/2511.21087},
  url={https://api.semanticscholar.org/CorpusID:283262006}
}

@article{Shen2025IMAGHarmonyCI,
  title={IMAGHarmony: Controllable Image Editing with Consistent Object Quantity and Layout},
  author={Fei Shen and Xiaoyu Du and Yutong Gao and Jian Yu and Yushe Cao and Xing Lei and Jinhui Tang},
  journal={ArXiv},
  year={2025},
  volume={abs/2506.01949},
  url={https://api.semanticscholar.org/CorpusID:279119734}
}

@article{Yang2026ControllableLI,
  title={Controllable Layered Image Generation for Real-World Editing},
  author={Jinrui Yang and Qing Liu and Yijun Li and Mengwei Ren and Letian Zhang and Zhe Lin and Cihang Xie and Yuyin Zhou},
  journal={arXiv preprint},
  year={2026}
}

@article{Mao2025VisualAM,
  title={Visual Autoregressive Modeling for Instruction-Guided Image Editing},
  author={Qingyang Mao and Qi Cai and Yehao Li and Yingwei Pan and Mingyue Cheng and Ting Yao and Qi Liu and Tao Mei},
  journal={ArXiv},
  year={2025},
  volume={abs/2508.15772},
  url={https://api.semanticscholar.org/CorpusID:280700028}
}

@article{Liu2025Step1XEditAP,
  title={Step1X-Edit: A Practical Framework for General Image Editing},
  author={Shiyu Liu and Yucheng Han and Peng Xing and Fukun Yin and Rui Wang and Wei Cheng and Jiaqi Liao and Yingming Wang and Honghao Fu and Chunrui Han and Guopeng Li and Yuang Peng and Quan Sun and Jingwei Wu and Yan Cai and Zheng Ge and Ranchen Ming and Lei Xia and Xianfang Zeng and Yibo Zhu and Binxing Jiao and Xiangyu Zhang and Gang Yu and Daxin Jiang},
  journal={ArXiv},
  year={2025},
  volume={abs/2504.17761},
  url={https://api.semanticscholar.org/CorpusID:278033726}
}

@article{Ma2025X2EditRA,
  title={X2Edit: Revisiting Arbitrary-Instruction Image Editing through Self-Constructed Data and Task-Aware Representation Learning},
  author={Jiancang Ma and Xujie Zhu and Zihao Pan and Qirong Peng and Xu Guo and Chen Chen and H. Lu},
  journal={ArXiv},
  year={2025},
  volume={abs/2508.07607},
  url={https://api.semanticscholar.org/CorpusID:280567028}
}

@article{Hu2025ImageEA,
  title={Image Editing As Programs with Diffusion Models},
  author={Yujia Hu and Songhua Liu and Zhenxiong Tan and Xingyi Yang and Xinchao Wang},
  journal={ArXiv},
  year={2025},
  volume={abs/2506.04158},
  url={https://api.semanticscholar.org/CorpusID:279154460}
}

@article{Yao2025ImplementationFF,
  title={Implementation Framework for Instruction-Driven Image Editing},
  author={Chaosheng Yao and Lei Cui and Jinbo Zhang and Changcai Lu and Ziyang Zhang and Zhengyan Fan},
  journal={2025 2nd International Symposium on AI and Cybersecurity (ISAICS)},
  year={2025},
  pages={1-5},
  url={https://api.semanticscholar.org/CorpusID:285106947}
}

@article{Fang2025TBStarEditFI,
  title={TBStar-Edit: From Image Editing Pattern Shifting to Consistency Enhancement},
  author={Hao Fang and Zechao Zhan and Weixin Feng and Ziwei Huang and Xubin Li and Tiezheng Ge},
  journal={ArXiv},
  year={2025},
  volume={abs/2510.04483},
  url={https://api.semanticscholar.org/CorpusID:281842655}
}

@article{ho2020denoising,
  title={Denoising diffusion probabilistic models},
  author={Ho, Jonathan and Jain, Ajay and Abbeel, Pieter},
  journal={Advances in neural information processing systems},
  volume={33},
  pages={6840--6851},
  year={2020}
}

@article{black2023training,
  title={Training diffusion models with reinforcement learning},
  author={Black, Kevin and Janner, Michael and Du, Yilun and Kostrikov, Ilya and Levine, Sergey},
  journal={arXiv preprint arXiv:2305.13301},
  year={2023}
}

@article{fan2023dpok,
  title={DPOK: Reinforcement learning for fine-tuning text-to-image diffusion models},
  author={Fan, Ying and Watkins, Olivia and Du, Yuqing and Liu, Hao and Ryu, Moonkyung and Boutilier, Craig and Abbeel, Pieter and Ghavamzadeh, Mohammad and Lee, Kangwook and Lee, Kimin},
  journal={Advances in Neural Information Processing Systems},
  volume={36},
  pages={79858--79885},
  year={2023}
}

@article{prabhudesai2023alignprop,
  title={Aligning text-to-image diffusion models with reward backpropagation},
  author={Prabhudesai, Mihir and Goyal, Anirudh and Pathak, Deepak and Fragkiadaki, Katerina},
  journal={arXiv preprint arXiv:2310.03739},
  year={2023}
}

@inproceedings{wallace2024diffusion,
  title={Diffusion model alignment using direct preference optimization},
  author={Wallace, Bram and Dang, Meihua and Rafailov, Rafael and Zhou, Linqi and Lou, Aaron and Purushwalkam, Senthil and Ermon, Stefano and Xiong, Caiming and Joty, Shafiq and Naik, Nikhil},
  booktitle={Proceedings of the IEEE/CVF Conference on Computer Vision and Pattern Recognition},
  pages={8228--8238},
  year={2024}
}

@inproceedings{yang2024dense,
  title={A dense reward view on aligning text-to-image diffusion with preference},
  author={Yang, Shentao and Chen, Tianqi and Zhou, Mingyuan},
  booktitle={Proceedings of the 41st International Conference on Machine Learning},
  year={2024}
}

@inproceedings{wu2025hybrid,
  title={Hybrid layout control for diffusion transformer: Fewer annotations, superior aesthetics},
  author={Wu, Keming and Chen, Junwen and Liang, Zhanhao and Wang, Yinuo and Li, Ji and Zhang, Chao and Wang, Bin and Yuan, Yuhui},
  booktitle={Proceedings of the IEEE/CVF International Conference on Computer Vision},
  pages={17930--17940},
  year={2025}
}

@inproceedings{liu2025videodpo,
  title={Videodpo: Omni-preference alignment for video diffusion generation},
  author={Liu, Runtao and Wu, Haoyu and Zheng, Ziqiang and Wei, Chen and He, Yingqing and Pi, Renjie and Chen, Qifeng},
  booktitle={Proceedings of the Computer Vision and Pattern Recognition Conference},
  pages={8009--8019},
  year={2025}
}

@article{liu2022flow,
  title={Flow straight and fast: Learning to generate and transfer data with rectified flow},
  author={Liu, Xingchao and Gong, Chengyue and Liu, Qiang},
  journal={arXiv preprint arXiv:2209.03003},
  year={2022}
}

@article{xue2025dancegrpo,
  title={Dancegrpo: Unleashing grpo on visual generation},
  author={Xue, Zeyue and Wu, Jie and Gao, Yu and Kong, Fangyuan and Zhu, Lingting and Chen, Mengzhao and Liu, Zhiheng and Liu, Wei and Guo, Qiushan and Huang, Weilin and others},
  journal={arXiv preprint arXiv:2505.07818},
  year={2025}
}

@article{liu2025flow,
  title={Flow-grpo: Training flow matching models via online rl},
  author={Liu, Jie and Liu, Gongye and Liang, Jiajun and Li, Yangguang and Liu, Jiaheng and Wang, Xintao and Wan, Pengfei and Zhang, Di and Ouyang, Wanli},
  journal={arXiv preprint arXiv:2505.05470},
  year={2025}
}

@article{kirstain2023pick,
  title={Pick-a-pic: An open dataset of user preferences for text-to-image generation},
  author={Kirstain, Yuval and Polyak, Adam and Singer, Uriel and Matiana, Shahbuland and Penna, Joe and Levy, Omer},
  journal={Advances in neural information processing systems},
  volume={36},
  pages={36652--36663},
  year={2023}
}

@inproceedings{ma2025hpsv3,
  title={Hpsv3: Towards wide-spectrum human preference score},
  author={Ma, Yuhang and Wu, Xiaoshi and Sun, Keqiang and Li, Hongsheng},
  booktitle={Proceedings of the IEEE/CVF International Conference on Computer Vision},
  pages={15086--15095},
  year={2025}
}

@inproceedings{zhang2024learning,
  title={Learning multi-dimensional human preference for text-to-image generation},
  author={Zhang, Sixian and Wang, Bohan and Wu, Junqiang and Li, Yan and Gao, Tingting and Zhang, Di and Wang, Zhongyuan},
  booktitle={Proceedings of the IEEE/CVF Conference on Computer Vision and Pattern Recognition},
  pages={8018--8027},
  year={2024}
}

@article{xu2024visionreward,
  title={Visionreward: Fine-grained multi-dimensional human preference learning for image and video generation},
  author={Xu, Jiazheng and Huang, Yu and Cheng, Jiale and Yang, Yuanming and Xu, Jiajun and Wang, Yuan and Duan, Wenbo and Yang, Shen and Jin, Qunlin and Li, Shurun and others},
  journal={arXiv preprint arXiv:2412.21059},
  year={2024}
}

@inproceedings{ku2024viescore,
  title={Viescore: Towards explainable metrics for conditional image synthesis evaluation},
  author={Ku, Max and Jiang, Dongfu and Wei, Cong and Yue, Xiang and Chen, Wenhu},
  booktitle={Proceedings of the 62nd Annual Meeting of the Association for Computational Linguistics (Volume 1: Long Papers)},
  pages={12268--12290},
  year={2024}
}

@article{wei2025skywork,
  title={Skywork unipic 2.0: Building kontext model with online rl for unified multimodal model},
  author={Wei, Hongyang and Xu, Baixin and Liu, Hongbo and Wu, Size and Liu, Jie and Peng, Yi and Wang, Peiyu and Liu, Zexiang and He, Jingwen and Xietian, Yidan and others},
  journal={arXiv preprint arXiv:2509.04548},
  year={2025}
}

@article{wu2025rewarddance,
  title={Rewarddance: Reward scaling in visual generation},
  author={Wu, Jie and Gao, Yu and Ye, Zilyu and Li, Ming and Li, Liang and Guo, Hanzhong and Liu, Jie and Xue, Zeyue and Hou, Xiaoxia and Liu, Wei and others},
  journal={arXiv preprint arXiv:2509.08826},
  year={2025}
}

@article{gong2025onereward,
  title={Onereward: Unified mask-guided image generation via multi-task human preference learning},
  author={Gong, Yuan and Wang, Xionghui and Wu, Jie and Wang, Shiyin and Wang, Yitong and Wu, Xinglong},
  journal={arXiv preprint arXiv:2508.21066},
  year={2025}
}

@article{wu2025editreward,
  title={Editreward: A human-aligned reward model for instruction-guided image editing},
  author={Wu, Keming and Jiang, Sicong and Ku, Max and Nie, Ping and Liu, Minghao and Chen, Wenhu},
  journal={arXiv preprint arXiv:2509.26346},
  year={2025}
}

@article{luo2025editscore,
  title={Editscore: Unlocking online rl for image editing via high-fidelity reward modeling},
  author={Luo, Xin and Wang, Jiahao and Wu, Chenyuan and Xiao, Shitao and Jiang, Xiyan and Lian, Defu and Zhang, Jiajun and Liu, Dong and others},
  journal={arXiv preprint arXiv:2509.23909},
  year={2025}
}

@article{zhang2025reasongen,
  title={ReasonGen-R1: CoT for Autoregressive Image generation models through SFT and RL},
  author={Zhang, Yu and Li, Yunqi and Yang, Yifan and Wang, Rui and Yang, Yuqing and Qi, Dai and Bao, Jianmin and Chen, Dongdong and Luo, Chong and Qiu, Lili},
  journal={arXiv preprint arXiv:2505.24875},
  year={2025}
}

@article{jiang2025t2i,
  title={T2i-r1: Reinforcing image generation with collaborative semantic-level and token-level cot},
  author={Jiang, Dongzhi and Guo, Ziyu and Zhang, Renrui and Zong, Zhuofan and Li, Hao and Zhuo, Le and Yan, Shilin and Heng, Pheng-Ann and Li, Hongsheng},
  journal={arXiv preprint arXiv:2505.00703},
  year={2025}
}

@article{guo2024gaussian,
  title={Gaussian mixture solvers for diffusion models},
  author={Guo, Hanzhong and Lu, Cheng and Bao, Fan and Pang, Tianyu and Yan, Shuicheng and Du, Chao and Li, Chongxuan},
  journal={Advances in Neural Information Processing Systems},
  volume={36},
  year={2024}
}

@inproceedings{xue2024sa,
  title={SA-Solver: Stochastic Adams Solver for Fast Sampling of Diffusion Models},
  author={Xue, Shuchen and Yi, Mingyang and Luo, Weijian and Zhang, Shifeng and Sun, Jiacheng and Li, Zhenguo and Ma, Zhi-Ming},
  booktitle={Advances in Neural Information Processing Systems},
  volume={36},
  year={2024}
}

@inproceedings{lu2022dpm,
  title={DPM-Solver: A Fast ODE Solver for Diffusion Probabilistic Model Sampling in Around 10 Steps},
  author={Lu, Cheng and Zhou, Yuhao and Bao, Fan and Chen, Jianfei and Li, Chongxuan and Zhu, Jun},
  booktitle={Advances in Neural Information Processing Systems},
  volume={35},
  year={2022}
}

@inproceedings{zhang2022fast,
  title={Fast Sampling of Diffusion Models with Exponential Integrator},
  author={Zhang, Qinsheng and Chen, Yongxin},
  booktitle={International Conference on Learning Representations},
  year={2023}
}

@article{jolicoeur2021gotta,
  title={Gotta Go Fast When Generating Data with Score-Based Models},
  author={Jolicoeur-Martineau, Alexia and Li, Ke and Pich\'{e}-Taillefer, R\'{e}mi and Kachman, Tal and Mitliagkas, Ioannis},
  journal={arXiv preprint arXiv:2105.14080},
  year={2021}
}

@inproceedings{karras2022elucidating,
  title={Elucidating the Design Space of Diffusion-Based Generative Models},
  author={Karras, Tero and Aittala, Miika and Aila, Timo and Laine, Samuli},
  booktitle={Advances in Neural Information Processing Systems},
  volume={35},
  year={2022}
}

@inproceedings{meng2022on,
  title={On fast sampling of diffusion models},
  author={Meng, Chenlin and Song, Jiaming and Ermon, Stefano},
  booktitle={International Conference on Machine Learning},
  year={2022}
}

@inproceedings{liu2022pseudo,
  title={Pseudo Numerical Methods for Diffusion Models on Manifolds},
  author={Liu, Luping and Ren, Yi and Lin, Zhijie and Zhao, Zhou},
  booktitle={International Conference on Learning Representations},
  year={2022}
}

@article{cai2025z,
  title={Z-image: An efficient image generation foundation model with single-stream diffusion transformer},
  author={Cai, Huanqia and Cao, Sihan and Du, Ruoyi and Gao, Peng and Hoi, Steven and Hou, Zhaohui and Huang, Shijie and Jiang, Dengyang and Jin, Xin and Li, Liangchen and others},
  journal={arXiv preprint arXiv:2511.22699},
  year={2025}
}

@article{wu2025qwen,
  title={Qwen-image technical report},
  author={Wu, Chenfei and Li, Jiahao and Zhou, Jingren and Lin, Junyang and Gao, Kaiyuan and Yan, Kun and Yin, Sheng-ming and Bai, Shuai and Xu, Xiao and Chen, Yilei and others},
  journal={arXiv preprint arXiv:2508.02324},
  year={2025}
}

@article{gao2025seedream,
  title={Seedream 3.0 technical report},
  author={Gao, Yu and Gong, Lixue and Guo, Qiushan and Hou, Xiaoxia and Lai, Zhichao and Li, Fanshi and Li, Liang and Lian, Xiaochen and Liao, Chao and Liu, Liyang and others},
  journal={arXiv preprint arXiv:2504.11346},
  year={2025}
}

@article{seedream2025seedream,
  title={Seedream 4.0: Toward next-generation multimodal image generation},
  author={Seedream, Team and Chen, Yunpeng and Gao, Yu and Gong, Lixue and Guo, Meng and Guo, Qiushan and Guo, Zhiyao and Hou, Xiaoxia and Huang, Weilin and Huang, Yixuan and others},
  journal={arXiv preprint arXiv:2509.20427},
  year={2025}
}

@article{cao2025hunyuanimage,
  title={Hunyuanimage 3.0 technical report},
  author={Cao, Siyu and Chen, Hangting and Chen, Peng and Cheng, Yiji and Cui, Yutao and Deng, Xinchi and Dong, Ying and Gong, Kipper and Gu, Tianpeng and Gu, Xiusen and others},
  journal={arXiv preprint arXiv:2509.23951},
  year={2025}
}

@article{wang2025seededit,
  title={Seededit 3.0: Fast and high-quality generative image editing},
  author={Wang, Peng and Shi, Yichun and Lian, Xiaochen and Zhai, Zhonghua and Xia, Xin and Xiao, Xuefeng and Huang, Weilin and Yang, Jianchao},
  journal={arXiv preprint arXiv:2506.05083},
  year={2025}
}

@article{team2025longcat,
  title={Longcat-image technical report},
  author={Team, Meituan LongCat and Ma, Hanghang and Tan, Haoxian and Huang, Jiale and Wu, Junqiang and He, Jun-Yan and Gao, Lishuai and Xiao, Songlin and Wei, Xiaoming and Ma, Xiaoqi and others},
  journal={arXiv preprint arXiv:2512.07584},
  year={2025}
}

@article{team2026firered,
  title={FireRed-Image-Edit-1.0 Technical Report},
  author={Team, Super Intelligence and Qiao, Changhao and Hui, Chao and Li, Chen and Wang, Cunzheng and Song, Dejia and Zhang, Jiale and Li, Jing and Xiang, Qiang and Wang, Runqi and others},
  journal={arXiv preprint arXiv:2602.13344},
  year={2026}
}

@misc{jdjoyaiimage,
  title={JoyAI-Image: Awakening Spatial Intelligence in Unified Multimodal Understanding and Generation},
  author={{Joy Future Academy, JD}},
  year={2026},
  howpublished={Technical report},
  note={Available at \url{https://joyai-image.s3.cn-north-1.jdcloud-oss.com/JoyAI-Image.pdf}, accessed 2026-04-22}
}

@article{team2026longcat,
  title={LongCat-Next: Lexicalizing Modalities as Discrete Tokens},
  author={Team, Meituan LongCat and Xiao, Bin and Wang, Chao and Li, Chengjiang and Zhang, Chi and Peng, Chong and Yu, Hang and Yang, Hao and Yan, Haonan and Sun, Haoze and others},
  journal={arXiv preprint arXiv:2603.27538},
  year={2026}
}

@article{mao2026wan,
  title={Wan-Image: Pushing the Boundaries of Generative Visual Intelligence},
  author={Mao, Chaojie and Xie, Chen-Wei and Zhong, Chongyang and Deng, Haoyou and Zhao, Jiaxing and Xiao, Jie and Xing, Jinbo and Zhang, Jingfeng and Zhou, Jingren and Zhang, Jingyi and others},
  journal={arXiv preprint arXiv:2604.19858},
  year={2026}
}

@article{tai2025investigating,
  title={Investigating Text Insulation and Attention Mechanisms for Complex Visual Text Generation},
  author={Tai, Ying and Du, Nikai and Xie, Rui and Chen, Zhennan and Wang, Qian and Jiang, Zhengkai and Zhang, Kai and Yang, Jian},
  journal={arXiv e-prints},
  pages={arXiv--2503},
  year={2025}
}

@inproceedings{ma2025glyphdraw2,
  title={Glyphdraw2: Automatic generation of complex glyph posters with diffusion models and large language models},
  author={Ma, Jian and Deng, Yonglin and Chen, Chen and Du, Nanyang and Lu, Haonan and Yang, Zhenyu},
  booktitle={Proceedings of the AAAI Conference on Artificial Intelligence},
  volume={39},
  number={6},
  pages={5955--5963},
  year={2025}
}

@article{xiao2026spatialedit,
  title={SpatialEdit: Benchmarking Fine-Grained Image Spatial Editing},
  author={Xiao, Yicheng and Zhang, Wenhu and Song, Lin and Chen, Yukang and Li, Wenbo and Jiang, Nan and Ren, Tianhe and Lin, Haokun and Huang, Wei and Huang, Haoyang and others},
  journal={arXiv preprint arXiv:2604.04911},
  year={2026}
}

@misc{aaarena2025,
  title={Artificial Analysis Image Arena},
  author={{Artificial Analysis}},
  howpublished={\url{https://artificialanalysis.ai/text-to-image}},
  year={2025}
}

@inproceedings{xu2023restart,
  title={Restart sampling for improving generative processes},
  author={Xu, Yilun and Deng, Mingyang and Cheng, Xiang and Tian, Yonglong and Liu, Ziming and Jaakkola, Tommi},
  booktitle={Advances in Neural Information Processing Systems},
  year={2023}
}

@inproceedings{fang2023structural,
  title={Structural Pruning for Diffusion Models},
  author={Fang, Gongfan and Ma, Xinyin and Wang, Xinchao},
  booktitle={Advances in Neural Information Processing Systems},
  volume={36},
  year={2023}
}

@inproceedings{castells2024ld,
  title={LD-Pruner: Efficient Pruning of Latent Diffusion Models using Task-Agnostic Insights},
  author={Castells, Thibault and Song, Hyoung-Kyu and Kim, Bo-Kyeong and Choi, Shinkook},
  booktitle={Proceedings of the IEEE/CVF Conference on Computer Vision and Pattern Recognition Workshops},
  pages={821--830},
  year={2024}
}

@inproceedings{kim2024layermerge,
  title={LayerMerge: Neural Network Depth Compression through Layer Pruning and Merging},
  author={Kim, Jinuk and El Halabi, Marwa and Ji, Mingi and Song, Hyun Oh},
  booktitle={International Conference on Machine Learning},
  year={2024}
}

@article{zhang2024laptop,
  title={Laptop-diff: Layer pruning and normalized distillation for compressing diffusion models},
  author={Zhang, Dingkun and Li, Sijia and Chen, Chen and Xie, Qingsong and Lu, Haonan},
  journal={arXiv preprint arXiv:2404.11098},
  year={2024}
}

@inproceedings{shang2023post,
  title={Post-training quantization on diffusion models},
  author={Shang, Yuzhang and Yuan, Zhihang and Xie, Bin and Wu, Bingzhe and Yan, Yan},
  booktitle={Proceedings of the IEEE/CVF conference on computer vision and pattern recognition},
  pages={1972--1981},
  year={2023}
}

@inproceedings{songDDIM,
  title        = {Denoising Diffusion Implicit Models},
  author       = {Song, Jiaming and Meng, Chenlin and Ermon, Stefano},
  year         = 2021,
  booktitle    = {International Conference on Learning Representations}
}

@article{lu2022dpm++,
  title={DPM-Solver++: Fast Solver for Guided Sampling of Diffusion Probabilistic Models},
  author={Lu, Cheng and Zhou, Yuhao and Bao, Fan and Chen, Jianfei and Li, Chongxuan and Zhu, Jun},
  journal={arXiv preprint arXiv:2211.01095},
  year={2022}
}

@inproceedings{zheng2023dpmsolvervF,
  title={DPM-Solver-v3: Improved Diffusion ODE Solver with Empirical Model Statistics},
  author={Zheng, Kaiwen and Lu, Cheng and Chen, Jianfei and Zhu, Jun},
  booktitle={Advances in Neural Information Processing Systems},
  volume={36},
  year={2023}
}

@inproceedings{10377259,
  title={Q-Diffusion: Quantizing Diffusion Models},
  author={Li, Xiuyu and Liu, Yijiang and Lian, Long and Yang, Huanrui and Dong, Zhen and Kang, Daniel and Zhang, Shanghang and Keutzer, Kurt},
  booktitle={Proceedings of the IEEE/CVF International Conference on Computer Vision},
  pages={17535--17545},
  year={2023}
}

@inproceedings{kim2025ditto,
  title={Ditto: Accelerating Diffusion Model via Temporal Value Similarity},
  author={Kim, Sungbin and Lee, Hyunwuk and Cho, Wonho and Park, Mincheol and Ro, Won Woo},
  booktitle={IEEE International Symposium on High-Performance Computer Architecture},
  year={2025}
}

@inproceedings{bolya2023tomesd,
  title={Token Merging for Fast Stable Diffusion},
  author={Bolya, Daniel and Hoffman, Judy},
  booktitle={Proceedings of the IEEE/CVF Conference on Computer Vision and Pattern Recognition Workshops},
  pages={4599--4603},
  year={2023}
}

@inproceedings{kim2024tofu,
  title={Token Fusion: Bridging the Gap between Token Pruning and Token Merging},
  author={Kim, Minchul and Gao, Shangqian and Hsu, Yen-Chang and Shen, Yilin and Jin, Hongxia},
  booktitle={Proceedings of the IEEE/CVF Winter Conference on Applications of Computer Vision},
  year={2024}
}

@inproceedings{ma2024deepcache,
  title        = {Deepcache: Accelerating diffusion models for free},
  author       = {Ma, Xinyin and Fang, Gongfan and Wang, Xinchao},
  year         = 2024,
  booktitle    = {Proceedings of the IEEE/CVF Conference on Computer Vision and Pattern Recognition},
  pages        = {15762--15772}
}

@article{chen2024delta-dit,
  title        = {$\Delta$-DiT: A Training-Free Acceleration Method Tailored for Diffusion Transformers},
  author       = {Chen, Pengtao and Shen, Mingzhu and Ye, Peng and Cao, Jianjian and Tu, Chongjun and Bouganis, Christos-Savvas and Zhao, Yiren and Chen, Tao},
  year         = 2024,
  journal      = {arXiv preprint arXiv:2406.01125}
}

@misc{liu2024timestep,
    title={Timestep Embedding Tells: It's Time to Cache for Video Diffusion Model},
    author={Feng Liu and Shiwei Zhang and Xiaofeng Wang and Yujie Wei and Haonan Qiu and Yuzhong Zhao and Yingya Zhang and Qixiang Ye and Fang Wan},
    year={2024},
    eprint={2411.19108},
    archivePrefix={arXiv},
    primaryClass={cs.CV}
}

@article{zou2024accelerating,
  title        = {Accelerating Diffusion Transformers with Token-wise Feature Caching},
  author       = {Zou, Chang and Liu, Xuyang and Liu, Ting and Huang, Siteng and Zhang, Linfeng},
  year         = 2024,
  journal      = {arXiv preprint arXiv:2410.05317}
}

@article{liuReusingForecastingAccelerating2025,
  title = {From Reusing to Forecasting: Accelerating Diffusion Models with TaylorSeers},
  author = {Liu, Jiacheng and Zou, Chang and Lyu, Yuanhuiyi and Chen, Junjie and Zhang, Linfeng},
  journal = {arXiv preprint arXiv:2503.06923},
  year = {2025}
}

@article{liu2025freqca,
  title={Freqca: Accelerating diffusion models via frequency-aware caching},
  author={Liu, Jiacheng and Cai, Peiliang and Zhou, Qinming and Lin, Yuqi and Kong, Deyang and Huang, Benhao and Pan, Yupei and Xu, Haowen and Zou, Chang and Tang, Junshu and others},
  journal={arXiv preprint arXiv:2510.08669},
  year={2025}
}

@inproceedings{liu2025speca,
  title={Speca: Accelerating diffusion transformers with speculative feature caching},
  author={Liu, Jiacheng and Zou, Chang and Lyu, Yuanhuiyi and Ren, Fei and Wang, Shaobo and Li, Kaixin and Zhang, Linfeng},
  booktitle={Proceedings of the 33rd ACM International Conference on Multimedia},
  pages={10024--10033},
  year={2025}
}

@article{zhu2026tap,
  title={TAP: A Token-Adaptive Predictor Framework for Training-Free Diffusion Acceleration},
  author={Zhu, Haowei and Huang, Tingxuan and Wang, Xing and Zhao, Tianyu and Wang, Jiexi and Chen, Weifeng and Peng, Xurui and Chen, Fangmin and Yong, Junhai and Wang, Bin},
  journal={arXiv preprint arXiv:2603.03792},
  year={2026}
}

@article{zhu2026diffsparse,
  title={DiffSparse: Accelerating Diffusion Transformers with Learned Token Sparsity},
  author={Zhu, Haowei and Liu, Ji and Liu, Ziqiong and Li, Dong and Yong, Junhai and Wang, Bin and Barsoum, Emad},
  journal={arXiv preprint arXiv:2604.03674},
  year={2026}
}

@article{ma2024learning,
  title={Learning-to-cache: Accelerating diffusion transformer via layer caching},
  author={Ma, Xinyin and Fang, Gongfan and Bi Mi, Michael and Wang, Xinchao},
  journal={Advances in Neural Information Processing Systems},
  volume={37},
  pages={133282--133304},
  year={2024}
}

@inproceedings{fang2025tinyfusion,
  title={Tinyfusion: Diffusion transformers learned shallow},
  author={Fang, Gongfan and Li, Kunjun and Ma, Xinyin and Wang, Xinchao},
  booktitle={Proceedings of the Computer Vision and Pattern Recognition Conference},
  pages={18144--18154},
  year={2025}
}

@article{yuan2024ditfastattn,
  title={Ditfastattn: Attention compression for diffusion transformer models},
  author={Yuan, Zhihang and Zhang, Hanling and Lu, Pu and Ning, Xuefei and Zhang, Linfeng and Zhao, Tianchen and Yan, Shengen and Dai, Guohao and Wang, Yu},
  journal={Advances in Neural Information Processing Systems},
  volume={37},
  pages={1196--1219},
  year={2024}
}

@article{zhu2024dip,
  title={Dip-go: A diffusion pruner via few-step gradient optimization},
  author={Zhu, Haowei and Tang, Dehua and Liu, Ji and Lu, Mingjie and Zheng, Jintu and Peng, Jinzhang and Li, Dong and Wang, Yu and Jiang, Fan and Tian, Lu and others},
  journal={Advances in Neural Information Processing Systems},
  volume={37},
  pages={92581--92604},
  year={2024}
}

@misc{goodfellow2014generativeadversarialnetworks,
      title={Generative Adversarial Networks}, 
      author={Ian J. Goodfellow and Jean Pouget-Abadie and Mehdi Mirza and Bing Xu and David Warde-Farley and Sherjil Ozair and Aaron Courville and Yoshua Bengio},
      year={2014},
      eprint={1406.2661},
      archivePrefix={arXiv},
      primaryClass={stat.ML},
      url={https://arxiv.org/abs/1406.2661}, 
}

@misc{arjovsky2017wassersteingan,
      title={Wasserstein GAN}, 
      author={Martin Arjovsky and Soumith Chintala and Léon Bottou},
      year={2017},
      eprint={1701.07875},
      archivePrefix={arXiv},
      primaryClass={stat.ML},
      url={https://arxiv.org/abs/1701.07875}, 
}

@misc{gulrajani2017improvedtrainingwassersteingans,
      title={Improved Training of Wasserstein GANs}, 
      author={Ishaan Gulrajani and Faruk Ahmed and Martin Arjovsky and Vincent Dumoulin and Aaron Courville},
      year={2017},
      eprint={1704.00028},
      archivePrefix={arXiv},
      primaryClass={cs.LG},
      url={https://arxiv.org/abs/1704.00028}, 
}

@misc{mao2017squaresgenerativeadversarialnetworks,
      title={Least Squares Generative Adversarial Networks}, 
      author={Xudong Mao and Qing Li and Haoran Xie and Raymond Y. K. Lau and Zhen Wang and Stephen Paul Smolley},
      year={2017},
      eprint={1611.04076},
      archivePrefix={arXiv},
      primaryClass={cs.CV},
      url={https://arxiv.org/abs/1611.04076}, 
}

@misc{miyato2018spectralnormalizationgenerativeadversarial,
      title={Spectral Normalization for Generative Adversarial Networks}, 
      author={Takeru Miyato and Toshiki Kataoka and Masanori Koyama and Yuichi Yoshida},
      year={2018},
      eprint={1802.05957},
      archivePrefix={arXiv},
      primaryClass={cs.LG},
      url={https://arxiv.org/abs/1802.05957}, 
}

@misc{radford2016unsupervisedrepresentationlearningdeep,
      title={Unsupervised Representation Learning with Deep Convolutional Generative Adversarial Networks}, 
      author={Alec Radford and Luke Metz and Soumith Chintala},
      year={2016},
      eprint={1511.06434},
      archivePrefix={arXiv},
      primaryClass={cs.LG},
      url={https://arxiv.org/abs/1511.06434}, 
}

@misc{brock2019largescalegantraining,
      title={Large Scale GAN Training for High Fidelity Natural Image Synthesis}, 
      author={Andrew Brock and Jeff Donahue and Karen Simonyan},
      year={2019},
      eprint={1809.11096},
      archivePrefix={arXiv},
      primaryClass={cs.LG},
      url={https://arxiv.org/abs/1809.11096}, 
}

@misc{karras2019stylebasedgeneratorarchitecturegenerative,
      title={A Style-Based Generator Architecture for Generative Adversarial Networks}, 
      author={Tero Karras and Samuli Laine and Timo Aila},
      year={2019},
      eprint={1812.04948},
      archivePrefix={arXiv},
      primaryClass={cs.NE},
      url={https://arxiv.org/abs/1812.04948}, 
}

@misc{chen2016infoganinterpretablerepresentationlearning,
      title={InfoGAN: Interpretable Representation Learning by Information Maximizing Generative Adversarial Nets}, 
      author={Xi Chen and Yan Duan and Rein Houthooft and John Schulman and Ilya Sutskever and Pieter Abbeel},
      year={2016},
      eprint={1606.03657},
      archivePrefix={arXiv},
      primaryClass={cs.LG},
      url={https://arxiv.org/abs/1606.03657}, 
}

@misc{zhu2020unpairedimagetoimagetranslationusing,
      title={Unpaired Image-to-Image Translation using Cycle-Consistent Adversarial Networks}, 
      author={Jun-Yan Zhu and Taesung Park and Phillip Isola and Alexei A. Efros},
      year={2020},
      eprint={1703.10593},
      archivePrefix={arXiv},
      primaryClass={cs.CV},
      url={https://arxiv.org/abs/1703.10593}, 
}

@misc{song2021scorebasedgenerativemodelingstochastic,
      title={Score-Based Generative Modeling through Stochastic Differential Equations}, 
      author={Yang Song and Jascha Sohl-Dickstein and Diederik P. Kingma and Abhishek Kumar and Stefano Ermon and Ben Poole},
      year={2021},
      eprint={2011.13456},
      archivePrefix={arXiv},
      primaryClass={cs.LG},
      url={https://arxiv.org/abs/2011.13456}, 
}

@misc{rombach2022stablediffusion,
      title={High-Resolution Image Synthesis with Latent Diffusion Models}, 
      author={Robin Rombach and Andreas Blattmann and Dominik Lorenz and Patrick Esser and Björn Ommer},
      year={2022},
      eprint={2112.10752},
      archivePrefix={arXiv},
      primaryClass={cs.CV},
      url={https://arxiv.org/abs/2112.10752}, 
}

@misc{esser2024stablediffusion3,
      title={Scaling Rectified Flow Transformers for High-Resolution Image Synthesis}, 
      author={Patrick Esser and Sumith Kulal and Andreas Blattmann and Rahim Entezari and Jonas Müller and Harry Saini and Yam Levi and Dominik Lorenz and Axel Sauer and Frederic Boesel and Dustin Podell and Tim Dockhorn and Zion English and Kyle Lacey and Alex Goodwin and Yannik Marek and Robin Rombach},
      year={2024},
      eprint={2403.03206},
      archivePrefix={arXiv},
      primaryClass={cs.CV},
      url={https://arxiv.org/abs/2403.03206}, 
}

@misc{saharia2022imagegen,
      title={Photorealistic Text-to-Image Diffusion Models with Deep Language Understanding}, 
      author={Chitwan Saharia and William Chan and Saurabh Saxena and Lala Li and Jay Whang and Emily Denton and Seyed Kamyar Seyed Ghasemipour and Burcu Karagol Ayan and S. Sara Mahdavi and Rapha Gontijo Lopes and Tim Salimans and Jonathan Ho and David J Fleet and Mohammad Norouzi},
      year={2022},
      eprint={2205.11487},
      archivePrefix={arXiv},
      primaryClass={cs.CV},
      url={https://arxiv.org/abs/2205.11487}, 
}

@misc{deng2025bagelemerging,
      title={Emerging Properties in Unified Multimodal Pretraining}, 
      author={Chaorui Deng and Deyao Zhu and Kunchang Li and Chenhui Gou and Feng Li and Zeyu Wang and Shu Zhong and Weihao Yu and Xiaonan Nie and Ziang Song and Guang Shi and Haoqi Fan},
      year={2025},
      eprint={2505.14683},
      archivePrefix={arXiv},
      primaryClass={cs.CV},
      url={https://arxiv.org/abs/2505.14683}, 
}

@misc{chen2025janusprounifiedmultimodalunderstanding,
      title={Janus-Pro: Unified Multimodal Understanding and Generation with Data and Model Scaling}, 
      author={Xiaokang Chen and Zhiyu Wu and Xingchao Liu and Zizheng Pan and Wen Liu and Zhenda Xie and Xingkai Yu and Chong Ruan},
      year={2025},
      eprint={2501.17811},
      archivePrefix={arXiv},
      primaryClass={cs.AI},
      url={https://arxiv.org/abs/2501.17811}, 
}

@misc{lipman2023flowmatchinggenerativemodeling,
      title={Flow Matching for Generative Modeling}, 
      author={Yaron Lipman and Ricky T. Q. Chen and Heli Ben-Hamu and Maximilian Nickel and Matt Le},
      year={2023},
      eprint={2210.02747},
      archivePrefix={arXiv},
      primaryClass={cs.LG},
      url={https://arxiv.org/abs/2210.02747}, 
}

@misc{schusterbauer2025diff2flowtrainingflowmatching,
      title={Diff2Flow: Training Flow Matching Models via Diffusion Model Alignment}, 
      author={Johannes Schusterbauer and Ming Gui and Frank Fundel and Björn Ommer},
      year={2025},
      eprint={2506.02221},
      archivePrefix={arXiv},
      primaryClass={cs.CV},
      url={https://arxiv.org/abs/2506.02221}, 
}

@inproceedings{peebles2023scalable,
  title={Scalable Diffusion Models with Transformers},
  author={Peebles, William and Xie, Saining},
  booktitle={Proceedings of the IEEE/CVF International Conference on Computer Vision (ICCV)},
  year={2023},
  url={https://arxiv.org/abs/2212.09748},
}

@inproceedings{tian2024visual,
  title={Visual Autoregressive Modeling: Scalable Image Generation via Next-Scale Prediction},
  author={Tian, Keyu and Jiang, Yi and Yuan, Zehuan and Peng, Bingyue and Wang, Liwei},
  booktitle={Advances in Neural Information Processing Systems (NeurIPS)},
  year={2024},
  url={https://arxiv.org/abs/2404.02905},
}

@article{sun2024autoregressive,
  title={Autoregressive Model Beats Diffusion: {Llama} for Scalable Image Generation},
  author={Sun, Peize and Jiang, Yi and Chen, Shoufa and Zhang, Shilong and Peng, Bingyue and Luo, Ping and Yuan, Zehuan},
  journal={arXiv preprint arXiv:2406.06525},
  year={2024},
  url={https://arxiv.org/abs/2406.06525},
}

@article{zou2025omnimamba,
  title={{OmniMamba}: Efficient and Unified Multimodal Understanding and Generation via State Space Models},
  author={Zou, Jialv and Liao, Bencheng and Zhang, Qian and Liu, Wenyu and Wang, Xinggang},
  journal={arXiv preprint arXiv:2503.08686},
  year={2025},
  url={https://arxiv.org/abs/2503.08686},
}

@article{wang2025bridge,
  title={Growing Visual Generative Capacity for Pre-Trained {MLLMs}},
  author={Wang, Hanyu and Han, Jiaming and Yang, Ziyan and Zhao, Qi and Lin, Shanchuan and Yue, Xiangyu and Shrivastava, Abhinav and Yang, Zhenheng and Chen, Hao},
  journal={arXiv preprint arXiv:2510.01546},
  year={2025},
  url={https://arxiv.org/abs/2510.01546},
}

@article{chen2025blip3onext,
  title={{BLIP3o-NEXT}: Next Frontier of Native Image Generation},
  author={Chen, Jiuhai and Xue, Le and Xu, Zhiyang and Pan, Xichen and Yang, Shusheng and Qin, Can and Yan, An and Zhou, Honglu and Chen, Zeyuan and Huang, Lifu and Zhou, Tianyi and Li, Junnan and Savarese, Silvio and Xiong, Caiming and Xu, Ran},
  journal={arXiv preprint arXiv:2510.15857},
  year={2025},
  url={https://arxiv.org/abs/2510.15857},
}

@article{li2025back,
  title={Back to Basics: Let Denoising Generative Models Denoise},
  author={Li, Tianhong and He, Kaiming},
  journal={arXiv preprint arXiv:2511.13720},
  year={2025},
  url={https://arxiv.org/abs/2511.13720},
}

@article{huang2025r3gan,
  title={The {GAN} is Dead; Long Live the {GAN}! {A} Modern {GAN} Baseline},
  author={Huang, Yiwen and Gokaslan, Aaron and Kuleshov, Volodymyr and Tompkin, James},
  journal={arXiv preprint arXiv:2501.05441},
  year={2025},
  url={https://arxiv.org/abs/2501.05441},
}

@article{zheng2025rae,
  title={Diffusion Transformers with Representation Autoencoders},
  author={Zheng, Boyang and Ma, Nanye and Tong, Shengbang and Xie, Saining},
  journal={arXiv preprint arXiv:2510.11690},
  year={2025},
  url={https://arxiv.org/abs/2510.11690},
}

@article{shi2025svg,
  title={Latent Diffusion Model without Variational Autoencoder},
  author={Shi, Minglei and Wang, Haolin and Zheng, Wenzhao and Yuan, Ziyang and Wu, Xiaoshi and Wang, Xintao and Wan, Pengfei and Zhou, Jie and Lu, Jiwen},
  journal={arXiv preprint arXiv:2510.15301},
  year={2025},
  url={https://arxiv.org/abs/2510.15301},
}

@article{yu2024representation,
  title={Representation Alignment for Generation: Training Diffusion Transformers Is Easier Than You Think},
  author={Yu, Sihyun and Kwak, Sangkyung and Jang, Huiwon and Jeong, Jongheon and Huang, Jonathan and Shin, Jinwoo and Xie, Saining},
  journal={arXiv preprint arXiv:2410.06940},
  year={2024},
  url={https://arxiv.org/abs/2410.06940},
}

@article{han2025vision,
  title={Vision as a Dialect: Unifying Visual Understanding and Generation via Text-Aligned Representations},
  author={Han, Jiaming and Chen, Hao and Zhao, Yang and Wang, Hanyu and Zhao, Qi and Yang, Ziyan and He, Hao and Yue, Xiangyu and Jiang, Lu},
  journal={arXiv preprint arXiv:2506.18898},
  year={2025},
  url={https://arxiv.org/abs/2506.18898},
}

@article{yan2025uae,
  title={Unified Multimodal Model as Auto-Encoder},
  author={Yan, Zhiyuan and Lin, Kaiqing and Li, Zongjian and Ye, Junyan and Han, Hui and Wang, Zhendong and Liu, Hao and Lin, Bin and Li, Hao and Xu, Xue and Xiao, Xinyan and Wang, Jingdong and Wang, Haifeng and Yuan, Li},
  journal={arXiv preprint arXiv:2509.09666},
  year={2025},
  url={https://arxiv.org/abs/2509.09666},
}

@inproceedings{li2023gligen,
  title={Gligen: Open-set grounded text-to-image generation},
  author={Li, Yuheng and Liu, Haotian and Wu, Qingyang and Mu, Fangzhou and Yang, Jianwei and Gao, Jianfeng and Li, Chunyuan and Lee, Yong Jae},
  booktitle={Proceedings of the IEEE/CVF conference on computer vision and pattern recognition},
  pages={22511--22521},
  year={2023}
}

@inproceedings{wang2024instancediffusion,
  title={Instancediffusion: Instance-level control for image generation},
  author={Wang, Xudong and Darrell, Trevor and Rambhatla, Sai Saketh and Girdhar, Rohit and Misra, Ishan},
  booktitle={Proceedings of the IEEE/CVF conference on computer vision and pattern recognition},
  pages={6232--6242},
  year={2024}
}

@inproceedings{zhou2024migc,
  title={Migc: Multi-instance generation controller for text-to-image synthesis},
  author={Zhou, Dewei and Li, You and Ma, Fan and Zhang, Xiaoting and Yang, Yi},
  booktitle={Proceedings of the IEEE/CVF conference on computer vision and pattern recognition},
  pages={6818--6828},
  year={2024}
}

@inproceedings{zhou2025dreamrenderer,
  title={Dreamrenderer: Taming multi-instance attribute control in large-scale text-to-image models},
  author={Zhou, Dewei and Li, Mingwei and Yang, Zongxin and Yang, Yi},
  booktitle={Proceedings of the IEEE/CVF International Conference on Computer Vision},
  pages={16712--16722},
  year={2025}
}

@inproceedings{zhang2023controlnet,
  title={Adding Conditional Control to Text-to-Image Diffusion Models},
  author={Zhang, Lvmin and Rao, Anyi and Agrawala, Maneesh},
  booktitle={Proceedings of the IEEE/CVF International Conference on Computer Vision (ICCV)},
  year={2023},
  url={https://arxiv.org/abs/2302.05543},
}

@article{ho2022classifier,
  title={Classifier-Free Diffusion Guidance},
  author={Ho, Jonathan and Salimans, Tim},
  journal={arXiv preprint arXiv:2207.12598},
  year={2022},
  url={https://arxiv.org/abs/2207.12598},
}

@article{yang2025mmada,
  title={{MMaDA}: Multimodal Large Diffusion Language Models},
  author={Yang, Ling and Tian, Ye and Li, Bowen and Zhang, Xinchen and Shen, Ke and Tong, Yunhai and Wang, Mengdi},
  journal={arXiv preprint arXiv:2505.15809},
  year={2025},
  url={https://arxiv.org/abs/2505.15809},
}

@article{xie2024showo,
  title={Show-o: One Single Transformer to Unify Multimodal Understanding and Generation},
  author={Xie, Jinheng and Mao, Weijia and Bai, Zechen and Zhang, David Junhao and Wang, Weihao and Lin, Kevin Qinghong and Gu, Yuchao and Chen, Zhijie and Yang, Zhenheng and Shou, Mike Zheng},
  journal={arXiv preprint arXiv:2408.12528},
  year={2024},
  url={https://arxiv.org/abs/2408.12528},
}

@article{zhou2024transfusion,
  title={Transfusion: Predict the Next Token and Diffuse Images with One Multi-Modal Model},
  author={Zhou, Chunting and Yu, Lili and Babu, Arun and Tirumala, Kushal and Yasunaga, Michihiro and Shamis, Leonid and Kahn, Jacob and Ma, Xuezhe and Zettlemoyer, Luke and Levy, Omer},
  journal={arXiv preprint arXiv:2408.11039},
  year={2024},
  url={https://arxiv.org/abs/2408.11039},
}

@article{geng2025xomni,
  title={{X-Omni}: Reinforcement Learning Makes Discrete Autoregressive Image Generative Models Great Again},
  author={Geng, Zigang and Wang, Yibing and Ma, Yeyao and Li, Chen and Rao, Yongming and Gu, Shuyang and Zhong, Zhao and Lu, Qinglin and Hu, Han and Zhang, Xiaosong and Wang, Di and Jiang, Jie},
  journal={arXiv preprint arXiv:2507.22058},
  year={2025},
  url={https://arxiv.org/abs/2507.22058},
}

@article{chameleon2024,
  title={Chameleon: Mixed-Modal Early-Fusion Foundation Models},
  author={{Chameleon Team}},
  journal={arXiv preprint arXiv:2405.09818},
  year={2024},
  url={https://arxiv.org/abs/2405.09818},
}

@article{wang2024emu3,
  title={Emu3: Next-Token Prediction is All You Need},
  author={Wang, Xinlong and Zhang, Xiaosong and Luo, Zhengxiong and Sun, Quan and Cui, Yufeng and Wang, Jinsheng and Zhang, Fan and Wang, Yueze and Li, Zhen and Yu, Qiying and Zhao, Yingli and Ao, Yulong and Min, Xuebin and Li, Tao and Wu, Boya and Zhao, Bo and Zhang, Bowen and Wang, Liangdong and Liu, Guang and He, Zheqi and Yang, Xi and Liu, Jingjing and Lin, Yonghua and Huang, Tiejun and Wang, Zhongyuan},
  journal={arXiv preprint arXiv:2409.18869},
  year={2024},
  url={https://arxiv.org/abs/2409.18869},
}

@article{ma2024janusflow,
  title={{JanusFlow}: Harmonizing Autoregression and Rectified Flow for Unified Multimodal Understanding and Generation},
  author={Ma, Yiyang and Liu, Xingchao and Chen, Xiaokang and Liu, Wen and Wu, Chengyue and Wu, Zhiyu and Pan, Zizheng and Xie, Zhenda and Zhang, Haowei and Yu, Xingkai and Zhao, Liang and Wang, Yisong and Liu, Jiaying and Ruan, Chong},
  journal={arXiv preprint arXiv:2411.07975},
  year={2024},
  url={https://arxiv.org/abs/2411.07975},
}

@article{xie2025showo2,
  title={Show-o2: Improved Native Unified Multimodal Models},
  author={Xie, Jinheng and Yang, Zhenheng and Shou, Mike Zheng},
  journal={arXiv preprint arXiv:2506.15564},
  year={2025},
  url={https://arxiv.org/abs/2506.15564},
}

@article{pan2025metaqueries,
  title={Transfer between Modalities with {MetaQueries}},
  author={Pan, Xichen and Shukla, Satya Narayan and Singh, Aashu and Zhao, Zhuokai and Mishra, Shlok Kumar and Wang, Jialiang and Xu, Zhiyang and Chen, Jiuhai and Li, Kunpeng and Juefei-Xu, Felix and Hou, Ji and Xie, Saining},
  journal={arXiv preprint arXiv:2504.06256},
  year={2025},
  url={https://arxiv.org/abs/2504.06256},
}

@article{qu2024tokenflow,
  title={{TokenFlow}: Unified Image Tokenizer for Multimodal Understanding and Generation},
  author={Qu, Liao and Zhang, Huichao and Liu, Yiheng and Wang, Xu and Jiang, Yi and Gao, Yiming and Ye, Hu and Du, Daniel K. and Yuan, Zehuan and Wu, Xinglong},
  journal={arXiv preprint arXiv:2412.03069},
  year={2024},
  url={https://arxiv.org/abs/2412.03069},
}

@article{yang2025hermesflow,
  title={{HermesFlow}: Seamlessly Closing the Gap in Multimodal Understanding and Generation},
  author={Yang, Ling and Zhang, Xinchen and Tian, Ye and Shang, Chenming and Xu, Minghao and Zhang, Wentao and Cui, Bin},
  journal={arXiv preprint arXiv:2502.12148},
  year={2025},
  url={https://arxiv.org/abs/2502.12148},
}

@inproceedings{yu2025anyedit,
  title={Anyedit: Mastering unified high-quality image editing for any idea},
  author={Yu, Qifan and Chow, Wei and Yue, Zhongqi and Pan, Kaihang and Wu, Yang and Wan, Xiaoyang and Li, Juncheng and Tang, Siliang and Zhang, Hanwang and Zhuang, Yueting},
  booktitle={Proceedings of the Computer Vision and Pattern Recognition Conference},
  pages={26125--26135},
  year={2025}
}

@inproceedings{pu2025art,
  title={Art: Anonymous region transformer for variable multi-layer transparent image generation},
  author={Pu, Yifan and Zhao, Yiming and Tang, Zhicong and Yin, Ruihong and Ye, Haoxing and Yuan, Yuhui and Chen, Dong and Bao, Jianmin and Zhang, Sirui and Wang, Yanbin and others},
  booktitle={Proceedings of the Computer Vision and Pattern Recognition Conference},
  pages={7952--7962},
  year={2025}
}

@article{chen2025blip3o,
  title={Blip3o-next: Next frontier of native image generation},
  author={Chen, Jiuhai and Xue, Le and Xu, Zhiyang and Pan, Xichen and Yang, Shusheng and Qin, Can and Yan, An and Zhou, Honglu and Chen, Zeyuan and Huang, Lifu and others},
  journal={arXiv preprint arXiv:2510.15857},
  year={2025}
}

@article{chang2025bytemorph,
  title={Bytemorph: Benchmarking instruction-guided image editing with non-rigid motions},
  author={Chang, Di and Cao, Mingdeng and Shi, Yichun and Liu, Bo and Cai, Shengqu and Zhou, Shijie and Huang, Weilin and Wetzstein, Gordon and Soleymani, Mohammad and Wang, Peng},
  journal={arXiv preprint arXiv:2506.03107},
  year={2025}
}

@inproceedings{zhang2025diffusion,
  title={Diffusion-4k: Ultra-high-resolution image synthesis with latent diffusion models},
  author={Zhang, Jinjin and Huang, Qiuyu and Liu, Junjie and Guo, Xiefan and Huang, Di},
  booktitle={Proceedings of the Computer Vision and Pattern Recognition Conference},
  pages={23464--23473},
  year={2025}
}

@article{bertazzini2025dragon,
  title={DRAGON: A Large-Scale Dataset of Realistic Images Generated by Diffusion Models},
  author={Bertazzini, Giulia and Baracchi, Daniele and Shullani, Dasara and Echizen, Isao and Piva, Alessandro},
  journal={arXiv preprint arXiv:2505.11257},
  year={2025}
}

@article{jang2025dreamgen,
  title={Dreamgen: Unlocking generalization in robot learning through video world models},
  author={Jang, Joel and Ye, Seonghyeon and Lin, Zongyu and Xiang, Jiannan and Bjorck, Johan and Fang, Yu and Hu, Fengyuan and Huang, Spencer and Kundalia, Kaushil and Lin, Yen-Chen and others},
  journal={arXiv preprint arXiv:2505.12705},
  year={2025}
}

@inproceedings{zeng2025editworld,
  title={Editworld: Simulating world dynamics for instruction-following image editing},
  author={Zeng, Bohan and Yang, Ling and Liu, Jiaming and Xu, Minghao and Zhang, Yuanxing and Wan, Pengfei and Zhang, Wentao and Yan, Shuicheng},
  booktitle={Proceedings of the 33rd ACM International Conference on Multimedia},
  pages={12674--12681},
  year={2025}
}

@article{hu2024ella,
  title={Ella: Equip diffusion models with llm for enhanced semantic alignment},
  author={Hu, Xiwei and Wang, Rui and Fang, Yixiao and Fu, Bin and Cheng, Pei and Yu, Gang},
  journal={arXiv preprint arXiv:2403.05135},
  year={2024}
}

@article{ma2025veomni,
  title={Veomni: Scaling any modality model training with model-centric distributed recipe zoo},
  author={Ma, Qianli and Zheng, Yaowei and Shi, Zhelun and Zhao, Zhongkai and Jia, Bin and Huang, Ziyue and Lin, Zhiqi and Li, Youjie and Yang, Jiacheng and Peng, Yanghua and others},
  journal={arXiv preprint arXiv:2508.02317},
  year={2025}
}

@article{wang2025unirl,
  title={UniRL-Zero: Reinforcement Learning on Unified Models with Joint Language Model and Diffusion Model Experts},
  author={Wang, Fu-Yun and Zhang, Han and Gharbi, Michael and Li, Hongsheng and Park, Taesung},
  journal={arXiv preprint arXiv:2510.17937},
  year={2025}
}

@misc{unirl_github,
  title        = {{UniRL: A Reinforcement Learning Framework for Unified Multimodal Models}},
  author       = {Haonan Wang and Linyu Wu and Qian Qiu and Lewei Jin and Bowen Ping and Jianghai Chen and Yiheng Du and Guangxin He and Yu Shi and Yongguang Lin and Zhuoxin Zhou and Zhanchao Zhou and Keming Wu and Rizhen Hu and Xuefei Ning and Lvfang Tao and Feiyu Hu and Xiangyan Liu and Siqi Kou and Jiarui Yao and Xiangxin Zhou and Liefeng Bo and Wenxi Zhu and Tianyu Pang},
  year         = {2026},
  howpublished = {\url{https://github.com/Tencent-Hunyuan/UniRL}},
  urldate      = {2026-06-05}
}

@article{wu2026worldreasonbench,
  title={WorldReasonBench: Human-Aligned Stress Testing of Video Generators as Future World-State Predictors},
  author={Wu, Keming and Cui, Yijing and Xue, Wenhan and Wang, Qijie and Luo, Xuan and Feng, Zhiyuan and Yang, Zuhao and Wang, Sudong and Jiang, Sicong and Zhu, Haowei and others},
  journal={arXiv preprint arXiv:2605.10434},
  year={2026}
}

@article{wang2026promptrl,
  title={PromptRL: Prompt Matters in RL for Flow-Based Image Generation},
  author={Wang, Fu-Yun and Zhang, Han and Gharbi, Michael and Li, Hongsheng and Park, Taesung},
  journal={arXiv preprint arXiv:2602.01382},
  year={2026}
}

@article{zheng2024open,
  title={Open-sora: Democratizing efficient video production for all},
  author={Zheng, Zangwei and Peng, Xiangyu and Yang, Tianji and Shen, Chenhui and Li, Shenggui and Liu, Hongxin and Zhou, Yukun and Li, Tianyi and You, Yang},
  journal={arXiv preprint arXiv:2412.20404},
  year={2024}
}

@misc{von-platen-etal-2022-diffusers,
  author = {Patrick von Platen and Suraj Patil and Anton Lozhkov and Pedro Cuenca and Nathan Lambert and Kashif Rasul and Mishig Davaadorj and Dhruv Nair and Sayak Paul and William Berman and Yiyi Xu and Steven Liu and Thomas Wolf},
  title = {Diffusers: State-of-the-art diffusion models},
  year = {2022},
  publisher = {GitHub},
  journal = {GitHub repository},
  howpublished = {\url{https://github.com/huggingface/diffusers}}
}

@misc{mmagic2023,
    title = {{MMagic}: {OpenMMLab} Multimodal Advanced, Generative, and Intelligent Creation Toolbox},
    author = {{MMagic Contributors}},
    howpublished = {\url{https://github.com/open-mmlab/mmagic}},
    year = {2023}
}

@article{zhuo2025factuality,
  title={Factuality Matters: When Image Generation and Editing Meet Structured Visuals},
  author={Zhuo, Le and Han, Songhao and Pu, Yuandong and Qiu, Boxiang and Paul, Sayak and Liao, Yue and Liu, Yihao and Shao, Jie and Chen, Xi and Liu, Si and others},
  journal={arXiv preprint arXiv:2510.05091},
  year={2025}
}

@article{fang2025flux,
  title={Flux-reason-6m \& prism-bench: A million-scale text-to-image reasoning dataset and comprehensive benchmark},
  author={Fang, Rongyao and Yu, Aldrich and Duan, Chengqi and Huang, Linjiang and Bai, Shuai and Cai, Yuxuan and Wang, Kun and Liu, Si and Liu, Xihui and Li, Hongsheng},
  journal={arXiv preprint arXiv:2509.09680},
  year={2025}
}

@inproceedings{katara2024gen2sim,
  title={Gen2sim: Scaling up robot learning in simulation with generative models},
  author={Katara, Pushkal and Xian, Zhou and Fragkiadaki, Katerina},
  booktitle={2024 IEEE International Conference on Robotics and Automation (ICRA)},
  pages={6672--6679},
  year={2024},
  organization={IEEE}
}

@inproceedings{kumari2025generating,
  title={Generating multi-image synthetic data for text-to-image customization},
  author={Kumari, Nupur and Yin, Xi and Zhu, Jun-Yan and Misra, Ishan and Azadi, Samaneh},
  booktitle={Proceedings of the IEEE/CVF International Conference on Computer Vision},
  pages={16524--16534},
  year={2025}
}

@article{ghosh2023geneval,
  title={Geneval: An object-focused framework for evaluating text-to-image alignment},
  author={Ghosh, Dhruba and Hajishirzi, Hannaneh and Schmidt, Ludwig},
  journal={Advances in Neural Information Processing Systems},
  volume={36},
  pages={52132--52152},
  year={2023}
}

@article{wang2025genexam,
  title={GenExam: A Multidisciplinary Text-to-Image Exam},
  author={Wang, Zhaokai and Yin, Penghao and Zhao, Xiangyu and Tian, Changyao and Qiao, Yu and Wang, Wenhai and Dai, Jifeng and Luo, Gen},
  journal={arXiv preprint arXiv:2509.14232},
  year={2025}
}

@article{wang2025gpt,
  title={Gpt-image-edit-1.5 m: A million-scale, gpt-generated image dataset},
  author={Wang, Yuhan and Yang, Siwei and Zhao, Bingchen and Zhang, Letian and Liu, Qing and Zhou, Yuyin and Xie, Cihang},
  journal={arXiv preprint arXiv:2507.21033},
  year={2025}
}

@article{hui2024hq,
  title={Hq-edit: A high-quality dataset for instruction-based image editing},
  author={Hui, Mude and Yang, Siwei and Zhao, Bingchen and Shi, Yichun and Wang, Heng and Wang, Peng and Zhou, Yuyin and Xie, Cihang},
  journal={arXiv preprint arXiv:2404.09990},
  year={2024}
}

@inproceedings{sani2026imagenworld,
  title={ImagenWorld: Stress-Testing Image Generation Models with Explainable Human Evaluation on Open-ended Real-World Tasks},
  author={Sani, Samin Mahdizadeh and Ku, Max and Jamali, Nima and Sani, Matina Mahdizadeh and Khoshtab, Paria and Sun, Wei-Chieh and Fazel, Parnian and Tam, Zhi Rui and Chong, Thomas and Chan, Edisy Kin Wai and others},
  booktitle={The Fourteenth International Conference on Learning Representations},
  year={2026}
}

@article{ye2025imgedit,
  title={Imgedit: A unified image editing dataset and benchmark},
  author={Ye, Yang and He, Xianyi and Li, Zongjian and Lin, Bin and Yuan, Shenghai and Yan, Zhiyuan and Hou, Bohan and Yuan, Li},
  journal={arXiv preprint arXiv:2505.20275},
  year={2025}
}

@article{lepert2025masquerade,
  title={Masquerade: Learning from in-the-wild human videos using data-editing},
  author={Lepert, Marion and Fang, Jiaying and Bohg, Jeannette},
  journal={arXiv preprint arXiv:2508.09976},
  year={2025}
}

@inproceedings{chen2025multiref,
  title={MultiRef: Controllable Image Generation with Multiple Visual References},
  author={Chen, Ruoxi and Chen, Dongping and Wu, Siyuan and Wang, Sinan and Lang, Shiyun and Sushko, Peter and Jiang, Gaoyang and Wan, Yao and Krishna, Ranjay},
  booktitle={Proceedings of the 33rd ACM International Conference on Multimedia},
  pages={13325--13331},
  year={2025}
}

@article{pakdamansavoji2025improving,
  title={Improving Robotic Manipulation Robustness via NICE Scene Surgery},
  author={Pakdamansavoji, Sajjad and Pourkeshavarz, Mozhgan and Sigal, Adam and Li, Zhiyuan and Yang, Rui Heng and Rasouli, Amir},
  journal={arXiv preprint arXiv:2511.22777},
  year={2025}
}

@inproceedings{wei2024omniedit,
  title={Omniedit: Building image editing generalist models through specialist supervision},
  author={Wei, Cong and Xiong, Zheyang and Ren, Weiming and Du, Xeron and Zhang, Ge and Chen, Wenhu},
  booktitle={The Thirteenth International Conference on Learning Representations},
  year={2024}
}

@article{chen2025opengpt,
  title={Opengpt-4o-image: A comprehensive dataset for advanced image generation and editing},
  author={Chen, Zhihong and Bai, Xuehai and Shi, Yang and Fu, Chaoyou and Zhang, Huanyu and Wang, Haotian and Sun, Xiaoyan and Zhang, Zhang and Wang, Liang and Zhang, Yuanxing and others},
  journal={arXiv preprint arXiv:2509.24900},
  year={2025}
}

@article{qian2025pico,
  title={Pico-banana-400k: A large-scale dataset for text-guided image editing},
  author={Qian, Yusu and Bocek-Rivele, Eli and Song, Liangchen and Tong, Jialing and Yang, Yinfei and Lu, Jiasen and Hu, Wenze and Gan, Zhe},
  journal={arXiv preprint arXiv:2510.19808},
  year={2025}
}

@inproceedings{sushko2025realedit,
  title={Realedit: Reddit edits as a large-scale empirical dataset for image transformations},
  author={Sushko, Peter and Bharadwaj, Ayana and Lim, Zhi Yang and Ilin, Vasily and Caffee, Ben and Chen, Dongping and Salehi, Mohammadreza and Hsieh, Cheng-Yu and Krishna, Ranjay},
  booktitle={Proceedings of the Computer Vision and Pattern Recognition Conference},
  pages={13403--13413},
  year={2025}
}

@inproceedings{yuan2025roboengine,
  title={Roboengine: Plug-and-play robot data augmentation with semantic robot segmentation and background generation},
  author={Yuan, Chengbo and Joshi, Suraj and Zhu, Shaoting and Su, Hang and Zhao, Hang and Gao, Yang},
  booktitle={2025 IEEE/RSJ International Conference on Intelligent Robots and Systems (IROS)},
  pages={7622--7629},
  year={2025},
  organization={IEEE}
}

@inproceedings{tao2025robopearls,
  title={RoboPearls: editable video simulation for robot manipulation},
  author={Tao, Tang and Zhang, Likui and Wen, Youpeng and Zhang, Kaidong and Bian, Jia-Wang and Zhou, Xia and Yan, Tianyi and Zhan, Kun and Jia, Peng and Wu, Hefeng and others},
  booktitle={Proceedings of the IEEE/CVF International Conference on Computer Vision},
  pages={10118--10129},
  year={2025}
}

@article{yu2023scaling,
  title={Scaling robot learning with semantically imagined experience},
  author={Yu, Tianhe and Xiao, Ted and Stone, Austin and Tompson, Jonathan and Brohan, Anthony and Wang, Su and Singh, Jaspiar and Tan, Clayton and Peralta, Jodilyn and Ichter, Brian and others},
  journal={arXiv preprint arXiv:2302.11550},
  year={2023}
}

@article{ge2024seed,
  title={Seed-data-edit technical report: A hybrid dataset for instructional image editing},
  author={Ge, Yuying and Zhao, Sijie and Li, Chen and Ge, Yixiao and Shan, Ying},
  journal={arXiv preprint arXiv:2405.04007},
  year={2024}
}

@article{chen2025sharegpt,
  title={Sharegpt-4o-image: Aligning multimodal models with gpt-4o-level image generation},
  author={Chen, Junying and Cai, Zhenyang and Chen, Pengcheng and Chen, Shunian and Ji, Ke and Wang, Xidong and Yang, Yunjin and Wang, Benyou},
  journal={arXiv preprint arXiv:2506.18095},
  year={2025}
}

@article{liu2025step1x,
  title={Step1x-edit: A practical framework for general image editing},
  author={Liu, Shiyu and Han, Yucheng and Xing, Peng and Yin, Fukun and Wang, Rui and Cheng, Wei and Liao, Jiaqi and Wang, Yingming and Fu, Honghao and Han, Chunrui and others},
  journal={arXiv preprint arXiv:2504.17761},
  year={2025}
}

@article{wang2025textatlas5m,
  title={Textatlas5m: A large-scale dataset for dense text image generation},
  author={Wang, Alex Jinpeng and Mao, Dongxing and Zhang, Jiawei and Han, Weiming and Dong, Zhuobai and Li, Linjie and Lin, Yiqi and Yang, Zhengyuan and Qin, Libo and Zhang, Fuwei and others},
  journal={arXiv preprint arXiv:2502.07870},
  year={2025}
}

@article{wei2025tiif,
  title={TIIF-Bench: How Does Your T2I Model Follow Your Instructions?},
  author={Wei, Xinyu and Zhang, Jinrui and Wang, Zeqing and Wei, Hongyang and Guo, Zhen and Zhang, Lei},
  journal={arXiv preprint arXiv:2506.02161},
  year={2025}
}

@article{zhao2024ultraedit,
  title={Ultraedit: Instruction-based fine-grained image editing at scale},
  author={Zhao, Haozhe and Ma, Xiaojian and Chen, Liang and Si, Shuzheng and Wu, Rujie and An, Kaikai and Yu, Peiyu and Zhang, Minjia and Li, Qing and Chang, Baobao},
  journal={Advances in Neural Information Processing Systems},
  volume={37},
  pages={3058--3093},
  year={2024}
}

@article{taesiri2025understanding,
  title={Understanding Generative AI Capabilities in Everyday Image Editing Tasks},
  author={Taesiri, Mohammad Reza and Collins, Brandon and Bolton, Logan and Lai, Viet Dac and Dernoncourt, Franck and Bui, Trung and Nguyen, Anh Totti},
  journal={arXiv preprint arXiv:2505.16181},
  year={2025}
}

@article{geng2025x,
  title={X-omni: Reinforcement learning makes discrete autoregressive image generative models great again},
  author={Geng, Zigang and Wang, Yibing and Ma, Yeyao and Li, Chen and Rao, Yongming and Gu, Shuyang and Zhong, Zhao and Lu, Qinglin and Hu, Han and Zhang, Xiaosong and others},
  journal={arXiv preprint arXiv:2507.22058},
  year={2025}
}

@article{huang2023t2i,
  title={T2i-compbench: A comprehensive benchmark for open-world compositional text-to-image generation},
  author={Huang, Kaiyi and Sun, Kaiyue and Xie, Enze and Li, Zhenguo and Liu, Xihui},
  journal={Advances in Neural Information Processing Systems},
  volume={36},
  pages={78723--78747},
  year={2023}
}

@article{wu2025omnigen2,
  title={Omnigen2: Exploration to advanced multimodal generation},
  author={Wu, Chenyuan and Zheng, Pengfei and Yan, Ruiran and Xiao, Shitao and Luo, Xin and Wang, Yueze and Li, Wanli and Jiang, Xiyan and Liu, Yexin and Zhou, Junjie and others},
  journal={arXiv preprint arXiv:2506.18871},
  year={2025}
}

@inproceedings{AnyEdit,
  author = {Qifan Yu and Wei Chow and Zhongqi Yue and Kaihang Pan and Yang Wu and Xiaoyang Wan and Juncheng Li and Siliang Tang and Hanwang Zhang and Yueting Zhuang},
  title = {AnyEdit: Mastering Unified High-Quality Image Editing for Any Idea},
  booktitle = {Proceedings of the IEEE/CVF Conference on Computer Vision and Pattern Recognition (CVPR)},
  month = {June},
  year = {2025}
}

@article{ART_plus_multi-layer,
  title = {ART: Anonymous Region Transformer for Variable Multi-Layer Transparent Image Generation},
  author = {Yifan Pu and Yiming Zhao and Zhicong Tang and Ruihong Yin and Haoxing Ye and Yuhui Yuan and Dong Chen and Jianmin Bao and Sirui Zhang and Yanbin Wang and Lin Liang and Lijuan Wang and Ji Li and Xiu Li and Zhouhui Lian and Gao Huang and Baining Guo},
  journal = {arXiv preprint arXiv:2502.18364},
  year = {2025}
}

@article{BLIP3o-NEXT,
  title = {BLIP3o-NEXT: Next Frontier of Native Image Generation},
  author = {Jiuhai Chen and Le Xue and Zhiyang Xu and Xichen Pan and Shusheng Yang and Can Qin and An Yan and Honglu Zhou and Zeyuan Chen and Lifu Huang and Tianyi Zhou and Junnan Li and Silvio Savarese and Caiming Xiong and Ran Xu},
  journal = {arXiv preprint arXiv:2510.15857},
  year = {2025}
}

@article{ByteMorph,
  title = {ByteMorph: Benchmarking Instruction-Guided Image Editing with Non-Rigid Motions},
  author = {Di Chang and Mingdeng Cao and Yichun Shi and Bo Liu and Shengqu Cai and Shijie Zhou and Weilin Huang and Gordon Wetzstein and Mohammad Soleymani and Peng Wang},
  journal = {arXiv preprint arXiv:2506.03107},
  year = {2025}
}

@inproceedings{Diffusion-4K,
  title = {Diffusion-4K: Ultra-High-Resolution Image Synthesis with Latent Diffusion Models},
  author = {Jinjin Zhang and Qiuyu Huang and Junjie Liu and Xiefan Guo and Di Huang},
  booktitle = {Proceedings of the IEEE/CVF Conference on Computer Vision and Pattern Recognition (CVPR)},
  year = {2025}
}

@article{DRAGON,
  title = {DRAGON: A Large-Scale Dataset of Realistic Images Generated by Diffusion Models},
  author = {Giulia Bertazzini and Daniele Baracchi and Dasara Shullani and Isao Echizen and Alessandro Piva},
  journal = {arXiv preprint arXiv:2505.11257},
  year = {2025}
}

@article{EditWorld,
  title={EditWorld: Simulating World Dynamics for Instruction-Following Image Editing},
  author={Yang, Ling and Zeng, Bohan and Liu, Jiaming and Li, Hong and Xu, Minghao and Zhang, Wentao and Yan, Shuicheng},
  journal={arXiv preprint arXiv:2405.14785},
  year={2024}
}

@article{FLUX-Reason-6M_PRISM-Bench,
  title={FLUX-Reason-6M \& PRISM-Bench: A Million-Scale Text-to-Image Reasoning Dataset and Comprehensive Benchmark},
  author={Fang, Rongyao and Yu, Aldrich and Duan, Chengqi and Huang, Linjiang and Bai, Shuai and Cai, Yuxuan and Wang, Kun and Liu, Si and Liu, Xihui and Li, Hongsheng},
  journal={arXiv preprint arXiv:2509.09680},
  year={2025}
}

@article{Gen2Sim,
  title={Gen2Sim: Scaling up Robot Learning in Simulation with Generative Models},
  author={Katara, Pushkal and Xian, Zhou and Fragkiadaki, Katerina},
  journal={arXiv preprint arXiv:2310.18308},
  year={2023}
}

@article{GenEval,
  title={GenEval: An Object-Focused Framework for Evaluating Text-to-Image Alignment},
  author={Ghosh, Dhruba and Hajishirzi, Hannaneh and Schmidt, Ludwig},
  journal={arXiv preprint arXiv:2310.11513},
  year={2023}
}

@article{GenExam,
  title={GenExam: A Multidisciplinary Text-to-Image Exam},
  author={Wang, Zhaokai and Yin, Penghao and Zhao, Xiangyu and Tian, Changyao and Qiao, Yu and Wang, Wenhai and Dai, Jifeng and Luo, Gen},
  journal={arXiv preprint arXiv:2509.14232},
  year={2025}
}

@misc{ImagenWorld,
  title={{ImagenWorld}: Stress-Testing Image Generation Models with Explainable Human Evaluation on Open-ended Real-World Tasks},
  author={Samin Mahdizadeh Sani and Max Ku and Nima Jamali and Matina Mahdizadeh Sani and Paria Khoshtab and Wei-Chieh Sun and Parnian Fazel and Zhi Rui Tam and Thomas Chong and Edisy Kin Wai Chan and Donald Wai Tong Tsang and Chiao-Wei Hsu and Ting Wai Lam and Ho Yin Sam Ng and Chiafeng Chu and Chak-Wing Mak and Keming Wu and Hiu Tung Wong and Yik Chun Ho and Chi Ruan and Zhuofeng Li and I-Sheng Fang and Shih-Ying Yeh and Ho Kei Cheng and Ping Nie and Wenhu Chen},
  year={2025},
  doi={10.5281/zenodo.17344183},
  url={https://zenodo.org/records/17344183}
}

@inproceedings{peng2025bizgen,
  title={Bizgen: Advancing article-level visual text rendering for infographics generation},
  author={Peng, Yuyang and Xiao, Shishi and Wu, Keming and Liao, Qisheng and Chen, Bohan and Lin, Kevin and Huang, Danqing and Li, Ji and Yuan, Yuhui},
  booktitle={Proceedings of the Computer Vision and Pattern Recognition Conference},
  pages={23615--23624},
  year={2025}
}

@inproceedings{tsai2025completeme,
  title={CompleteMe: Reference-based Human Image Completion},
  author={Tsai, Yu-Ju and Price, Brian and Liu, Qing and Figueroa, Luis and Pakhomov, Daniil and Ding, Zhihong and Cohen, Scott and Yang, Ming-Hsuan},
  booktitle={Proceedings of the IEEE/CVF International Conference on Computer Vision},
  pages={18252--18261},
  year={2025}
}

@article{ImgEdit,
  title={{ImgEdit}: A Unified Image Editing Dataset and Benchmark},
  author={Ye, Yang and He, Xianyi and Li, Zongjian and Lin, Bin and Yuan, Shenghai and Yan, Zhiyuan and Hou, Bohan and Yuan, Li},
  journal={arXiv preprint arXiv:2505.20275},
  year={2025}
}

@inproceedings{MultiRef,
  title={{MultiRef}: Controllable Image Generation with Multiple Visual References},
  author={Chen, Ruoxi and Chen, Dongping and Wu, Siyuan and Wang, Sinan and Lang, Shiyun and Sushko, Petr and Jiang, Gaoyang and Wan, Yao and Krishna, Ranjay},
  booktitle={Proceedings of the 33rd ACM International Conference on Multimedia},
  year={2025},
  url={https://dl.acm.org/doi/10.1145/3746027.3758292}
}

@article{NICE,
  title={Improving Robotic Manipulation Robustness via {NICE} Scene Surgery},
  author={Pakdamansavoji, Sajjad and Pourkeshavarz, Mozhgan and Sigal, Adam and Li, Zhiyuan and Yang, Rui Heng and Rasouli, Amir},
  journal={arXiv preprint arXiv:2511.22777},
  year={2025}
}

@article{OneIG-Bench,
  title={{OneIG-Bench}: Omni-dimensional Nuanced Evaluation for Image Generation},
  author={Chang, Jingjing and Fang, Yixiao and Xing, Peng and Wu, Shuhan and Cheng, Wei and Wang, Rui and Zeng, Xianfang and Yu, Gang and Chen, Hai-Bao},
  journal={arXiv preprint arXiv:2506.07977},
  year={2025}
}

@article{OpenGPT-4o-Image,
  title={OpenGPT-4o-Image: A Comprehensive Dataset for Advanced Image Generation and Editing},
  author={Chen, Zhihong and Bai, Xuehai and Shi, Yang and Fu, Chaoyou and Zhang, Huanyu and Wang, Haotian and Sun, Xiaoyan and Zhang, Zhang and Wang, Liang and Zhang, Yuanxing and Wan, Pengfei and Zhang, Yi-Fan},
  journal={arXiv preprint arXiv:2509.24900},
  year={2025}
}

@inproceedings{REALEDIT,
  title={RealEdit: Reddit Edits As a Large-scale Empirical Dataset for Image Transformations},
  author={Sushko, Peter and Bharadwaj, Ayana and Lim, Zhi Yang and Ilin, Vasily and Caffee, Ben and Chen, Dongping and Salehi, Mohammadreza and Hsieh, Cheng-Yu and Krishna, Ranjay},
  booktitle={Proceedings of the IEEE/CVF Conference on Computer Vision and Pattern Recognition (CVPR)},
  pages={13403--13413},
  year={2025}
}

@article{RoboEngine,
  title={RoboEngine: Plug-and-Play Robot Data Augmentation with Semantic Robot Segmentation and Background Generation},
  author={Yuan, Chengbo and Joshi, Suraj and Zhu, Shaoting and Su, Hang and Zhao, Hang and Gao, Yang},
  journal={arXiv preprint arXiv:2503.18738},
  year={2025}
}

@inproceedings{RoboPearls,
  title={RoboPearls: Editable Video Simulation for Robot Manipulation},
  author={Tang, Tao and Zhang, Likui and Wen, Youpeng and Zhang, Kaidong and Bian, Jia-Wang and Zhou, Xia and Yan, Tianyi and Zhan, Kun and Jia, Peng and Wu, Hefeng and Lin, Liang and Liang, Xiaodan},
  booktitle={Proceedings of the IEEE/CVF International Conference on Computer Vision (ICCV)},
  year={2025}
}

@article{SEED-Data-Edit,
  title={SEED-Data-Edit Technical Report: A Hybrid Dataset for Instructional Image Editing},
  author={Ge, Yuying and Zhao, Sijie and Li, Chen and Ge, Yixiao and Shan, Ying},
  journal={arXiv preprint arXiv:2405.04007},
  year={2024}
}

@article{ShareGPT-4o-Image,
  title={ShareGPT-4o-Image: Aligning Multimodal Models with GPT-4o-Level Image Generation},
  author={Junying Chen and Zhenyang Cai and Pengcheng Chen and Shunian Chen and Ke Ji and Xidong Wang and Yunjin Yang and Benyou Wang},
  journal={arXiv preprint arXiv:2506.18095},
  year={2025}
}

@article{TextAtlas5M,
  title={TextAtlas5M: A Large-scale Dataset for Dense Text Image Generation},
  author={Alex Jinpeng Wang and Dongxing Mao and Jiawei Zhang and Weiming Han and Zhuobai Dong and Linjie Li and Yiqi Lin and Zhengyuan Yang and Libo Qin and Fuwei Zhang and Lijuan Wang and Min Li},
  journal={arXiv preprint arXiv:2502.07870},
  year={2025}
}

@article{TIIF-Bench,
  title={TIIF-Bench: How Does Your T2I Model Follow Your Instructions?},
  author={Xinyu Wei and Jinrui Zhang and Zeqing Wang and Hongyang Wei and Zhenzhao Guo and Lei Zhang},
  journal={arXiv preprint arXiv:2506.02161},
  year={2025}
}

@article{UltraEdit,
  title={UltraEdit: Instruction-based Fine-Grained Image Editing at Scale},
  author={Haozhe Zhao and Xiaojian Ma and Liang Chen and Shuzheng Si and Rujie Wu and Kaikai An and Peiyu Yu and Minjia Zhang and Qing Li and Baobao Chang},
  journal={arXiv preprint arXiv:2407.05282},
  year={2024}
}

@article{X-Omni,
  title={X-Omni: Reinforcement Learning Makes Discrete Autoregressive Image Generative Models Great Again},
  author={Zigang Geng and Yibing Wang and Yeyao Ma and Chen Li and Yongming Rao and Shuyang Gu and Zhao Zhong and Qinglin Lu and Han Hu and Xiaosong Zhang and Linus and Di Wang and Jie Jiang},
  journal={arXiv preprint arXiv:2507.22058},
  year={2025}
}

@inproceedings{lin2014microsoft,
  title={Microsoft coco: Common objects in context},
  author={Lin, Tsung-Yi and Maire, Michael and Belongie, Serge and Hays, James and Perona, Pietro and Ramanan, Deva and Doll{\'a}r, Piotr and Zitnick, C Lawrence},
  booktitle={European conference on computer vision},
  pages={740--755},
  year={2014},
  organization={Springer}
}

@inproceedings{agrawal2019nocaps,
  title={Nocaps: Novel object captioning at scale},
  author={Agrawal, Harsh and Desai, Karan and Wang, Yufei and Chen, Xinlei and Jain, Rishabh and Johnson, Mark and Batra, Dhruv and Parikh, Devi and Lee, Stefan and Anderson, Peter},
  booktitle={Proceedings of the IEEE/CVF international conference on computer vision},
  pages={8948--8957},
  year={2019}
}

@inproceedings{gurari2018vizwiz,
  title={Vizwiz grand challenge: Answering visual questions from blind people},
  author={Gurari, Danna and Li, Qing and Stangl, Abigale J and Guo, Anhong and Lin, Chi and Grauman, Kristen and Luo, Jiebo and Bigham, Jeffrey P},
  booktitle={Proceedings of the IEEE conference on computer vision and pattern recognition},
  pages={3608--3617},
  year={2018}
}

@inproceedings{sidorov2020textcaps,
  title={Textcaps: a dataset for image captioning with reading comprehension},
  author={Sidorov, Oleksii and Hu, Ronghang and Rohrbach, Marcus and Singh, Amanpreet},
  booktitle={European conference on computer vision},
  pages={742--758},
  year={2020},
  organization={Springer}
}

@article{schuhmann2022laion,
  title={Laion-5b: An open large-scale dataset for training next generation image-text models},
  author={Schuhmann, Christoph and Beaumont, Romain and Vencu, Richard and Gordon, Cade and Wightman, Ross and Cherti, Mehdi and Coombes, Theo and Katta, Aarush and Mullis, Clayton and Wortsman, Mitchell and others},
  journal={Advances in neural information processing systems},
  volume={35},
  pages={25278--25294},
  year={2022}
}

@inproceedings{deng2009imagenet,
  title={Imagenet: A large-scale hierarchical image database},
  author={Deng, Jia and Dong, Wei and Socher, Richard and Li, Li-Jia and Li, Kai and Fei-Fei, Li},
  booktitle={2009 IEEE conference on computer vision and pattern recognition},
  pages={248--255},
  year={2009},
  organization={Ieee}
}

@misc{kakaobrain2022coyo-700m,
  title         = {COYO-700M: Image-Text Pair Dataset},
  author        = {Minwoo Byeon and Beomhee Park and Haecheon Kim and Sungjun Lee and Woonhyuk Baek and Saehoon Kim},
  year          = {2022},
  howpublished  = {\url{https://github.com/kakaobrain/coyo-dataset}},
}

@article{gadre2023datacomp,
  title={Datacomp: In search of the next generation of multimodal datasets},
  author={Gadre, Samir Yitzhak and Ilharco, Gabriel and Fang, Alex and Hayase, Jonathan and Smyrnis, Georgios and Nguyen, Thao and Marten, Ryan and Wortsman, Mitchell and Ghosh, Dhruba and Zhang, Jieyu and others},
  journal={Advances in Neural Information Processing Systems},
  volume={36},
  pages={27092--27112},
  year={2023}
}

@inproceedings{yu2023mvimgnet,
  title={Mvimgnet: A large-scale dataset of multi-view images},
  author={Yu, Xianggang and Xu, Mutian and Zhang, Yidan and Liu, Haolin and Ye, Chongjie and Wu, Yushuang and Yan, Zizheng and Zhu, Chenming and Xiong, Zhangyang and Liang, Tianyou and others},
  booktitle={Proceedings of the IEEE/CVF conference on computer vision and pattern recognition},
  pages={9150--9161},
  year={2023}
}

@inproceedings{sharma2018conceptual,
  title={Conceptual captions: A cleaned, hypernymed, image alt-text dataset for automatic image captioning},
  author={Sharma, Piyush and Ding, Nan and Goodman, Sebastian and Soricut, Radu},
  booktitle={Proceedings of the 56th Annual Meeting of the Association for Computational Linguistics (Volume 1: Long Papers)},
  pages={2556--2565},
  year={2018}
}

@article{wang2023internvid,
  title={Internvid: A large-scale video-text dataset for multimodal understanding and generation},
  author={Wang, Yi and He, Yinan and Li, Yizhuo and Li, Kunchang and Yu, Jiashuo and Ma, Xin and Li, Xinhao and Chen, Guo and Chen, Xinyuan and Wang, Yaohui and others},
  journal={arXiv preprint arXiv:2307.06942},
  year={2023}
}

@inproceedings{damen2018scaling,
  title={Scaling egocentric vision: The epic-kitchens dataset},
  author={Damen, Dima and Doughty, Hazel and Farinella, Giovanni Maria and Fidler, Sanja and Furnari, Antonino and Kazakos, Evangelos and Moltisanti, Davide and Munro, Jonathan and Perrett, Toby and Price, Will and others},
  booktitle={Proceedings of the European conference on computer vision (ECCV)},
  pages={720--736},
  year={2018}
}

@inproceedings{deitke2023objaverse,
  title={Objaverse: A universe of annotated 3d objects},
  author={Deitke, Matt and Schwenk, Dustin and Salvador, Jordi and Weihs, Luca and Michel, Oscar and VanderBilt, Eli and Schmidt, Ludwig and Ehsani, Kiana and Kembhavi, Aniruddha and Farhadi, Ali},
  booktitle={Proceedings of the IEEE/CVF conference on computer vision and pattern recognition},
  pages={13142--13153},
  year={2023}
}

@inproceedings{liu2023zero,
  title={Zero-1-to-3: Zero-shot one image to 3d object},
  author={Liu, Ruoshi and Wu, Rundi and Van Hoorick, Basile and Tokmakov, Pavel and Zakharov, Sergey and Vondrick, Carl},
  booktitle={Proceedings of the IEEE/CVF international conference on computer vision},
  pages={9298--9309},
  year={2023}
}

@article{ma2024shapesplat,
  title={Shapesplat: A large-scale dataset of gaussian splats and their self-supervised pretraining},
  author={Ma, Qi and Li, Yue and Ren, Bin and Sebe, Nicu and Konukoglu, Ender and Gevers, Theo and Van Gool, Luc and Paudel, Danda Pani},
  journal={arXiv preprint arXiv:2408.10906},
  year={2024}
}

@inproceedings{liu24uco3d,
	Author = {Liu, Xingchen and Tayal, Piyush and Wang, Jianyuan and Zarzar, Jesus and Monnier, Tom and Tertikas, Konstantinos and Duan, Jiali and Toisoul, Antoine and Zhang, Jason Y. and Neverova, Natalia and Vedaldi, Andrea and Shapovalov, Roman and Novotny, David},
	Booktitle = {arXiv},
	Title = {UnCommon Objects in 3D},
	Year = {2025},
}

@article{sun2023journeydb,
  title={Journeydb: A benchmark for generative image understanding},
  author={Sun, Keqiang and Pan, Junting and Ge, Yuying and Li, Hao and Duan, Haodong and Wu, Xiaoshi and Zhang, Renrui and Zhou, Aojun and Qin, Zipeng and Wang, Yi and others},
  journal={Advances in neural information processing systems},
  volume={36},
  pages={49659--49678},
  year={2023}
}

@article{young2014image,
  title={From image descriptions to visual denotations: New similarity metrics for semantic inference over event descriptions},
  author={Young, Peter and Lai, Alice and Hodosh, Micah and Hockenmaier, Julia},
  journal={Transactions of the Association for Computational Linguistics},
  volume={2},
  pages={67--78},
  year={2014},
  publisher={MIT Press}
}

@article{GenAI-Arena,
  title={Genai arena: An open evaluation platform for generative models},
  author={Jiang, Dongfu and Ku, Max and Li, Tianle and Ni, Yuansheng and Sun, Shizhuo and Fan, Rongqi and Chen, Wenhu},
  journal={Advances in Neural Information Processing Systems},
  volume={37},
  pages={79889--79908},
  year={2024}
}

@inproceedings{ChatbotArena,
  title={Chatbot arena: An open platform for evaluating llms by human preference},
  author={Chiang, Wei-Lin and Zheng, Lianmin and Sheng, Ying and Angelopoulos, Anastasios Nikolas and Li, Tianle and Li, Dacheng and Zhu, Banghua and Zhang, Hao and Jordan, Michael and Gonzalez, Joseph E and others},
  booktitle={Forty-first International Conference on Machine Learning},
  year={2024}
}

@inproceedings{salimans2022progressive,
  title={Progressive distillation for fast sampling of diffusion models},
  author={Salimans, Tim and Ho, Jonathan and Dhariwal, Prafulla and others},
  booktitle={International Conference on Machine Learning},
  year={2022}
}

@article{geng2025mean,
  title={Mean flows for one-step generative modeling},
  author={Geng, Zhengyang and Deng, Mingyang and Bai, Xingjian and Kolter, J Zico and He, Kaiming},
  journal={arXiv preprint arXiv:2505.13447},
  year={2025}
}

@article{frans2024one,
  title={One step diffusion via shortcut models},
  author={Frans, Kevin and Hafner, Danijar and Levine, Sergey and Abbeel, Pieter},
  journal={arXiv preprint arXiv:2410.12557},
  year={2024}
}

@inproceedings{song2023consistency,
  title={Consistency Models},
  author={Song, Yang and Dhariwal, Prafulla and Chen, Mark and Sutskever, Ilya},
  booktitle={International Conference on Machine Learning},
  pages={32211--32252},
  year={2023},
  organization={PMLR}
}

@article{lu2024simplifying,
  title={Simplifying, stabilizing and scaling continuous-time consistency models},
  author={Lu, Cheng and Song, Yang},
  journal={arXiv preprint arXiv:2410.11081},
  year={2024}
}

@article{zheng2025rcm,
  title={Large Scale Diffusion Distillation via Score-Regularized Continuous-Time Consistency},
  author={Zheng, Kaiwen and Wang, Yuji and Ma, Qianli and Chen, Huayu and Zhang, Jintao and Balaji, Yogesh and Chen, Jianfei and Liu, Ming-Yu and Zhu, Jun and Zhang, Qinsheng},
  journal={arXiv preprint arXiv:2510.08431},
  year={2025}
}

@inproceedings{sauer2024adversarial,
  title={Adversarial diffusion distillation},
  author={Sauer, Axel and Lorenz, Dominik and Blattmann, Andreas and Rombach, Robin},
  booktitle={European Conference on Computer Vision},
  pages={87--103},
  year={2024},
  organization={Springer}
}

@inproceedings{yin2024one,
  title={One-step diffusion with distribution matching distillation},
  author={Yin, Tianwei and Gharbi, Micha{\"e}l and Zhang, Richard and Shechtman, Eli and Durand, Fredo and Freeman, William T and Park, Taesung},
  booktitle={Proceedings of the IEEE/CVF conference on computer vision and pattern recognition},
  pages={6613--6623},
  year={2024}
}

@article{yin2024improved,
  title={Improved distribution matching distillation for fast image synthesis},
  author={Yin, Tianwei and Gharbi, Micha{\"e}l and Park, Taesung and Zhang, Richard and Shechtman, Eli and Durand, Fredo and Freeman, Bill},
  journal={Advances in neural information processing systems},
  volume={37},
  pages={47455--47487},
  year={2024}
}

@article{nie2026tmd,
  title={Transition Matching Distillation for Fast Video Generation},
  author={Nie, Weili and Berner, Julius and Ma, Nanye and Liu, Chao and Xie, Saining and Vahdat, Arash},
  journal={arXiv preprint arXiv:2601.09881},
  year={2026}
}

@article{li2024mar,
  title={Autoregressive image generation without vector quantization},
  author={Li, Tianhong and Tian, Yonglong and Li, He and Deng, Mingyang and He, Kaiming},
  journal={Advances in Neural Information Processing Systems},
  volume={37},
  pages={56424--56445},
  year={2024}
}

@article{team2025nextstep1,
  title={Nextstep-1: Toward autoregressive image generation with continuous tokens at scale},
  author={Team, NextStep and Han, Chunrui and Li, Guopeng and Wu, Jingwei and Sun, Quan and Cai, Yan and Peng, Yuang and Ge, Zheng and Zhou, Deyu and Tang, Haomiao and others},
  journal={arXiv preprint arXiv:2508.10711},
  year={2025}
}
\clearpage
\end{document}